%% file: main.tex
\documentclass[a4paper]{article}

\usepackage{authblk}

\title{Synthetic Data for Deep Learning}
\author[1,2]{Sergey I. Nikolenko}
\affil[1]{Synthesis.ai, San Francisco, CA}
\affil[2]{Steklov Institute of Mathematics at St. Petersburg, Russia \vspace{.2cm} \newline \texttt{snikolenko@synthesis.ai}}

\input{defs}

\begin{document}

\maketitle

\begin{abstract}
Synthetic data is an increasingly popular tool for training deep learning models, especially in computer vision but also in other areas. In this work, we attempt to provide a comprehensive survey of the various directions in the development and application of synthetic data. First, we discuss synthetic datasets for basic computer vision problems, both low-level (e.g., optical flow estimation) and high-level (e.g., semantic segmentation), synthetic environments and datasets for outdoor and urban scenes (autonomous driving), indoor scenes (indoor navigation), aerial navigation, simulation environments for robotics, applications of synthetic data outside computer vision (in neural programming, bioinformatics, NLP, and more); we also survey the work on improving synthetic data development and alternative ways to produce it such as GANs. Second, we discuss in detail the synthetic-to-real domain adaptation problem that inevitably arises in applications of synthetic data, including synthetic-to-real refinement with GAN-based models and domain adaptation at the feature/model level without explicit data transformations. Third, we turn to privacy-related applications of synthetic data and review the work on generating synthetic datasets with differential privacy guarantees. We conclude by highlighting the most promising directions for further work in synthetic data studies.
\end{abstract}

\pagebreak

{
\setstretch{.96}

\tableofcontents
}

\pagebreak

\include{intro}

\include{static}

\include{env}

\include{other}

\include{bettersyn}

\include{domain}

\include{privacy}

\include{future}

\include{conclusion}

\bibliographystyle{plain}
\bibliography{synthetic}


\end{document}

%% file: defs.tex
\usepackage[utf8]{inputenc}
\usepackage{amsmath,amssymb,amsthm}
\usepackage{hyperref}
\usepackage{graphicx}
\usepackage{textcomp}
\usepackage{tikz}
\usepackage{pgfplots}
\usetikzlibrary{plotmarks,shapes,snakes,pgfplots.dateplot}
\usetikzlibrary{calc,trees,positioning,arrows,chains,shapes.geometric,%
    decorations.pathreplacing,decorations.pathmorphing,shapes,%
    matrix,shapes.symbols}
\usepackage{enumerate}
\usepackage{makecell}
\usepackage{multirow}
\usepackage{cite}
\usepackage{setspace}

\usepackage{array}
\newcolumntype{P}[1]{>{\centering\arraybackslash}p{#1}}

\newcommand\myp[2][\linewidth]{\includegraphics[width=#1]{picdraft/#2}}

\def\X{\mathcal{X}}

\def\LL{\mathcal{L}}
\def\U{\mathcal{U}}

\def\E{\mathbb{E}}

\newcommand\EE[1]{{\mathbb E}\left[ #1 \right]}
\newcommand\EEE[2]{{\mathbb E}_{#1}\left[ #2 \right]}

\def\x{\mathbf{x}}
\def\s{\mathbf{s}}
\def\y{\mathbf{y}}
\def\bd{\mathbf{d}}

\def\z{\mathbf{z}}
\def\h{\mathbf{h}}
\def\xs{\x_S}
\def\xt{\x_T}
\def\zs{\z_S}
\def\zt{\z_T}
\def\ys{\y_S}

\def\D{\mathcal{D}}

\def\X{\mathcal{X}}

\def\Xs{\mathcal{X}_S}
\def\Xt{\mathcal{X}_T}

\def\z{\mathbf{z}}

\def\btheta{\mathbf{\theta}}

\def\hx{{\hat{\mathbf{x}}}}
\def\bphi{\mathbf{\phi}}
\def\l{\mathcal{L}}

\def\hx{\hat{\x}}
\def\hy{\hat{\y}}
\def\td{\tilde{\bd}}
\def\ty{\tilde{\y}}
\def\hxs{\hx_S}
\def\hxt{\hx_T}
\newcommand\xssub[1]{\x_{S,#1}}
\newcommand\xtsub[1]{\x_{T,#1}}
\newcommand\hxssub[1]{\hx_{S,#1}}

\def\m{\mathbf{m}}

\def\grt{G^{\textsc{Ref}}_{\btheta}}
\def\ddp{D^{\textsc{Ref}}_{\bphi}}
\def\ldp{\LL^{\textsc{Ref}}_{D}(\bphi)}
\def\lgreal{\LL^{\textsc{Ref}}_{\mathrm{real}}}
\def\lgreg{\LL^{\textsc{Ref}}_{\mathrm{reg}}}
\def\lgr{\LL^{\textsc{Ref}}_{G}(\btheta)}

\def\ggaze{G^{\textsc{Gz}}}
\def\fgaze{F^{\textsc{Gz}}}
\def\egaze{E^{\textsc{Gz}}}
\def\dsgaze{D^{\textsc{Gz}}_S}
\def\drgaze{D^{\textsc{Gz}}_R}
\def\lgaze{\LL^{\textsc{Gz}}}

\def\gda{G^{\textsc{DA}}_{\btheta}}
\def\dda{D^{\textsc{DA}}_{\bphi}}
\def\fda{C^{\textsc{DA}}_{\bphi}}
\def\lda{\LL^{\textsc{DA}}}

\def\dsyn{D_{\mathrm{syn}}}
\def\psyn{p_{\mathrm{syn}}}
\def\dreal{D_{\mathrm{real}}}
\def\preal{p_{\mathrm{real}}}
\def\pdata{p_{\mathrm{data}}}

\def\gtp{G^{\mathrm{TP}}}
\def\dtp{D^{\mathrm{TP}}}
\def\ltp{\LL^{\mathrm{TP}}}

\def\god{G^{\textsc{OD}}}
\def\dod{D^{\textsc{OD}}}
\def\fod{F^{\textsc{OD}}}
\def\lod{\LL^{\textsc{OD}}}
\def\lodgan{\LL^{\textsc{OD}}_{\mathrm{GAN}}}
\def\lodcyc{\LL^{\textsc{OD}}_{\mathrm{cyc}}}
\def\lodbg{\LL^{\textsc{OD}}_{\mathrm{bg}}}
\def\lodfg{\LL^{\textsc{OD}}_{\mathrm{fg}}}

\def\gssim{G^{\textsc{SSIM}}}
\def\dssim{D^{\textsc{SSIM}}}
\def\fssim{F^{\textsc{SSIM}}}
\def\lssim{\LL^{\textsc{SSIM}}}
\def\lssimgan{\LL^{\textsc{SSIM}}_{\mathrm{GAN}}}
\def\lssimcyc{\LL^{\textsc{SSIM}}_{\mathrm{cyc}}}
\def\lssimse{\LL^{\textsc{SSIM}}_{\mathrm{SE}}}

\def\CE{\mathrm{CE}}

\def\gptp{G^{\textsc{p2p}}}
\def\dptp{D^{\textsc{p2p}}}
\def\eptp{E^{\textsc{p2p}}}
\def\lptpg{\LL^{\textsc{p2p}}_{\mathrm{GAN}}}
\def\lptpfm{\LL^{\textsc{p2p}}_{\mathrm{FM}}}
\def\lptppr{\LL^{\textsc{p2p}}_{\mathrm{PR}}}

\def\ggioimg{G^{\textsc{GIO}}_{\mathrm{img}}}
\def\ggiotask{G^{\textsc{GIO}}_{\mathrm{task}}}

\def\dgioimg{D^{\textsc{GIO}}_{\mathrm{img}}}
\def\dgioout{D^{\textsc{GIO}}_{\mathrm{out}}}

\def\lgioseg{\LL^{\textsc{GIO}}_{\mathrm{seg}}}
\def\lgiodepth{\LL^{\textsc{GIO}}_{\mathrm{depth}}}
\def\lgioimage{\LL^{\textsc{GIO}}_{\mathrm{image}}}
\def\lgioout{\LL^{\textsc{GIO}}_{\mathrm{out}}}

\def\gmed{G^{\textsc{med}}}
\def\dmed{D^{\textsc{med}}}
\def\decmed{\mathrm{Dec}^{\textsc{med}}}
\def\encmed{\mathrm{Enc}^{\textsc{med}}}

\def\eswayve{E^{\textsc{wve}}_S}
\def\etwayve{E^{\textsc{wve}}_T}
\def\gswayve{G^{\textsc{wve}}_S}
\def\gtwayve{G^{\textsc{wve}}_T}
\def\dswayve{D^{\textsc{wve}}_S}
\def\dtwayve{D^{\textsc{wve}}_T}
\def\cwayve{C^{\textsc{wve}}}
\def\lwayve{\LL^{\textsc{wve}}}
\def\lrecwayve{\LL^{\textsc{wve}}_{\mathrm{rec}}}
\def\lcycwayve{\LL^{\textsc{wve}}_{\mathrm{cyc}}}
\def\lcontrolwayve{\LL^{\textsc{wve}}_{\mathrm{ctrl}}}
\def\lccycwayve{\LL^{\textsc{wve}}_{\mathrm{cyctrl}}}
\def\llswayve{\LL^{\textsc{wve}}_{\mathrm{LSGAN}}}
\def\lpercwayve{\LL^{\textsc{wve}}_{\mathrm{perc}}}
\def\lzrecwayve{\LL^{\textsc{wve}}_{\mathrm{zrec}}}

\def\gvol{G^{\textsc{vol}}}
\def\dvol{D^{\textsc{vol}}}
\def\svol{S^{\textsc{vol}}}
\def\lvol{\LL^{\textsc{vol}}}
\def\lvolg{\LL^{\textsc{vol}}_{\mathrm{GAN}}}
\def\lvolcyc{\LL^{\textsc{vol}}_{\mathrm{cyc}}}
\def\lvolshape{\LL^{\textsc{vol}}_{\mathrm{shape}}}

\def\gttwo{G^{\textsc{T2}}}
\def\gsttwo{\gttwo_S}
\def\dttwo{D^{\textsc{T2}}}
\def\drttwo{\dttwo_T}
\def\dfttwo{\dttwo_{f}}
\def\fttwo{f^{\textsc{T2}}_{\mathrm{task}}}
\def\lttwo{\LL^{\textsc{T2}}}
\def\lttwog{\LL^{\textsc{T2}}_{\mathrm{GAN}}}
\def\lttwogf{\LL^{\textsc{T2}}_{\mathrm{GAN}_f}}

\def\gsda{G^{\textsc{SDA}}}
\def\ghssda{{\hat{G}}^{\textsc{SDA}}}
\def\dsda{D^{\textsc{SDA}}}

\def\fsda{f^{\textsc{SDA}}}
\def\tsda{T^{\textsc{SDA}}}
\def\lsda{\LL^{\textsc{SDA}}}
\def\lsdag{\LL^{\textsc{SDA}}_{\mathrm{GAN}}}

\def\gvrg{G^{\textsc{VRG}}}

\def\dvrg{D^{\textsc{VRG}}}

\def\fvrg{f^{\textsc{VRG}}}

\def\lvrg{\LL^{\textsc{VRG}}}
\def\lvrgg{\LL^{\textsc{VRG}}_{\mathrm{GAN}}}
\def\lvrgcyc{\LL^{\textsc{VRG}}_{\mathrm{cyc}}}
\def\lvrgsem{\LL^{\textsc{VRG}}_{\mathrm{sem}}}
\def\lvrgshift{\LL^{\textsc{VRG}}_{\mathrm{shift}}}

\def\ffcnw{f^{\textsc{FCNW}}}

\def\lfcnw{\LL^{\textsc{FCNW}}}

\def\lcca{\LL^{\textsc{CCA}}}

\def\ddsn{D^{\textsc{DSN}}}
\def\edsn{E^{\textsc{DSN}}}

\def\fdsn{f^{\textsc{DSN}}}

\def\ldsn{\LL^{\textsc{DSN}}}

\def\gfuse{G^{\textsc{Fus}}}

\def\dfuse{D^{\textsc{Fus}}}

\def\lfuse{\LL^{\textsc{Fus}}}

\def\gfil{G^{\textsc{fil}}}
\def\dfil{D^{\textsc{fil}}}

\def\lfil{\LL^{\textsc{fil}}}
\def\lfilgan{\LL^{\textsc{fil}}_{\mathrm{GAN}}}
\def\lfilcont{\LL^{\textsc{fil}}_{\mathrm{cont}}}
\def\lfilsty{\LL^{\textsc{fil}}_{\mathrm{sty}}}
\def\lfiltv{\LL^{\textsc{fil}}_{\mathrm{TV}}}

\def\gpix{G^{\textsc{pix}}}
\def\dpix{D^{\textsc{pix}}}
\def\tpix{T^{\textsc{pix}}}

\def\lpixd{\LL^{\textsc{pix}}_{\mathrm{dom}}}
\def\lpixt{\LL^{\textsc{pix}}_{\mathrm{task}}}
\def\lpixc{\LL^{\textsc{pix}}_{\mathrm{cont}}}

\def\expectation{\mathop{\mathbb{E}}}

\def\exps{\expectation\nolimits_{S}}
\def\expt{\expectation\nolimits_{T}}

\newcommand\esb[1]{\exps\left[#1\right]}
\newcommand\etb[1]{\expt\left[#1\right]}

\pgfdeclarelayer{background}
\pgfdeclarelayer{foreground}
\pgfsetlayers{background,main,foreground}

\tikzset{every picture/.style={semithick},every path/.style={very thick,rounded corners,->}}

\tikzset{
  ne/.style={
    draw=none, fill=none,
    font=\sffamily, 
    minimum height=0em,
    text centered},
  nd/.style={
    font=\sffamily, 
    text centered},
  diablo/.style={
    rectangle, 
    rounded corners, 
    draw=black, thick,
    text width=10em, 
    font=\sffamily,
    minimum height=3em, 
    text centered},
  branch/.style ={circle,inner sep=0pt,minimum size=1.5mm,fill=black,draw=black},
  diablo2/.style={
    rectangle, 
    rounded corners, 
    fill=red!10,
    draw=black!80,thick,
    text width=3em, 
    font=\sffamily,
    minimum height=2.7em, 
    text centered},
  dialoss/.style={
    diablo2,
    fill=green!10,
  },
  diablo3/.style={
    rectangle, 
    rounded corners, 
    fill=blue!10,
    draw=blue!40,thick,
    text width=3.5em, 
    font=\sffamily\bfseries,
    text=blue,
    minimum height=1.5em,
    text centered},
  line/.style={draw=red,rounded corners,thick, ->, decoration={markings,mark=at position 1 with %
    {\arrow[scale=4,>=stealth]{>}}},postaction={decorate}},
  element/.style={
    tape,
    top color=white,
    bottom color=blue!50!black!60!,
    minimum width=8em,
    draw=blue!40!black!90, very thick,
    text width=10em, 
    minimum height=3.5em, 
    text centered, 
    on chain},
  every join/.style={->,rounded corners,thick,shorten >=1pt},
  decoration={brace},
  lineblue/.style={
  	join,line width=.07cm,->,blue!20
  }
}

%% file: intro.tex
\section{Introduction}\label{sec:intro}

\begin{figure}[!t]\centering
\setlength{\tabcolsep}{3pt}
\begin{tabular}{P{.38\linewidth}P{.19\linewidth}P{.4\linewidth}}
\myp{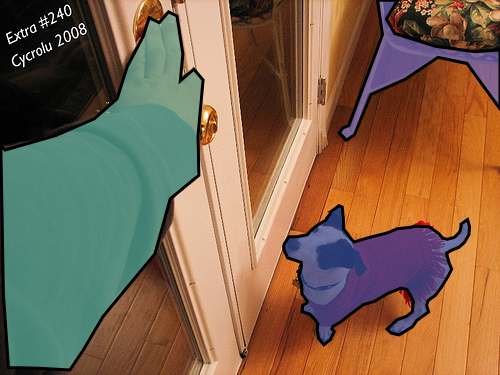} & \myp{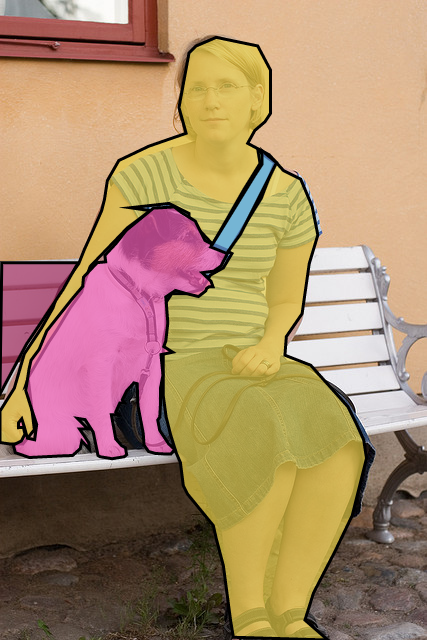} & \myp{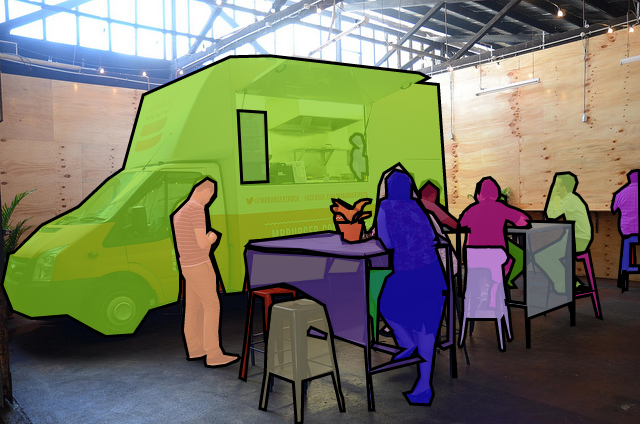} \\
(a) & (b) & (c) \\
\multicolumn{3}{c}{\myp{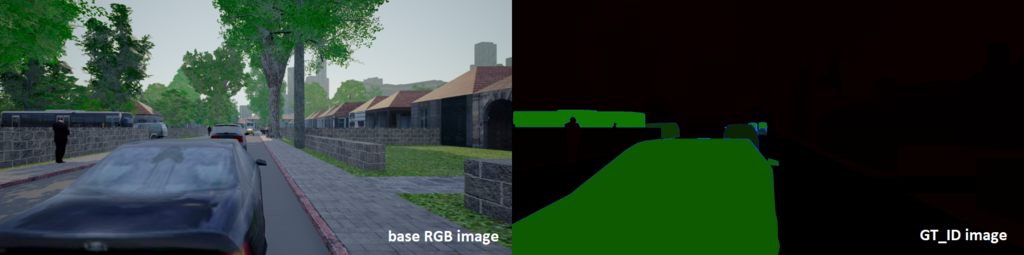}}\\
\multicolumn{3}{c}{
 \begin{tabular}{P{.49\linewidth}P{.49\linewidth}}
	(d) & (e) \\
	\end{tabular} } \\
\end{tabular}

\caption{Sample images: (a-c) MS COCO~\cite{DBLP:journals/corr/LinMBHPRDZ14} real data samples with ground truth segmentation maps overlaid; (d-e) \emph{ProcSy}~\cite{Khan_2019_CVPR_Workshops}: (d) RGB image, (e) ground truth segmentation map.}\label{pic:intro}
\end{figure}

Consider segmentation, a standard computer vision problem. How does one produce a labeled dataset for image segmentation? At some point, all images have to be manually processed: humans have to either draw or at least verify and correct segmentation masks. Making the result pixel-perfect is so laborious that it is commonly considered to be not worth the effort. Figure~\ref{pic:intro}a-c shows samples from the industry standard \emph{Microsoft Common Objects in Context} (MS COCO) dataset~\cite{DBLP:journals/corr/LinMBHPRDZ14}; you can immediately see that the segmentation mask is a rather rough polygon and misses many finer features. It did not take us long to find such rough segmentation maps, by the way; these are some of the first images found by the ``dog'' and ``person'' queries.

How can one get a higher quality segmentation dataset? To manually correct all of these masks in the MS COCO dataset would probably cost hundreds of thousand dollars. Fortunately, there is a different solution: \emph{synthetic data}. In the context of segmentation, this means that the dataset developers create a 3D environment with modes of the objects they want to recognize and their surroundings and then render the result. Figure~\ref{pic:intro}d-e shows a sample frame from a synthetic dataset called \emph{ProcSy}~\cite{Khan_2019_CVPR_Workshops} (we discuss it in more detail in Section~\ref{sec:visiongeneral}): note how the segmentation map is now perfectly correct. While 3D modeling is still mostly manual labor, this is a one-time investment, and as a result one can get a potentially unlimited number of pixel-perfect labeled data: not only RGB images and segmentation maps but also depth images, stereo pairs produced from different viewpoints, point clouds, synthetic video clips, and other modalities.

In general, many problems of modern AI come down to insufficient data: either the available datasets are too small or, also very often, even while capturing unlabeled data is relatively easy the costs of manual labeling are prohibitively high. \emph{Synthetic data} is an important approach to solving the data problem by either producing artificial data from scratch or using advanced data manipulation techniques to produce novel and diverse training examples. The synthetic data approach is most easily exemplified by standard computer vision problems, as we have done above, but it is also relevant in other domains. Naturally, other problems arise, the most important of them being the problem of domain transfer: synthetic images, as you can see from Figure~\ref{pic:intro}, do not look exactly like real images, and one has to make them as photorealistic as possible (a common theme in synthetic data research is whether realism is actually necessary; we will encounter this question several times in this survey) and/or devise techniques that help models transfer from synthetic training sets to real test sets; thus, domain adaptation becomes a major topic in synthetic data research and in this survey as well.

We begin with a few general remarks regarding synthetic data. First, note that synthetic data can be produced and supplied to machine learning models on the fly, during training, with software synthetic data generators, thus alleviating the need to ever store huge datasets; see, e.g., Mason et al.~\cite{Mason2019AnT} who discuss this ``on the fly'' generation in detail. Second, while synthetic data is a rising field we know of no satisfactory general overview of the field, and this was our primary motivation for writing this survey. We note surveys that attempt to cover applications of synthetic data~\cite{Chum2019} and a special issue of the \emph{International Journal of Computer Vision}~\cite{Gaidon2018}, but hope that the present work paints a more comprehensive picture.

Third, we distinguish between synthetic data and \emph{data augmentation}; the latter is a set of techniques intended to modify real data rather than create new synthetic data. These days, data augmentation is a crucial part of virtually every computer vision pipeline; we refer to the surveys~\cite{Shorten2019,DBLP:journals/corr/abs-1904-11685} and especially recommend the \emph{Albumentations} library~\cite{DBLP:journals/corr/abs-1809-06839} that has proven invaluable in our practice, but in this survey we concentrate on synthetic data rather than augmentation. Admittedly, the line between them is blurry, and some techniques discussed here could instead be classified as ``smart augmentation''.

Fourth, we note a natural application of synthetic data in machine learning: testing hypotheses and comparing methods and algorithms in a controlled synthetic setting. Toy examples and illustrative examples are usually synthetic, with a known data distribution so that machine learning models can be evaluated on how well they learn this distribution. This approach is widely used throughout the field, sometimes for entire meta-analyses~\cite{Bolon-Canedo:2013:RFS:3225633.3225767}, and we do not dwell on it here; our subject is synthetic data used to transfer to real data rather than direct comparisons between models on synthetic datasets.

% ...

In the survey, we cover three main directions for the use of synthetic data in machine learning; we discuss all three, and below we give references to specific parts of the survey related to these directions.

\begin{enumerate}
\item Using synthetically generated datasets to train machine learning models directly. This is the approach often taken in computer vision, and most of this survey is devoted to variations of this approach. In particular, one can:
\begin{itemize}
	\item train models on synthetic data with the intention to use them on real data; we discuss through most of Sections~\ref{sec:cv},~\ref{sec:direct}, and~\ref{sec:other};
	\item train (usually generative) models that change (refine) synthetic data in order to make it more suitable for training; Section~\ref{sec:domain} is devoted to this kind of models.
\end{itemize}
\item Using synthetic data to augment existing real datasets so that the resulting hybrid datasets are better for training the models. In this case, the synthetic data is usually employed to cover parts of the data distribution that are not sufficiently represented in the real dataset, with the main purpose being to alleviate dataset bias. The synthetic data can either
\begin{itemize}
	\item be generated separately with e.g., CGI-based methods for computer vision (see examples in Sections~\ref{sec:cv} and~\ref{sec:direct});
	\item or be generated from existing real data with the help of generative models (see Section~\ref{sec:synfromreal}).
\end{itemize}
\item Using synthetic data to resolve privacy or legal issues that make the use of real data impossible or prohibitively hard. This becomes especially relevant for certain specific fields of application, among which we discuss:
\begin{itemize}
	\item synthetic data in healthcare, which is not restricted to imaging but also extends to medical records and the like (Sections~\ref{sec:finance},~\ref{sec:medical});
	\item synthetic data in finance and social science, where direct applications are hard but privacy-related ones do begin to appear (see Section~\ref{sec:finance});
	\item synthetic data with privacy guarantees: many applications are sensitive enough to require a guarantee of privacy, for example from the standpoint of the differential privacy framework, and there has been an important line of work that makes synthetic data generation provide such guarantees (see Section~\ref{sec:privacy}).
\end{itemize}
\end{enumerate}

The survey is organized as follows. Section~\ref{sec:cv} presents synthetic datasets and results for basic computer vision problems, including low-level problems such as optical flow or stereo disparity estimation (Section~\ref{sec:datalow}), basic high-level problems such as object detection or segmentation (Section~\ref{sec:datahigh}), human-related synthetic data (Section~\ref{sec:datapeople}), character and text recognition (Section~\ref{sec:ocr}), and visual reasoning problems (Section~\ref{sec:datareason}). In Section~\ref{sec:direct}, we proceed to synthetic datasets that are more akin to full-scale simulated environments, covering outdoor and urban environments (Section~\ref{sec:dataoutdoor}), indoor scenes (Section~\ref{sec:dataindoor}), synthetic simulators for robotics (Section~\ref{sec:robotics}) and autonomous flying (Section~\ref{sec:cvnav}), and computer games as simulation environments (Section~\ref{sec:games}). 

Section~\ref{sec:other} is devoted to other domains of application for synthetic data, including neural programming (Section~\ref{sec:neuroprog}), bioinformatics (Section~\ref{sec:bio}), and natural language processing (Section~\ref{sec:nlp}). Section~\ref{sec:syn} discusses research intended to improve synthetic data generation: domain randomization (Section~\ref{sec:domrand}), development of methods for CGI-based generation (Section~\ref{sec:cgi}), synthetic data produced by ``cutting and pasting'' parts of real data samples (Section~\ref{sec:cutpaste}), and direct generation of synthetic data by generative models (Section~\ref{sec:datagan}).

Section~\ref{sec:domain} deals with the synthetic-to-real domain adaptation problem that we discussed above; there are many approaches here that can be broadly classified into synthetic-to-real refinement, where domain adaptation models are used to make synthetic data more realistic (Section~\ref{sec:refine}), and domain adaptation at the feature/model level, where the model and/or the training process is adapted rather than the data itself (Section~\ref{sec:damodel}); we also discuss case studies of domain adaptation for control and robotics (Section~\ref{sec:darobot}) and medical imaging (Section~\ref{sec:medical}).

Section~\ref{sec:privacy} is devoted to the privacy side of synthetic data: Section~\ref{sec:privdl} introduces differential privacy, Section~\ref{sec:privgan} shows how to generate synthetic data with differential privacy guarantees, and Section~\ref{sec:finance} presents a case study about private synthetic data in finance and related fields. 

In an attempt to look forward, we devote Section~\ref{sec:future} to directions for further work related to synthetic data that seem most promising: procedural generation of synthetic data (Section~\ref{sec:procedural}), closing the generation feedback loop (Section~\ref{sec:feedback}), introducing domain knowledge into domain adaptation (Section~\ref{sec:knowledge}), and improving domain adaptation models with additional modalities that are easy to obtain in synthetic datasets (Section~\ref{sec:futureadd}). Section~\ref{sec:concl} concludes the paper.

%% file: static.tex
\section{Synthetic data for basic computer vision problems}\label{sec:cv}

In this section, we present an overview of several directions for using synthetic data in computer vision, surveying both popular synthetic datasets that have been widely used in recent studies and the studies themselves. We organize this section by classifying datasets and models with respect to use cases, from generic object detection and segmentation problems to specific domains such as face recognition. All of these domains benefit highly from pixel-perfect labeling available by default in synthetic data, both in the form of classical computer vision labeling---bounding boxes for objects, segmentation masks---and labeling that would be very hard or impossible to do by hand: depth estimation, stereo image matching, 3D labeling in voxel space, and others.

In Section~\ref{sec:datalow}, we begin with low-level computer vision problems such as optical flow or stereo disparity estimation. Section~\ref{sec:datahigh} is devoted to basic high-level computer vision problems, including recognition of basic objects (Section~\ref{sec:objects}) and improving general problems such as object detection or segmentation with synthetic data (Section~\ref{sec:visiongeneral}). We also discuss several more specialized directions: human-related computer vision problems such as face recognition or crowd counting in Section~\ref{sec:datapeople}, character and text recognition in Section~\ref{sec:ocr}, and visual reasoning problems in Section~\ref{sec:datareason}. We also refer to Table~\ref{tbl:datacv} for a brief overview of the major datasets considered in this chapter.

\subsection{Low-level computer vision}\label{sec:datalow}

\begin{table}[!t]\centering\small
\setlength{\tabcolsep}{3pt}
\begin{tabular}{|rccl|}\hline
\bf Name & \bf Year & \bf Ref & \bf Size / comments \\
\hline\multicolumn{4}{|c|}{\it Low-level computer vision} \\\hline
Tsukuba Stereo & 2012 & \cite{6460313} & 1800 high-res stereo image pairs \\
MPI-Sintel & 2012 & \cite{10.1007/978-3-642-33783-3_44} & Optical flow from an animated movie \\
Middlebury 2014 & 2014 & \cite{conf/dagm/ScharsteinHKKNWW14} & 33 high-res stereo datasets \\
Flying Chairs & 2015 & \cite{7410673} & 22K frame pairs with ground truth flow \\
Flying Chairs 3D & 2015 & \cite{DBLP:journals/corr/MayerIHFCDB15} & 22K stereo frames \\
Monkaa & 2015 & \cite{DBLP:journals/corr/MayerIHFCDB15} & 8591 stereo frames \\
Driving & 2015 & \cite{DBLP:journals/corr/MayerIHFCDB15} & 4392 stereo frames \\
UnrealStereo & 2016 & \cite{Zhang2016UnrealStereoAS} & Data generation software \\
Underwater & 2018 & \cite{Olson2018SyntheticDG} & Underwater synthetic stereo pairs generator \\
\hline\multicolumn{4}{|c|}{\it Datasets of basic objects} \\\hline
YCB & 2015 & \cite{qiu2017unrealcv} & 77 objects in 5 categories \\
ShapeNet & 2015 & \cite{DBLP:journals/corr/ChangFGHHLSSSSX15} & $>$3M models, 3135 categories, rich annotations \\
ShapeNetCore & 2017 & \cite{DBLP:journals/corr/abs-1710-06104} & 51K manually verified models from 55 categories \\
UnrealCV & 2017 & \cite{qiu2017unrealcv} & Plugin for UE4 to generate synthetic data \\
VANDAL & 2017 & \cite{7989162} & 4.1M depth images, $>$9K objects in 319 categories \\
SceneNet & 2015 & \cite{7487797} & Automated indoor synthetic data generator \\
SceneNet RGB-D & 2017 & \cite{8237554} & 5M RGB-D images from 16K 3D trajectories \\
DepthSynth & 2017 & \cite{8374552} & Framework for realistic simulation of depth sensors \\
PartNet & 2018 & \cite{DBLP:journals/corr/abs-1812-02713} & 26671 models, 573535 annotated part instances \\
Falling Things & 2018 & \cite{8575443} & 61.5K images of YCB objects in virtual envs \\
ADORESet & 2019 & \cite{Bayraktar2019} & Hybrid dataset for object recognition testing \\
\hline\multicolumn{4}{|c|}{\it Datasets of synthetic people} \\\hline
ViHASi & 2008 & \cite{4635730} & Silhouette-based action recognition \\
Agoraset & 2014 & \cite{COURTY2014161} & Crowd scenes generator \\
LCrowdV & 2016 & \cite{10.1007/978-3-319-48881-3_50} & 1M videos, 20M frames with crowds \\
PHAV & 2017 & \cite{8099761} & 40K videos for action recognition (35 categories) \\
SURREAL & 2017 & \cite{DBLP:journals/corr/Varol0MMBLS17} & 145 subjects, 2.6K sequences, 6.5M frames \\
SyRI & 2018 & \cite{Bak2018DomainAT} & Virtual humans in UE4 with realistic lighting \\
GCC & 2019 & \cite{DBLP:journals/corr/abs-1903-03303} & 15K images with 7.6M subjects \\
\hline
\end{tabular}

\caption{An overview of synthetic datasets discussed in Section~\ref{sec:cv}.}\label{tbl:datacv}
\end{table}

Low-level computer vision problems include, in particular, \emph{optical flow estimation}, i.e., estimating the distribution of apparent velocities of movement along the image, \emph{stereo image matching}, i.e., finding the correspondence between the points of two images of the same scene from different viewpoints, \emph{background subtraction}, and so on. Algorithms for solving these problems can serve as the foundation for computer vision systems; for example, optical flow is important for motion estimation and video compression. Low-level problems can usually be approached with methods that do not require modern large-scale datasets or much learning at all, e.g., classical differential methods for optical flow estimation. However, at the same time, ground truth datasets are very hard to label manually, and hardware sensors that would provide direct measurements of optical flow or stereo image correspondence are difficult to construct (e.g., commodity optical flow sensors simply run the same estimation algorithms).

\begin{figure}[!t]\centering
\setlength{\tabcolsep}{3pt}
\begin{tabular}{P{.32\linewidth}P{.32\linewidth}P{.32\linewidth}}
\multicolumn{3}{c}{\myp{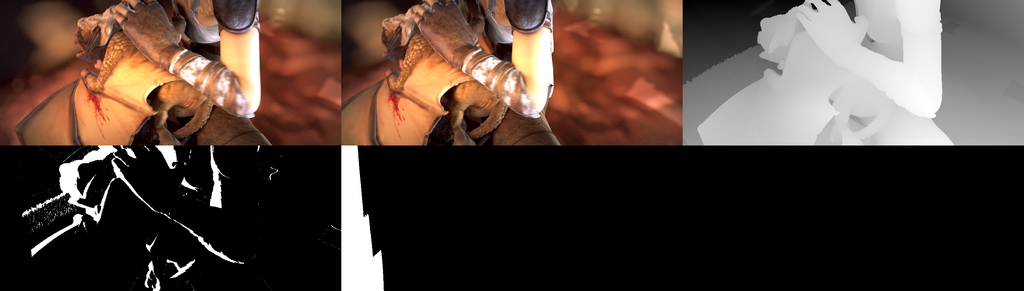}} \\
\multicolumn{3}{c}{(a)} \\
\myp{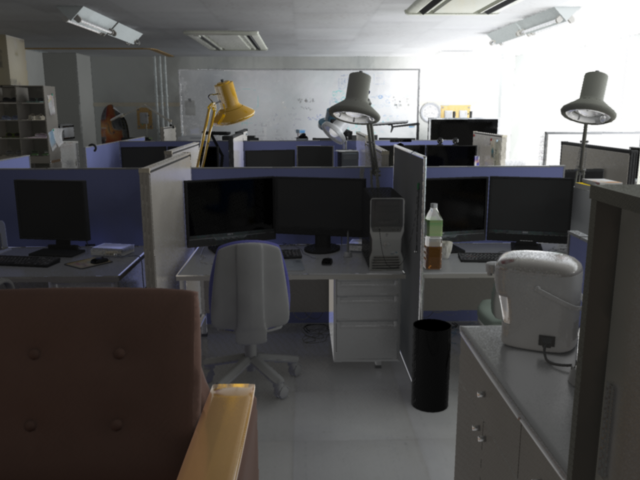} & \myp{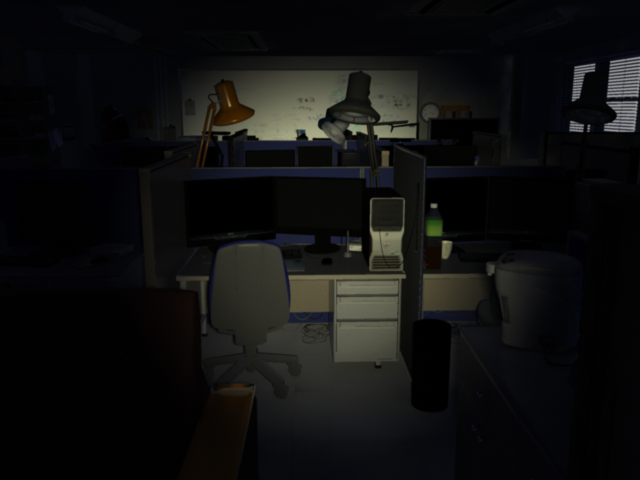} & \myp{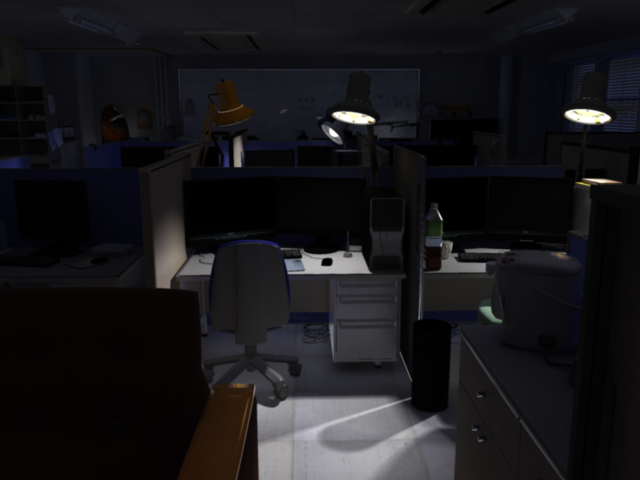} \\
(b) & (c) & (d) \\
\myp{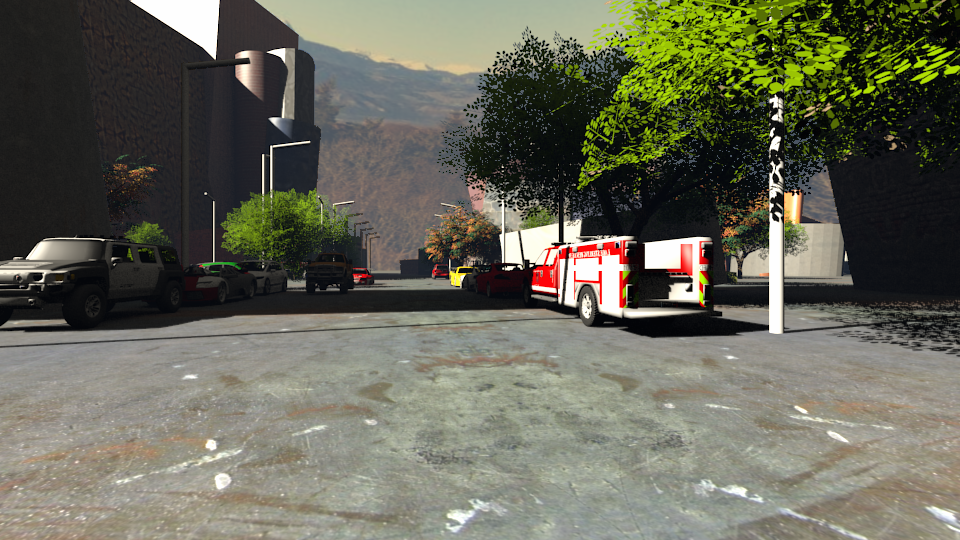} & \myp{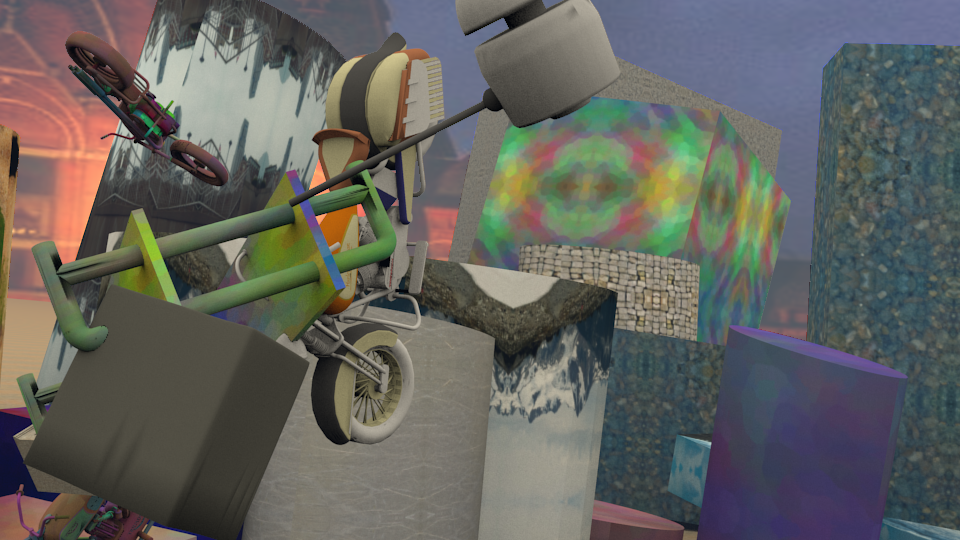} & \myp{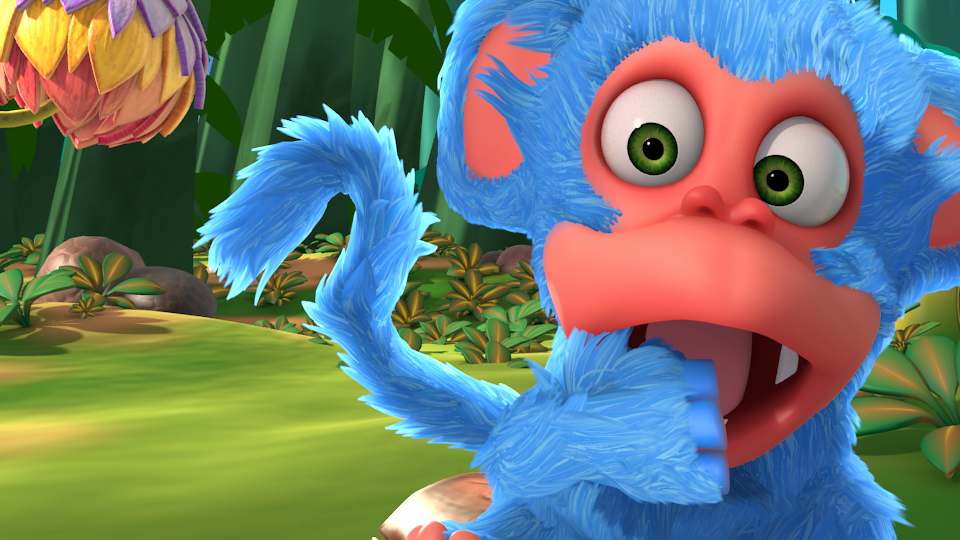} \\
\centering (e) & \centering (f) & \centering (g) \\
\end{tabular}

\caption{Sample images from synthetic low-level datasets: (a) \emph{MPI-Sintel}~\cite{10.1007/978-3-642-33783-3_44} (left to right: left view, right view, disparities; bottom row shows occluded and out-of-frame pixels); (b-d) \emph{Tsukuba CG Stereo Dataset}~\cite{6460313} with different illumination conditions: (b) daylight, (c) flashlight, (d) lamps;
(e) \emph{Driving}~\cite{DBLP:journals/corr/MayerIHFCDB15}; (f) \emph{FlyingThings3D}~\cite{DBLP:journals/corr/MayerIHFCDB15}; (g) \emph{Monkaa}~\cite{DBLP:journals/corr/MayerIHFCDB15}.}\label{pic:low}
\end{figure}

All of these reasons make low-level computer vision one of the oldest problems where synthetic data was successfully used, originally mostly for evaluation. Works as far back as late 1980s~\cite{47107} and early 1990s~\cite{Barron1994} presented and used synthetic datasets to evaluate different optical flow estimation algorithms. In 1999, Freeman et al.~\cite{790414} presented a synthetically generated world of images, with labeling derived from the corresponding 3D scenes, designed to train and evaluate low-level computer vision algorithms.

A modern dataset for low-level vision is \emph{Middlebury} presented by Baker et al.~\cite{Baker2011}; in addition to ground truth real life measurements taken with specially constructed computer-controlled lighting, they provide realistic synthetic imagery as part of the dataset and include it in the large-scale evaluation of optical flow and stereo correspondence estimation algorithms that they undertake. The \emph{Middlebury} dataset played an important role in the development of low-level computer vision algorithms~\cite{988771,Seitz:2006:CEM:1153170.1153518}, but its main emphasis was still on real imagery, as evidenced by its next version, \emph{Middlebury 2014}~\cite{conf/dagm/ScharsteinHKKNWW14}.

Peris et al.~\cite{6460313} present the \emph{Tsukuba CG Stereo Dataset} with synthetic data and ground truth disparity maps and show improvements in disparity classification quality. Butler et al.~\cite{10.1007/978-3-642-33783-3_44} presented a synthetic optical flow dataset \emph{MPI-Sintel} derived from the short animated movie Sintel\footnote{\url{http://www.sintel.org/}} produced as part of the Durian Open Movie Project. The main characteristic feature of \emph{MPI-Sintel} is that it contains the same scenes with different render settings, varying quality and complexity; this approach can provide a deeper understanding of where different optical flow algorithms break down. This is an interesting idea that has not yet found its way into synthetic data for deep learning-based computer vision but might be worthwhile to investigate. An interesting study by Meister and Kondermann~\cite{5936557} shows that while real and (high-quality, produced with ray tracing) synthetic data yield approximately the same results for optical flow detection in terms of mean endpoint error, the spatial distributions of errors are different, so synthetic data in this case may supplement real data in unexpected ways.

As the field moved from classical unsupervised approaches to deep learning, state of the art models began to require large datasets that could not be produced in real life, and after the transition to deep learning synthetic datasets started to dominate. Dosovitsky et al.~\cite{7410673} present a large synthetic dataset called \emph{Flying Chairs} from a public database of 3D chair models, adding them on top of real backgrounds to train a CNN-based optical flow estimation model. Mayer et al.~\cite{DBLP:journals/corr/MayerIHFCDB15} extended this work from optical flow to disparity and scene flow estimation, presenting three synthetic datasets produced in Blender (similar to Sintel):
\begin{itemize}
	\item \emph{FlyingThings3D} with everyday objects flying along randomized trajectories, 
	\item \emph{Monkaa} from its namesake animated short film with soft nonrigid motion and complex details such as fur, and 
	\item \emph{Driving} with naturalistic dynamic outdoor scenes from the viewpoint of a driving car (for more outdoor datasets see Section~\ref{sec:dataoutdoor}).
\end{itemize}
The \emph{Flying Chairs} dataset was also later extended with additional modalities to \emph{ChairsSDHom}~\cite{IMKDB17} with optical flow ground trugh and \emph{Flying Chairs 2}~\cite{ISKB18} with occlusion weights and motion boundaries.

The \emph{UnrealStereo} dataset by Zhang et al.~\cite{Zhang2016UnrealStereoAS} is a data generation framework for stereo scene analysis based on the \emph{Unreal Engine 4}, designed to evaluate the robustness of stereo vision algorithms to changes in material and other scene parameters. Many datasets that we describe below for high-level problems, such as SceneNet RGB-D~\cite{8237554} or SYNTHIA~\cite{7780721}, also contain labeling for optical flow and have been used to train the corresponding models.

Olson et al.~\cite{Olson2018SyntheticDG} consider an unusual special case for this problem: \emph{underwater} disparity estimation. Their work is also interesting in the way they produce synthetic data: Olson et al. project real underwater images on randomized synthetic surfaces produced in Blender, and then use rendering tools developed to mimic the underwater sensors and characteristic underwater effects such as fast light decay and backscattering. They produce synthetic stereo image pairs and use the dataset to train disparity estimation models, with successful transfer to real images.

In a recent work, Mayer et al.~\cite{Mayer:2018:MGS:3270334.3270407} provide an overview of different synthetic datasets for low-level computer vision and compare them from the standpoint of training optical flow models. They come to interesting conclusions:
\begin{itemize}
	\item first, for low-level vision synthetic data does not have to be realistic, \emph{Flying Chairs} works just fine; 
	\item second, it is best to combine different synthetic datasets and train in a variety of situations and domains; this ties into the domain randomization idea that we discuss in Section~\ref{sec:domrand};
	\item third, while realism itself is not needed, it does help to simulate the flaws of a specific real camera; Mayer et al. show that simulating, e.g., lens distortion and blur or Bayer interpolation artifacts in synthetic data improves the results on a real test set afterwards.
\end{itemize}
The question of realism remains open for synthetic data, and we will touch upon it many times in this survey. While it does seem plausible that for low-level problems such as optical flow estimation ``low-level realism'' (simulating camera idiosyncrasies) is much more important than high-level scene realism, the answer may be different for other problems.

\subsection{Basic high-level computer vision}\label{sec:datahigh}


Basic high-level computer vision problems, such as object detection or segmentation, fully enjoy the benefits of perfect labeling provided by synthetic data, and there is plenty of effort devoted to making synthetic data work for these problems. Since making synthetic data requires the development of 3D models, datasets usually also feature 3D-related labeling such as the depth map, labeled 3D parts of a shape, volumetric 3D data, and so on. There are many applications of these problems, including object detection for everyday objects and retail items (where a high number of classes and frequently appearing new classes make using real data impractical), counting and detection of small objects, basically all applications of semantic and instance segmentation (where manual labeling is especially hard to obtain), and more.

\subsubsection{Datasets of basic objects}\label{sec:objects}

\begin{figure}[!t]\centering
\setlength{\tabcolsep}{3pt}
\begin{tabular}{P{.32\linewidth}P{.32\linewidth}P{.32\linewidth}}
\myp{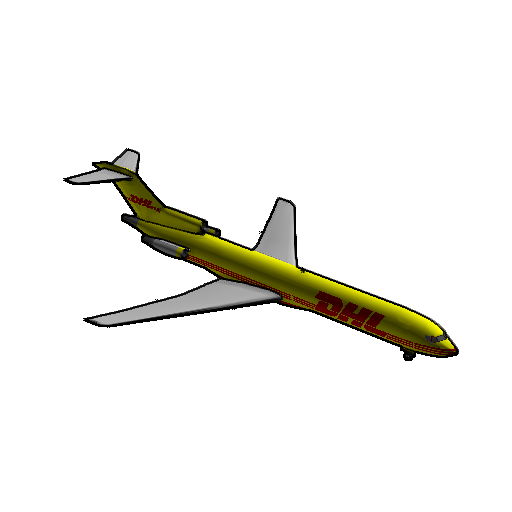} & \myp{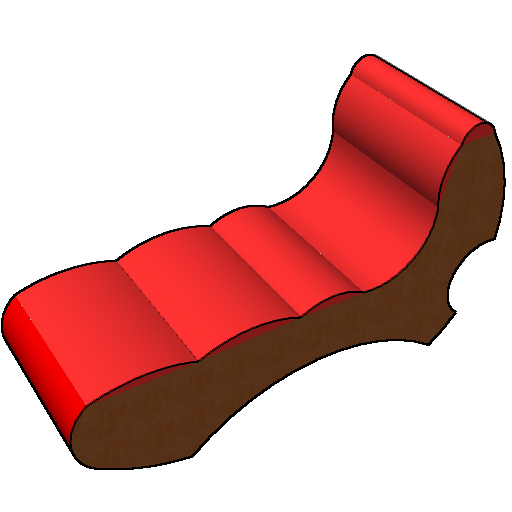} & \myp{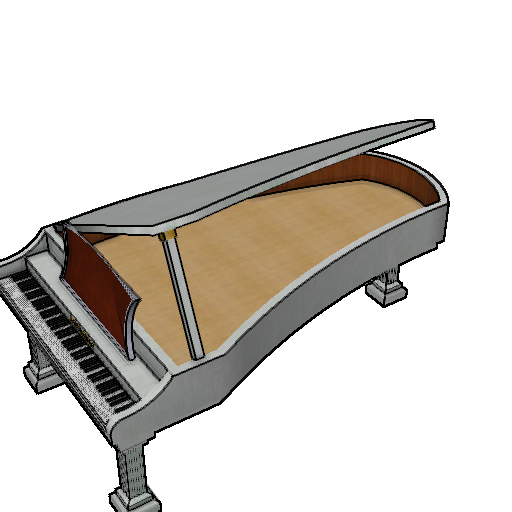} \\
(a) & (b) & (c) \\
\multicolumn{3}{c}{
 \begin{tabular}{P{.49\linewidth}P{.49\linewidth}}
	\myp{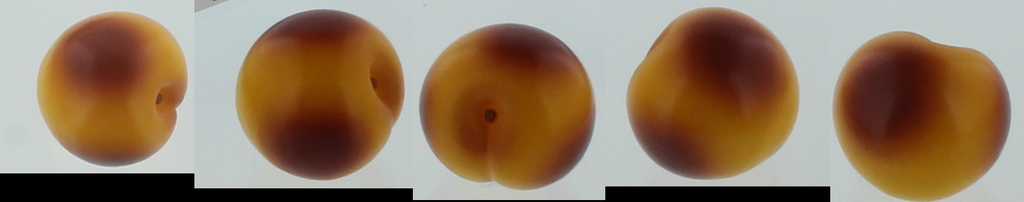} & \myp{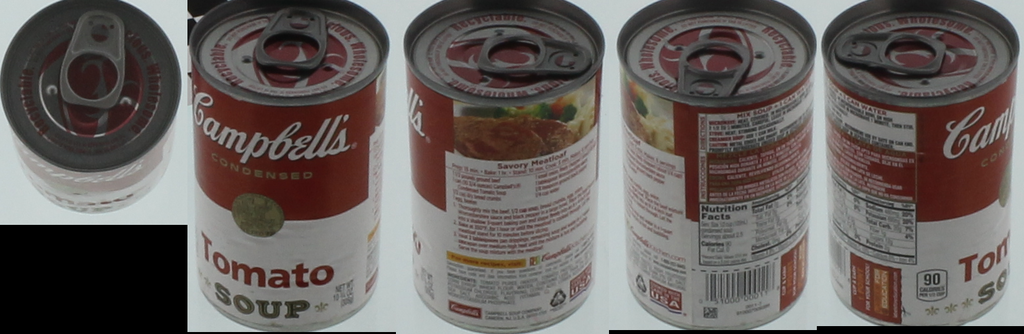} \\
	(d) & (e) \\
	\myp{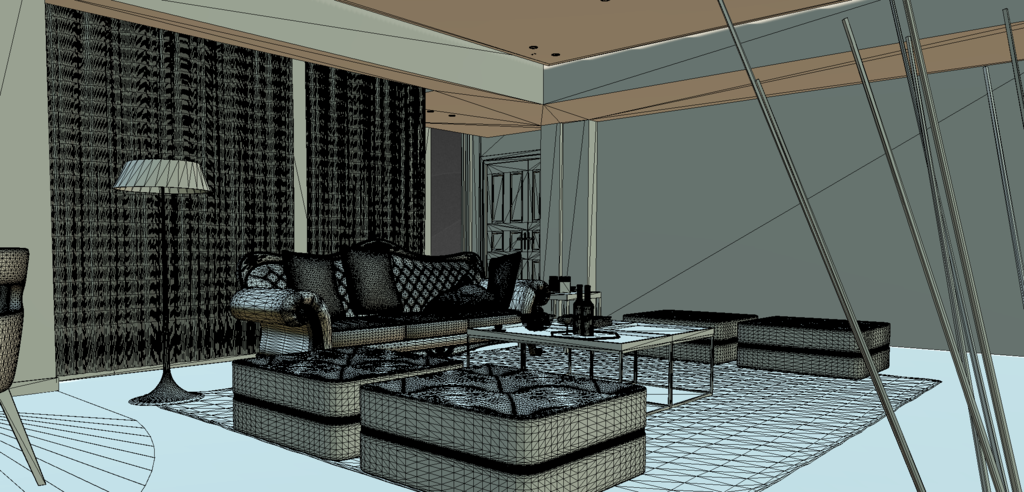} & \myp{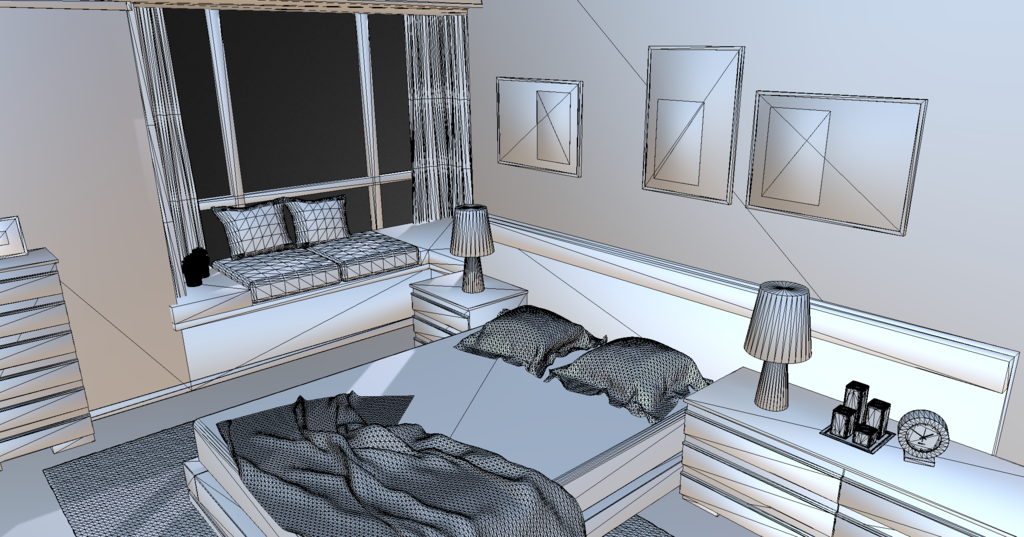} \\
	(f) & (g) \\
	\end{tabular} } \\
\myp{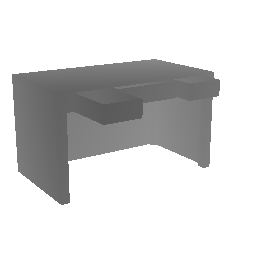} & \myp{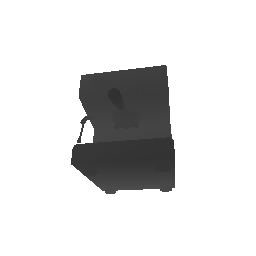} & \myp{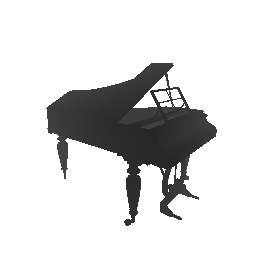} \\
(h) & (i) & (k) \\	
\end{tabular}

\caption{Sample images from synthetic datasets of basic objects: (a-c) shapes from \emph{ShapeNet}~\cite{DBLP:journals/corr/ChangFGHHLSSSSX15}: (a) airplane, (b) chair, (c) grand piano); (d-e) from YCB Object and Model set~\cite{7251504}: (d) peach, (e) tomato soup can; (f-g) 3D scenes from \emph{SceneNet}~\cite{7487797,DBLP:journals/corr/HandaPBSC15a}: (f) living room, (g) bedroom; (h-k) depth images from VANDAL~\cite{7989162}: (h) desk, (i) coffee maker, (k) grand piano.}\label{pic:basic}
\end{figure}

Many works apply synthetic data to recognizing everyday objects such as retail items, food, or furniture, and most of them draw upon the same database for 3D models. Developed by Chang et al.~\cite{DBLP:journals/corr/ChangFGHHLSSSSX15}, \emph{ShapeNet}\footnote{\url{https://www.shapenet.org/}} indexes more than three million models, with $220{,}000$ of them classified into $3{,}135$ categories that match WordNet synsets. Apart from class labels, \emph{ShapeNet} also includes geometric, functional, and physical annotations, including planes of symmetry, part hierarchies, weight and materials, and more. Researchers often use the clean and manually verified \emph{ShapeNetCore} subset that covers $55$ common object categories with about $51{,}000$ unique 3D models~\cite{DBLP:journals/corr/abs-1710-06104}. 

\emph{ShapeNet} has become the basis for further efforts devoted to improving labelings. In particular, region annotation (e.g., breaking an airplane into wings, body, and tail) is a manual process even in a synthetic dataset, while shape segmentation models increasingly rely on synthetic data~\cite{Yi:2017:LHS:3072959.3073652}; this also relates to the 3D mesh segmentation problem~\cite{doi:10.1111/j.1467-8659.2007.01103.x}. Based on \emph{ShapeNet}, Yi et al.~\cite{Yi:2016:SAF:2980179.2980238} developed a framework for scalable region annotation in 3D models based on active learning, and Chen et al.~\cite{Chen:2009:BMS:1531326.1531379} released a benchmark dataset for 3D mesh segmentation. A recent important effort related to \emph{ShapeNet} is the release of \emph{PartNet}~\cite{DBLP:journals/corr/abs-1812-02713}, a large-scale dataset of 3D objects annotated with fine-grained, instance-level, and hierarchical 3D part information; it contains $573{,}585$ part instances across $26{,}671$ 3D models from $24$ object categories. PartNet is mostly intended as a benchmark for 3D object and scene understanding, but the corresponding 3D models will no doubt be widely used to generate synthetic data.

One common approach to generating synthetic data is to reuse the work of 3D artists that went into creating the virtual environments of video games. For example, Richter et al.~\cite{DBLP:journals/corr/abs-1709-07322,DBLP:journals/corr/RichterVRK16} captured datasets from the \emph{Grand Theft Auto~V} video game (see also Section~\ref{sec:dataoutdoor}). They concentrated on semantic segmentation; note that getting pixel-wise labels for segmentation still required manual labor, but the authors claim that by capturing the communication between the game and the graphics hardware, they have been able to cut the labeling costs (in annotation time) by orders of magnitude. Once the annotator has worked through the first frame, the same combinations of meshes, textures, and shaders reused on subsequent frames can be automatically recognized and labeled, and the annotators are only asked to label new combinations. In essence, the game engine provides perfect superpixels that are persistent across frames.

As \emph{Grand Theft Auto V} and other games became popular for collecting synthetic datasets (see also Section~\ref{sec:dataoutdoor}), more specialized solutions began to appear. One such solution is \emph{UnrealCV} developed by Qiu et al.~\cite{DBLP:journals/corr/QiuY16,qiu2017unrealcv}, an open-source plugin for the popular game engine \emph{Unreal Engine~4} that provides commands that allow to get and set camera location and field of view, get the set of objects in a scene together with their positions, set lighting parameters, modify properties of the objects such as material, and capture from the engine the image and depth ground truth for the current camera and lighting parameters. This allows to create synthetic image datasets from realistic virtual worlds.

Robotics has motivated the appearance of synthetic datasets with objects that might be subject for manipulation, usually with fairly accurate models of their physical properties. The computer vision objectives in these datasets usually relate to robotic perception and include segmentation, depth estimation, object pose estimation, and object tracking. In particular, Choi et al.~\cite{6696485} present a dataset of 3D models of household objects for their tracking filter, while Hodan et al.~\cite{DBLP:journals/corr/HodanHOMLZ17} provide a real dataset of textureless objects supplemented with 3D models of these objects that provide the 6D ground truth poses. Lee et al.~\cite{10.1007/978-94-007-5699-1_15} test existing tracking methods with simulated video sequences with occlusion effects. Papon and Schoeler~\cite{7410452} consider the problem of object pose and depth estimation in indoor scenes. They have developed a synthetic data generator and trained on 7000 randomly generated scenes with $\approx$60K instances of 2842 pose-aligned models from the \emph{ModelNet10} dataset~\cite{Wu20143DSF}, showing excellent results in transfer to real test data.

\begin{figure}[!t]\centering
\setlength{\tabcolsep}{3pt}
\begin{tabular}{P{.49\linewidth}P{.49\linewidth}}
\myp{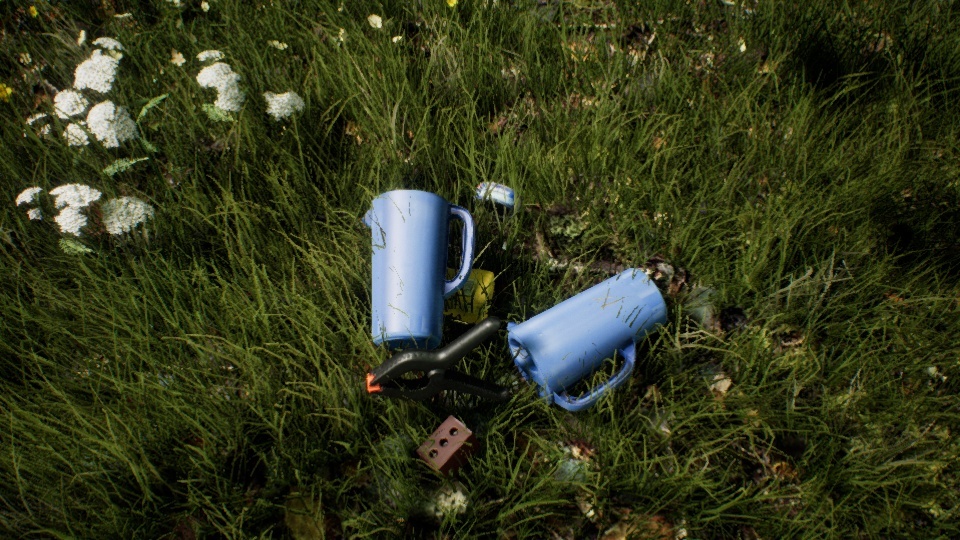} & \myp{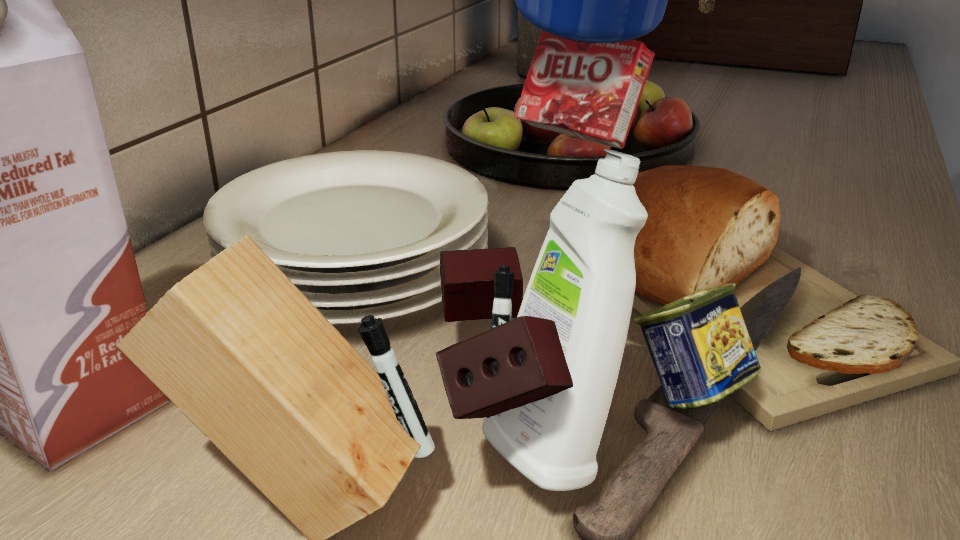} \\
(a) & (b) \\
\myp{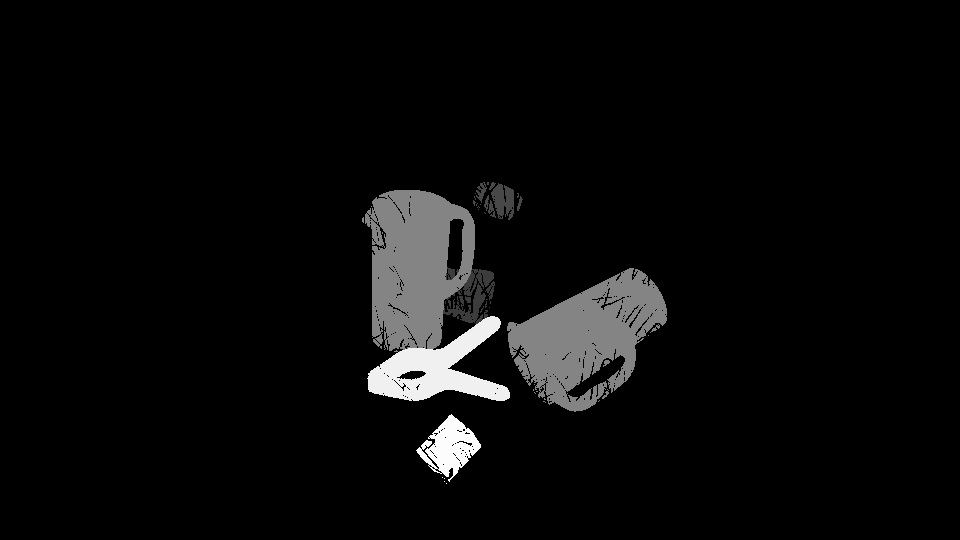} & \myp{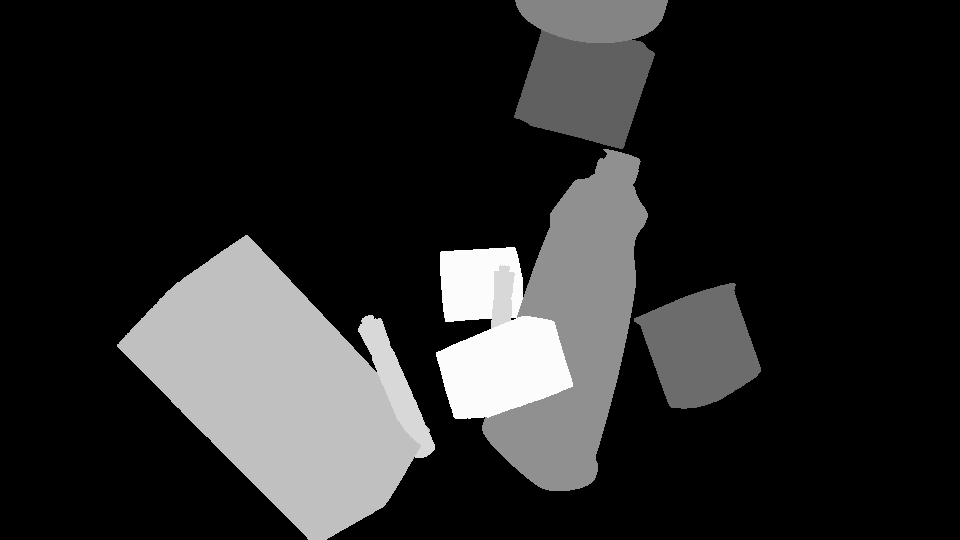} \\
(c) & (d) \\
\myp{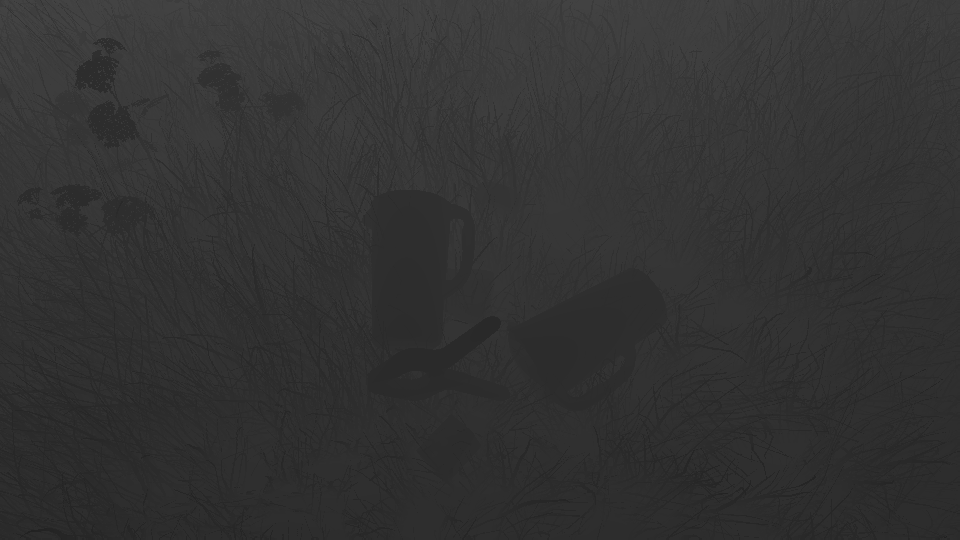} & \myp{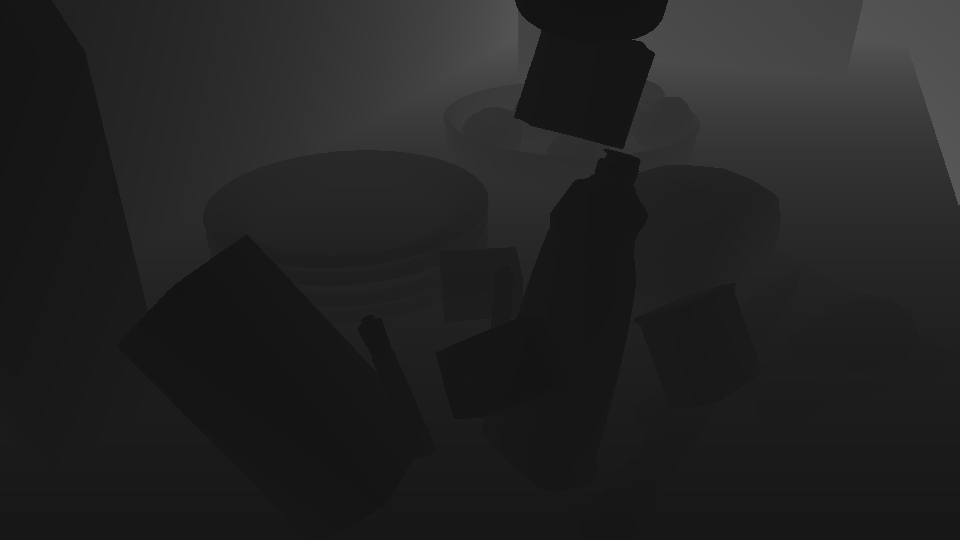} \\
(e) & (f) \\
\end{tabular}

\caption{Sample images from the \emph{Falling Things} dataset~\cite{8575443}: (a-b) RGB images, (c-d) ground truth segmentation maps; (e-f) depth maps.}\label{pic:fat}
\end{figure}

The \emph{Yale-CMU-Berkeley} (YCB) Object and Model set presented by Calli et al.~\cite{7251504} contains a set of 3D models of objects commonly used for robotic grasping together with a database of real RGB-D scans and physical properties of the objects, which makes it possible to use them in simulations. The \emph{Falling Things} (FAT) dataset by NVIDIA researchers Tremblay et al.~\cite{8575443} contains about 61500 images of 21 household objects taken from the YCB dataset and placed into virtual environments under a wide variety of lighting conditions, with 3D poses, pixel-perfect segmentation, depth images, and 2D/3D bounding box coordinates for each object; we show sample images from FAT on Figure~\ref{pic:fat}.

Recent works begin to use synthetic datasets of everyday objects in more complex ways, in particular by placing them in real surroundings. Abu Alhaija et al.~\cite{AbuAlhaija2018} and Georgakis et al.~\cite{Georgakis2017SynthesizingTD} propose procedures to augment real backgrounds with synthetic objects (see also Section~\ref{sec:cutpaste} where we discuss placing real objects on real backgrounds). In~\cite{AbuAlhaija2018}, the backgrounds come from the KITTI dataset of outdoor scenes and the objects are synthetic models of cars, while in~\cite{Georgakis2017SynthesizingTD} the authors place synthetic objects into indoor scenes with an eye towards home service robots. Synthetic objects have been used on real backgrounds many times before, but the main distinguishing feature of~\cite{AbuAlhaija2018} and~\cite{Georgakis2017SynthesizingTD} is that they are able to paste synthetic objects on real surfaces in a way consistent with the rest of the background scene. Abu Alhaija et al. developed a pipeline for automated analysis that recognized road surfaces on 360\textdegree{} panoramic images, but at the same time they conclude that the best (w.r.t. to the quality of the resulting segmentation model) way to insert the cars was to do it manually, and almost all their experiments used manual car placement. These experiments showed that state of the art models for instance segmentation and object detection yield better results on real validation tests when trained on scenes augmented with synthetic cars. Georgakis et al. use the algorithm from~\cite{Taylor2012ParsingIS} to extract supporting surfaces from an image and place synthetic objects on such surfaces with proper scale; they show significant improvements by training on hybrid real+synthetic datasets. One of the latest and currently most advanced pipelines in this direction for autonomous driving is AADS~\cite{Li2019AADSAA} that we discuss in Section~\ref{sec:dataoutdoor}.

In general, by now researchers have relatively easy access to large datasets of 3D models of everyday objects to generate synthetic environments (we will see more of this in Sections~\ref{sec:dataoutdoor} and~\ref{sec:dataindoor}), add synthetic objects as distractors to real images, place synthetic objects on real backgrounds in smarter ways, and so on. Although RGBD datasets with real scans are also increasingly available as the corresponding hardware becomes available (see, e.g., the survey~\cite{7789578} and~\cite{Choi2016ALD}), they cannot compete with synthetic data in terms of the quality of labeling and diversity of environments (see also Section~\ref{sec:domrand}). In the next section, we will see how this progress helps to solve the basic computer problems: after all, recognizing \emph{synthetic} objects is never the end goal.


\subsubsection{Improving high-level computer vision with synthetic data}\label{sec:visiongeneral}

A long line of work has used synthetic data for object detection. Peng et al.~\cite{Peng:2015:LDO:2919332.2920015} trained an object detection framework on non-realistic images of 3D models superimposed on real backgrounds, noting, in particular, that the results improve when synthetic data is varied along 3D pose, texture and color. Bochinski et al.~\cite{7738056} were one of the first to train object detection CNNs on purely synthetic datasets and show that the results transfer to real world evaluation data. Rajpura et al.~\cite{DBLP:journals/corr/RajpuraHB17} developed a Blender-based synthetic scene generator for recognizing objects inside a refrigerator, showing improved results with a fully convolutional version of GoogLeNet~\cite{7298594} adapted for object detection. Bayraktar et al.~\cite{Bayraktar2018AHI} show improvements in object detection on a hybrid dataset in the context of robotics, extending a real dataset with images generated by the \emph{Gazebo} simulation environment (see Section~\ref{sec:robotics}). In a recent work, Bayraktar et al.~\cite{Bayraktar2019} test modern object recognition architectures such as \emph{VGGNet}~\cite{Krizhevsky:2012:ICD:2999134.2999257}, \emph{Inception v3}~\cite{7780677}, \emph{ResNet}~\cite{DBLP:journals/corr/HeZRS15}, and \emph{Xception}~\cite{DBLP:journals/corr/Chollet16a} by fine-tuning them on the \emph{ADORESet} dataset that contains 2500 real and 750 synthetic images for each of 30 object categories in the context of robotic manipulation; they find that a hybrid dataset achieves much better recognition quality compared to purely synthetic or purely real datasets. Recent applications of synthetic data for object detection include the detection of objects in vending machines~\cite{DBLP:journals/corr/abs-1904-12294}, objects in piles for training robotic arms~\cite{Buls2019GenerationOS}, computer game objects~\cite{Struckmeier2019LeagueAIIO}, smoke detection~\cite{Xu2019AdversarialAF}, deformable part models~\cite{7495710}, face detection in biomedical literature~\cite{Dawson:2017:MFB:3107411.3107476}, drone detection~\cite{8310033}, and more. 

Recently, Nowruzi et al.~\cite{Nowruzi2019HowMR} studied the question of how much real data is actually needed for object detection in comparison to synthetic data. Using SSD-MobileNet~\cite{DBLP:journals/corr/HowardZCKWWAA17}, they have compared different modes of training for a number of synthetic and real datasets. They conclude that fine-tuning models trained on synthetic datasets with a small amount of real data is preferable to mixed training on a hybrid dataset with the same amount of real data, and that photorealism appears to be less important than the diversity of synthetic data; this runs contrary to the conclusions of the works~\cite{DBLP:journals/corr/abs-1710-06270,DBLP:journals/corr/abs-1810-08705} that we discuss below.

An interesting method of using synthetic data for object detection was proposed by Hinterstoisser et al.~\cite{10.1007/978-3-030-11009-3_42}. They note that training on purely synthetic data may give sub-par results due to the low-level differences between synthetic (rendered) images and real photographs. To avoid this, they propose to simply freeze the lower layers of, say, a pretrained object detection architecture and only train the top layers on synthetic data; in this way, basic features will remain suited for the domain of real photos while the classification part (top layers) can be fine-tuned for new classes. Otherwise, this is a straightforward test of synthetic data: Hinterstoisser et al. superimpose synthetic renderings on randomly selected backgrounds and fine-tune pretrained Faster-RCNN~\cite{7485869}, R-FCN~\cite{DBLP:journals/corr/DaiLHS16}, and Mask R-CNN~\cite{8237584} object detection architectures with freezed feature extraction layers. They report that freezing the layers helps significantly, and different steps in the synthetic data generation pipeline (different domain randomization steps, see also Section~\ref{sec:domrand}) help as well, obtaining results close to training on a large real dataset.

Segmentation is another classical computer vision problem with obvious benefits to be had from pixel-perfect synthetic annotations. The above-mentioned \emph{SceneNet RGB-D} dataset by McCormac et al.~\cite{8237554} comes with a study showing that an RGB-only CNN for semantic segmentation pretrained from scratch on purely synthetic data can improve over CNNs pretrained on ImageNet; as far as we know, this was the first time synthetic data managed to achieve such an improvement. The dataset is actually an extension of \emph{SceneNet}~\cite{7487797,DBLP:journals/corr/HandaPBSC15a}, an annotated model generator for indoor scene understanding that can use existing datasets of 3D object models and place them in 3D environments with synthetic annotation. By now, segmentation models are commonly trained with synthetic data: semantic segmentation is the main problem for most automotive driving models (Section~\ref{sec:dataoutdoor}) and indoor navigation models (Section~\ref{sec:dataindoor}), Grard et al.~\cite{10.1007/978-3-319-89327-3_16} do it for object segmentation in depth maps of piles of bulk objects, and so on.

Saleh et al.~\cite{DBLP:journals/corr/abs-1807-06132} note that not all classes in a semantic segmentation problem are equally suited for synthetic data. Foreground classes that correspond to objects (people, cars, bikes etc., i.e., \emph{things} in the terminology of~\cite{10.1007/978-3-540-88682-2_4}) are well suited for object detectors (that use shape a lot) but suffer from the synthetic-to-real transfer for segmentation networks because their textures (which segmentation models usually rely upon) are hard to make photorealistic. On the other hand, background classes (grass, road surface, sky etc., i.e., \emph{stuff} in the terminology of~\cite{10.1007/978-3-540-88682-2_4}) look very realistic on synthetic images due to their high degree of ``texture realism'', and a semantic segmentation network can be successfully trained on synthetic data for background classes. Therefore, Saleh et al. propose a pipeline that combines detection-based masks by Mask R-CNN~\cite{8237584} for foreground classes and semantic segmentation masks by \emph{DeepLab}~\cite{7913730} for background classes.

Although most works on synthetic data use large and well-known synthetic datasets, there are many efforts to bring synthetic data to novel applications by developing synthetic datasets from scratch. For instance, O'Byrne et al.~\cite{jmse6030093} develop a synthetic dataset for biofouling detection on marine structures, i.e., segmenting various types of marine growth on underwater images. Ward et al.~\cite{Ward2018DeepLS} improve leaf segmentation for \emph{Arabidopsis} plants for the CVPPP Leaf Segmentation Challenge by augmenting real data with a synthetic dataset produced with Blender. For a parallel challenge of leaf counting, Ubbens et al.~\cite{Ubbens2018} produce synthetic data based on an L-system plant model; they report improved counting results. Moiseev et al.~\cite{10.1007/978-3-319-02895-8_52} propose a method to generate synthetic street signs, showing improvements in their recognition. Neff et al.~\cite{Neff18} use GANs to produce synthetically augmented data for small segmentation datasets (see Section~\ref{sec:medical}). We also note that in other problems, such as video stream summarization, researchers are also beginning to use synthetic data~\cite{Al-Musawi2015}.

Another important class of applications for CGI-based synthetic data relates to problems such as 3D pose, viewpoint, and depth estimation, where manual labeling of real data is very difficult and sometimes close to impossible. One of the basic problems here is 2D-3D alignment, the problem of finding correspondences between regions in a 2D image and a 3D model (this also implies pose estimation for objects). In an early work, Aubry et al.~\cite{Aubry14} solved the 2D-3D alignment problem for chairs with a dataset of synthetic CAD models. Gupta et al.~\cite{7299105} train a CNN to detect and segment object instances for 3D model alignment with synthetic data with renderings of synthetic objects. Su et al.~\cite{7410471} learn to recognize 3D shapes from several 2D images, training their multi-view CNNs on synthetic 2D views. Triyonoputro et al.~\cite{Triyonoputro2019QuicklyIP} train a deep neural network on multi-view synthetic images to help visual servoing for an industrial robot. Liu et al.~\cite{Liu:2017:ISM:3140137.3140204} perform indoor scene modeling from a single RGB image by training on a dataset of 3D models, and in other works~\cite{LIU2018108} do 2D-3D alignment from a single image for indoor basic objects. Shoman et al.~\cite{8699232} use synthetic data for camera localization (a crucial part of tracking and augmented reality systems), using synthetic data to cover a wide variety of lighting and weather conditions. They use an autoencoder-like architecture to bring together the features extracted from real and synthetic data and report significantly improved results.

3D position and orientation estimation for objects, known as the 6-DoF (degrees of freedom) pose estimation, is another important computer vision problem related to robotic grasping and manipulation. NVIDIA researchers Tremblay et al.~\cite{Tremblay2018DeepOP} approach it with synthetic data: using the synthetic data generation techniques we described in Section~\ref{sec:datahigh}, they train a deep neural network and report the first state of the art network for 6-DoF pose estimation trained purely on synthetic data. The novelty was that Tremblay et al. train on a mixture of domain randomized images, where distractor objects are placed randomly in front of a random background, and photorealistic images, where the foreground objects are placed in 3D background scenes obeying physical constraints; domain randomized images provide the diversity needed to cover real data (see Section~\ref{sec:domrand}) while realistic images provide proper context for the objects and are easier to transfer to real data. Latest results~\cite{DBLP:journals/corr/abs-1712-03904,8490961,Mora19} show that synthetic data, especially with proper domain randomization for the data and domain adaptation for the features, can indeed successfully transfer 3D pose estimation from synthetic to real objects.

This also relates to depth estimation; synthetic renderings are easy to augment with pixel-perfect depth maps, and many synthetic datasets include RGB-D data. Carlucci et al.~\cite{7989162} created VANDAL, one of the first synthetic depth image databases, collecting 3D models from public CAD repositories for about 480 \emph{ImageNet} categories of common objects; the authors showed that features extracted from these depth images by common CNN architectures improve object classification and are complementary to features extracted by the same architectures trained on \emph{ImageNet}. Liebelt et al.~\cite{4587614} used 3D models to extract a set of 3D feature maps, then used a nearest neighbors approach to do multi-view object class detection and 3D pose estimation. Lee and Moloney~\cite{Lee:2017:ESD:3150978.3150982} present a synthetic dataset with high quality stereo pairs and show that deep neural networks for stereo vision can perform competitively with networks trained on real data. \emph{Siemens} researchers Planche et al.~\cite{8374552} consider the problem of more realistic simulation of depth data from real sensors and present \emph{DepthSynth}, an end-to-end framework able to generate realistic depth data rather than purely synthetic perfect depth maps; they show that this added realism leads to improvements with modern 2.5D recognition methods.

Easy variations and transformations provided by synthetic data can not only directly improve the results by training, but also represent a valuable tool for studying the properties of neural networks and other feature extractors. In particular, Pinto et al.~\cite{5711540} used synthetic data to study the invariance of different existing visual feature sets to variation in position, scale, pose, and illumination, while Kaneva et al.~\cite{6126508} used a photorealistic virtual environment to evaluate image feature descriptors. Peng et al.~\cite{Peng2014ExploringII}, Pepik et al.~\cite{DBLP:journals/corr/PepikBRS15}, and Aubry and Russell~\cite{Aubry:2015:UDF:2919332.2920060} used synthetic data to study the properties of deep convolutional networks, in particular robustness to various transformations, since synthetic data is easy to manipulate in a predefined way.

Earlier works recognized that the domain gap between synthetic and real images does not allow to expect state of the art results when training on synthetic data only, so many of them concentrated on bridging this gap by constructing hybrid datasets. In particular, V{\'a}zquez et al.~\cite{6587038} considered pedestrian detection and proposed a scheme based on active learning: they initially train a detector on virtual data and then use selective sampling~\cite{Cohn1994} to choose a small subset of real images for manual labeling, achieving results on par with training on a large real datasets while using 10x less real data. 

Purely synthetic approaches were also used in early works, although mostly for problems where manual labeling would be even harder and noisier than for object detection or segmentation. The \emph{Render for CNN} approach by Su et al.~\cite{7410665} outperformed real data with a hybrid synthetic+real dataset on the viewpoint estimation problem. Synthetic data helped improve 3D object pose estimation in Gupta et al.~\cite{DBLP:journals/corr/GuptaAGM15} and multi-view object class detection in Liebelt and Schmid~\cite{5539836} and Stark et al.~\cite{Stark2010BackTT}; as an intermediate step, the latter work used synthetic data to learn shape models. Hattori et al.~\cite{7299006} trained scene-specific pedestrian detectors on a purely synthetic dataset, superimposing rendered pedestrians onto a fixed real scene background; synthetic data has also been used for pedestrian detection by Marin et al.~\cite{5540218}.

We finish this section by returning to an important question for direct applications: how realistic must synthetic data be in order to help with the underlying computer vision problem? Early works often argued that photorealism is not necessary for good domain transfer results; see, e.g.,~\cite{BMVC.28.82}. This question was studied in detail by Movshovitz-Attias et al.~\cite{10.1007/978-3-319-49409-8_18}. With the example of the viewpoint estimation problem for cars, they showed that photorealistic rendering does indeed help, showed that the gap between models trained on synthetic and real data can often be explained by domain adaptation (i.e., adapting from a different real dataset would be just as hard as adapting from a synthetic one), and hybrid synthetic+real datasets can significantly outperform training on real data only. 

Another data point is provided by Tsirikoglou et al.~\cite{DBLP:journals/corr/abs-1710-06270} who present a very realistic effort for the rendering of synthetic data, including Monte Carlo-based light transport simulation and simulation of optics and sensors, within the domain of rendering outdoor scenes (see also Section~\ref{sec:dataoutdoor}, where we discuss a continuation of this work by Wrenninge and Unger~\cite{DBLP:journals/corr/abs-1810-08705}). They show improved results in object detection over other synthetic datasets and conclude that ``a focus on maximizing variation and realism is well worth the effort''.

\subsection{Synthetic people}\label{sec:datapeople}

\begin{figure}[!t]\centering
\setlength{\tabcolsep}{3pt}
\begin{tabular}{P{.32\linewidth}P{.32\linewidth}P{.32\linewidth}}
\multicolumn{3}{c}{\myp{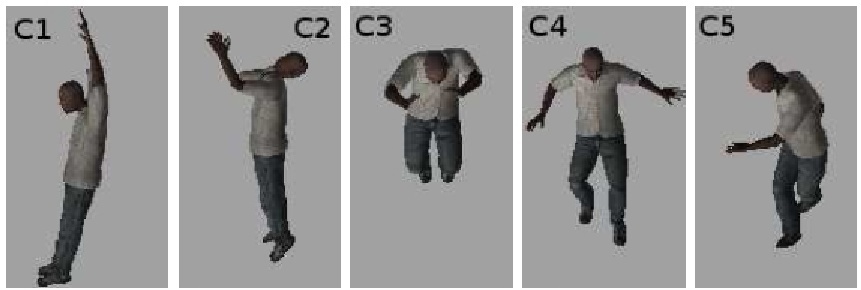}} \\
\multicolumn{3}{c}{(a)} \\
\multicolumn{3}{c}{\myp{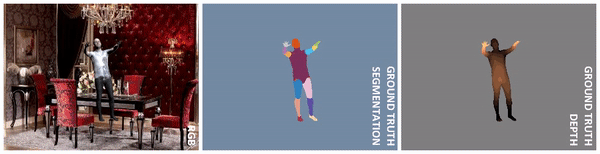}} \\
(b) & (c) & (d) \\
\myp{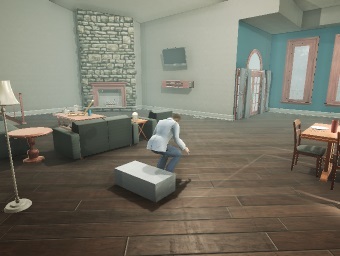} & \myp{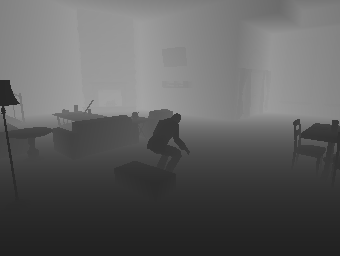} & \myp{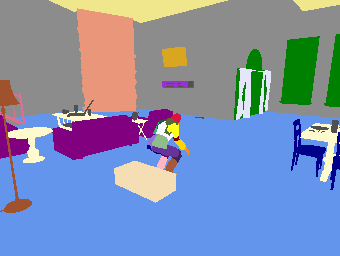} \\
(e) & (f) & (g) \\
\myp{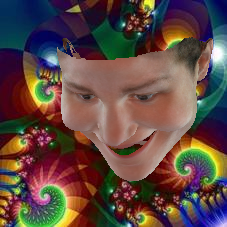} & \myp{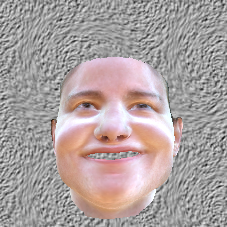} & \myp{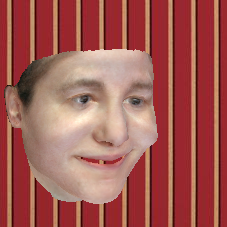} \\
(h) & (i) & (j) \\
\end{tabular}

\caption{Sample images from human-related synthetic datasets: (a) video frames from \emph{ViHASi}~\cite{4635730}; (b-d) a frame with ground truth from SURREAL~\cite{DBLP:journals/corr/Varol0MMBLS17}: (b) RGB image, (c) segmentation map, (d) depth map; (e-g) a frame from PHAV~\cite{8099761}: (e) RGB image, (f) segmentation map, (g) depth map; (h-j) sample synthetic faces with randomized backgrounds from~\cite{DBLP:journals/corr/abs-1811-08565,Kortylewski_2019_CVPR_Workshops} (based on the Basel Face Model).}\label{pic:human}
\end{figure}

Synthetic models and images of people (both faces and full bodies) are an especially interesting subject for synthetic data. On one hand, real datasets here are even harder to collect due to several reasons:
\begin{itemize}
	\item there are privacy issues involved in the collection of real human faces;
	\item the labeling for some basic computer vision problems is especially complex: while pose estimation is doable, facial keypoint detection (a key element for facial recognition and image manipulation for faces) may require to specify several dozen landmarks on a human face, which becomes very hard for human labeling~\cite{6909616,DBLP:journals/corr/abs-1905-00641};
	\item finally, even in the presence of a large dataset it often contains biases in its composition of genders, races, or other parameters, sometimes famously so~\cite{DBLP:journals/corr/abs-1712-01619,Tommasi2017}.
\end{itemize} 
On the other hand, there are complications as well:
\begin{itemize}
	\item synthetic 3D models people and especially synthetic faces are much harder to create than models of basic objects, especially if sufficient fidelity is required to ;
	\item basic human-related tasks are very important in practice, so there already exist large real datasets for face recognition~\cite{Cao18,DBLP:journals/corr/GuoZHHG16,Li:2011:HFR:2073486}, pose estimation~\cite{andriluka14cvpr,6682899,Trumble17,Liu:2015:SHP:2829388.2829749}, and other problems, which often limits synthetic data to covering corner cases, augmenting real datasets, or serving more exotic use cases.
\end{itemize}
This creates a tension between the quality of available synthetic faces and improvements in face recognition and other related tasks that they can provide. In this section, we review how synthetic people have been used to improve computer vision models in this domain.

In an early effort, Queiroz et al.~\cite{5720343} presented a pipeline for generating synthetic videos with automatic ground truth for human faces and the resulting \emph{Virtual Human Faces Database} (VHuF) dataset with realistic face skin textures that can be extracted from real photos. Bak et al.~\cite{Bak2018DomainAT} present the \emph{Synthetic Data for person Re-Identification} (SyRI) dataset with virtual 3D humans designed with \emph{Adobe Fuse CC} to make the models and \emph{Unreal Engine 4} for high-speed rendering. Interestingly, they model realistic lighting conditions by using real HDR environment maps collected with light probes and panoramic photos.

While face recognition for full-face frontal high-quality photos has been mostly solved in 2D, achieving human and superhuman performance both for classification~\cite{6909616} and retrieval via embeddings~\cite{Schroff2015FaceNetAU}, pose-invariant face recognition \emph{in the wild}~\cite{Huang2008LabeledFI,6778340,Ding:2016:CSP:2885506.2845089}, i.e., under arbitrary angles and imperfect conditions, remains challenging. Here synthetic data is often used to augment a real dataset, where frontal photos usually prevail, with more diverse data points; we refer to Section~\ref{sec:synfromreal} for a detailed overview of the works by Huang et al.~\cite{8237529} and Zhao et al.~\cite{Zhao2018,ijcai2018-165} on GAN-based refinement.

An interesting approach to creating synthetic data for face recognition is provided by Hu et al.~\cite{8049355}. In their ``Frankenstein'' pipeline, they combine automatically detected body parts (eyes, mouth, nose etc.) from different subjects; interestingly, they report that the inevitable artifacts in the resulting images, both boundary effects and variations between facial patches, do not hinder training on synthetic data and may even improve the robustness of the resulting model.

There is also a related field of \emph{3D-aided face recognition}. This approach uses a morphable synthetic 3D model of an abstract human face that has a number of free parameters; the model learns to tune these parameters so that the 3D model fits a given photo and then uses the model and texture from the photo either to produce a frontal image or to directly recognize photos taken from other angles. This is a classic approach, dating back to late 1990s~\cite{Blanz:1999:MMS:311535.311556,1227983} and developed in many subsequent works with new morphable models~\cite{Hu2016,7350989}, deep learning used to perform the regression for morphing parameters~\cite{Tran2017RegressingRA}, extended to 3D face scans~\cite{4409029}, and so on; see, e.g., the survey~\cite{Ding:2016:CSP:2885506.2845089} for more details. Xu et al.~\cite{DBLP:journals/corr/abs-1709-06532} use synthetic data to train their 3D-aided model for pose-invariant face recognition as well. Recent works used GANs to produce synthetic data for 3D-aided face recognition~\cite{Zhao2018}; we discuss this approach in detail in Section~\ref{sec:synfromreal}.

In a large-scale effort to combat dataset bias in face recognition and related problems with synthetic data, Kortylewski et al.~\cite{DBLP:journals/corr/abs-1811-08565,Kortylewski_2019_CVPR_Workshops} have developed a pipeline to directly create synthetic faces. They use the \emph{Basel Face Model 2017}~\cite{8373814}, a 3D morphable model of face shape~\cite{Blanz:1999:MMS:311535.311556,4409029}, and take special care to randomize the pose, camera location, illumination conditions, and background. They report significantly improved results for face recognition and facial landmark detection with the \emph{OpenFace} framework~\cite{Amos2016OpenFaceAG,SANTOSO2018510} and state of the art models for face detection and alignment~\cite{7553523} and landmark detection~\cite{8170321}.

Human pose estimation is a very well known and widely studied problem~\cite{Liu:2015:SHP:2829388.2829749,doi:10.1080/10798587.2015.1095419,8727761} with many direct applications, so it is no wonder that the field does not suffer from lack of real data, with large-scale datasets available~\cite{DBLP:journals/corr/LinMBHPRDZ14,6909866,BMVC.24.12} and state of the art models achieving impressive results~\cite{DBLP:journals/corr/abs-1708-01101,DBLP:journals/corr/abs-1902-07837,DBLP:journals/corr/abs-1902-09212}. However, synthetic data still can help. Ludl et al.~\cite{8569489} show that in corner cases, corresponding to rare activities not covered by available datasets, existing pose estimation models produce errors, but augmenting the training set with synthetic data that covers these corner cases helps improve pose estimation. Another specialized use-case has been considered by Rematas et al. in a very interesting application of pose estimation called ``Soccer on Your Tabletop''~\cite{8578596}. They trained specialized pose and depth estimation models for soccer players and produced a unified model that maps 2D footage of a soccer match into a 3D model suitable for rendering on a real tabletop through augmented reality devices. For training, Rematas et al. used synthetic data captured from the \emph{FIFA} video game series. These are model examples of how synthetic data can improve the results even when comprehensive real datasets are available.

Moving from still images to videos, we begin with human action recognition~\cite{Poppe:2010:SVH:1751678.1751964,ZhangHumanAR19}. ViHASi by Ragheb et al.~\cite{4635730} is a virtual environment and dataset for silhouette-based human action recognition. De Souza et al.~\cite{8099761} present PHAV (Procedural Human Action Videos), a synthetic dataset that contains 39{,}982 videos with more than 1{,}000 examples for each action of 35 categories. Inria and MPI researchers Varol et al.~\cite{DBLP:journals/corr/Varol0MMBLS17} present the SURREAL (Synthetic hUmans
foR REAL tasks) dataset. They generate photorealistic synthetic images with labeling for human part segmentation and depth estimation, producing 6.5M frames in 67.5K short clips (about 100 frames each) of 2.6K action sequences with 145 different synthetic subjects. 

\emph{Microsoft} researchers Khodabandeh et al.~\cite{8575353} present the \emph{DIY Human Action} generator for human actions. Their framework consists of a generative model, called the \emph{Skeleton Trajectory GAN}, that learns to generate a sequence of frames with human skeletons conditioned on the label for the desired action, and a \emph{Frame GAN} that generates photorealistic frames conditioned on a skeleton and a reference image of the person. As a result, they can generate realistic videos of people defined with a reference image that perform the necessary actions, and, moreover, the \emph{Frame GAN} is trained on an unlabeled set of human action videos.

We also note here some privacy-related applications of synthetic data that are not about differential privacy (which we discuss in Section~\ref{sec:privacy}). For example, Ren et al.~\cite{10.1007/978-3-030-01246-5_38} present an adversarial architecture for video face anonymization; their model learns to modify the original real video to remove private information while at the same time still maximizing the performance of action recognition models.

As the problems become dynamic rather than static, e.g., as we move to recognizing human movements on surveillance cameras, synthetic data takes the form of full-scale simulated environments. This direction started a long time ago: already in 2007, The \emph{ObjectVideo Virtual Video} (OVVV) system by Taylor et al.~\cite{4270516} used the \emph{Half-Life 2} game engine with additional camera parameters designed to simulate real-world surveillance cameras to detect a variety of different events. Fernandez et al.~\cite{FERNANDEZ2011878} place virtual agents onto real video surveillance footage in a kind of augmented reality to simulate rare events. Qureshi and Terzopoulos~\cite{4270096} present a multi-camera virtual reality surveillance system. 

An interesting human-related video analysis problem, important for autonomous vehicles, is to predict pedestrian trajectories in an urban environment. Anderson et al.~\cite{Anderson2019StochasticSS} develop a method for stochastic sampling-based simulation of pedestrian trajectories. They then train the \emph{SocialGAN} model by Gupta et al.~\cite{DBLP:journals/corr/abs-1803-10892} that generates pedestrian trajectories with a recurrent architecture and uses a recurrent discriminator to distinguish fake trajectories from real ones; Anderson et al. show that synthetic trajectories significantly improve the results for a predictive model such as \emph{SocialGAN}.

\begin{figure}[!t]\centering
\setlength{\tabcolsep}{3pt}
\begin{tabular}{P{.32\linewidth}P{.32\linewidth}P{.32\linewidth}}
\multicolumn{3}{c}{\myp{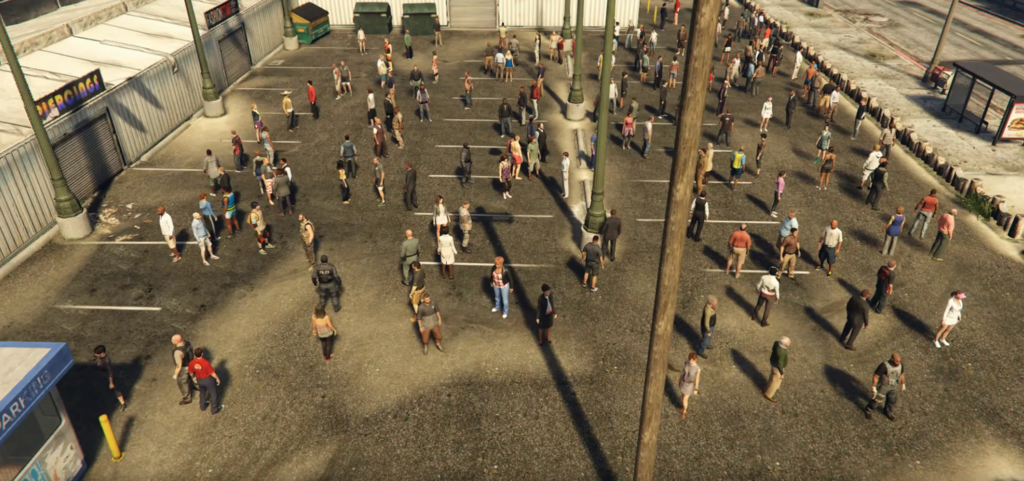}} \\
\multicolumn{3}{c}{(a)} \\
\myp{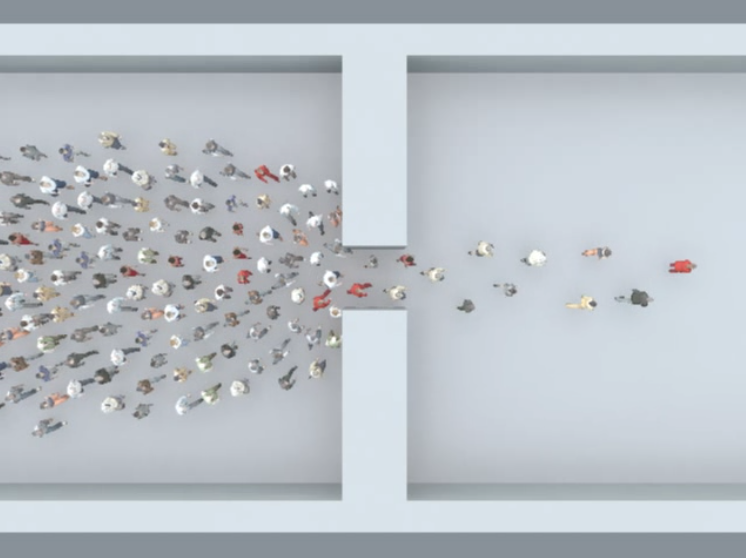} & \myp{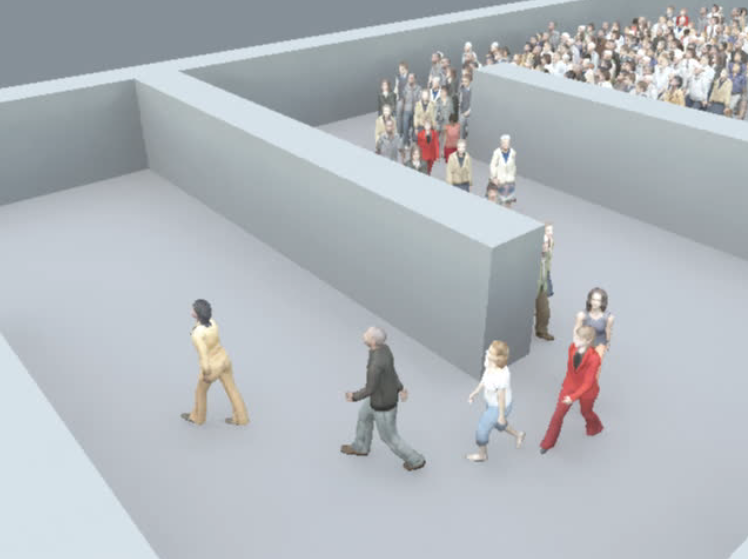} & \myp{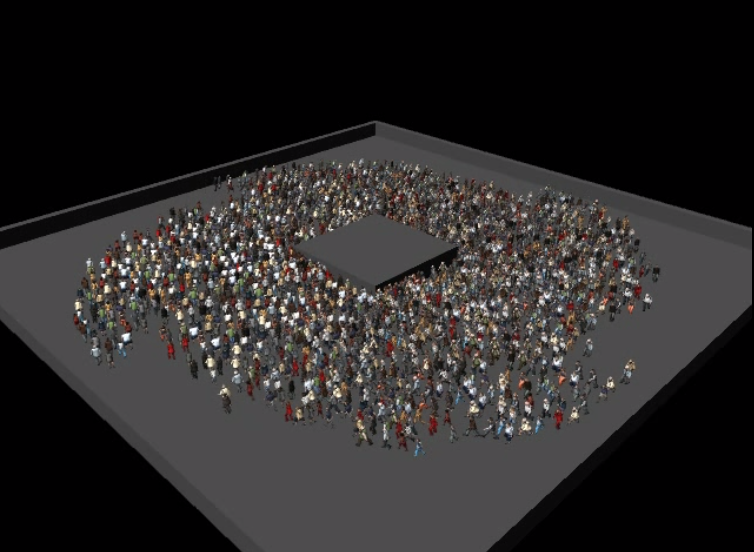} \\
\multicolumn{3}{c}{(b)} \\
\myp{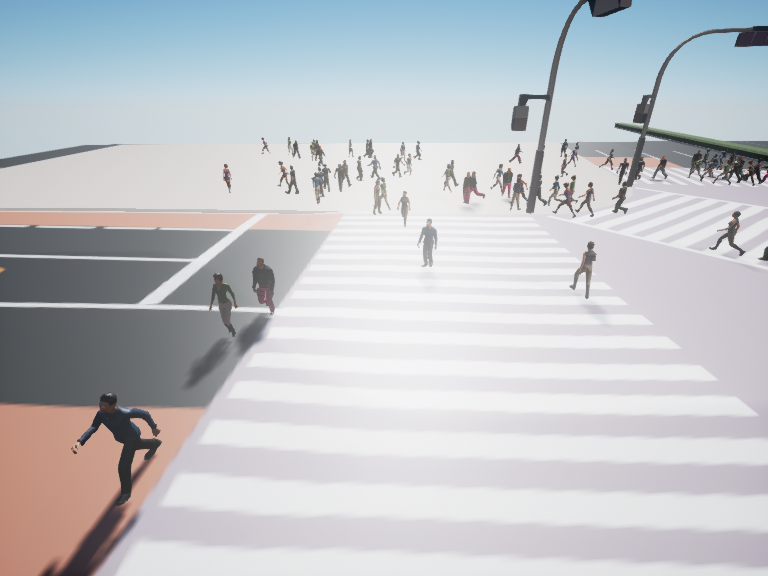} & \myp{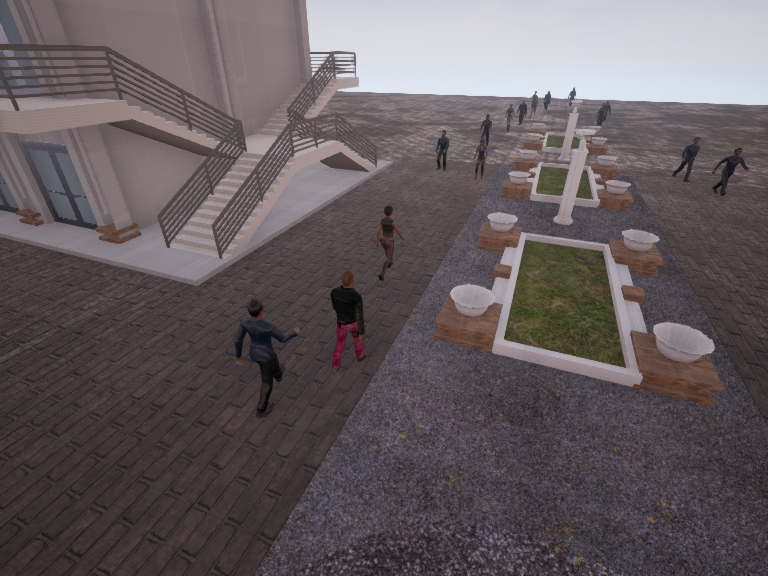} & \myp{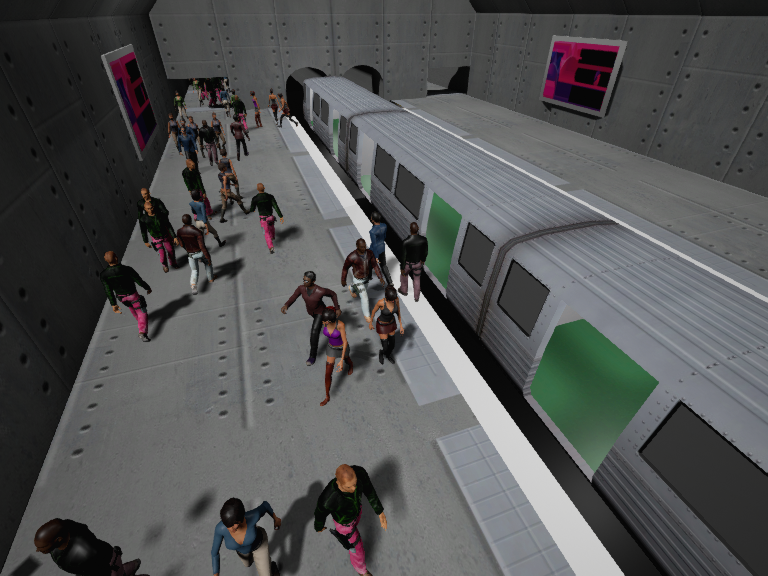} \\
\multicolumn{3}{c}{(c)} \\
\end{tabular}

\caption{Sample images from synthetic crowd counting datasets: (a) \emph{GTA5 Crowd Counting}~\cite{DBLP:journals/corr/abs-1903-03303}; (b) \emph{Agoraset}~\cite{COURTY2014161}; (c) \emph{LCrowdV}~\cite{10.1007/978-3-319-48881-3_50}.}\label{pic:crowd}
\end{figure}

Another important application of datasets of synthetic people is crowd counting. In this case, collecting ground truth labels, especially if the model is supposed to do segmentation in addition to simple counting, is especially labor-intensive since crowd counting scenes are often highly congested and contain hundreds, if not thousands of people. Existing real datasets are either relatively small or insufficiently diverse; e.g., the UCSD dataset~\cite{4587569} and the Mall dataset~\cite{Chen2012FeatureMF} have both about 50{,}000 pedestrians but in each case collected from a single surveillance camera, the ShanghaiTech dataset~\cite{7780439} has about 330{,}000 heads but only about 1200 images, again collected on the same event, and the UCF-QNRF dataset~\cite{DBLP:journals/corr/abs-1808-01050}, while more diverse than previous ones, is limited to extremely congested scenes, with up to 12{,}000 people on the same image, and has only about 1500 images. The \emph{LCrowdV} system~\cite{10.1007/978-3-319-48881-3_50} generates labeled crowd videos and shows that augmenting real data with a produced synthetic dataset improves the accuracy of pedestrian detection. 

To provide sufficient diversity and scale, Wang et al.~\cite{DBLP:journals/corr/abs-1903-03303} presented a synthetic \emph{GTA5 Crowd Counting} dataset collected with the help of the \emph{Grand Theft Auto V} engine; the released dataset contains about 15{,}000 synthetic images with more than 7{,}5 million annotated people in a wide variety of scenes. They compare various approaches to crowd counting as a supervised problem, in particular their new spatial fully convolutional network (SFCN) model that directly predicts the density map of people on a crowded image. They report improved results when pretraining on GCC and then fine-tuning on a real dataset; they also consider GAN-based approaches that we discuss in Section~\ref{sec:damodel}. A more direct approach to generating synthetic data has been developed by Ekbatani et al.~\cite{icpram17}, who extract real pedestrians from images and add them at various locations on other backgrounds, with special improvement procedures for added realism; they also report improved counting results. Khadka et al.~\cite{Khadka:2019:LAC:3328756.3328773} also present a synthetic crowd dataset, showing improvements in crowd counting.

This ties into crowd analysis, where synthetic data is used to model crowds and train visual crowd analysis tools on rendered images~\cite{5562657}. Huang et al.~\cite{7279793} present virtual crowd models that could be used for such simulations. Courty et al.~\cite{COURTY2014161} present the \emph{Agoraset} dataset for crowd analysis research that aims to provide realistic agent trajectories (done through the social force model by Heibling and Moln\'ar~\cite{PhysRevE.51.4282}) and high-quality rendering with the Mental Ray renderer~\cite{Driemeyer01}; the dataset has 26 different characters and provides a variety of different scenes: corridor, flow around obstacles, escape through a bottleneck, and so on.

In general, we summarize that while the most popular problems such as frontal face recognition or human pose estimation are already being successfully solved with models trained on real datasets (because there has been sufficient interest for these problems to collect and manually label large-scale datasets), synthetic data remains very important for alleviating the effect of dataset bias in real collections, covering corner cases, and tackling other problems or basic problems with different kinds of data, e.g., in different modalities (such as face recognition with an IR sensor). We believe that there are important opportunities for synthetic data in human-related computer vision problems and expect this field to grow in the near future.

\subsection{Character and text recognition}\label{sec:ocr}

Various tasks related to text recognition, including optical character recognition (OCR), text detection, layout analysis and text line segmentation for document digitization, and others have often been attacked with the help of synthetic data, usually with synthetic text superimposed on real images. This is a standard technique in the field because text pasted in a randomized way often looks quite reasonable even with minimal additional postprocessing. Synthetic data was used for character detection and recognition in, e.g.,~\cite{6460871,Campos09,10.1007/978-3-642-31919-8_25,7469575} and for text block detection in, e.g.,~\cite{DBLP:journals/corr/JaderbergSVZ14,Jaderberg:2016:RTW:2877061.2877148}. Krishnan and Jawahar~\cite{Krishnan2019} use synthetic data to pretrain deep neural networks for learning efficient representations of handwritten word images. Jo et al.~\cite{Jo2019} train an end-to-end convolutional architecture that can digitize documents with a mixture of handwritten and printed text; to train the network, they produce a synthetic dataset with real handwritten text superimposed on machine-printed forms, with Otsu binarization applied before pasting.

There exist published datasets of synthetic text and software to produce them, in particular MJSynth~\cite{DBLP:journals/corr/JaderbergSVZ14} and SynthText in the Wild~\cite{DBLP:journals/corr/GuptaVZ16}. In the latter work, Gupta et al. use available depth estimation and segmentation solutions to find regions (planes) of a natural image suitable for placing synthetic text and find the correct rotation of text for a given plane. Moreover, recent works have used GAN-based refinement (see Section~\ref{sec:refine}) to make synthetic text more realistic~\cite{Efimova19}. There are also synthetic handwriting generation models based on GANs that are conditioned on character sequences and produce excellent results~\cite{DBLP:journals/corr/abs-1907-11845,DBLP:journals/corr/abs-1903-00277}.

\subsection{Visual reasoning}\label{sec:datareason}

\begin{figure}[!t]\centering\noindent
\myp[.8\linewidth]{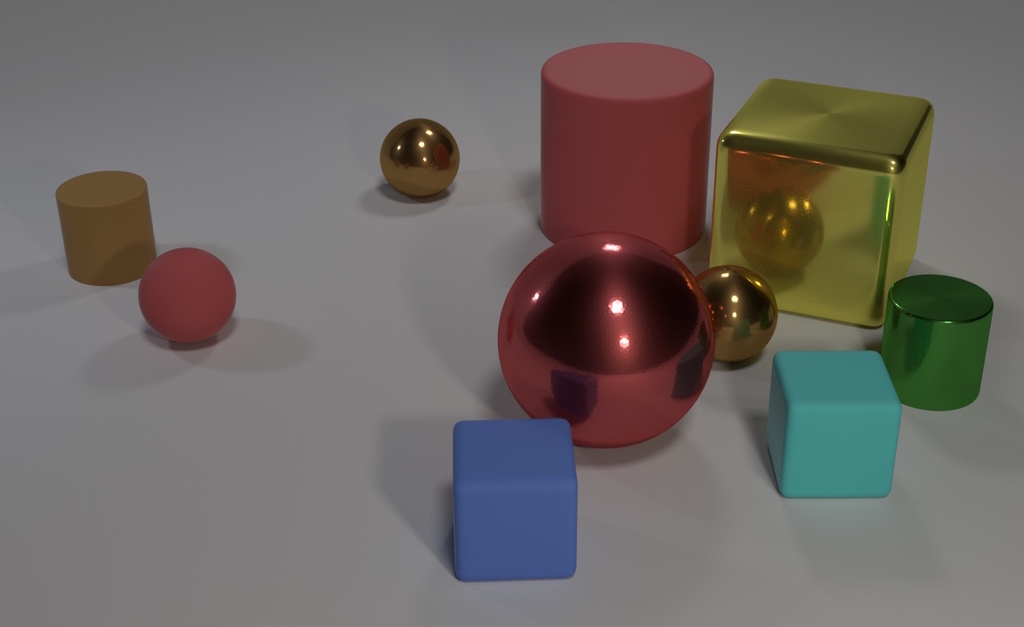}

\noindent
{\small (a)}

\noindent
\input{tikz_clevr}

\caption{The CLEVR dataset~\cite{Johnson2017CLEVRAD}: (a) sample image; (b-c) sample visual reasoning questions.}\label{fig:clevr}
\end{figure}

\emph{Visual reasoning} is the field of artificial intelligence where models are trained to reason and answer questions about visual data. It is usually studied in the form of visual question answering (VQA), when models are trained to answer questions about a picture such as ``What is the color of the small metal sphere?'' or ``Is there an equal number of balls and boxes?''.

There exist datasets for visual question answering based on real photographs, collected and validated by human labelers; they include the first large dataset called VQA~\cite{Agrawal2017} and its recent extension, VQA v2.0~\cite{Goyal2019}. However, the problem yields itself naturally to automated generation, so it is no wonder that synthetic datasets are important in the field.

The most important synthetic VQA dataset is CLEVR (Compositional Language and Elementary Visual Reasoning), created by Johnson et al. in a collaboration between Stanford and Facebook Research~\cite{Johnson2017CLEVRAD}. It contains 100K rendered images with scenes composed of simple geometric shapes and about 1M (853K unique) automatically generated questions about these images. The intention behind this dataset was to enable detailed analysis of VQA models, simplifying visual recognition and concentrating on reasoning about the objects.

In CLEVR, scenes are represented as scene graphs~\cite{Krishna2017,Johnson2015ImageRU}, where the nodes are objects annotated with attributes (shape, size, material, and color) and edges correspond to spatial relations between objects (``left'', ``right'', ``behind'', and ``in front''). A scene can be rendered based on its scene graph with randomized positions of the objects. The questions are represented as functional programs that can be executed on scene graphs, e.g., ``What color is the cube to the right of the white sphere?''. Different question types include querying attributes (``what color''), comparing attributes (``are they the same size''), existence (``are there any''), counting (``how many''), and integer comparison (``are there fewer''). When generating questions, special care is taken to ensure that the answer exists and is unique, and then the natural language question is generated with a relatively simple grammar. Figure~\ref{fig:clevr} shows two sample questions and their functional programs: on Fig.~\ref{fig:clevr}a the program is a simple chain of filters, and Fig.~\ref{fig:clevr}b adds a logical connective, which makes the graph a tree.

We also note a recently published COG dataset produced by Google Brain researchers~\cite{Yang2018ADA} that extends CLEVR's ideas to video processing. It also contains synthetic visual inputs and questions generated from functional programs, but now questions can refer to time (e.g., ``what is the color of the latest triangle?''). The authors also released a generator that can produce synthetic video-question pairs that are progressively more challenging and that have minimal response bias, an important problem for synthetic datasets (in this case, the generator begins with a balanced set of target responses and then generates videos and questions for them rather than the other way around).

%% file: tikz_clevr.tex
\begin{tikzpicture}[node distance=.3cm,start chain=going right, every node/.append style={on chain}]
	\node[diablo2,join] (n1) {Filter color};
	\node[diablo2,join] (n2) {Filter shape};
	\node[diablo2,join] (n3) {Unique};
	\node[diablo2,join] (n4) {Relate};
	\node[diablo2,join] (n5) {Filter shape};
	\node[diablo2,join] (n6) {Unique};
	\node[diablo2,join] (n7) {Query color};

	\node[nd,above=.15cm of n4,font=\footnotesize\sffamily] (title) {(b) Chain-structured question: \textit{What color is the cube to the right of the red cylinder?}};
	\node[diablo3,below=.3cm of n1] (w1) {red};
	\draw[lineblue] (w1) -> (n1);
	\node[diablo3,below=.3cm of n2] (w2) {cylinder};
	\draw[lineblue] (w2) -> (n2);
	\node[diablo3,below=.3cm of n4] (w4) {right};
	\draw[lineblue] (w4) -> (n4);
	\node[diablo3,below=.3cm of n5] (w5) {cube};
	\draw[lineblue] (w5) -> (n5);
\end{tikzpicture}\vspace{.2cm}

\begin{tikzpicture}[node distance=1.6cm]
	\node[font=\footnotesize\sffamily,text width=25em] (title) {(c) Tree-structured question: \textit{How many spheres are behind the blue thing and on the left side of the left cylinder?}};

	\node[diablo2,below left=.15cm and 0cm of title] (m1) {Filter shape};
	\node[diablo2,left of=m1] (m0) {Filter color};
	\node[diablo2,right of=m1] (m2) {Unique};
	\node[diablo2,right of=m2] (m3) {Relate};
	\node[diablo3,below=.3cm of m0] (v0) {green};
	\node[diablo3,below=.3cm of m1] (v1) {cylinder};
	\draw[lineblue] (v1) -> (m1);
	\draw[lineblue] (v0) -> (m0);
	\node[diablo3,below=.3cm of m3] (v3) {left};
	\draw[lineblue] (v3) -> (m3);

	\node[diablo2,below=.15cm of v1] (m4) {Filter color};
	\node[diablo2,right of=m4] (m5) {Unique};
	\node[diablo2,right of=m5] (m6) {Relate};
	\node[diablo3,below=.3cm of m4] (v4) {blue};
	\draw[lineblue] (v4) -> (m4);
	\node[diablo3,below=.3cm of m6] (v6) {behind};
	\draw[lineblue] (v6) -> (m6);

	\node[diablo2,below right=0cm and .5cm of m3,fill=green!10] (m7) {AND};
	\node[diablo2,right of=m7] (m8) {Filter shape};
	\node[diablo2,right of=m8] (m9) {Count};
	\node[diablo3,below=.3cm of m8] (v8) {sphere};
	\draw[lineblue] (v8) -> (m8);

	\draw[thick,shorten >=1pt,->,rounded corners] (m0) edge (m1) (m1) edge (m2) (m2) edge (m3) (m3) edge (m7);
	\draw[thick,shorten >=1pt,->,rounded corners] (m4) edge (m5) (m5) edge (m6) (m6) edge (m7);
	\draw[thick,shorten >=1pt,->,rounded corners] (m7) edge (m8) (m8) edge (m9);
\end{tikzpicture}

%% file: env.tex
\section{Synthetic simulated environments}\label{sec:direct}

While collecting synthetic datasets is a challenging task by itself, it is insufficient to train, e.g., an autonomous vehicle such as a self-driving car or a drone, or an industrial robot. Learning to control a vehicle or robot often requires \emph{reinforcement learning}, where an agent has to learn from interacting with the environment, and real world experiments to train a self-driving car or a robotic arm are completely impractical. Fortunately, this is another field where synthetic data shines: once one has a fully developed 3D environment that can produce datasets for computer vision or other sensory readings, it is only one more step to active interaction with this environment. Therefore, in most domains considered below we can see the shift from static synthetic datasets to interactive simulation environments. 

Reinforcement learning (RL) agents are commonly trained on simulations because the interactive nature of reinforcement learning makes training in the real world extremely expensive. We discuss synthetic-to-real domain adaptation in this context in Section~\ref{sec:darobot}. However, in many works, there is no explicit domain adaptation: robots are trained on simulators and later fine-tuned on real data or simply transferred to the real world. 

Table~\ref{tbl:datasim} shows a brief summary of datasets and simulators that we review in this section. To make the exposition more clear, we group together both environments and ``static'' synthetic datasets for outdoor (Section~\ref{sec:dataoutdoor}) and indoor (Section~\ref{sec:dataindoor}) scenes, including some works that use them to improve RL agents and other models. Next, we consider synthetic robotic simulators (Section~\ref{sec:robotics}) and vision-based simulators for autonomous flying (Section~\ref{sec:cvnav}), finishing with an idea of using computer games as simulation environments in Section~\ref{sec:games}. Reinforcement learning in virtual environments remains a common thread throughout this section.



\subsection{Urban and outdoor environments: learning to drive}\label{sec:dataoutdoor}

An important direction of applications for synthetic data is related to navigation, localization and mapping (SLAM), or similar problems intended to improve the motion of autonomous robots. Possible applications include SLAM, motion planning, and motion for control for self-driving cars (urban navigation)~\cite{Lategahn11,Milz_2018_CVPR_Workshops,Sensing17,Paden16}, unmanned aerial vehicles~\cite{KanellakisN17,ALKAFF2018447,COURBON2010789}, and more; see also general surveys of computer vision for mobile robot navigation~\cite{982903,Bonin-Font2008} and perception and control for autonomous driving~\cite{machines5010006}.

Before proceeding to current state of the art, we note an interesting historical fact: one of the first autonomous driving attempts based on neural networks, ALVINN~\cite{Pomerleau:1989:AAL:89851.89891}, which used as input $30\times 32$ videos supplemented with $8\times 32$ range finder data, was already training on synthetic data. As early as 1989, the authors remark that ``training on actual road images is logistically difficult because... the network must be presented with a large number of training exemplars... under a wide variety of conditions'' and proceed to describing a simulator. One of the first widely adopted full-scale visual simulation environments for robotics, \emph{Gazebo}~\cite{1389727} (see Section~\ref{sec:robotics}), provided both indoor and outdoor environments for robotic control training.

% ...

\begin{table}[!t]\centering\small
\setlength{\tabcolsep}{3pt}
\begin{tabular}{|rcccl|}\hline
\bf Name & \bf Year & \bf Ref & \bf Engine & \bf Size / comments \\
\hline\multicolumn{5}{|c|}{\it Outdoor urban environments, driving} \\\hline
TORCS & 2014 & \cite{TORCS} & Custom & Game-based simulation engine \\
Virtual KITTI & 2016 & \cite{Gaidon:Virtual:CVPR2016} & Unity & 5 environments, 50 videos \\
GTAVision & 2016 & \cite{7989092} & GTA V & GTA plugin, 200K images \\
SYNTHIA & 2016 & \cite{7780721} & Unity  & 213K images \\
GTAV & 2016 & \cite{DBLP:journals/corr/RichterVRK16} & GTA V & 25K images \\
VIPER & 2017 & \cite{DBLP:journals/corr/abs-1709-07322} & GTA V & 254K images \\
CARLA & 2017 & \cite{Dosovitskiy17} & UE & Simulator \\
VIES & 2018 & \cite{DBLP:journals/corr/abs-1807-06132} & Unity3D & 61K images, 5 environments \\
ParallelEye & 2018 & \cite{Tian18} & Esri & Procedural gen, import from OSM \\
VIVID & 2018 & \cite{Lai:2018:VVE:3240508.3243653} & UE & Urban sim with emphasis on people \\
DeepDrive & 2018 & \cite{craig_quiter_2018_1248998} & UE & Driving sim + 8.2h of videos \\
PreSIL & 2019 & \cite{Hurl2019PreciseSI} & GTA V & 50K images with LIDAR point clouds \\
AADS & 2019 & \cite{Li2019AADSAA} & Custom & 3D models of cars on real backgrounds \\
WoodScape & 2019 & \cite{DBLP:journals/corr/abs-1905-01489} & Custom & 360\textdegree{} panoramas with fisheye cameras \\
\emph{ProcSy} & 2019 & \cite{Khan_2019_CVPR_Workshops} & Esri & Procedural generation with varying conditions \\
\hline\multicolumn{5}{|c|}{\it Robotic simulators and aerial navigation} \\\hline
Gazebo & 2004 & \cite{1389727} & Custom & Industry standard robotic sim \\
MuJoCo & 2012 & \cite{Todorov2012MuJoCoAP} & Custom & Common physics engine for robotics \\
AirSim & 2017 & \cite{airsim2017fsr} & UE & Sensor readings, hardware-in-the-loop \\
$\mathrm{CAD}^2\mathrm{RL}$ & 2017 & \cite{Sadeghi2017CAD2RLRS} & Custom & Indoor flying sim \\
X-Plane & 2019 & \cite{DBLP:journals/corr/abs-1811-11067} & X-Plane & 8K landings, 114 runways \\
Air Learning & 2019 & \cite{DBLP:journals/corr/abs-1906-00421} & --- & Platform for flying sims \\
VRGym & 2019 & \cite{Xie2019VRGymAV} & UE & VR for human-in-the-loop training \\
ORRB & 2019 & \cite{DBLP:journals/corr/abs-1906-11633} & Unity & Accurate sim used to train real robots \\
\hline\multicolumn{5}{|c|}{\it Indoor environments} \\\hline
ICL-NUIM & 2014 & \cite{6907054} & Custom & RGB-D with noise models, 2 scenes \\
SUNCG & 2016 & \cite{song2016ssc} & Custom & 45K floors, 3D models \\
MINOS & 2017 & \cite{savva2017minos} & SUNCG & Indoor sim based on SUNCG \\
AI2-THOR & 2017 & \cite{DBLP:journals/corr/abs-1712-05474} & Unity3D & Indoor sim with actionable objects \\
House3D & 2018 & \cite{DBLP:journals/corr/abs-1801-02209} & SUNCG & Indoor sim based on SUNCG \\
Habitat & 2019 & \cite{habitat19arxiv} & Custom & Indoor sim platform and library \\
\hline
\end{tabular}

\caption{An overview of synthetic datasets and virtual environments discussed in Section~\ref{sec:direct}.}\label{tbl:datasim}
\end{table}

In a much more recent effort, \emph{Xerox} researchers Gaidon et al.~\cite{Gaidon:Virtual:CVPR2016} presented a photorealistic synthetic video dataset Virtual KITTI\footnote{The name comes from the KITTI dataset~\cite{Geiger2013IJRR,Menze2015CVPR} created in a joint project of the Karlsuhe Institute of Technology and Toyota Technological Institute at Chicago.} intended for object detection and multi-object tracking, scene-level and instance-level semantic segmentation, optical flow, and depth estimation. The dataset contains five different virtual outdoor environments created with the Unity game engine and 50 photorealistic synthetic videos. Gaidon et al. studied existing multi-object trackers, e.g., based on an improved min-cost flow algorithm~\cite{5995604} and on Markov decision processes~\cite{7410891}; they found minimal real-to-virtual gap. Note, however, that experiments in~\cite{Gaidon:Virtual:CVPR2016} were done on trackers trained on real data and evaluated on synthetic videos (and that's where they worked well), not the other way around. In general, the \emph{Virtual KITTI} dataset is much too small to train a model on it, it is intended for evaluation, which also explains the experimental setup.

Johnson-Robertson et al.~\cite{7989092}, on the other hand, presented a method to \emph{train} on synthetic data. They collected a large dataset by capturing scene information from the \emph{Grand Theft Auto V} video game that provides sufficiently realistic graphics and at the same time stores scene information such as depth maps and rough bounding boxes in the GPU stencil buffer, which can also be captured; the authors developed an automated pipeline to obtain tight bounding boxes. Three datasets were generated, with 10K, 50K, and 200K images respectively. The main positive result of~\cite{7989092} is that a standard Faster R-CNN architecture~\cite{7485869} trained on 50K and 200K images outperformed on a real validation set (KITTI) the same architecture trained on a real dataset. The real training set was \emph{Cityscapes}~\cite{Cordts2016Cityscapes} that contains 2{,}975 images, so while the authors used more synthetic data than real, the difference is only 1-2 orders of magnitude. The VIPER and GTAV datasets by Richter et al.~\cite{DBLP:journals/corr/abs-1709-07322,DBLP:journals/corr/RichterVRK16} were also captured from \emph{Grand Theft Auto V}; the latter provides more than 250K $1920\times 1080$ images fully annotated with optical flow, instance segmentation masks, 3D scene layout, and visual odometry.

\begin{figure}[!t]\centering
\setlength{\tabcolsep}{3pt}
\begin{tabular}{P{.49\linewidth}P{.49\linewidth}}
\myp{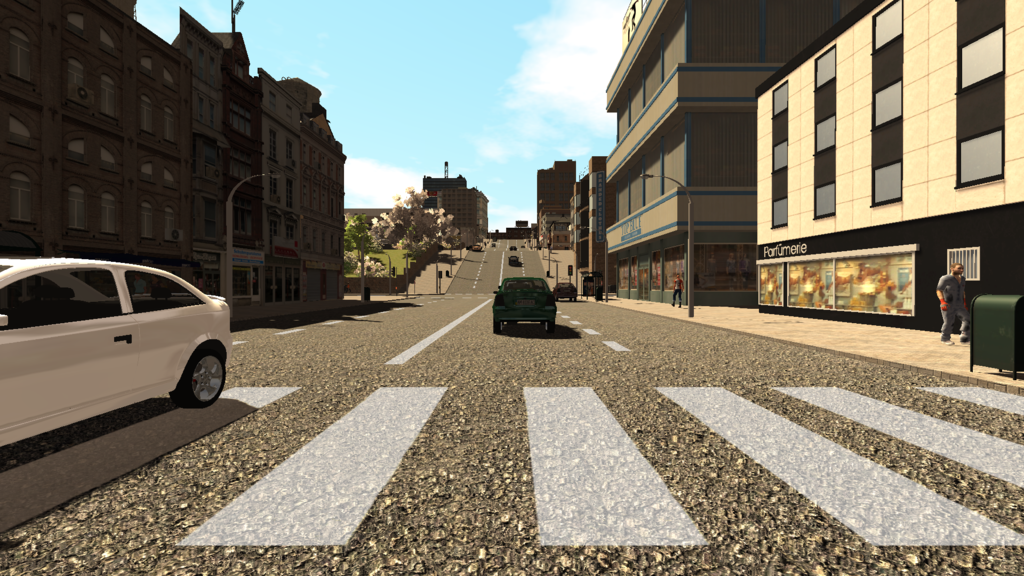} & \myp{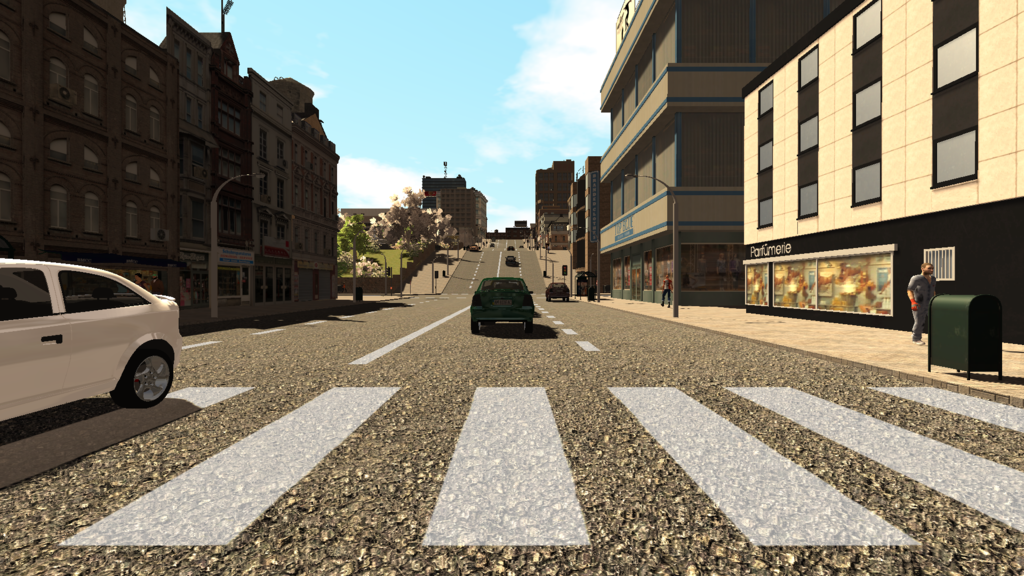} \\
(a) & (b) \\
\myp{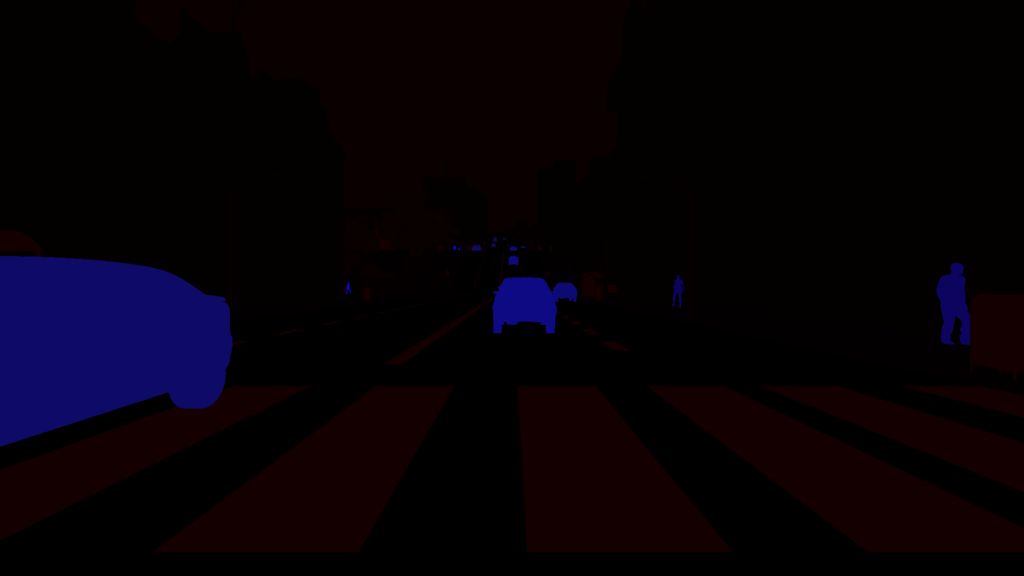} & \myp{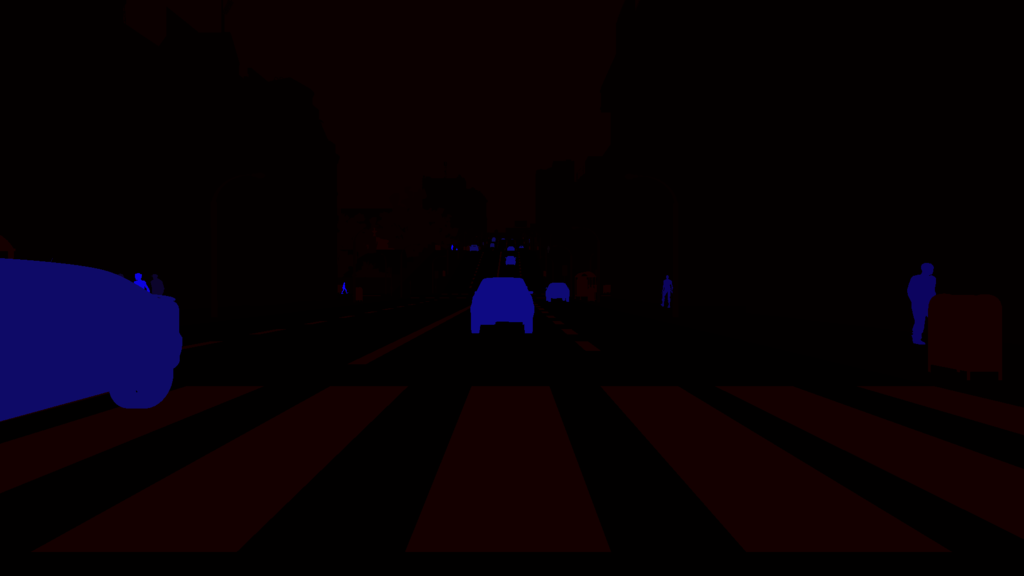} \\
(c) & (d) \\
\myp{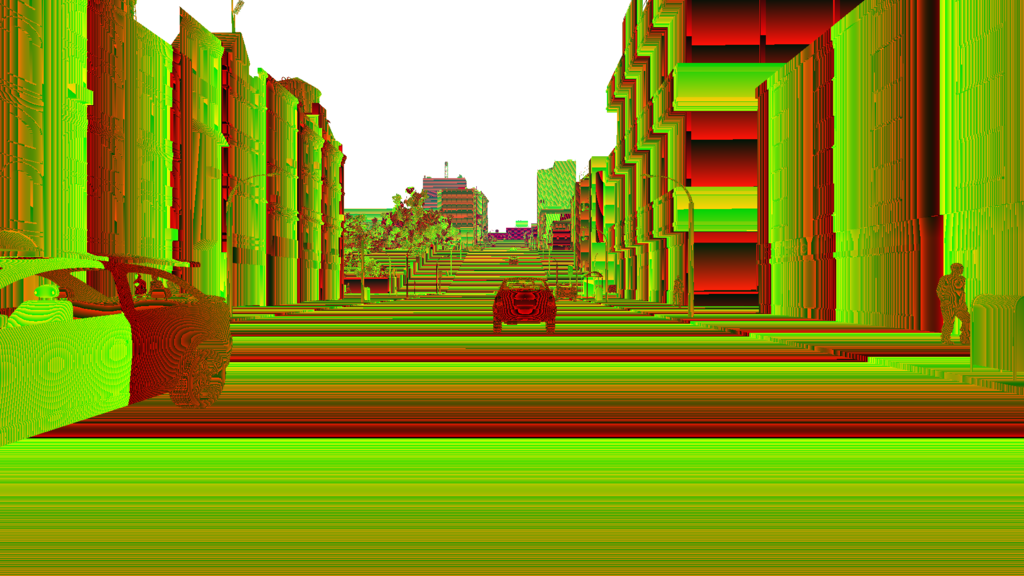} & \myp{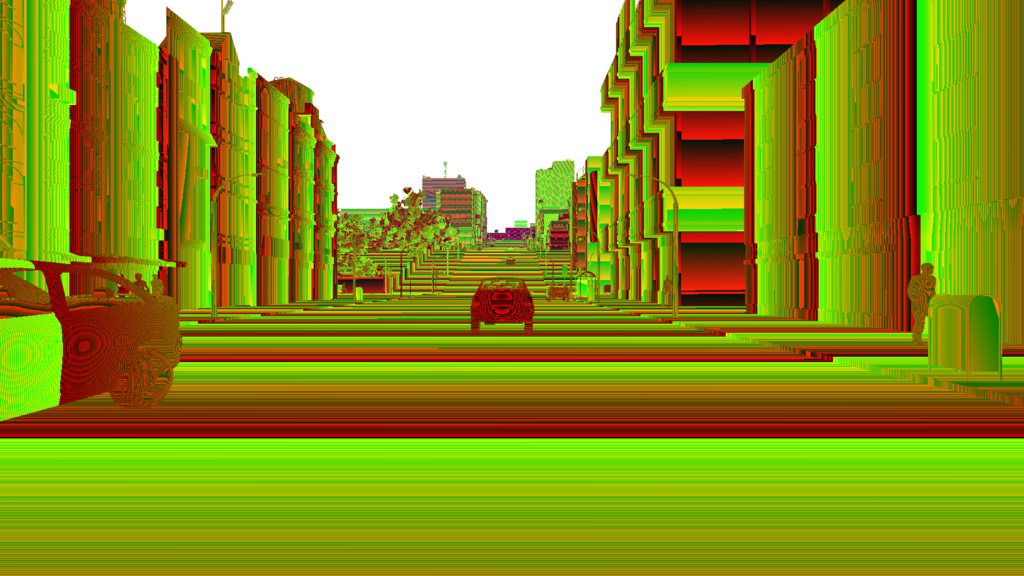} \\
(e) & (f) \\
\end{tabular}

\caption{Sample images from SYNTHIA-SF~\cite{HernandezBMVC17}: (a-b) RGB ground truth (left and right camera); (c-d) ground truth segmentation maps; (e-f) depth maps (depth is encoded in the color as $R+256\cdot G+256^2\cdot B$).}\label{pic:synthiasf}
\end{figure}

The SYNTHIA dataset presented by Ros et al.~\cite{7780721} provides synthetic images of urban scenes labeled for semantic segmentation. It consists of renderings of a virtual New York City constructed by the authors with the \emph{Unity} platform and includes segmentation annotations for 13 classes such as pedestrians, cyclists, buildings, roads, and so on. The dataset contains more than $213{,}000$ synthetic images covering a wide variety of scenes and environmental conditions; experiments in~\cite{7780721} show that augmenting real datasets with SYNTHIA leads to improved segmentation. Later, Hernandez-Juarez et al.~\cite{HernandezBMVC17} presented SYNTHIA-SF, the San Francisco version of SYNTHIA. We illustrate SYNTHIA with a sample frame (that is, two frames since the dataset contains two cameras) from the SYNTHIA-SF dataset on Figure~\ref{pic:synthiasf}.

\begin{figure}[!t]\centering
\setlength{\tabcolsep}{3pt}
\begin{tabular}{P{.49\linewidth}P{.49\linewidth}}
\multicolumn{2}{c}{\myp{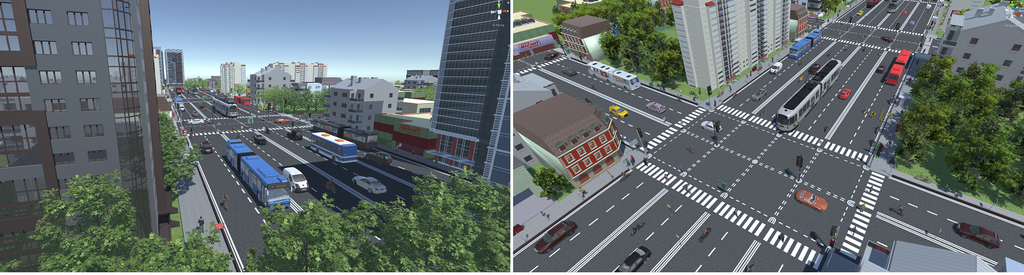}} \\
\multicolumn{2}{c}{(a)} \\
\multicolumn{2}{c}{\myp{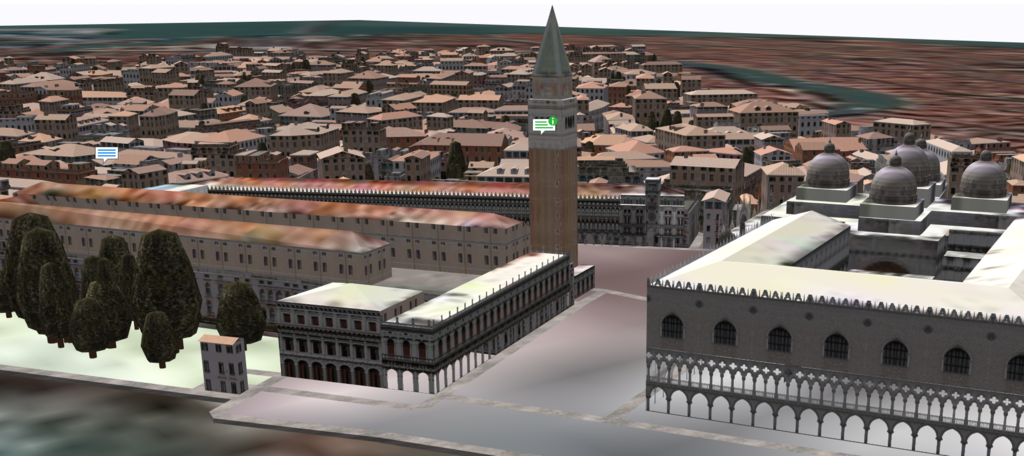}} \\
\multicolumn{2}{c}{(b)} \\
\multicolumn{2}{c}{\myp{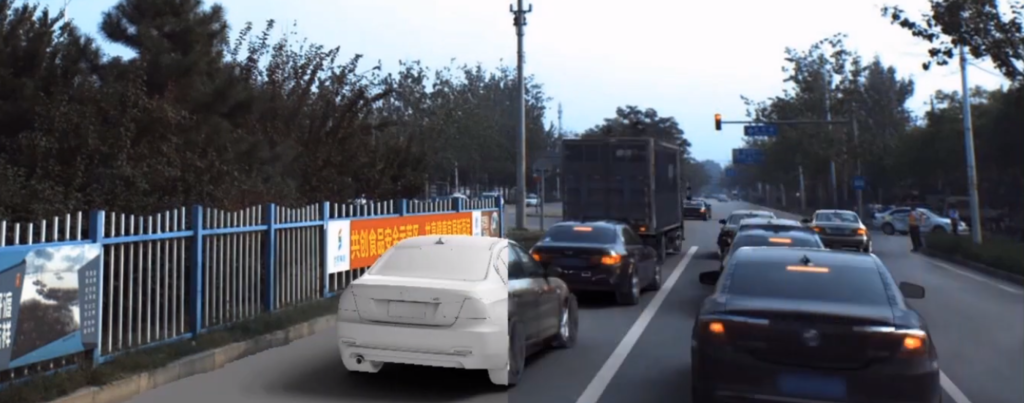}} \\
\multicolumn{2}{c}{(c)} \\
\end{tabular}

\caption{Sample images from synthetic outdoor datasets: (a) VEIS~\cite{DBLP:journals/corr/abs-1807-06132}; (b) \emph{Esri CityEngine} Venice sample scene~\cite{6737786}; (c) AADS~\cite{Li2019AADSAA} (part of a frame from a showcase video).}\label{pic:outdoor}
\end{figure}

\begin{figure}[!t]\centering
\setlength{\tabcolsep}{3pt}
\begin{tabular}{P{.49\linewidth}P{.49\linewidth}}
\myp{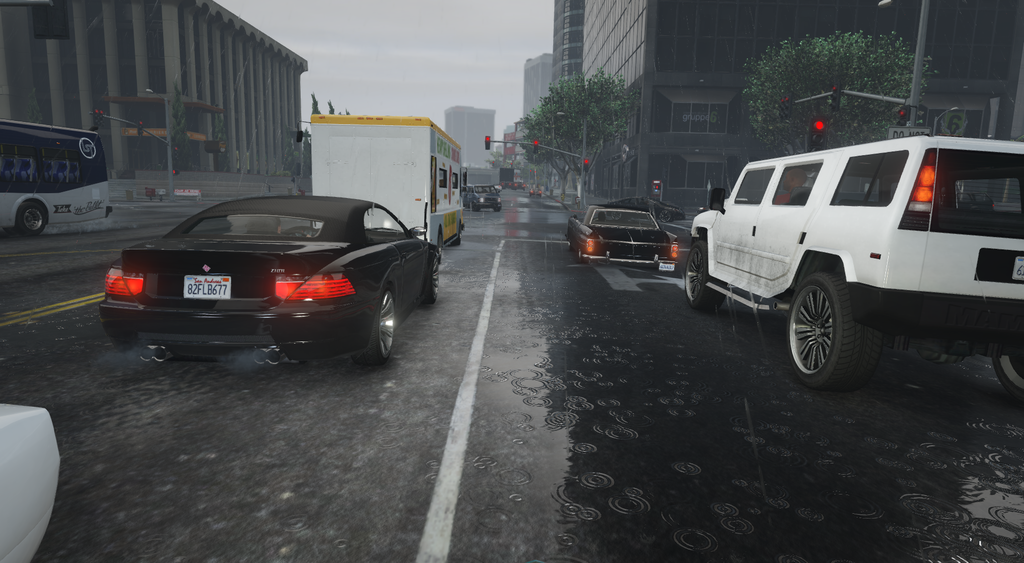} & \myp{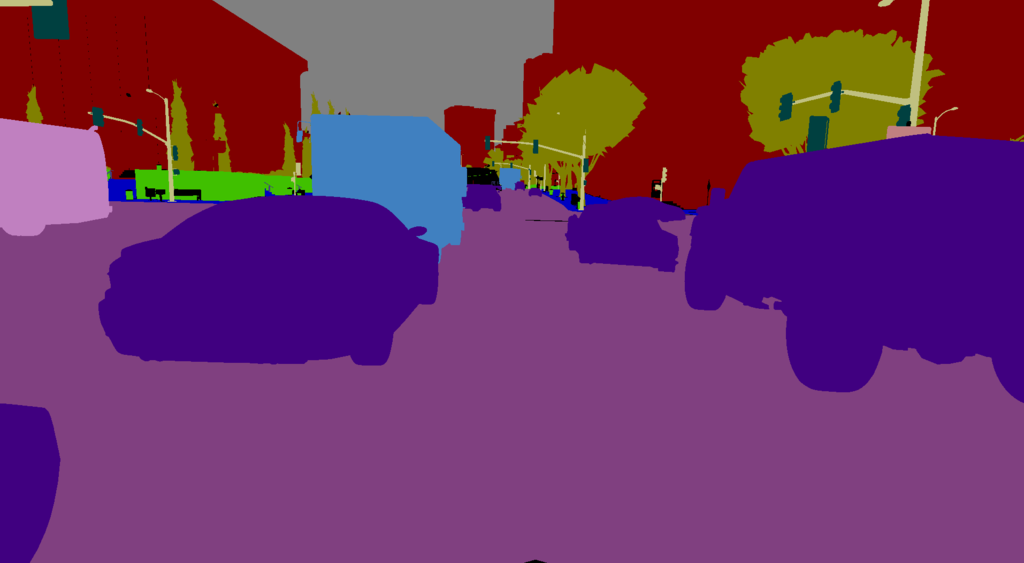} \\
(a) & (b) \\
\myp{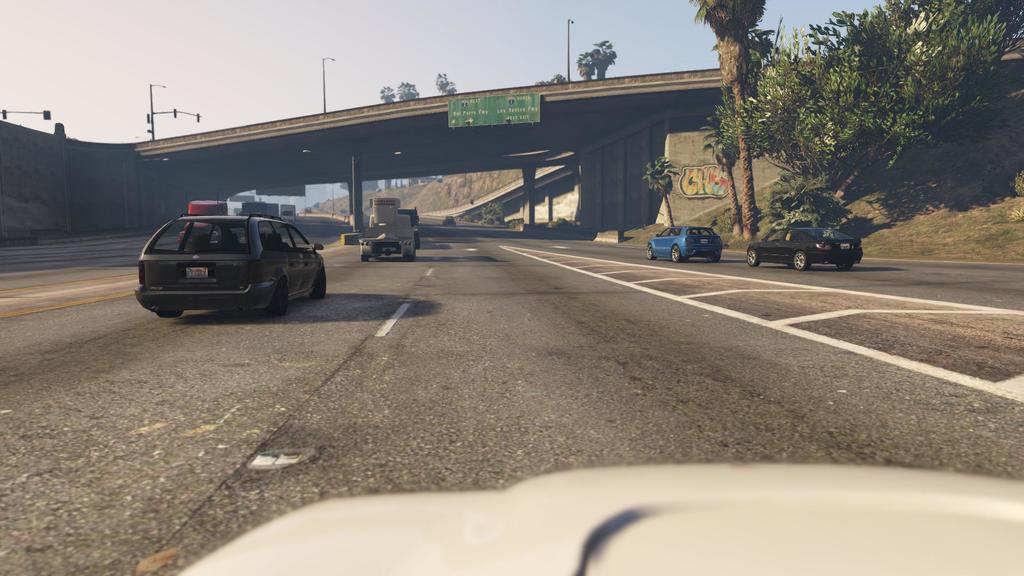} & \myp{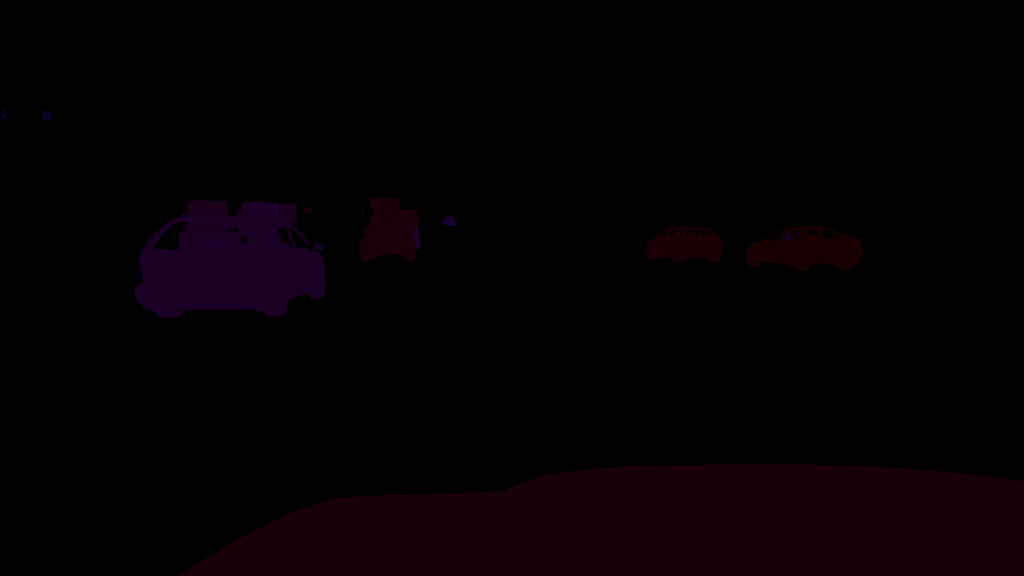} \\
(c) & (d) \\
\myp{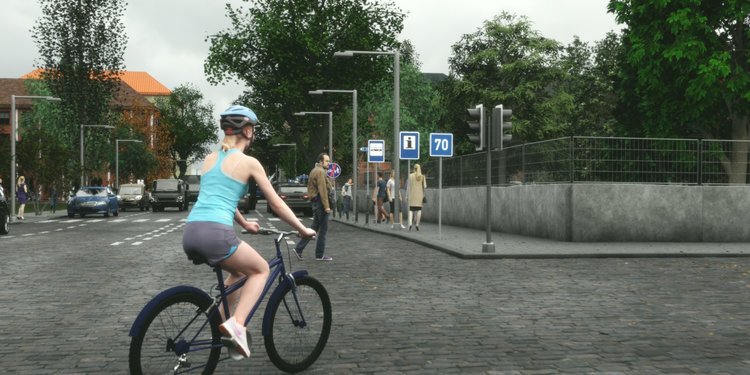} & \myp{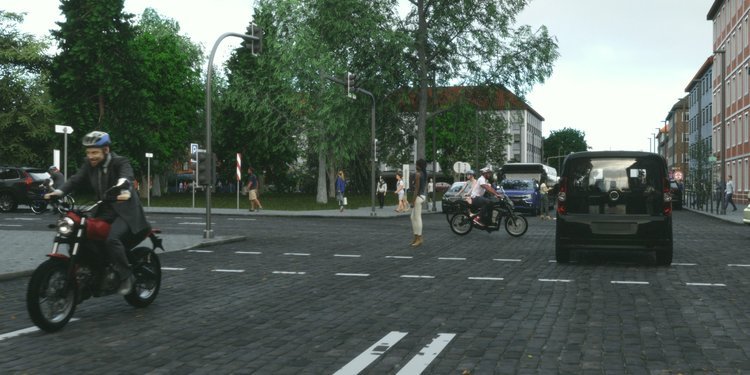} \\
(e) & (f) \\
\end{tabular}

\caption{Sample images from synthetic outdoor datasets: (a-b) GTAV~\cite{DBLP:journals/corr/RichterVRK16}: (a) RGB image, (b) ground truth segmentation; (c-d) VIPER~\cite{DBLP:journals/corr/abs-1709-07322}: (c) RGB image, (d) ground truth segmentation; (e-f) \emph{Synscapes}~\cite{DBLP:journals/corr/abs-1810-08705}.}\label{pic:outdoor2}
\end{figure}

Saleh et al.~\cite{DBLP:journals/corr/abs-1807-06132} presented a Unity3D framework called VEIS (Virtual Environment for Instance Segmentation); while not very realistic, it worked well with their detection-based pipeline (see Section~\ref{sec:visiongeneral}). Li et al.~\cite{8265275} present a synthetic dataset with foggy images to simulate difficult driving conditions. We note the work of Lopez et al.~\cite{Lopez2017} whose experiments suggest that the level of realism achieved in SYNTHIA and GTAV is already sufficient for successful transfer of object detection methods.

Tian et al.~\cite{Tian18} present the \emph{ParallelEye} synthetic dataset for urban outdoor scenes. Their approach is rather flexible and relies on previously developed \emph{Esri CityEngine} framework~\cite{6737786} that provides capabilities for batch generation of 3D city scenes based on terrain data. In~\cite{Tian18}, this data is automatically extracted from the OpenStreetMap platform\footnote{\url{https://www.openstreetmap.org/}}. The 3D scene is then imported into the Unity3D game engine, which helped add urban vehicles on the roads, set up traffic rules, add support for different weather and lighting conditions. Tian et al. showed improvements in object detection quality for state of the art architectures trained on \emph{ParallelEye} and tested on the real KITTI test set as compared to training on the real KITTI training set.

Li et al.~\cite{Li2019AADSAA} develop the \emph{Augmented Autonomous Driving Simulation} (AADS) environment that is able to insert synthetic traffic on real-life RGB images. Starting from the real-life \emph{ApolloScape} dataset for autonomous driving~\cite{DBLP:journals/corr/abs-1803-06184} that contains LIDAR point clouds, the authors remove moving objects, restore backgrounds by inpainting, estimate illumination conditions, simulate traffic conditions and trajectories of synthetic cars, preprocess the textures of the models according to lighting and other conditions, and add synthetic cars in realistic places on the road. In this way, a single real image can be reused many times in different synthetic traffic situations. This is similar to the approach of Abu Alhaija et al.~\cite{AbuAlhaija2018} (Section~\ref{sec:objects}) but due to available 3D information AADS can also change the observation viewpoint and even be used in a closed-loop simulator such as CARLA or \emph{AirSim} (see below). We do not go into details on the already large and diverse field of virtual traffic simulation and refer to a recent survey~\cite{doi:10.1111/cgf.13803}.

Wrenninge and Unger~\cite{DBLP:journals/corr/abs-1810-08705} present the \emph{Synscapes} dataset that continues the work of Tsirikoglou et al.~\cite{DBLP:journals/corr/abs-1710-06270} (see Section~\ref{sec:visiongeneral}) and contains accurate photorealistic renderings of urban scenes (Fig.~\ref{pic:outdoor2}e-f), with unbiased path tracing for rendering, special models for light scattering effects in camera optics, motion blur, and more. They find that their additional efforts for photorealism do indeed result in significant improvements in object detection over GTA-based datasets, even though the latter have a wider variety of scenes and pedestrian and car models.

Khan et al.~\cite{Khan_2019_CVPR_Workshops} introduce \emph{ProcSy}, a procedurally generated synthetic dataset aimed at semantic segmentation (we showed a sample frame on Fig.~\ref{pic:intro}c-d). It is modeling a real world urban environment, and its main emphasis is on simulating various weather and lighting conditions for the same scenes. The authors show that, e.g., adding a mere 3\% of rainy images in the training set improves the mIoU of a state of the art segmentation network (in this case, Deeplab v3+~\cite{10.1007/978-3-030-01234-2_49}) by as much as 10\% on rainy test images. This again supports the benefits from using synthetic data to augment real datasets and cover rare cases; for a discussion of the procedural side of this work see Section~\ref{sec:procedural}.

Synthetic datasets with explicit 3D data (with simulated sensors) for outdoor environments are less common, although such sensors seem to be straightforward to include into self-driving car hardware. In their development of the \emph{SqueezeSeg} architecture, Wu et al.~\cite{DBLP:journals/corr/abs-1710-07368,DBLP:journals/corr/abs-1809-08495} added a LiDAR simulator to \emph{Grand Theft Auto V} and collected a synthetic dataset from the game. \emph{SynthCity} by Griffiths and Boehm~\cite{2019arXiv190704758G} is a large-scale open synthetic dataset which is basically a huge point cloud of an urban/suburban environment. It simulates \emph{Mobile Laser Scanner} (MLS) readings with a \emph{Blender} plugin~\cite{10.1007/978-3-642-24031-7_20} and is specifically intended for pretraining deep neural networks. Yogamani et al.~\cite{DBLP:journals/corr/abs-1905-01489} present \emph{WoodScape}, a multi-camera fisheye dataset for autonomous driving that concentrates on getting 360\textdegree{} sensing around a vehicle through panoramic fisheye images with a large field of view. They record 4 fisheye cameras with 190\textdegree{} horizontal field of view, a rotating LiDAR, GNSS and IMU sensors, and odometry signals with 400K frames with depth labeling and 10K frames with semantic segmentation labeling. Importantly for us, together with their real dataset they also released a synthetic part (10K frames) that matches their fisheye cameras, with the explicit purpose of helping synthetic-to-real transfer learning. This further validates the importance of synthetic data in autonomous driving.

Simulated environments rather than datasets are, naturally, also an important part of the outdoor navigation scene; below we describe the main players in the field and also refer to surveys~\cite{8507288,Kang2019TestYS} for a more in-depth analysis of some of them. There is also a separate line of work related to developing more accurate modeling in such simulators, e.g., sensor noise models~\cite{8570015}, that falls outside the scope of this survey.

\begin{figure}[!t]\centering
\setlength{\tabcolsep}{3pt}
\begin{tabular}{P{.49\linewidth}P{.49\linewidth}}
\myp{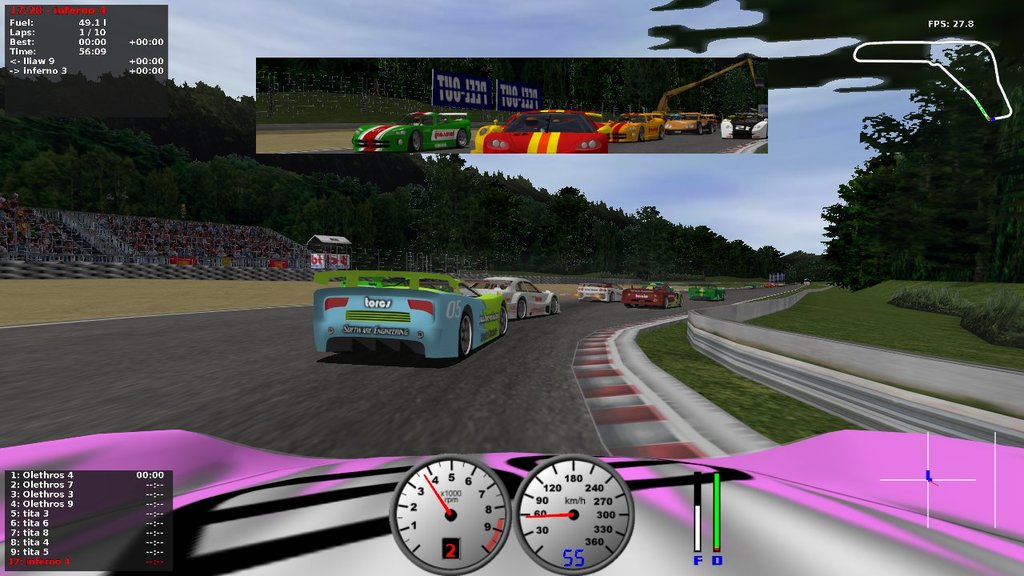} & \myp{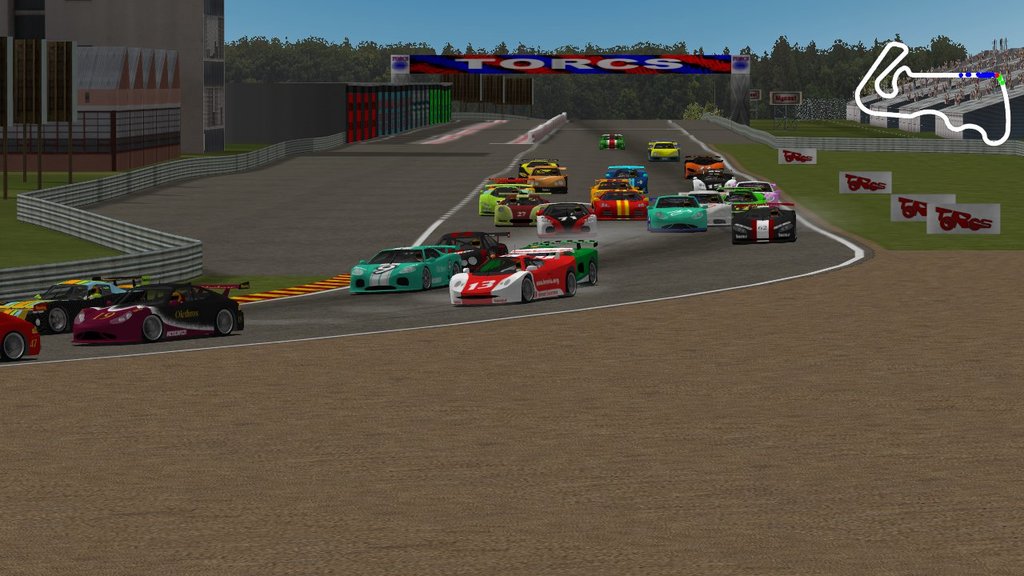} \\
(a) & (b) \\
\end{tabular}

\begin{tabular}{P{.38\linewidth}P{.29\linewidth}P{.29\linewidth}}
\myp{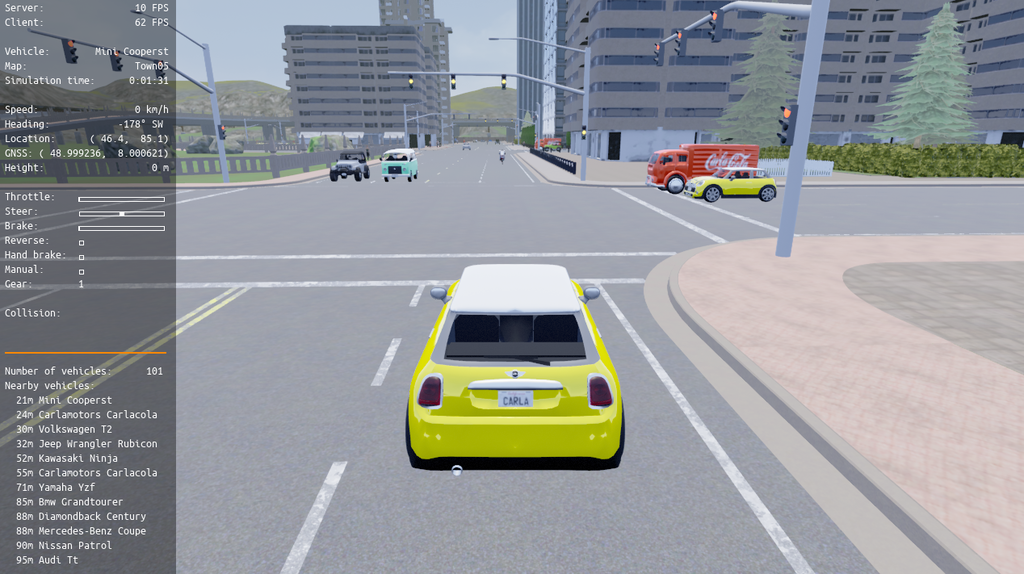} & \myp{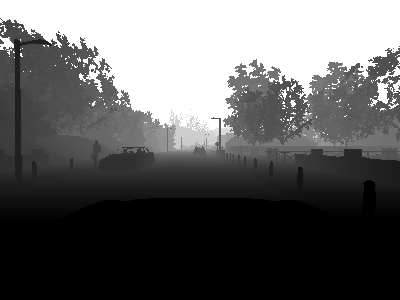} & \myp{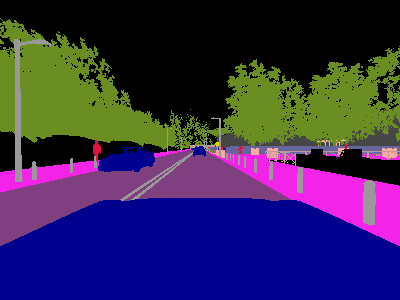} \\
(c) & (d) & (e) \\
\end{tabular}

\begin{tabular}{P{.32\linewidth}P{.32\linewidth}P{.32\linewidth}}
\myp{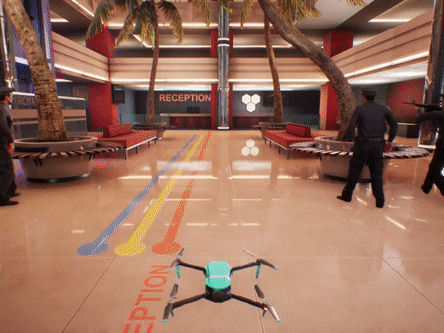} & \myp{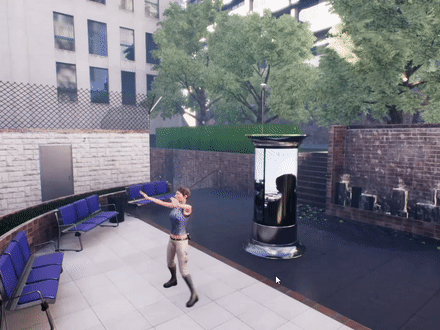} & \myp{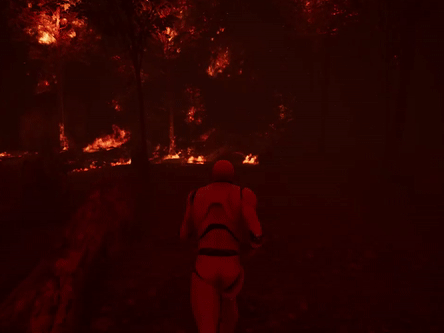} \\
(f) & (g) & (h) \\
\end{tabular}

\caption{Sample images from outdoor environments: (a-b) TORCS~\cite{TORCS}; (c-e) CARLA~\cite{Dosovitskiy17}; (f-h) VIVID~\cite{Lai:2018:VVE:3240508.3243653}.}\label{pic:env}
\end{figure}

TORCS~\footnote{\url{http://torcs.sourceforge.net/}} (The Open Racing Car Simulator)~\cite{TORCS} is an open source 3D car racing simulator that started as a game for \emph{Linux} in the late 1990s but became increasingly popular as a virtual simulation platform for driving agents and intelligent control systems for various car components. TORCS provides a sufficiently involved simulation of racing physics, including accurate basic properties (mass, rotational inertia), mechanical details (suspension types etc.), friction profiles of tyres, and a realistic aerodynamic model, so it is widely accepted as useful as a source of synthetic data. TORCS has become the basis for the annual Simulated Car Racing Championship~\cite{DBLP:journals/corr/abs-1304-1672} and has been used in hundreds of works on autonomous driving and control systems (see Section~\ref{sec:robotics}).

CARLA (CAR Learning to Act)~\cite{Dosovitskiy17} is an open simulator for urban driving, developed as an open-source layer over \emph{Unreal Engine~4}~\cite{Karis:2013}. Technically, it operates similarly to~\cite{7989092}, as an open source layer over \emph{Unreal Engine~4} that provides sensors in the form of RGB cameras (with customizable positions), ground truth depth maps, ground truth semantic segmentation maps with 12 semantic classes designed for driving (road, lane marking, traffic sign, sidewalk and so on), bounding boxes for dynamic objects in the environment, and measurements of the agent itself (vehicle location and orientation). \emph{DeepDrive} \cite{craig_quiter_2018_1248998} is a simulator designed for training self-driving AI models, also developed as an \emph{Unreal Engine} plugin; it provides 8 RGB cameras with $512\times 512$ resolution at close to real time rates (20Hz), as well as a generated 8.2 hour video dataset.

VIVID (VIrtual environment for VIsual Deep learning), developed by Lai et al.~\cite{Lai:2018:VVE:3240508.3243653}, tackles a more ambitious problem: adding people interacting in various ways and a much wider variety of synthetic environments, they present a universal dataset and simulator of outdoor scenes such as outdoor shooting, forest fires, drones patrolling a warehouse, pedestrian detection on the roads, and more. VIVID is also based on the \emph{Unreal Engine} and uses the wide variety of assets available for it; for example, NPCs acting in the scenes are programmed using \emph{Blueprint}, an \emph{Unreal} scripting engine, and the human models are animated by the \emph{Unreal} animation editor. VIVID provides the ability to record video simulations and can communicate with deep learning libraries via the TCP/IP protocol, through the \emph{Microsoft Remote Procedure Call} (RPC) library originally developed for \emph{AirSim} (see Section~\ref{sec:robotics}).


\begin{figure}[!t]\centering
\setlength{\tabcolsep}{3pt}
\begin{tabular}{P{.49\linewidth}P{.49\linewidth}}
\myp{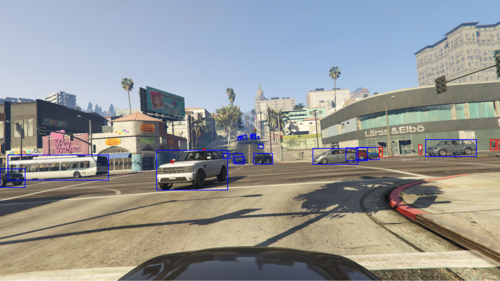} & \myp{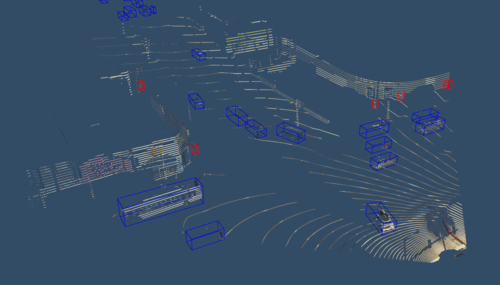} \\
(a) & (b) \\
\myp{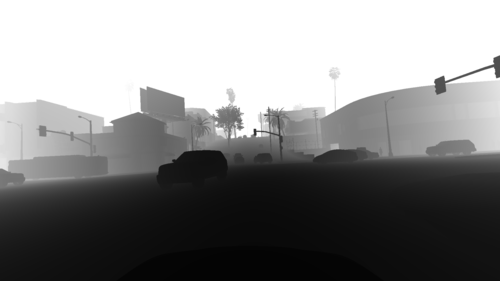} & \myp{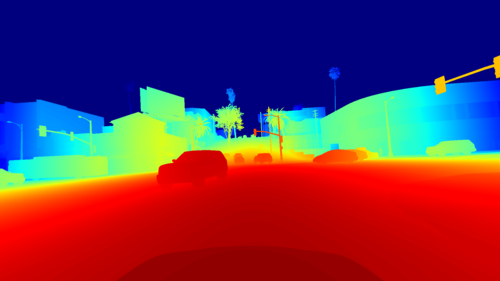} \\
(c) & (d) \\
\myp{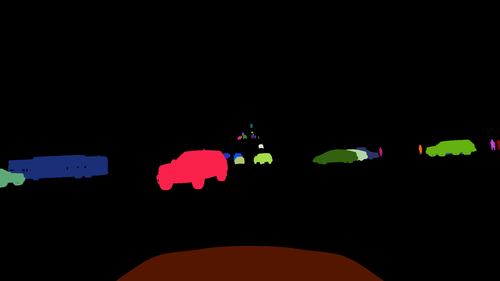} & \myp{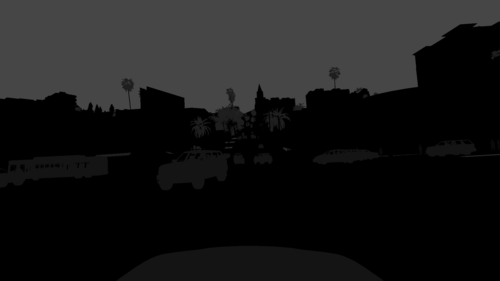} \\
(e) & (f) \\
\end{tabular}

\caption{Sample images from the \emph{PreSIL} dataset~\cite{Hurl2019PreciseSI}: (a) RGB image, (b) point cloud, (c) black-and-white depth map, (d) color depth map, (e) segmentation map, (f) stencil buffer.}\label{pic:presil}
\end{figure}

As for reinforcement learning (RL) in autonomous driving, the original paper on the CARLA simulator~\cite{Dosovitskiy17} also provides a comparison on synthetic data between conditional imitation learning, deep reinforcement learning, and a modular pipeline with separated perception, local planning, and continuous control, with limited success but generally best results obtained by the modular pipeline. Many works on autonomous driving use TORCS~\cite{TORCS} as a testbed, both in virtual autonomous driving competitions and simply as a well-established research platform. We do not aim to provide a full in-depth survey of the entire field and only note that despite its long history TORCS is being actively used for research purposes up to this day. In particular, Sallab et al.~\cite{Sallab2016EndtoEndDR,Sallab2017DeepRL} use it in their deep reinforcement learning frameworks for lane keeping assist and autonomous driving, Xiong et al.~\cite{DBLP:journals/corr/XiongWZL16} add safety-based control on top of deep RL, Wang et al.~\cite{DBLP:journals/corr/abs-1811-11329} train a deep RL agent for autonomous driving in TORCS, Barati et al.~\cite{DBLP:journals/corr/abs-1905-03985} use it to add multi-view inputs for deep RL agents, Li et al.~\cite{DBLP:journals/corr/abs-1810-12778} develop \emph{Visual TORCS}, a deep RL environment based on TORCS, Ando, Lubashevsky et al.~\cite{2015arXiv151104640A,2016arXiv160901812L} use TORCS to study the statistical properties of human driving, Glassner et al.~\cite{DBLP:journals/corr/abs-1901-00114} shift the emphasis to trajectory learning, Luo et al.~\cite{DBLP:journals/corr/abs-1811-06151} use TORCS as the main test environment for a new variation of the policy gradient algorithm, Liu et al.~\cite{DBLP:journals/corr/LiuSPVK17} make use of the multimodal sensors available in TORCS for end-to-end learning, Xu et al.~\cite{DBLP:journals/corr/abs-1801-05299} train a segmentation network and feed segmentation results to the RL agent in order to unify synthetic imagery from TORCS and real data, and so on. In an interesting recent work, Choi et al.~\cite{DBLP:journals/corr/abs-1809-01822} consider the driving experience transfer problem but consider a transfer not from a synthetic simulator to the real domain but from one simulator (TORCS) to another (GTA V). Tai et al.~\cite{8202134} learn continuous control for mapless navigation with asynchronous deep RL in virtual environments.

Synthetic data in autonomous driving extends to other sensor modalities as well. Thieling et al.~\cite{8603563} discuss the issues of physically realistic simulation for various robot sensors.  Yue et al.~\cite{Yue2018ALP} present a LIDAR point cloud generator based on the \emph{Grand Theft Auto V} engine, showing significant improvements in point cloud segmentation when augmenting the KITTI dataset with their synthetic data. Sanchez et al.~\cite{8722866} generate synthetic 3D point clouds with the robotic simulator \emph{Gazebo} (see Section~\ref{sec:robotics}). Wang et al.~\cite{8691584} develop a separate open source plugin for LIDAR point cloud generation. Fang et al.~\cite{Fang2018SimulatingLP} present an augmented LIDAR point cloud simulator that can generate simulated point clouds from real 3D scanner data, extending it with synthetic objects (additional cars). The work based on GTA V has recently been continued by Hurl et al.~\cite{Hurl2019PreciseSI} who have developed a precise LIDAR simulator within the GTA V engine and published the \emph{PreSIL} (Precise Synthetic Image and LIDAR) dataset with over 50000 frames with depth information, point clouds, semantic segmentation, and detailed annotations; we use \emph{PreSIL} to showcase on Fig.~\ref{pic:presil} the modalities available in modern synthetic datasets. There are also works on synthesizing specific elements of the environment, thus augmenting real data with synthetic elements; for example, Bruls et al.~\cite{Bruls2019GeneratingAT} generate synthetic road marking layouts which improves road marking segmentation, especially in corner cases.

Outdoor simulated environments go beyond driving, however, with simulators and synthetic datasets successfully used for autonomous aerial vehicles. Most of them are intended for unmanned aerial vehicles (UAVs) and have an emphasis on plugging in robotic controllers, possibly even with hardware-in-the-loop approaches; we discuss these simulators in Sections~\ref{sec:robotics} and~\ref{sec:cvnav}.

\subsection{Datasets and simulators of indoor scenes}\label{sec:dataindoor}

Although, as we have seen in the previous section, the main emphasis of many influential applications remains in the outdoors, \emph{indoor navigation} is also an important field where synthetic datasets are required. The main problems remain the same---SLAM and navigation---but the potential applications are now more in the field of home robotics, industrial robots, and embodied AI~\cite{Mautz11,zhou17}. There are large-scale efforts to create real annotated datasets of indoor scenes~\cite{dai2017scannet,7298655,10.1007/978-3-642-33715-4_54,6751312,DBLP:journals/corr/abs-1709-06158,DBLP:journals/corr/abs-1808-10654}, but synthetic data is increasingly being used in the field~\cite{Zhang2019ASO}. 

\begin{figure}[!t]\centering
\setlength{\tabcolsep}{3pt}
\begin{tabular}{P{.49\linewidth}P{.49\linewidth}}
\myp{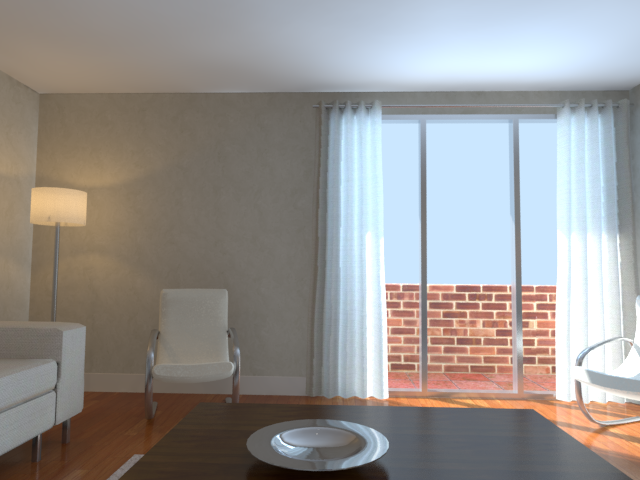} & \myp{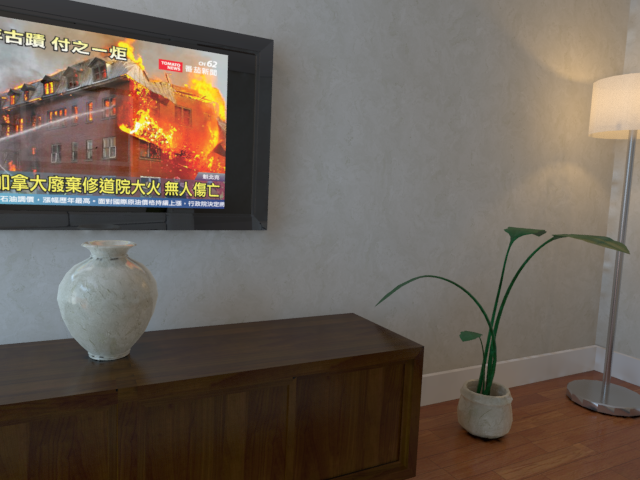} \\
(a) & (b) \\
\multicolumn{2}{c}{\myp{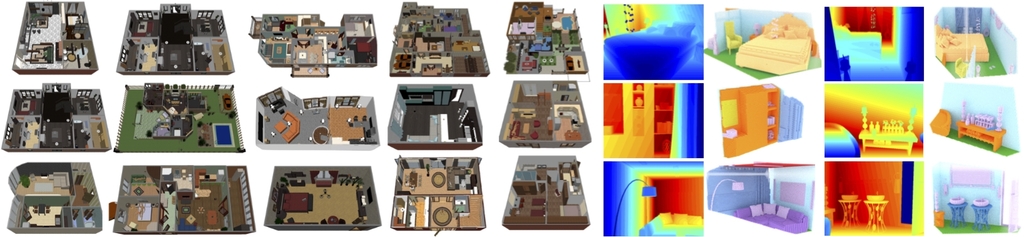}} \\
\multicolumn{2}{c}{(c)} \\
\myp{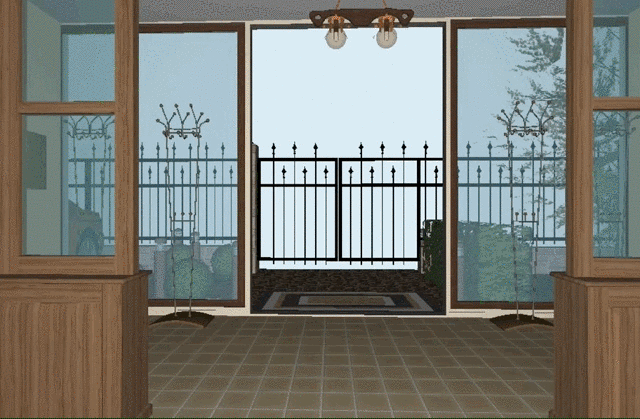} & \myp{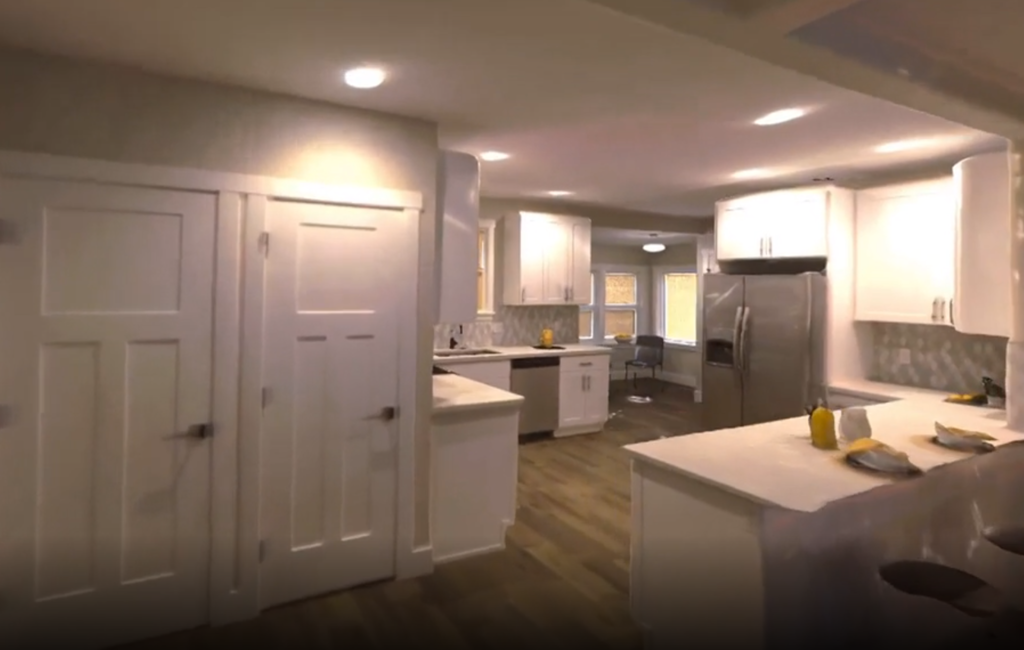} \\
(d) & (e) \\
\myp{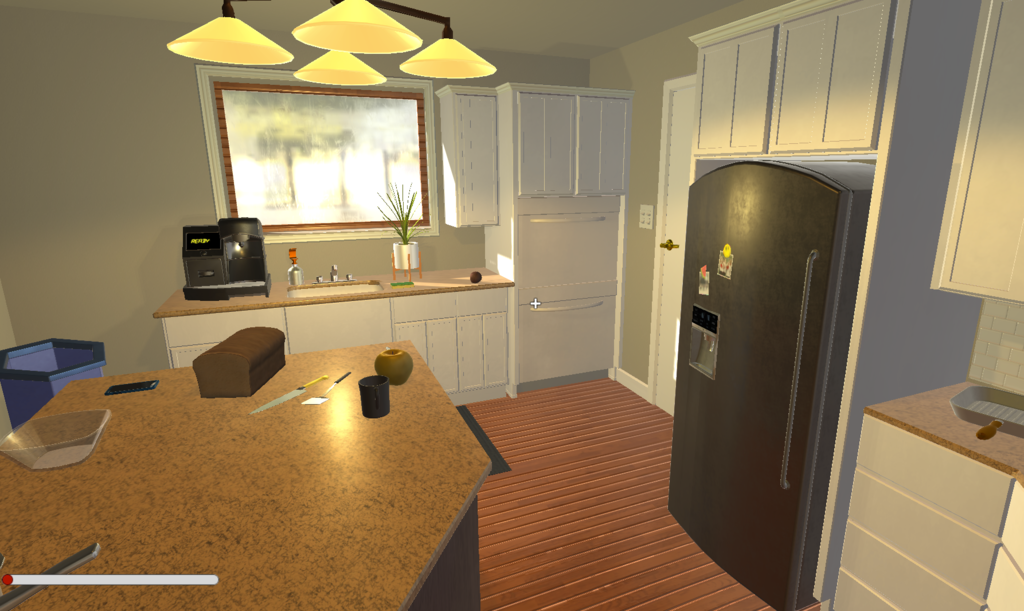} & \myp{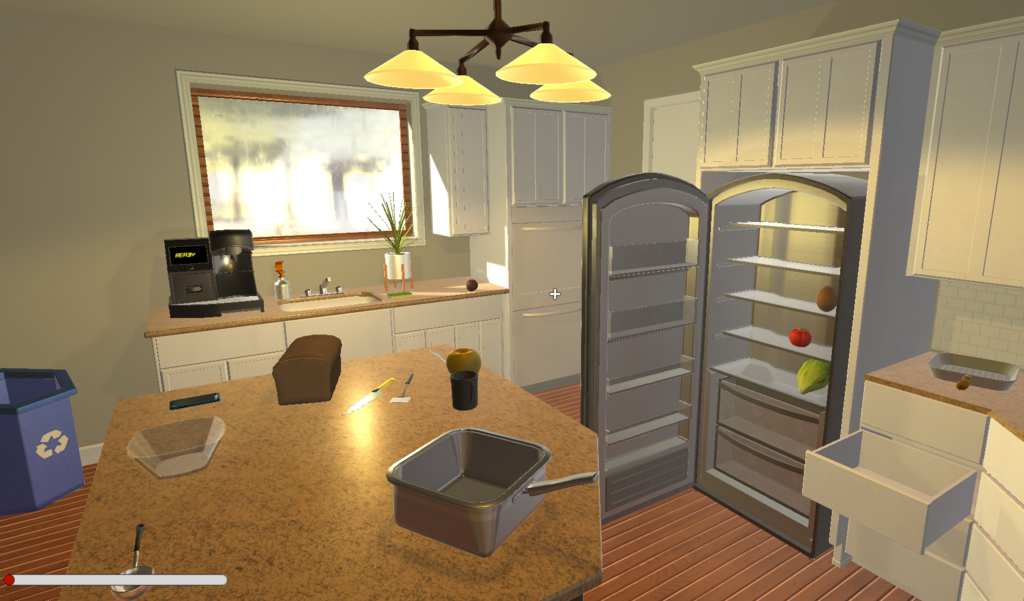} \\
(f) & (g) \\
\end{tabular}

\caption{Sample images from indoor datasets and simulation environments: (a-b) ICL-NUIM~\cite{6907054}; (c) SUNCG~\cite{song2016ssc}; (d) House3D~\cite{DBLP:journals/corr/abs-1801-02209}; (e) Habitat~\cite{habitat19arxiv}; (f-g) AI2THOR~\cite{DBLP:journals/corr/abs-1712-05474}.}\label{pic:indoor}
\end{figure}

Currently, the main synthetic dataset for indoor navigation is SUNCG\footnote{\url{http://suncg.cs.princeton.edu/}} presented by Song et al.~\cite{song2016ssc}. It contains over 45{,}000 different scenes (floors of private houses) with manually created realistic room layouts, 3D models of the furniture, realistic textures, and so on. All of the scenes are semantically annotated at the object level, and the dataset provides synthetic depth maps and volumetric ground truth data for the scenes. The original paper~\cite{song2016ssc} presented state of the art results in semantic scene completion, but, naturally, SUNCG has been used for many different tasks related to depth estimation, indoor navigation, SLAM, and others~\cite{Han2019DeepRL,Qi20173DGN,Chen20193DSS,Abbasi2018Deep3S,Ma2018LanguagedrivenSO}, and it often serves as the basis for scene understanding competitions~\cite{sumo2019}. 

Interestingly, at the time of writing this survey the SUNCG website was down and the dataset itself unavailable due to a legal controversy over the data\footnote{\url{https://futurism.com/tech-suing-facebook-princeton-data}}; that is why Fig.~\ref{pic:indoor}c shows a standard showcase picture from~\cite{song2016ssc} instead of data samples. While synthetic data can solve a lot of legal issues with real data (see Section~\ref{sec:privacy} for a discussion of privacy concerns, for example), the data, and especially handmade or manually collected 3D models, are still intellectual property and can bring about problems of its own unless properly released to the public domain.

SUNCG has given rise to a number of simulation environments. Before SUNCG, we note the \emph{Gazebo} platform mentioned above~\cite{1389727} (see also Section~\ref{sec:robotics}) and the V-REP robot simulation framework~\cite{6696520} that not only provided visual information but also simulated a number of actual robot types, further simplifying control deployment and development. Handa et al.~\cite{6907054} provide their own simulated living room environment and dataset ICL-NUIM (Imperial College London and National University of Ireland Maynooth) with special emphasis on visual odometry and SLAM; they render high-quality RGB-D images with ray tracing and take special care to model the noise in both depth and RGB channels.

MINOS by Savva et al.~\cite{savva2017minos} is a multimodal simulator for indoor navigation (later superceded by Habitat, see below). Wu et al.~\cite{DBLP:journals/corr/abs-1801-02209} made SUNCG into a full-scale simulation environment named House3D\footnote{\url{http://github.com/facebookresearch/House3D}}, with high-speed rendering suitable for large-scale reinforcement learning. In House3D, a virtual agent can freely explore 3D environments taken from SUNCG while providing all the modalities of SUNCG. House3D has been famously used for navigation control with natural language: Wu et al~\cite{DBLP:journals/corr/abs-1801-02209} presented \emph{RoomNav}, a task of navigation from natural language instructions and some models able to do it, while Das et al.~\cite{8575449} presented an embodied question answering model, where a robot is supposed to answer natural language questions by navigating an indoor environment. The AI2-THOR (The House Of inteRactions) framework~\cite{DBLP:journals/corr/abs-1712-05474} provides near photorealistic interactive environments with actionable objects (doors that can be opened, furniture that can be moved etc.) based on the \emph{Unity3D} game engine (see example of the same scene after some actions applied to objects on Fig.~\ref{pic:indoor}f-g).

Zhang et al.~\cite{DBLP:journals/corr/ZhangSYSLJF16} studied the importance of synthetic data realism for various indoor vision tasks. They fixed some problems with 3D models from SUNCG, improved their geometry and materials, sampled a diverse set of cameras for each scene, and compared OpenGL rendering against physically-based rendering (with Metropolis light transport models~\cite{Veach:1997:MLT:258734.258775} and the Mitsuba renderer\footnote{\url{http://www.mitsuba-renderer.org/}}) across a variety of lighting conditions. Their main conclusion is that, again, added realism is worth the effort: the quality gains are quite significant.

At the time of writing, the last (and very recent) major advance in the field is \emph{Habitat}, a simulation platform for embodied AI developed by Facebook researchers Savva et al.~\cite{habitat19arxiv}. Its simulator, called \emph{Habitat-Sim}, presents a number of important improvements over previous work that we have surveyed in this section:
\begin{itemize}	
	\item dataset support: \emph{Habitat-Sim} supports both synthetic datasets such as SUNCG~\cite{song2016ssc} and real-world datasets such as Matterport3D~\cite{DBLP:journals/corr/abs-1709-06158} and Gibson~\cite{DBLP:journals/corr/abs-1808-10654};
	\item rendering performance: \emph{Habitat-Sim} can render thousands of frames per second, 10-100x faster than previous simulators; the authors claim that ``it is often faster to generate images using \emph{Habitat-Sim} than to load images from disk''; this is important because simulation stops being a bottleneck in large-scale model training;
	\item humans-as-agents: humans can function as agents in the simulated environment, which allows to use real human behaviour in agent training and evaluation;
	\item accompanying library: the \emph{Habitat-API} library defines embodied AI tasks and implements metrics for easy agent development.
\end{itemize}
Savva et al. also provide a large-scale experimental study of various state of the art agents, arriving at the conclusion (counter to previous research) that reinforcement learning-based agents outperform SLAM-based ones, and RL agents generalize best across datasets, including the synthetic-to-real generalization from SUNCG to Matterport3D and Gibson. This is an important finding for synthetic data in indoor navigation, and we expect it to be confirmed in later studies; moreover, we expect \emph{Habitat} and its successors to become the new standard for indoor navigation and embodied AI research.


\subsection{Robotic simulators}\label{sec:robotics}

We have seen autonomous driving sims that mostly concentrate on accurately reflecting the outside world, modeling additional sensors such as LIDAR, and physics of the driving process. Simulators for indoor robots and unmanned aerial vehicles (UAV) add another complication: embedded hardware for such robots may be relatively weak and needs to be taken into account. Hence, robotic simulators usually support the \emph{Robot Operating System}\footnote{\url{https://www.ros.org/}} (ROS), a common framework for writing robot software. In some cases, simulators go as far as provide \emph{hardware-in-the-loop} capabilities, where a real hardware controller can be plugged into the simulator; for example, hardware-in-the-loop approaches to testing UAVs have been known for a long time and represent an important methodology in flight controller development~\cite{DBLP:journals/corr/abs-0804-3874,7849648,7352505}. We also refer to the surveys~\cite{8450505,MAIRAJ2019100}.

For a brief review, we highlight four works, starting with two standard references. \emph{Gazebo}, originally presented by Koenig and Howard~\cite{1389727} and now being developed by OSRF (Open Source Robotics Foundation), is probably the best-known robotic simulation platform. It supports ROS integration out of the box, has been used in the DARPA Robotics Challenge~\cite{Aguero-2015-VRC}, NASA Space Robotics Challenge, and others, and has been instrumental for thousands of research and industrial projects in robotics. \emph{Gazebo} uses a realistic physical engine (actually, several different engines) that supports illumination and lighting effects, gravity, inertia, and so on; it can be integrated with robotic hardware via ROS and provides realistic simulation that often leads to successful transfer to the real world.

\emph{MuJoCo} (Multi-Joint Dynamics with Contact) developed by Todorov~\cite{Todorov2012MuJoCoAP} is a physics engine specializing on contact-rich behaviours, which abound in robotics. Both \emph{Gazebo} and \emph{MuJoCo} have become industry standards for robotics research, and surveying the full range of their applications goes far beyond the scope of this work.
There are, of course, other platforms as well. For example, Gupta and Jarvis~\cite{10.1007/978-3-642-10331-5_22} present a simulation platform for training mobile robots based on the \emph{Half-Life 2} game engine.

\begin{figure}[!t]\centering
\setlength{\tabcolsep}{3pt}
\begin{tabular}{P{.49\linewidth}P{.49\linewidth}}
\myp{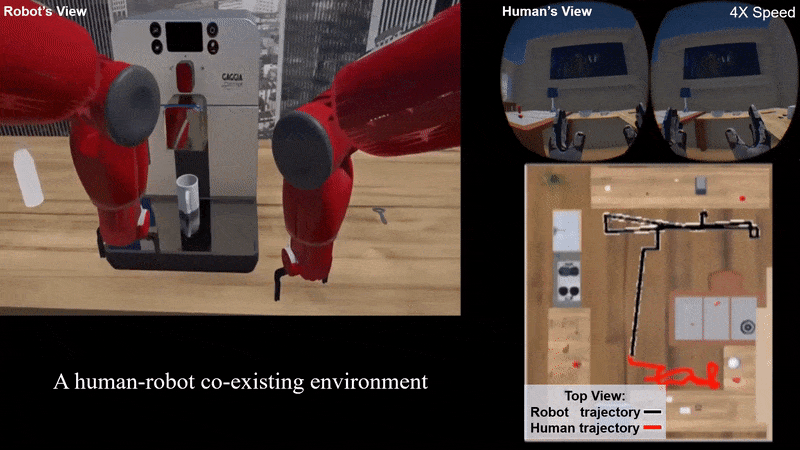} & \myp{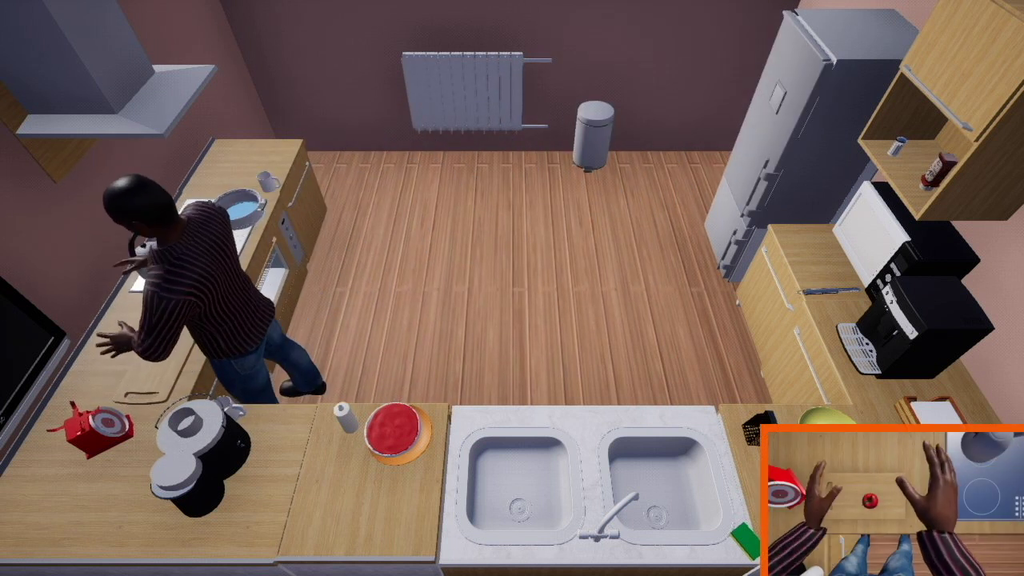} \\
(a) & (b) \\
\multicolumn{2}{c}{\myp{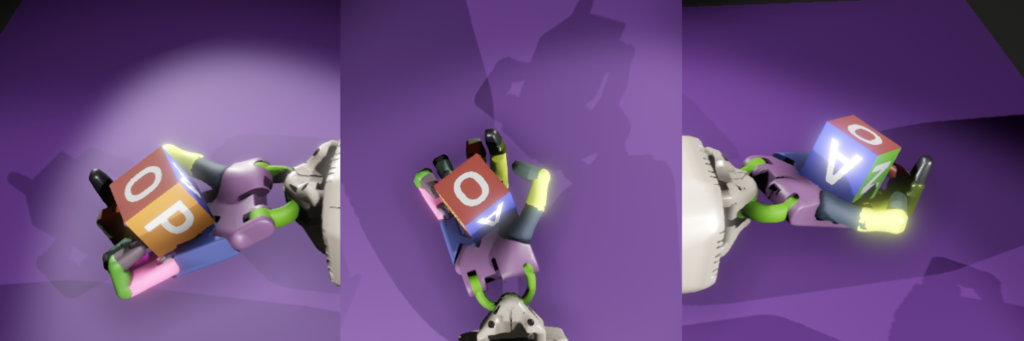}} \\
\multicolumn{2}{c}{(c)} \\
\end{tabular}

\caption{Sample images from robotic simulation environments: (a) \emph{VRGym}~\cite{Xie2019VRGymAV}; (b) \emph{VRKitchen}~\cite{DBLP:journals/corr/abs-1903-05757}; (c) ORRB~\cite{DBLP:journals/corr/abs-1906-11633}.}\label{pic:robots}
\end{figure}

The other two works are, on the contrary, very recent and may well define a new industry standard in the near future. First, Xie et al. present \emph{VRGym}~\cite{Xie2019VRGymAV}, a virtual reality testbed for physical and interactive AI agents. Their main difference from previous work is the support of human input via VR hardware integration. The rendering and physics engine are based on \emph{Unreal Engine~4}, additional multi-sensor hardware is capable of full body sensing and integration of human subjects to virtual environments, while a ROS bridge allows to easily communicate with robotic hardware. Xie et al. benchmark RL algorithms and show possibilities for socially aware models and intention prediction; \emph{VRGym} is already being further extended into, e.g., \emph{VRKitchen} by Gao et al.~\cite{DBLP:journals/corr/abs-1903-05757} designed to learn cooking tasks. 

The second work is the \emph{OpenAI Remote Rendering Backend} (ORRB) developed by OpenAI researchers Chociej et al.~\cite{DBLP:journals/corr/abs-1906-11633} able to render robotic environments and provide depth and segmentation maps in its renderings. ORRB emphasizes diversity, aiming to provide domain randomization effects (see Section~\ref{sec:domrand}); it is based on \emph{Unity3D} for rendering and \emph{MuJoCo} for physics, and it supports distributed cloud environments. ORRB has been used to train \emph{Dactyl}~\cite{openai2018learning}, a robotic hand for multi-finger small object manipulation developed by OpenAI; \emph{Dactyl}'s RL-based policies have been trained entirely in ORRB simulation and then have been successfully transferred to the physical robot. We expect more exciting developments in such synthetic-to-real transfer for robotic applications in the near future.

\subsection{Vision-based applications in unmanned aerial vehicles}\label{sec:cvnav}

A long line of research deals with vision-based approaches to operating unmanned aerial vehicles (UAV)~\cite{ALKAFF2018447}. Since in this case it is almost inevitable to use simulated environments for training, and often testing the model is also restricted to synthetic simulators (real world experiments are expensive outdoors and simply prohibited in urban environments), almost the entire field uses some kind of synthetic datasets or simulators. Classical vision-related problems for UAVs include three major tasks that UAVs often solve with computer vision:
\begin{itemize}
	\item localization and pose estimation, i.e., estimating the UAV position and orientation in both 2D (on the map) and in the 3D space; for this problem, real datasets collected by real world UAVs are available~\cite{10.1007/978-3-319-46484-8_33,Bonetto2015PrivacyIM,sensefly,Sturm2012ABF}, and the field is advanced enough to use actual field tests, so synthetic-only results are viewed with suspicion, but existing research still often employs synthetic simulators, either handmade for a specific problem~\cite{6696652} or based on professional flight simulators~\cite{DBLP:conf/fusion/AngelinoBC13};
	\item obstacle detection and avoidance, where real-world experiments are often too expensive even for testing~\cite{5152487};
	\item visual servoing, i.e., using feedback from visual sensors in order to maintain position, stability, and perform maneuvers; here synthetic data has usually been used in the form of hardware-in-the-loop simulators for the developed controllers~\cite{6224828}, sometimes augmented with full-scale flight simulators~\cite{Kurnaz:2007:ANI:1418707.1418711} or specially designed ``virtual reality'' environments~\cite{8460692}.
\end{itemize}

During the latest years, these classical applications of synthetic data for UAVs have been extended and taken to new heights with modern approaches to synthetic data generation. For example, in~\cite{10.1007/978-3-030-11012-3_4} the problem is to locate safe landing areas for UAVs, which requires depth estimation and segmentation into ``horizontal'', ``vertical'', and ``uncertain'' regions to distinguish horizontal areas that would be safe for landing. To train a convolutional architecture for this segmentation and depth estimation task, the authors propose an interesting approach to generating synthetic data: they begin with \emph{Google Earth} data\footnote{\url{https://www.google.com/earth/}} and extract 3D scenes from it. However, since 3D meshes in \emph{Google Earth} are far from perfect, the authors then map textures to the 3D scenes to obtain less realistic-looking images but ones for which the depth maps are known perfectly. The authors show that from a bird's eye view the resulting images look quite realistic, and compare different segmentation architectures on the resulting synthetic dataset. Oyuki Rojas-Perez et al.~\cite{Oyuki18} also compared the results obtained by training on synthetic and real datasets, with the results in favor of synthetic data due to the availability of depth maps for synthetic images. In a recent work, Castagno et al.~\cite{Castagno2019RealtimeRL} solve the landing site selection problem with a high-fidelity visual synthetic model of Manhattan rooftops, rendered with the \emph{Unreal Engine} and provided to the simulated robots via \emph{AirSim}.

\begin{figure}[!t]\centering
\setlength{\tabcolsep}{3pt}
\begin{tabular}{P{.49\linewidth}P{.49\linewidth}}
\multicolumn{2}{c}{\myp{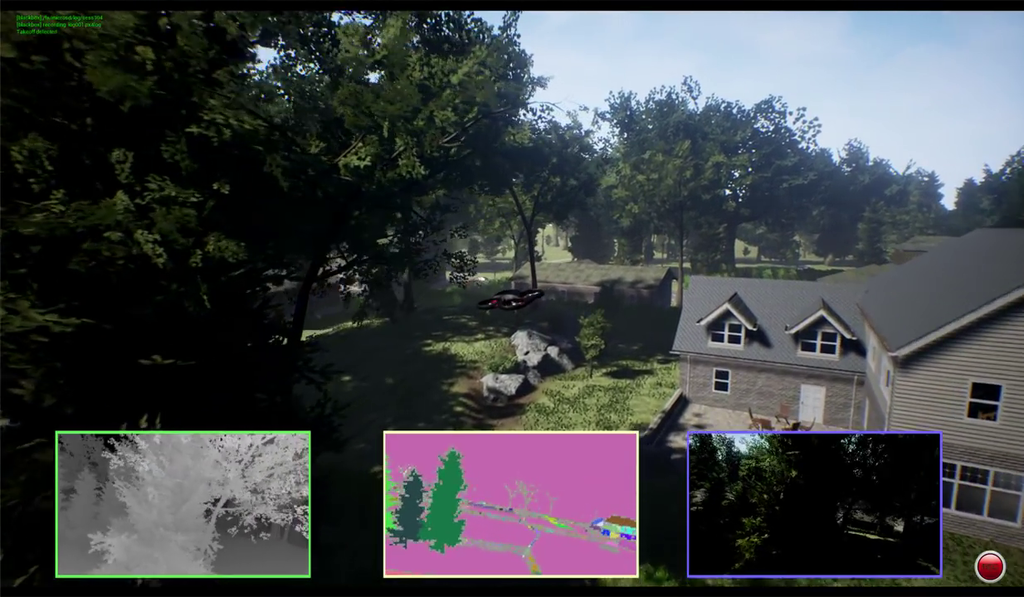}} \\
\multicolumn{2}{c}{(a)} \\
\myp{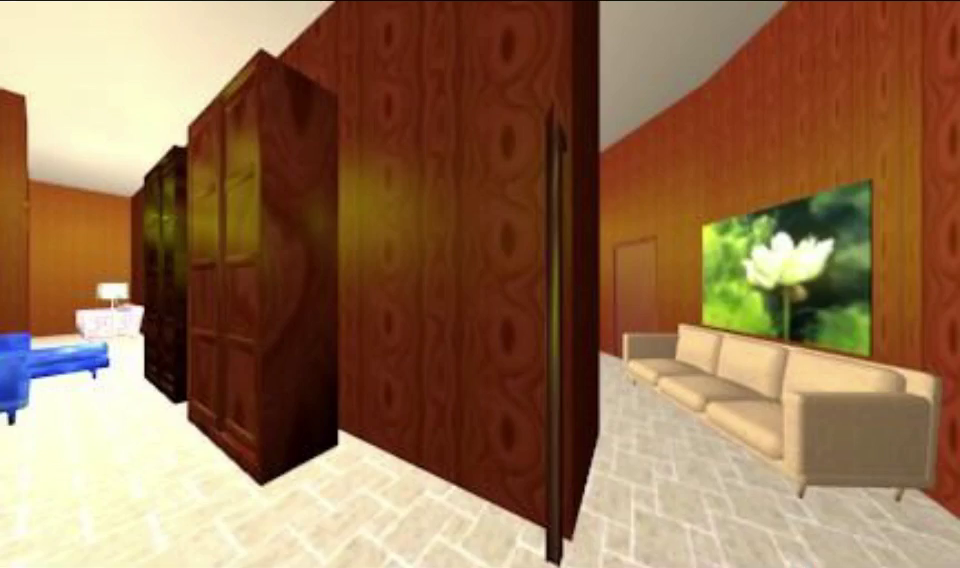} & \myp{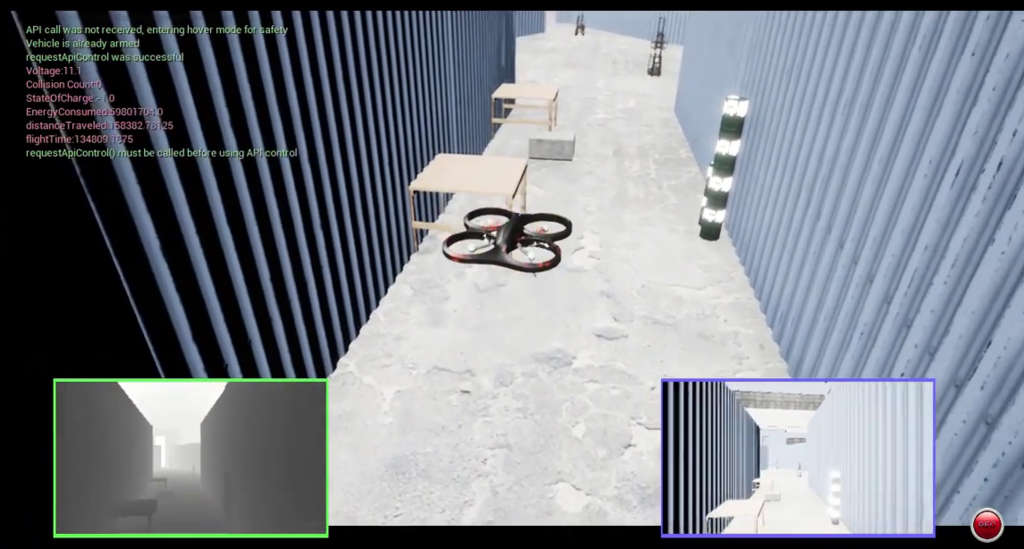} \\
(b) & (c) \\
\myp{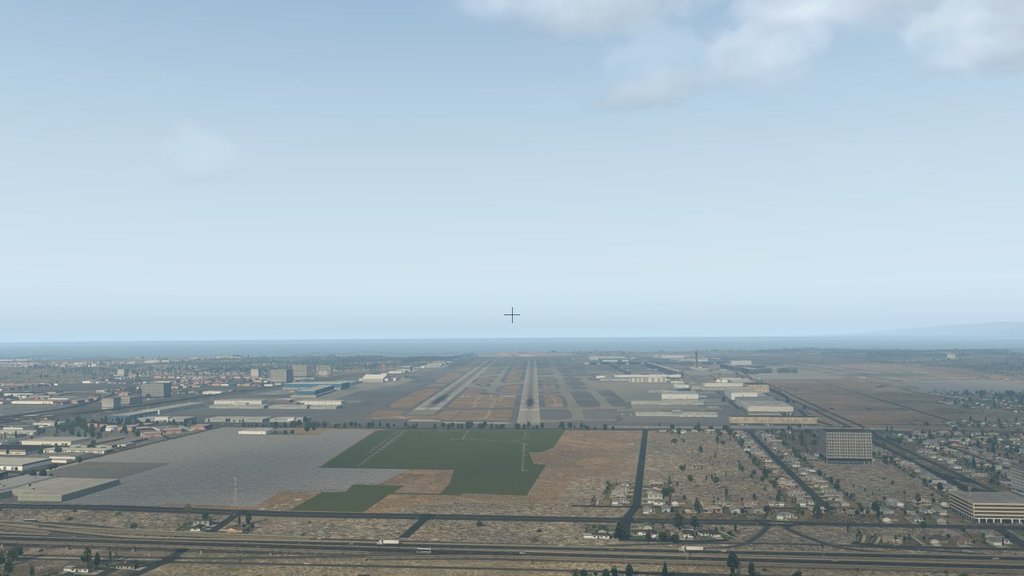} & \myp{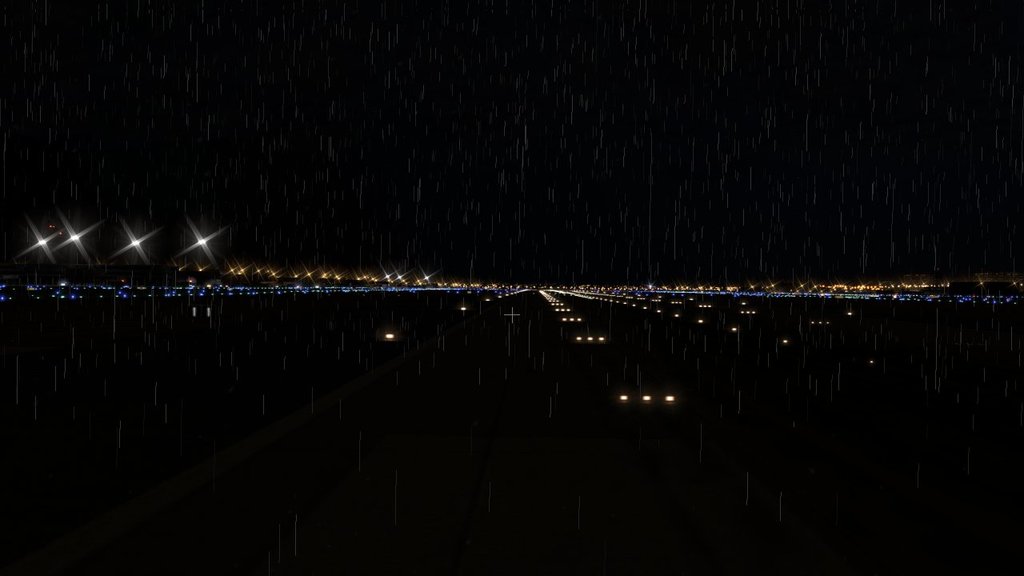} \\
(d) & (e) \\
\end{tabular}

\caption{Sample images from flight simulators: (a) \emph{AirSim}~\cite{airsim2017fsr} drone demo with depth, segmentation, and RGB drone view on the bottom; (b) $\mathrm{CAD}^2\mathrm{RL}$~\cite{Sadeghi2017CAD2RLRS}; (c) \emph{Air Learning}~\cite{DBLP:journals/corr/abs-1906-00421} (depth image and drone camera view on the bottom); (d-e) XPlane dataset~\cite{DBLP:journals/corr/abs-1811-11067}.}
\end{figure}

\emph{Gazebo} has been used for UAV simulations~\cite{Furrer2016}, but there are more popular specialized simulators in the field. \emph{AirSim}~\cite{airsim2017fsr} by \emph{Microsoft} is a flight simulator that operates more like a robotic simulator such as \emph{Gazebo} than an accurate flight sim such as \emph{Microsoft Flight Simulator} or \emph{X-Plane}: it contains a detailed physics engine designed to interact with specific flight controllers and providing a realistic physics-based vehicle model but also supports ROS to interact with the drones' software and/or hardware. Apart from visualizations rendered with \emph{Unreal Engine~4}, \emph{AirSim} provides sensor readings, including a barometer, gyroscope, accelerometer, magnetometer, and GPS. \emph{AirSim} is also available as a plugin for \emph{Unreal Engine~4} which enables extensions and new projects based on \emph{AirSim} simulations of autonomous vehicles. There are plenty of extensions and projects that use \emph{AirSim} and provide interesting synthetic simulated environments (see also the survey~\cite{MAIRAJ2019100}), in particular:
\begin{itemize}
	\item Chen et al.~\cite{Chen2018LearningTU} add realistic forests to use UAVs for forest visual perception;
	\item Bondi et al.~\cite{Bondi2018AirSimWAS} concentrate on wildlife preservation, extending the engine with realistic renderings of, e.g., the African savanna;
	\item in two different projects, Smyth et al.~\cite{8516527,Smyth2018AVE} and Ullah et al.~\cite{10.1007/978-3-030-13453-2_18} simulate critical incidents (such as chemical, biological, or nuclear incidents or attacks) with an explicit goal of training autonomous drones to collect data in these virtual environments;
	\item Huang et al.~\cite{flightcamera} extend \emph{AirSim} with a natural language understanding model to simulate natural language commands for drones, and so on.
\end{itemize}
We also note that \emph{Microsoft} itself recently extended \emph{AirSim} to include autonomous car simulations, making the engine applicable to all the tasks we discussed in Section~\ref{sec:dataoutdoor}.

While \emph{AirSim} and \emph{Gazebo} have been the most successful simulation frameworks, we also note other frameworks for multi-agent simulation that have been used for autonomous vehicles, usually without a realistic 3D rendered environment (see also a comparison of the frameworks in~\cite{MUALLA2018791} and of their perception systems in~\cite{s19030648}): JaSIM~\cite{Galland2009EnvironmentMF}, a multiagent 3D environment model based on the \emph{Janus} platform~\cite{10.1007/978-3-642-03278-3_7}, \emph{Repast Simphony}~\cite{North2013}, an agent-based simulation library for complex adaptive systems, FLAME~\cite{Kiran:2010:FSL:1838206.1838517} that concentrates on parallel simulations, and JADE~\cite{Bellifemine:2007:DMS:1197665}, a popular library for multi-agent simulations.

Sadeghi and Levine~\cite{Sadeghi2017CAD2RLRS} present the $\mathrm{CAD}^2\mathrm{RL}$ framework for avoiding collisions while flying in indoor environments. They motivate the use of synthetic data by the domain randomization idea (see Section~\ref{sec:domrand}) and produce a wide variety of indoor simulated environments constructed in \emph{Blender}. They do not use any real images during training, learning the Q-function for the reinforcement learning agent entirely on simulated data, with a fully convolutional neural network. The authors report improvements in a number of complex settings such as flying around corners, navigating through narrow corridors, flying up a staircase, avoiding dynamic obstacles, and so on.

The \emph{Air Learning} platform recently presented by Krishnan et al.~\cite{DBLP:journals/corr/abs-1906-00421} is an end-to-end simulation environment for autonomous aerial robots that combines pluggable environment and physics engine (such as \emph{AirSim} or \emph{Gazebo}), learning algorithms, policies for robot control (implemented in, e.g., \emph{TensorFlow} or \emph{PyTorch}), and ``hardware-in-the-loop'' controllers, where a real flight controller can be plugged in and evaluated on the \emph{Air Learning} platform. As an example, Krishnan et al. benchmark several reinforcement learning approaches to point-to-point obstacle avoidance tasks and arrive at important conclusions: for instance, it turns out that having more onboard compute (a desktop CPU vs. a \emph{Rapsberry Pi}) can significantly improve the result, producing almost 2x shorter trajectories.

Among vision-based synthetic datasets, we note the work by Solovev et al.~\cite{DBLP:journals/corr/abs-1811-11067} who present a synthetic dataset for airplane landing based on the XPlane flight simulator~\cite{xplane}, which has been used for other UAV simulations as well~\cite{Bittar2014,Garcia2009}. The main intention of Solovev et al. is to present a benchmark for representation learning by combining different modalities (RGB images and various sensors), but the resulting synthetic dataset is sufficiently large to be used for other applications: 93GB of images and sensor readings from 8K landings on 114 different runways.

Another interesting application where real and synthetic data come together is provided by Madaan et al.~\cite{8206190}, who solve a truly life-or-death problem for UAVs: wire detection. They use real background images and superimpose them with renderings of realistic 3D models of wires. The authors vary different properties of the wires (material, wire sag, camera angle etc.) but do not make any attempts to adapt the wires to the semantics of the background image itself, simply pasting wires onto the images. Nevertheless, Madaan et al. report good results with training first on synthetic data and then fine-tuning on a small real dataset; synthetic pretraining proves to be helpful.

\subsection{Computer games as virtual environments}\label{sec:games}

Computer games and game engines have been a very important source of problems and virtual environments for deep RL and AI in general~\cite{Capo2018StateOT,DBLP:journals/corr/abs-1109-1314,8632747}. Much of the foundational work in modern deep RL has been done in the environments of 2D arcade games, usually classical \emph{Atari} games~\cite{Bellemare:2013:ALE:2566972.2566979,mnih2015humanlevel}. 
First, many new ideas for RL architectures or training procedures for robotic navigation were introduced as direct applications of deep RL to navigation in complex synthetic environments, in particular to navigating mazes in video games such as \emph{Doom}~\cite{DBLP:journals/corr/BhattiDMNST16} and \emph{Minecraft}~\cite{Oh:2016:CMA:3045390.3045684} or specially constructed 3D mazes, e.g., from the \emph{DeepMind Lab} environment~\cite{DBLP:journals/corr/BeattieLTWWKLGV16,8460646}. \emph{Doom} became something of an industry standard, with the \emph{VizDoom} framework developed by Kempka et al.~\cite{DBLP:journals/corr/KempkaWRTJ16} and later used in many important works~\cite{Alvernaz2017AutoencoderaugmentedNF,Ratcliffe2017ClydeAD,Lample:2017:PFG:3298483.3298548,Wu2017TrainingAF}.

Games of other genres, in particular real-time strategy games~\cite{8490409,Tang2018ARO}, also represent a rich source of synthetic environments for deep RL; we note the \emph{TorchCraft} library for machine learning research in \emph{StarCraft}~\cite{DBLP:journals/corr/SynnaeveNACLLRU16}, synthetic datasets extracted from \emph{StarCraft}~\cite{DBLP:journals/corr/abs-1708-02139,DBLP:journals/corr/abs-1712-10179,Wu2017MSCAD}, the \emph{ELF} (Extensive, Lightweight, and Flexible) library platform with three simplified real-time strategy games~\cite{DBLP:journals/corr/TianGSWZ17}, and the SC2LE (StarCraft II Learning Environment) library for \emph{Starcraft II}~\cite{DBLP:journals/corr/abs-1708-04782}. Racing games have been used as environments for training end-to-end RL driving agents~\cite{8014798}. While data from computer games could technically be considered synthetic data, we do not go into further details on game-related research and concentrate on virtual environments specially designed for machine learning and/or transfer to real world tasks. Note, however, that synthetic data is already making inroads even into learning to play computer games: Justesen et al.~\cite{DBLP:journals/corr/abs-1806-10729} show that using procedurally generated levels improves generalization and final results for \emph{Atari} games and can even produce models that work well when evaluated on a completely new level every time they play.

All of the above does not look too much in line with the general topic of synthetic data: while game environments are certainly ``synthetic'', there is usually no goal to transfer, say, an RL agent playing \emph{StarCraft} to a real armed conflict (thankfully). However, recent works suggest that there is potential in this direction. For example, while navigation in a first-person shooter is very different from real robotic navigation, successful attempts at transfer learning from computer games to the real world are already starting to appear. Karttunen et al.~\cite{Karttunen2019FromVG} present an RL agent for navigation trained on the \emph{Doom} environment and transferred to a real life Turtlebot by freezing most of the weights in the DQN network and only fine-tuning a small subset of them. As computer games get more realistic, we expect such transfer to become easier.



%% file: other.tex
\section{Synthetic data outside computer vision}\label{sec:other}

While computer vision remains the main focus of synthetic data applications, other fields also begin to use synthetic datasets, with some directions entirely dependent on synthetic data. In this section, we survey some of these fields. Specifically, in Section~\ref{sec:neuroprog} we consider neural programming, in Section~\ref{sec:bio} discuss synthetic data generation and use in bioinformatics, and Section~\ref{sec:nlp} reviews the (admittedly limited) applications of synthetic data in natural language processing.

\subsection{Synthetic data for neural programming}\label{sec:neuroprog}

One interesting domain where synthetic data is paramount is neural program synthesis and neural program induction. The basic idea of teaching a machine learning model to program can be broadly divided into two subfields: program induction aims to train an end-to-end differentiable model to capture an algorithm~\cite{NIPS2017_6803}, while program synthesis tries to teach a model the semantics of a domain-specific language (DSL), so that the model is able to generate programs according to given specifications~\cite{DBLP:journals/corr/abs-1802-02353}. Basically, in program induction the network \emph{is} the program, while in program synthesis the network \emph{writes} the program. Naturally, both tasks require large datasets of programs together with their input-output pairs; since no such large datasets exist, and generating synthetic programs and running them in this case is relatively easy (arguably even easier than generating synthetic data for computer vision), all modern works use synthetic data to train the ``neural computers''.

In program induction, the tasks are so far relatively simple, and synthetic data generation does not present too many difficulties. For example, Joulin and Mikolov~\cite{Joulin:2015:IAP:2969239.2969261} present a new architecture (stack-augmented recurrent networks) to learn regularities in algorithmically generated sequences of symbols; the training data, as in previous such works~\cite{10.1007/3540634932_25,Hochreiter:1997:LSM:1246443.1246450,Wiles1995LearningTC,963769}, is synthetically generated by hand-crafted simple algorithms, including generating sequences from a pattern such as $a^nb^{2n}$ or $a^nb^mc^{n+m}$, binary addition (supervised problem asking to continue a string such as $110+10=$), and similar.  Zaremba and Sutskever~\cite{DBLP:journals/corr/ZarembaS14} train a model to execute programs, i.e., map their textual representations to outputs; they generate training data as Python-style programs with addition, subtraction, multiplication, variable assignments, if-statements, and for-loops (but without nested loops), each ending with a \texttt{print} statement; see Fig.~\ref{fig:prog} for an illustration. Neural RAM machines~\cite{45398} are trained on a number of simple tasks (access, increment, copy etc.) whose specific instances are randomly synthesized. The same goes for neural Turing machines~\cite{DBLP:journals/corr/GravesWD14} and neural GPUs~\cite{Kaiser2016NeuralGL,Freivalds2017ImprovingTN}: they are trained and evaluated on synthetic examples generated for simple basic problems such as copying, sorting, or arithmetic operations.

In neural program synthesis, again, the programs are usually simple and also generated automatically; attempts to collect natural datasets for program synthesis have begun only very recently~\cite{DBLP:journals/corr/abs-1807-03168}. Learning-based program synthesis (earlier attempts were based on logical inference~\cite{Manna:1971:TAP:362566.362568}) began with learning string manipulation programs~\cite{Gulwani:2011:ASP:1925844.1926423} and soon branched into deep learning, using recurrent architectures to guide search-based methods~\cite{DBLP:journals/corr/abs-1804-01186} or to generate programs in encoder-decoder architectures~\cite{DBLP:journals/corr/DevlinUBSMK17,chen2019}. In~\cite{DBLP:journals/corr/abs-1805-04276,si2019}, reinforcement learning is added on top of a recurrent architecture in order to alleviate the \emph{program aliasing} problem, i.e., the fact that many different equivalent programs provide equally correct answers while the training set contains only one. All of the above-mentioned models were trained on synthetic datasets of randomly generated programs (usually in the form of abstract syntax trees) run on randomized sets of inputs.

As a separate thread, we mention works on program synthesis for visual reasoning and, generally speaking, question answering~\cite{Johnson2017,Santoro2017ASN,DBLP:journals/corr/HuARDS17}. To do visual question answering~\cite{Agrawal2017}, models are trained to compose a short program based on the natural language question that will lead to the answer, usually in the form of an execution graph or a network architecture. This line of work is based on neural module networks~\cite{DBLP:journals/corr/AndreasRDK15,DBLP:journals/corr/abs-1809-08697} and similar constructions~\cite{DBLP:journals/corr/HuRADS16,DBLP:journals/corr/AndreasRDK16}, where the network learns to create a composition of modules (in QA, based on parsing the question) that are also neural networks, all learned jointly (see, however, the critique in~\cite{SEJNOVA2018481}). Latest works use architectures based on self-attention that have proven their worth across a wide variety of NLP tasks~\cite{lewis2018generative}. Naturally, most of these works use the CLEVR synthetic dataset for evaluation (see Section~\ref{sec:datareason}).

Generation of synthetic data itself has not been given much attention in this field until very recently, but the interest is rising. The work by Shin et al.~\cite{shin2018synthetic} presents a study of contemporary synthetic data generation techniques for neural program synthesis and concludes that the resulting models do not capture the full semantics of the language, even if they do well on the test set. Shin et al. show that synthetic data generation algorithms, including the one from the popular \emph{tensor2tensor} library~\cite{tensor2tensor}, have biases that fail to cover important parts of the program space and deteriorate the final result. To fix this problem, they propose a novel methodology for generating distributions over the space of datasets for program induction and synthesis, showing significant improvements for two important domains: \emph{Calculator} (which computes the results of arithmetic expressions) and \emph{Karel} (which achieves a given objective with a virtual robot moving on a two-dimensional grid with walls and markers). We expect more research into synthetic data generation for neural program induction and synthesis to follow in the near future.

\begin{figure}[!t]\centering
\begin{minipage}[b]{.36\linewidth}
\textbf{Input}:
\begin{verbatim}
j=8584
for x in range(8):
    j+=920
b=(1500+j)
print((b+7567))
\end{verbatim}
\textbf{Target}: \verb$25011$.
\end{minipage}
~
\begin{minipage}[b]{.6\linewidth}
\textbf{Input}:
\begin{verbatim}
i=8827
c=(i-5347)
print((c+8704) if 2641<8500 else 5308)
\end{verbatim}
\textbf{Target}: \verb$12184$.
\end{minipage}

\caption{Sample synthetic programs from~\cite{DBLP:journals/corr/ZarembaS14}.}\label{fig:prog}
\end{figure}

\subsection{Synthetic data in bioinformatics}\label{sec:bio}

We use examples from the heathcare and biomedical domain throughout this survey; see, e.g., Sections~\ref{sec:medical} and~\ref{sec:finance}. In this section, we concentrate on applications of synthetic data in bioinformatics that fall outside either producing synthetic medical images (usually through GANs, see Section~\ref{sec:medical}) or providing privacy guarantees for sensitive data through synthetic datasets (Section~\ref{sec:finance}). It turns out that there are still plenty, and synthetic data is routinely used and generated throughout bioinformatics; see also a survey in~\cite{doi:10.1098/rsif.2017.0387}.

For many of these methods, generated synthetic data is the end goal rather than a tool to improve machine learning models. In particular, \emph{de novo} drug design~\cite{doi:10.1021/acs.jmedchem.5b01849,doi:10.1002/wcms.49} is a field that searches for molecules with desirable properties in a search space of about $10^{60}$ synthesizable molecules~\cite{ChEMBL,doi:10.1002/wcms.1104}, and the goal is to find (which in a space of this size rather means to \emph{generate}) candidate molecules that would later have to be explored further in lab studies and then clinical trials. First modern attempts at \emph{de novo} drug design used rule-based methods that simulate chemical reactions~\cite{doi:10.1021/acs.jmedchem.5b01849}, but the field soon turned to generative models, in particular based on deep learning~\cite{Gawehn15,CHEN20181241}. In this context, molecules are usually represented in the SMILES format~\cite{doi:10.1021/ci00057a005} that encodes molecular graphs as strings in a certain formal grammar, which makes it possible to use sequence learning models to generate new SMILES strings. Segler et al.~\cite{DBLP:journals/corr/SeglerKTW17} used LSTM-based RNNs to learn a chemical language model, while G{\'o}mez-Bombarelli et al.~\cite{DBLP:journals/corr/Gomez-Bombarelli16} trained a variational autoencoder (VAE) which is already a generative model capable of generating new candidate molecules. To further improve generation, Kusner et al.~\cite{2017arXiv170301925K} developed a novel extension of VAEs called Grammar Variational Autoencoders that can take into account the rules of the SMILES formal grammar and make VAE outputs conform to the grammar. To then use the models to obtain molecules with desired properties, researchers used either a small set of labeled positive examples~\cite{DBLP:journals/corr/SeglerKTW17} or augment the RNN training procedure with reinforcement learning~\cite{DBLP:journals/corr/JaquesGTE16}. In particular, Olivecrona et al.~\cite{Olivecrona2017} use a recurrent neural network trained to generate SMILES representations: first, a prior network (3 layers of 1024 GRU) is trained in a supervised way on the RDKit subset~\cite{Landrum2016RDKit2016_09_4} of ChEMBL database~\cite{ChEMBL}, then an RL agent (with the same structure, initialized from the prior network) is fine-tuned with the REINFORCE algorithm to improve the resulting SMILES encoding. A similar architecture, but with a stack-augmented RNN~\cite{DBLP:journals/corr/JoulinM15} as the basis, which enables more long-term dependencies, was presented by Popova et al.~\cite{Popovaeaap7885}. We note a series of works by \emph{Insilico} researchers Kadurin, Polykovskiy and others who applied different generative models to this problem:
\begin{itemize}
	\item the work~\cite{Kadurin16} trains a supervised adversarial autoencoder~\cite{44904} with the condition (in this case, growth inhibition percentage for tumor cells after treatment) added as a separate neuron to the latent layer;
	\item in~\cite{KNKAZ17}, Kadurin et al. compared adversarial autoencoders (AAE) with variational autoencoders (VAE) for the same problem, with new modifications to the architecture that result in improved generation;
	\item in~\cite{doi:10.1021/acs.molpharmaceut.8b00839}, Polykovskiy et al. introduced a new AAE modification, entangled conditional adversarial autoencoder, to ensure the disentanglement of latent features; in this case, which is still quite rare for deep learning in drug discovery, a newly discovered molecule (a new inhibitor of Janus kinase 3) was actually tested in the lab and showed good activity and selectivity \emph{in vitro};
	\item Kuzmynikh et al.~\cite{KPKZBNSZ18} presented a novel 3D molecular representation based on the wave transform that led to improved performance for CNN-based autoencoders and improved MACCS fingerprint prediction;
	\item Polykovskiy et al.~\cite{PZSGTBKAAVKNAZ18} presented MOSES (Molecular Sets), a benchmarking platform for molecular generation models, which implemented and compared various generative models for molecular generation including CharRNN~\cite{doi:10.1021/acscentsci.7b00512}, VAE, AAE, and Junction Tree VAE~\cite{2018arXiv180204364J}, together with a variety of evaluation metrics for generation results.
\end{itemize}
The above works can be thought of as generation of synthetic data (e.g., molecular structures) that could be of direct use for practical applications.

Johnson et al.~\cite{2017arXiv170500092J} undertake an ambitious project: they learn a generative model of the variation in cell and nuclear morphology based on fluorescence microscopy images. Their model is based on two adversarial autoencoders~\cite{44904}, one learning a probabilistic model of cell and nuclear shape and the other learning the interrelations between subcellular structures conditional on an encoding of the cell and nuclear shape from the first autoencoder. The resulting model produces plausible synthetic images of the cell with known localizations of subcellular structures.

One interesting variation of ``synthetic data'' in bioinformatics concerns learning from live experiments on synthetically generated biological material. For example, Rosenberg et al.~\cite{ROSENBERG2015698} study alternative RNA splicing, in particular the functional effects of genetic variation on the molecular phenotypes through alternative splicing. To do that, they create a large-scale gene library with more than two million randomly generated synthetic DNA sequences, then used massively parallel reporter assays (MPRA) to measure the isoform ratio for all mini-genes in the experiment, and then used it to learn a (simple linear) machine learning model for alternative splicing. It turned out that this approach significantly improved prediction quality, outperforming state of the art deep learning models for alternative splicing trained on the actual human genome~\cite{Xiong1254806} in predicting the results of \emph{in vivo} experiments. There is also a related field of imitational modeling for bioinformatics data that often results in realistic synthetic generators; e.g., Van den Bulcke et al.~\cite{VandenBulcke2006} provide a generator for synthetic gene expression data, able to produce synthetic transcriptional regulatory networks and simulated gene expression data while closely matching real statistics of biological networks.

\subsection{Synthetic data in natural language processing}\label{sec:nlp}

Synthetic data has not been widely used in natural language processing (NLP). In our opinion, there is a conceptual reason for this. Compare with computer vision: there, the process of synthetic data generation can be done separately from learning the models, and the essence of what the models are learning is, in a way, ``orthogonal'' to the difference between real and synthetic data. If I show you a cartoon ad featuring a new bottle of soda, you will be able to find it in a supermarket even though you would never confuse the cartoon with a real photo. In natural language processing, on the other hand, text generation is the hard problem itself. The problem of generating meaningful synthetic text with predefined target variables such as topic or sentiment is the subject of many studies in NLP, we are still a long way to go before it is solved, and it is quite probable that when text generation finally reaches near-human levels, discriminative models for the target variables will easily follow from it, rendering synthetic data useless.

Nevertheless, there have been works that use data augmentation for NLP in a fashion that borders on using synthetic data. There have been simple augmentation approaches such as to simply drop out certain words~\cite{W16-2323}. A development of this idea shown in~\cite{D18-1100} switches out certain words, replacing them 
with random words from the vocabulary. The work~\cite{Xie2017DataNA} develops methods of data noising for language models, adding noise to word counts in a way reminiscent of smoothing in language models based on $n$-grams.

A more developed approach is to do data augmentation with synonyms: to expand a dataset, one can replace words with their synonyms, getting ``synthetic sentences'' that can still preserve target variables such as the topic of the text, its sentiment, and so on. The work~\cite{NIPS2015_5782} used this method directly to train a character-level network for text classification, while~\cite{GAN16} tested augmentation with synonyms for morphology-rich languages such as Russian. In~\cite{P17-2090}, augmentation with synonyms was used for low-resource machine translation, with an auxiliary LSTM-based language model used to recognize whether the synonym substitution is correct. The work~\cite{Wang2015ThatsSA}, which concentrated on studying tweets, proposed to use embedding-based data augmentation, using neighboring words in the word vector space as synonyms. The work~\cite{N18-2072} extends augmentation with synonyms by replacing words in sentences with other words in paradigmatic relations with the original words, as predicted by a bi-directional language model at the word positions. 

Techniques for generating synthetic text are constantly evolving. First, modern language models based on multi-head self-attention from the Transformer family, starting from the Transformer itself~\cite{NIPS2017_7181} and then further developed by BERT~\cite{DBLP:conf/naacl/DevlinCLT19}, OpenAI GPT~\cite{radford2018improving}, Transformer-XL~\cite{dai-etal-2019-transformer}, OpenAI GPT-2~\cite{openaigpt2}, and GROVER~\cite{DBLP:journals/corr/abs-1905-12616}, generate increasingly coherent text. Actually, Zellers et al.~\cite{DBLP:journals/corr/abs-1905-12616} claim that their GROVER model for conditional generation (e.g., generating the text of a news article given its title, domain, and author) outperforms human-generated text in the ``fake news''/``propaganda'' category in terms of style and content (evaluated by humans).

Moreover, recently developed models allow to generate text with GANs. This is a challenging problem because unlike, say, images, text is discrete and hence the generator output is not differentiable. There are several approaches to solving this problem:
\begin{itemize}
	\item training with the REINFORCE algorithm and other techniques from reinforcement learning that are able to handle discrete outputs; this path has been taken in the pioneering model named SeqGAN~\cite{Yu:2017:SSG:3298483.3298649}, LeakGAN for generating long text fragments~\cite{DBLP:journals/corr/abs-1709-08624}, and MaskGAN that learns to fill in missing text with an actor-critic conditional GAN~\cite{fedus2018maskgan}, among others;
	\item approximating discrete sampling with a continuous function; these approaches include the Gumbel Softmax trick~\cite{2016arXiv161104051K}, TextGAN that approximates the $\arg\max$ function~\cite{2017arXiv170603850Z}, TextKD-GAN that uses an autoencoder to smooth the one-hot representation into a softmax output~\cite{DBLP:journals/corr/abs-1905-01976}, and more;
	\item generating elements of the latent space for an autoencoder instead of directly generating text; this field started with adversarially regularized autoencoders by Zhao et al.~\cite{DBLP:journals/corr/ZhaoKZRL17} and has been extended into text style transfer by disentangling style and content in the latent space~\cite{john-etal-2019-disentangled}, disentangling syntax and semantics~\cite{bao-etal-2019-generating}, DialogWAE for dialog modeling~\cite{gu2018dialogwae}, Bilingual-GAN able to generate parallel sentences in two languages~\cite{rashid-etal-2019-bilingual}, and other works.
\end{itemize}
This abundance of generative models for text has not, however, led to any significant use of synthetic data for training NLP models; this has been done only in very restricted domains such as electronic medical records~\cite{8621223} (we discuss this field in detail in Section~\ref{sec:finance}). We have suggested the reasons for this in the beginning of this section, and so far the development of natural language processing supports our view.

%% file: bettersyn.tex
\section{Directions in synthetic data development}\label{sec:syn}

In this section, we outline the main directions that intend to further improve synthetic data, making it more useful for a wide variety of applications in computer vision and other fields. In particular, we discuss the idea of domain randomization (Section~\ref{sec:domrand}) intended to improve the applications of synthetic datasets, methods to improve CGI-based synthetic data generation itself (Section~\ref{sec:cgi}), ways to create synthetic data from real images by cutting and pasting (Section~\ref{sec:cutpaste}), and finally possibilities to produce synthetic data by generative models (Section~\ref{sec:datagan}). The latter means generating useful synthetic data from scratch rather than domain adaptation and refinement, which we consider in a separate Section~\ref{sec:domain}.

\subsection{Domain randomization}\label{sec:domrand}

\emph{Domain randomization} is one of the most promising approaches to make straightforward transfer learning from synthetic data to real actually work. Consider a model that is supposed to train on $\dsyn\sim\psyn$ and later be applied to $\dreal\sim\preal$. The basic idea of domain randomization had been known since the 1990s~\cite{10.1007/3-540-59496-5_337} but was probably first explicitly presented and named in~\cite{8202133}. The idea is simple: let us try to make the synthetic data distribution $\psyn$ sufficiently wide and varied so that the model trained on $\psyn$ will be robust enough to work well on $\preal$.

Ideally, we would like to cover $\preal$ with $\psyn$, but in reality this is never achieved directly. Instead, synthetic data in computer vision can be randomized and made more diverse in a number of different ways at the level of either constructing a 3D scene or rendering 2D images from it:
\begin{itemize}
	\item at the scene construction level, a synthetic data generator (SDG) can randomize the number of objects, its relative and absolute positions, number and shape of distractor objects, contents of the scene background, textures of all objects participating in the scene, and so on;
	\item at the rendering level, SDG can randomize lighting conditions, in particular the position, orientation, and intensity of light sources, change the rendering quality by modifying image resolution, rendering type such as ray tracing or other options, add random noise to the resulting images, and so on.
\end{itemize}

The work~\cite{8202133} did the first steps to show that domain randomization works well; they used simple geometric shapes (polyhedra) as both target and distractor objects, random textures such as gradient fills or checkered patterns. The authors found that synthetic pretraining is indeed very helpful when only a small real training set is available, but helpful only if sufficiently randomized, in particular when using a large number of random textures. This approach was subsequently applied to a more ambitious domain by NVIDIA researchers Tremblay et al.~\cite{Tremblay2018TrainingDN}, who trained object detection models on synthetic data with the following procedure:
\begin{itemize}
	\item create randomized 3D scenes, adding objects of interest on top of random surfaces in the scenes;
	\item add so-called ``flying distractors'', diverse geometric shapes that are supposed to serve as negative examples for object detection;
	\item add random textures to every object, randomize the camera parameters, lighting, and other parameters.
\end{itemize}
The resulting images were completely unrealistic, yet diverse enough that the networks had to concentrate on the shape of the objects in question. Tremblay et al. report improved car detection results for R-FCN~\cite{DBLP:journals/corr/DaiLHS16} and SSD~\cite{10.1007/978-3-319-46448-0_2} architectures (but failing to improve Faster R-CNN~\cite{7485869}) on their dataset compared to Virtual KITTI (see Section~\ref{sec:dataoutdoor}), as well as improved results on hybrid datasets (adding a domain-randomized training set to COCO~\cite{DBLP:journals/corr/LinMBHPRDZ14}), a detailed ablation study, and extensive experiments showing the effect of various hyperparameters. 

Since then, domain randomization has been used and further developed in many works. Borrego et al.~\cite{Borrego2018ApplyingDR} aim to improve object detection for common objects, showing that domain randomization in the synthetic part of the dataset significantly improves the results. Tobin et al.~\cite{8593933} consider robotic grasping, a problem where the lack of real data is especially dire (see also Sections~\ref{sec:robotics} and~\ref{sec:darobot}). They use domain randomization to generate a wide variety of unrealistic procedurally generated object meshes and textured objects for grasping, so that a model trained on them would generalize to real objects as well. They show that a grasping model trained entirely on non-realistic procedurally generated objects can be successfully transferred to realistic objects.

Up until recently, domain randomization had operated under the assumption that realism is not necessary in synthetic data. Prakash et al.~\cite{Prakash2018StructuredDR} take the next logical step, continuing this effort to \emph{structured} domain randomization. They still randomize all of the settings mentioned above, but only within realistic ranges, taking into account the structure and context of a specific scene.

Finally, another important direction is learning \emph{how} to randomize. Van Vuong et al~\cite{Vuong2019HowTP} provide one of the first works in this direction, concentrating on picking the best possible domain randomization parameters for sim-to-real transfer of reinforcement learning policies. They show that the parameters that control sampling over Markov decision processes is important for the quality of transferring the learned policy to a real environment and that these parameters can be optimized. We mark this as a first attempt and expect more works devoted to structuring and honing the parameters of domain randomization.

\subsection{Improving CGI-based generation}\label{sec:cgi}

The basic workflow of synthetic data in computer vision is relatively straightforward: prepare the 3D models, place them in a controlled scene, set up the environment (camera type, lighting etc.), and render synthetic images to be used for training. However, some works on synthetic data present additional ways to enhance the data not by domain adaptation/refinement to real images (we will discuss this approach in Section~\ref{sec:domain}) but directly on the stage of CGI generation. 
	
There are two different directions for this kind of added realism in CGI generation. The first direction is to make more realistic objects. For example, Wang et al.~\cite{DBLP:journals/corr/abs-1904-12294} recognize retail items in a smart vending machine; to simulate natural deformations in the objects, they use a surface-based mesh deformation algorithm proposed in~\cite{doi:10.3722/cadaps.2012.345-359}, introducing and minimizing a global energy function for the object's mesh that accounts for random deformations and rigidity properties of the material (Wang et al. also use GAN-based refinement, see Section~\ref{sec:synrefine}). Another approach, initiated by Rozantsev et al.~\cite{ROZANTSEV201524}, is to estimate the rendering parameters required to synthetize similar images from data; this approach ties into the synthetic data generation feedback loop that we discuss in Section~\ref{sec:feedback}.

The second direction is to make more realistic ``sensors'', introducing synthetic data postprocessing that mimics the noise characteristics of real cameras/sensors. For example, we discussed \emph{DepthSynth} by Planche et al.~\cite{8374552} (see Section~\ref{sec:visiongeneral}), a system that makes simulated depth data more realistic, more similar to real depth sensors, while the OVVV system by Taylor et al.~\cite{4270516} (Section~\ref{sec:datapeople}) and the ICL-NUIM dataset by Handa et al.~\cite{6907054} (Section~\ref{sec:dataoutdoor}) take special care to simulate the noise of real cameras. There is even a separate area of research completely devoted to better modeling of the noise and distortions in real world cameras~\cite{8553409}

Apart from added realism on the level of images, there is also the question of high-level coherence and realism of the scenes. While there is no problem with coherence when the scenes are done by hand, the scale of modern datasets requires to automate scene composition as well. We note a recent joint effort in this direction by NVIDIA, University of Toronto, and MIT: Kar et al.~\cite{DBLP:journals/corr/abs-1904-11621} present \emph{Meta-Sim}, a general framework that learns to generate synthetic urban environments (see also Section~\ref{sec:dataoutdoor}). \emph{Meta-Sim} represents the composition of a 3D scene with a \emph{scene graph} and a \emph{probabilistic scene grammar}, a common representation in computer graphics~\cite{Zhu:2006:SGI:1315336.1315337}. The goal is to learn how to transform samples coming from the probabilistic grammar so that the distribution of synthetic scenes becomes similar to the distribution of scenes in a real dataset; this is known as bridging the \emph{distribution gap}. What's more, \emph{Meta-Sim} can also learn these transformations with the objective of improving the performance of networks trained on the resulting synthetic data for a specific task such as object detection (see also Section~\ref{sec:feedback}).

There are also a number of domain-specific developments that improve synthetic data generation for specific fields. For example, Cheung et al.~\cite{10.1007/978-3-319-48881-3_50} present \emph{LCrowdV}, a generation framework for crowd videos that combines a procedural simulation framework that concentrates of movements and human behaviour and a rendering framework for image/video generation, while Anderson et al.~\cite{Anderson2019StochasticSS} develop a method for stochastic sampling-based simulation of pedestrian trajectories (see Section~\ref{sec:datapeople}). 

In general, while computer graphics is increasingly using machine learning to speed up rendering (by, e.g., learning approximations to complex computationally intensive transformations) and improve the resulting 3D graphics, works on synthetic data seldom make use of these advances; a need to improve CGI-based synthetic data is usually considered in the direction of making it more realistic with refinement models (see Section~\ref{sec:refine}). However, we do expect further interesting developments in specific domains, especially in situations where the characteristics of specific sensors are important (such as, e.g., LIDARs in autonomous vehicles).

\subsection{Compositing real data to produce synthetic datasets}\label{sec:cutpaste}

Another notable line of work that, in our opinion, lies at the boundary between synthetic data and data augmentation is to use combinations and fusions of different real images to produce a larger and more diverse set of images for training. This does not require the use of CGI for rendering the synthetic images, but does require a dataset of real images.

Early works in this direction were limited by the quality of segmentation needed to cut out real objects. For some problems, however, it was easy enough to work. For example, Eggert et al.~\cite{Eggert:2015:BSD:2733373.2806407} concentrate on company logo detection. To generate synthetic images, they use a small number of real base images where the logos are clearly visible and supplied with segmentation masks, apply random warping, color transformations, and blurring, and then paste the modified (segmented) logo onto a new background image, improving logo detection results. In Section~\ref{sec:datapeople} we have discussed the ``Frankenstein'' pipeline for compositing human faces~\cite{8049355}.

The field started in earnest with the \emph{Cut, Paste, and Learn} approach by Dwibedi et al.~\cite{Dwibedi2017CutPA}, which is based on the assumption that only \emph{patch-level realism} is needed to train, e.g., an object detector. They take a collection of object instance images, cut them out with a segmentation model (assuming that the instance images are simple enough that segmentation will work almost perfectly), and paste them onto randomized background scenes, with no regard to preserving scale or scene composition. Dwibedi et al. compare different classical computer vision blending approaches (e.g., Gaussian and Poisson blending~\cite{Perez:2003:PIE:882262.882269}) to alleviate the influence of boundary artifacts after the paste; they report improved instance detection results. The work on cut-and-paste was later extended with GAN-based models (used for more realistic pasting and inpainting) and continued in the direction of unsupervised segmentation by Remez et al.~\cite{Remez2018LearningTS} and Ostyakov et al.~\cite{DBLP:journals/corr/abs-1811-07630}.

Subsequent works extend this approach for generating more realistic synthetic datasets. Dvornik et al.~\cite{DBLP:journals/corr/abs-1807-07428} argue that an important problem for this type of data augmentation is to preserve visual context, i.e., make the environment around the objects more or less realistic. They describe a preliminary experiment where they placed segmented object at completely random positions in new scenes and not only did not see significant improvements for object detection on the VOC'12 dataset but actually saw the performance deteriorate, regardless of the distractors or strategies used for blending and boundary artifact removal. Therefore, they added a separate model (also a CNN) that predicts what kind of objects can be placed in a given bounding box of an image from the rest of the image with this bounding box masked out; then the trained model is used to evaluate potential bounding boxes for data augmentation, choose the ones with the best object category score, and then paste a segmented object of this category in the bounding box. The authors report improved object detection results on VOC'12.

Wang et al.~\cite{Wang2019DataAF} develop this into an even simpler idea of \emph{instance switching}: let us switch only instances of the same class between different images in the training set; in this way, the context is automatically right, and shape and scale can also be taken into account. Wang et al. also propose to use instance switching to adjust the distribution of instances across classes in the training set and account for class importance by adding more switching for classes with lower scores. The resulting PSIS (Progressive and Selective Instance Switching) system provides improved results on the MS COCO dataset for various object detectors including Faster-RCNN~\cite{7485869}, FPN~\cite{8099589}, Mask R-CNN~\cite{8237584}, and SNIPER~\cite{NIPS2018_8143}.

With the development of conditional generative models, this field has blossomed into more complex conditional generation, usually called \emph{image fusion}, that goes beyond cut-and-paste; we discuss these extensions in Section~\ref{sec:synfromreal}.

\subsection{Synthetic data produced by generative models}\label{sec:datagan}

Generative models, especially Generative Adversarial Networks (GAN)~\cite{NIPS2014_5423}, are increasingly being used for domain adaptation, either in the form of refining synthetic images to make them more realistic or in the form of ``smart augmentation'', making nontrivial transformations on real data. We discuss these techniques in Section~\ref{sec:domain}. Producing synthetic data directly from random noise for classical computer vision applications generally does not sound promising: GANs can only try to approximate what is already in the data, so why can't the model itself do it?  However, in a number of applications synthetic data produced by GANs directly from random noise, usually with an abstract condition such as a segmentation mask, can help; in this section, we consider several examples of these approaches.

Counting (objects on an image) is a computer vision problem that, formally speaking, reduce to object detection or segmentation but in practice is significantly harder: to count correctly the model needs to detect all objects on the image, missing no one. Large datasets are helpful for counting, and synthetic data generated with a GAN conditioned on the number of objects or a segmentation mask with known number of objects, either produced at random or taken from a labeled real dataset, proves to be helpful. In particular, there is a line of work that deals with leaf counting on images of plants: ARIGAN by Giuffrida et al.~\cite{DBLP:journals/corr/abs-1709-00938} generates images of arabidopsis plants conditioned on the number of leaves, Zhu et al. generate the same conditioned on segmentation masks~\cite{Zhu2018DataAU}, and Kuznichov et al.~\cite{Kuznichov2019DataAF} generate synthetically augmented data that preserves the geometric structure of the leaves; all works report improved counting.

Santana and Hotz~\cite{Santana2016LearningAD} present a generative model that can learn to generate realistic looking images and even {videos} of the road for potential training of self-driving cars. Their model is a VAE+GAN autoencoder based on the architecture from~\cite{pmlr-v48-larsen16} that is combined with a recurrent transition model that learns realistic transitions in the embedded space. The resulting model produces synthetic videos that preserve road texture, lane markings, and car edges, keeping the road structure for at least 100 frames of the video. This interesting approach, however, has not yet led to any improvements in the training of actual driving agents.

It is hard to find impressive applications where synthetic data is generated purely from scratch by generative models; as we have discussed, this may be a principled limitation. Still, even a small amount of additional supervision may do. For example, Alonso et al.~\cite{DBLP:journals/corr/abs-1903-00277} consider adversarial generation of handwritten text (see also Section~\ref{sec:ocr}). They condition the generator on the text itself (sequence of characters), then generate handwritten instances for various vocabulary words and augment the real RIMES dataset~\cite{5277783} with the resulting synthetic dataset. Alonso et al. report improved performance in terms of both edit distance and word error rate. This example shows that synthetic data does not need to involve complicated 3D modeling to work and improve results; in this case, all information Alonso et al. provided for the generative model was a vocabulary of words.

A related but different field considers unsupervised approaches to segmentation and other computer vision problems based on adversarial architectures, including learning to segment via cut-and-paste~\cite{Remez2018LearningTS}, unsupervised segmentation by moving objects between pairs of images with inpainting~\cite{DBLP:journals/corr/abs-1811-07630}, segmentation learned from unannotated medical images~\cite{10.1007/978-3-319-66179-7_47}, and more~\cite{Bielski2019EmergenceOO}. While this is not synthetic data \emph{per se}, in general we expect unsupervised approaches to computer vision to be an important trend in the use of synthetic data.

%% file: domain.tex
\section{Synthetic-to-real domain adaptation and refinement}\label{sec:domain}

So far, we have discussed direct applications where synthetic data has been used to augment real datasets of insufficient size or to create virtual environments for training. In this section, we proceed to methods that can make the use of synthetic data much more efficient. \emph{Domain adaptation} is a set of techniques designed to make a model trained on one domain of data, the \emph{source} domain, work well on a different, \emph{target} domain. This is a natural fit for synthetic data: in almost all applications, we would like to train the model in the source domain of synthetic data but then apply the results in the target domain of real data.

In this section, we give a survey of domain adaptation approaches that have been used for such synthetic-to-real adaptation. We broadly divide the methods outlined in this section into two groups. Approaches from the first group operates on the data level, which makes it possible to extract synthetic data ``refined'' in order to work better on real data, while approaches from the second group operate directly on the model, its feature space or training procedure, leaving the data itself unchanged. We concentrate mostly on recent work related to deep neural networks and refer to, e.g., the survey~\cite{7078994} for an overview of earlier work.

In Section~\ref{sec:refine}, we discuss synthetic-to-real refinement, where a model learns to make synthetic ``fake'' data more realistic with an adversarial framework; we begin with a case study on gaze estimation (Section~\ref{sec:gaze}) where this field has mostly originated from and then proceed to other applications of such refiners (Section~\ref{sec:synrefine}) and GAN-based models that work in the opposite direction, making real data more ``synthetic-like'' (Section~\ref{sec:synfromreal}). In Section~\ref{sec:damodel} we proceed to domain adaptation at the feature and model level, i.e., to methods that perform synthetic-to-real domain adaptation but do not necessarily yield more realistic synthetic data as a by-product. Section~\ref{sec:darobot} is devoted to domain adaptation in control and robotics, and in Section~\ref{sec:medical} we present a case study of adversarial architectures for medical imaging, one of the fields where synthetic data produced with GANs can significantly improve results.

\subsection{Synthetic-to-real refinement}\label{sec:refine}

The first group of approaches for synthetic-to-real domain adaptation work with the data itself. The models below can take a synthetic image and ``refine'' it, making it better for subsequent model training. Note that while in most works we discuss here the objective is basically to make synthetic data more realistic (and it is supported by discriminators that aim to distinguish refined synthetic data from real samples), this does not necessarily have to be the case; some early works on synthetic data concluded that, e.g., synthetic imagery may work better if it is less realistic, resulting in better generalization of the models; we discuss this, e.g., in Section~\ref{sec:domrand}.

We begin with a case study on a specific problem that kickstarted synthetic-to-real refinement and then proceed to other approaches, both refining already existing synthetic data and generating new synthetic data from real by generative manipulation.

\subsubsection{Case study: GAN-based refinement for gaze estimation}\label{sec:gaze}

One of the first successful examples of straightforward synthetic-to-real refinement was given by Apple researchers Shrivastava et al. in~\cite{Shrivastava2017LearningFS}, so we begin by considering this case study in more detail and show how the research progressed afterwards. The underlying problem here is \emph{gaze estimation}: recognizing the direction where a human eye is looking. Gaze estimation methods are usually divided into \emph{model-based}, which model the geometric structure of the eye and adjacent regions, and \emph{appearance-based}, which use the eye image directly as input; naturally, synthetic data is made and refined for the latter class of approaches.

Before~\cite{Shrivastava2017LearningFS}, this problem had already been tackled with synthetic data.
Wood et al.~\cite{7410785,Wood:2016:LAG:2857491.2857492} presented a large dataset of realistic renderings of human eyes and showed improvements on real test sets over previous work done with the \emph{MPIIgaze} dataset of real labeled images~\cite{7299081}. Note that the usual increase in scale here is manifested as an increase in variability: \emph{MPIIgaze} contains about 214K images, and the synthetic training set was only about 1M images, but all images in \emph{MPIIgaze} come from the same 15 participants of the experiment, while the \emph{UnityEyes} system developed in~\cite{Wood:2016:LAG:2857491.2857492} can render every image in a different randomized environment, which makes the model significantly more robust.

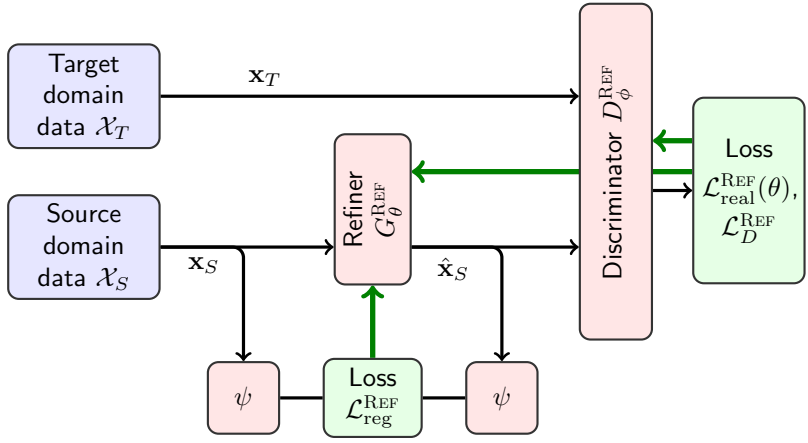
\begin{figure}[t]\centering
\input{tikz_simgan}

\caption{The architecture of SimGAN, a GAN-based refiner for synthetic data~\cite{Shrivastava2017LearningFS}.}\label{fig:simgan}
\end{figure}

Shrivastava et al. further improve upon this result by presenting a GAN-based system trained to improve synthesized images of the eyes, making them more realistic. They call this idea \emph{Simulated+Unsupervised learning}, and in it they learn a transformation implemented with a \emph{Refiner} network with the \emph{SimGAN} adversarial architecture. SimGAN consists of a generator (refiner) $\grt$ with parameters $\btheta$ and a discriminator $\ddp$ with parameters $\bphi$; see Fig.~\ref{fig:simgan} for an illustration. The discriminator learns to distinguish between real and refined images with standard binary classification loss function
$$\ldp = -\esb{\log\ddp(\hxs)} - \etb{\log\left(1 - \ddp(\xt)\right)},$$
where $\hxs=\grt(\xs)$ is the refined version of $\xs$ produced by $\grt$. The generator, in turn, is trained with a combination of the realism loss $\lgreal$ that makes $\grt$ learn to fool $\ddp$ and regularization loss $\lgreg$ that captures the similarity between the refined image and the original one in order to preserve the target variable (gaze direction in~\cite{Shrivastava2017LearningFS}):
\begin{align*}
\lgr &= \esb{\lgreal(\btheta; \xs) + \lambda\lgreg(\btheta; \xs)},\text{ where}\\
\lgreal(\btheta; \xs) &= -\log\left(1 - \ddp(\grt(\xs))\right),\\
\lgreg(\btheta; \xs) &= \left\|\psi(\grt(\xs)) - \psi(\xs)\right\|_{1},
\end{align*}
where $\psi(\x)$ is a mapping to a feature space (that can contain the image itself, image derivatives, statistics of color channels, or features produced by a fixed extractor such as a pretrained CNN), and $\|\cdot\|_1$ denotes the $L_1$ distance. On Fig.~\ref{fig:simgan}, black arrows denote the data flow and green arrows show the gradient flow (on subsequent pictures, we omit the gradient flow to avoid clutter); $\lgreal(\btheta)$ and $\ldp$ are shown in the same block since it is the same loss function differentiated with respect to different weights for $G$ and $D$ respectively.

In SimGAN, the generator is a fully convolutional neural network that consists of several ResNet blocks~\cite{DBLP:journals/corr/HeZRS15} and does not contain any striding or pooling, which makes it possible to operate on pixel level while preserving the global structure. The training proceeds by alternating between minimizing $\lgr$ and $\ldp$, with an additional trick of drawing training samples for the discriminator from a stored history of refined images in order to keep it effective against all versions of the generator. Another important feature is the locality of adversarial loss: $\ddp$ outputs a probability map on local patches of the original image, and $\ldp$ is summed over the patches.

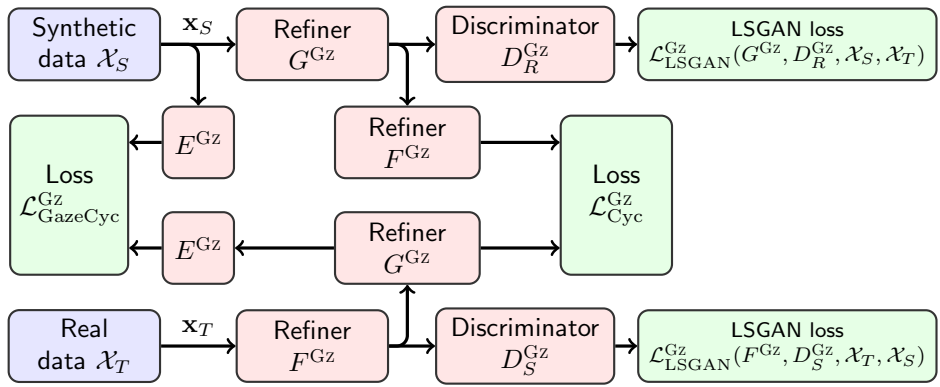
\begin{figure}[t]\centering
\input{tikz_gazegan}

\caption{The architecture of GazeGAN~\cite{Sela2017GazeGANU}. Blocks with identical labels have shared weights.}\label{fig:gazegan}
\end{figure}

SimGAN's ideas were later picked up and extended in many works. A direct successor of SimGAN, GazeGAN developed by Sela et al.~\cite{Sela2017GazeGANU}, applied to synthetic data refinement the idea of CycleGAN for unpaired image-to-image translation~\cite{8237506}.
The structure of GazeGAN contains four networks: $\ggaze$ is the generator that learns to map images from the synthetic domain S to the real domain R, $\fgaze$ learns the opposite mapping, from R to S, and two discriminators $\dsgaze$ and $\drgaze$ learn to distinguish between real and fake images in the synthetic and real domains respectively. An overview of the GazeGAN architecture is shown on Fig.~\ref{fig:gazegan}. It uses the following loss functions:
\begin{itemize}
	\item the LSGAN~\cite{8237566} loss for the generator with label smoothing to $0.9$~\cite{DBLP:journals/corr/PereyraTCKH17} to stabilize training:
	$$\lgaze_{\mathrm{LSGAN}}(G,D,S,R) = \EEE{\x_S\sim\psyn}{\left(D(G(\x_S))-0.9\right)^2} + \EEE{\x_T\sim\preal}{D(\x_T)^2};$$
	this loss is applied to both directions, as $\lgaze_{\mathrm{LSGAN}}(\ggaze,\drgaze,\X_S,\X_T)$ and $\lgaze_{\mathrm{LSGAN}}(\fgaze,\dsgaze,\X_T,\X_S)$;
	\item the cycle consistency loss~\cite{8237506} designed to make sure both $F\circ G$ and $G\circ F$ are close to identity:
	\begin{align*}
	\lgaze_{\mathrm{Cyc}}(\ggaze,\fgaze) =& \EEE{\x_S\sim\psyn}{\|\fgaze(\ggaze(\x_S))-\x_S\|_1} + \\
	& \EEE{\x_T\sim\preal}{\|\ggaze(\fgaze(\x_T))-\x_T\|_1};
	\end{align*}
	\item finally, a special gaze cycle consistency loss to preserve the gaze direction (so that the target variable can be transferred with no change); for this, the authors train a separate gaze estimation network $\egaze$ designed to overfit and predict the gaze very accurately on synthetic data; the loss makes sure $\egaze$ still works after applying $F\circ G$:
	$$\lgaze_{\mathrm{GazeCyc}}(\ggaze,\fgaze) = \EEE{\x_S\sim\psyn}{\|\egaze(\fgaze(\ggaze(\x_S)))-\egaze(\x_S)\|_2^2}.$$
\end{itemize}
Sela et al. report improved gaze estimation results. Importantly for us, they operate not on the $30\times 60$ grayscale images as in~\cite{Shrivastava2017LearningFS}, but on $128\times 128$ color images, and GazeGAN actually refines not only the eye itself but parts of the image (e.g., nose and hair) that were not part of the 3D model of the eye.

Finally, a note of caution: GAN-based refinement is not the only way to go. Kan et al.~\cite{Kan2018EffectivenessOD} compared three approaches to data augmentation for pupil center point detection, an important subproblem in gaze estimation: affine transformations of real images, synthetic images from \emph{UnityEyes}, and GAN-based refinement. In their experiments, real data augmentation with affine transformations was a clear winner, with the GAN improving over \emph{UnityEyes} but falling short of the augmented real dataset. This is one example of a general common wisdom: in cases where a real dataset is available, one should squeeze out all the information available in it and apply as much augmentation as possible, regardless of whether the dataset is augmented with synthetic data or not.


\subsubsection{Refining synthetic data with GANs}\label{sec:synrefine}

Gaze estimation is a convenient problem for GAN-based refining because the images of eyes used for gaze estimation have relatively low resolution, and scaling GANs up to high-resolution images has proven to be a difficult task in many applications. Nevertheless, in this section we consider a wider picture of other GAN-based refiners applied for synthetic-to-real domain adaptation.

We begin with an early work in refinement, parallel to~\cite{Shrivastava2017LearningFS}, which was done by \emph{Google} researchers Bousmalis et al.~\cite{Bousmalis2017UnsupervisedPD}. They train a GAN-based architecture for pixel-level domain adaptation (PixelDA), using a basic style transfer GAN, i.e., they find by alternating optimization steps
$$\min_{\btheta_G,\btheta_T}\max_{\bphi}\lambda_1\lpixd(\dpix, \gpix) + \lambda_2\lpixt(\gpix, \tpix) + \lambda_3\lpixc(\gpix),\quad\text{where:}$$
\begin{itemize}
	\item $\lpixd(\dpix, \gpix)$ is the domain loss,
	\begin{align*}
	\lpixd(\dpix, \gpix) =& \EEE{\x_S\sim\psyn}{\log\left(1-\dpix(\gpix(\x_S;\btheta_G);\bphi)\right)} + \\
	& \EEE{\x_T\sim\preal}{\log\dpix(\x_T;\bphi)};
	\end{align*}
	\item $\lpixt(\gpix, \tpix)$ is the task-specific loss, which in~\cite{Bousmalis2017UnsupervisedPD} was the image classification cross-entropy loss provided by a classifier $\tpix(\x;\btheta_T)$ which is also trained as part of the model:
	\begin{multline*}
	\lpixt(\gpix, \tpix) = \\ \EEE{\x_S,\y_S\sim\psyn}{-\y_S^\top\log \tpix(\gpix(\x_S;\btheta_G);\btheta_T)-\y_S^\top\log \tpix(\x_S;\btheta_T)};
	\end{multline*}
	\item $\lpixc(\gpix)$ is the content similarity loss, intended to make $\gpix$ preserve the parts of the image related to the target variables; in~\cite{Bousmalis2017UnsupervisedPD}, $\lpixc$ was used to preserve the foreground objects (that would later need to be classified) with a mean squared error applied to their masks:
	\begin{align*}
	\lpixc(\gpix) = \E_{\x_S\sim\psyn} & \left[ \frac 1k\left\|(\x_S-\gpix(\x_S;\btheta_G))\odot\m(\x)\right\|^2_2-\right. \\
	&\left. -\frac 1{k^2}\left((\x_S-\gpix(\x_S;\btheta_G))^\top\m(\x)\right)^2\right],
	\end{align*}
	where $\m(\x_S)$ is a segmentation mask for the foreground object extracted from the synthetic data renderer; note that this loss does not ``insist'' on preserving pixel values in the object but rather encourages the model to change object pixels in a consistent way, preserving their pairwise differences.
\end{itemize}
Bousmalis et al. applied this GAN to the \emph{Synthetic Cropped LineMod} dataset, a synthetic version of a small object classification dataset~\cite{Wohlhart2015LearningDF}, doing both classification and pose estimation for the objects. They report improved results in both metrics compared to both training on purely synthetic data and a number of previous approaches to domain adaptation.

Many modern approaches to synthetic data refinement include the ideas of CycleGAN~\cite{8237506}. The most direct application is the \emph{GeneSIS-RT} framework by Stein and Roy~\cite{Stein2018GeneSISRtGS} that refines synthetic data directly with the CycleGAN trained on unpaired datasets of synthetic and real images. They show that a training set produced by image-to-image translation learned by CycleGAN improves the results of training machine learning systems for real-world tasks such as obstacle avoidance and semantic segmentation.

\begin{figure}[t]\centering
\input{tikz_t2net}

\caption{The architecture of $\text{T}^2$Net~\cite{10.1007/978-3-030-01234-2_47}. Blocks with identical labels have shared weights.}\label{fig:t2net}
\end{figure}

\emph{$\text{T}^2$Net} by Zheng et al.~\cite{10.1007/978-3-030-01234-2_47} uses synthetic-to-real refinement for depth estimation from a single image. This work also uses the general ideas of CycleGAN with a translation network that makes the images more realistic. The new idea here is that $\text{T}^2$Net asks the synthetic-to-real generator $\gsttwo$ not only to translate one specific domain (synthetic data) to another (real data) but also to work across a number of different input domains, making the input image ``more realistic'' in every case, as shown on Figure~\ref{fig:t2net}. In essence, this means that $\gsttwo$ aims to learn the minimal transformation necessary to make an image realistic, in particular, it should not change real images much. In total, $\text{T}^2$Net has the generator loss function
\begin{multline*}
\lttwo = \lttwog(\gsttwo,\drttwo)+\lambda_1\lttwogf(\fttwo,\dfttwo) + \lambda_2\lttwo_r(\gsttwo) \\ 
		+ \lambda_3\lttwo_t(\fttwo)+\lambda_4\lttwo_s(\fttwo),\text{ where}
\end{multline*}
\begin{itemize}
	\item $\lttwog(\gsttwo,\drttwo)$ is the usual GAN loss for synthetic-to-real transfer with discriminator $\drttwo$:
	\begin{align*}
	\lttwog(\gsttwo,\drttwo) = & \EEE{\x_S\sim\psyn}{\log(1-\drttwo(\gsttwo(\x_S)))} + \\ & \EEE{\x_T\sim\preal}{\log\drttwo(\x_T)};
	\end{align*}
	\item $\lttwogf(\fttwo,\dfttwo)$ is the feature-level GAN loss for the features extracted from translated and real images with discriminator $\dfttwo$:
	\begin{align*}
	\lttwogf(\fttwo,\drttwo) = & \EEE{\x_S\sim\psyn}{\log\dfttwo(\fttwo(\gsttwo(\x_S)))} + \\ & \EEE{\x_T\sim\preal}{\log(1-\dfttwo(\fttwo(\x_T)))};
	\end{align*}
	\item $\lttwo_r(\gsttwo)=\left\|\gsttwo(\xt)-\xt\right\|_1$ is the reconstruction loss for real images;
	\item $\lttwo_r(\fttwo)=\left\|\fttwo(\hxs)-\y_S\right\|_1$ is the \emph{task loss} for depth estimation on synthetic images, namely the $L_1$-norm of the difference between the predicted depth map for a translated synthetic image $\hxs$ and the original ground truth synthetic depth map $\y_S$; this loss ensures that the translation does not change the depth map;
	\item $\lttwo_s(\fttwo)=\left|\partial_x\fttwo(\xt)\right|^{-|\partial_x\xt|} + \left|\partial_y\fttwo(\xt)\right|^{-|\partial_y\xt|}$, where $\partial_x$ and $\partial_y$ are image gradients, is the task loss for depth estimation on real images; since ground truth depth maps are not available now, this regularizer is a locally smooth loss intended to optimize object boundaries, a common tool in depth estimation models~\cite{monodepth17}.
\end{itemize}
Zheng et al. show that $\text{T}^2$Net can produce realistic images from synthetic ones, conclude that end-to-end training is preferable over separated training (of the translation network and depth estimation network), and note that $\text{T}^2$Net can achieve good results for depth estimation with no access to real paired data, even outperforming some (but not all) supervised approaches.

We note a few more interesting applications of refiner-based architectures. Wang et al.~\cite{Wang2018CombiningRN} use a classical refiner modeled after~\cite{Shrivastava2017LearningFS} for human motion synthesis and control. Their model first generates a motion sequence from a recurrent neural network and then refines it with a GAN; since the goal is to model and refine sequences, both generator and discriminator in the refiner also have RNN-based architectures.
Dilipkumar~\cite{Dilipkumar2017GenerativeAI} applied SimGAN to improve handwriting recognition. They generated synthetic handwriting images and applied SimGAN to refine them, with significantly improved recognition of real handwriting after training on a hybrid dataset.

A recent example that applies GAN-based refinement in a classical computer vision setting is provided by Wang et al.~\cite{DBLP:journals/corr/abs-1904-12294}. They consider the problem of recognizing objects inside an automatic vending machines; this is a basic functionality needed for monotoring the state of supplies and is usually done based on object detection. Wang et al. begin by scanning the objects, adding random deformations to the resulting 3D models (see Section~\ref{sec:cgi}), setting up scenes and rendering with settings matching the fisheye cameras used in smart vending machines. Then they refine rendered images with virtual-to-real style transfer done by a CycleGAN-based architecture. The novelty here is that Wang et al. separate foreground and background losses, arguing that style transfer needed for foreground objects is very different from (much stronger than) the style transfer for backgrounds. Thus, they use the overall loss function
\begin{align*}
\lod =& \lodgan(\god, \dod_T, \X_S, \X_T) + \lodgan(\fod, \dod_S, \X_T, \X_S) + \\
   +& \lambda_1\lodcyc(\god,\fod) + \lambda_2\lodbg + \lambda_3\lodfg,\quad\text{where:}
\end{align*}
\begin{itemize}
	\item $\lodgan(G,D,X,Y)$ is the standard adversarial loss for generator $G$ mapping from domain $X$ to domain $Y$ and discriminator $D$ distinguishing real images from fake ones in the domain $Y$;
	\item $\lodcyc(G, F)$ is the cycle consistency loss as used in CycleGAN~\cite{8237506} and detailed above;
	\item $\lodbg$ is the background loss, which is the cycle consistency loss computed only for the background part of the images as defined by the mask $\m_{\mathrm{bg}}$:
	\begin{align*}
	\lodbg =& \EEE{\x_T\sim\preal}{\left\|\left(\god(\fod(\x_T))-\x_T\right)\odot\m_{\mathrm{bg}}(\x_T)\right\|_2} \\
	+& \EEE{\x_S\sim\psyn}{\left\|\left(\fod(\god(\x_S))-\x_S\right)\odot\m_{\mathrm{bg}}(\x_S)\right\|_2};
	\end{align*}
	\item $\lodfg$ is the foreground loss, similar to $\lodbg$ but computed only for the hue channel in the HSV color space (the authors argue that color and profile are the most critical for recognition and thus need to be preserved the most), as denoted by $\cdot^H$ below:
	\begin{align*}
	\lodfg =& \EEE{\x_T\sim\preal}{\left\|\left(\god(\fod(\x_T))^H-\x_T^H\right)\odot\m_{\mathrm{fg}}(\x_T)\right\|_2} \\
	 +& \EEE{\x_S\sim\psyn}{\left\|\left(\fod(\god(\x_S))^H-\x_S^H\right)\odot\m_{\mathrm{fg}}(\x_S)\right\|_2}.
	 \end{align*}
\end{itemize}
Segmentation into foreground and background is done automatically in synthetic data and is made easy in~\cite{DBLP:journals/corr/abs-1904-12294} for real data since the camera position is fixed, and the authors can collect a dataset of real background templates from the vending machines they used in the experiments and then simply subtract the backgrounds to get the foreground part. As a result, Wang et al. report significantly improved results when using hybrid datasets of real and synthetic data for all three tested object detection architectures: PVANET~\cite{DBLP:journals/corr/KimCHRP16}, SSD~\cite{10.1007/978-3-319-46448-0_2}, and YOLOv3~\cite{DBLP:journals/corr/abs-1804-02767}. Even more importantly, they report a comparison between basic and refined synthetic data with clear gains achieved by refinement across all architectures.

\begin{figure}[t]\centering
\input{tikz_wanggan}

\caption{The architecture of the refiner used in~\cite{DBLP:journals/corr/abs-1904-12294}. Blocks with identical labels have shared weights.}\label{fig:wanggan}
\end{figure}
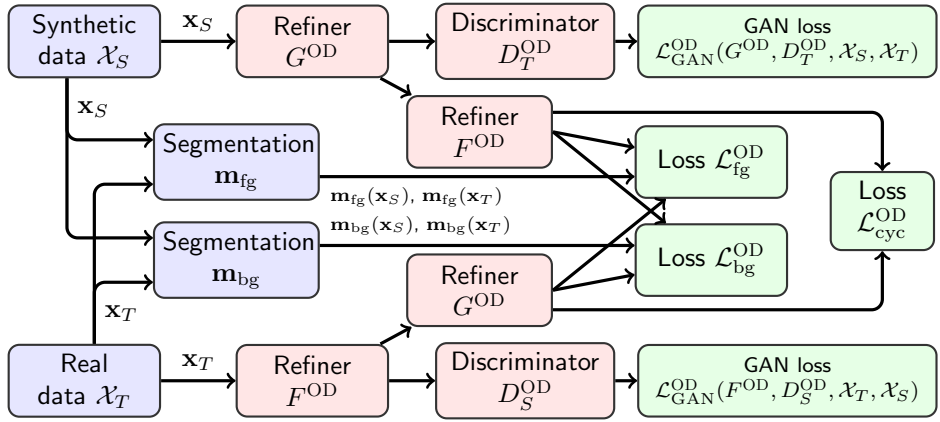

Wang et al.~\cite{DBLP:journals/corr/abs-1903-03303} discuss synthetic-to-real domain adaptation in the context of crowd counting, a domain where synthetic data has been successfully used for a long time. They collect synthetic data with the \emph{Grand Theft Auto V} engine, producing the so-called \emph{GTA5 Crowd Counting} (GCC) dataset (see also Section~\ref{sec:datapeople}). They also use a CycleGAN-based refiner from the domain of synthetic images $\X_S$ to the domain of real images $\X_T$ but remark that CycleGAN can easily lose local patterns and textures which is exactly what is important for crowd counting. Therefore, they modify the CycleGAN loss function with the Structural Similarity Index (SSIM)~\cite{1284395} that computes the similarity between images in terms of local patterns. Their final loss function is
\begin{align*}\lssim =& \lssimgan(\gssim, \dssim_T, \X_S, \X_T)+\lssimgan(\fssim, \dssim_S, \X_T, \X_S) \\
&+\lambda\lssimcyc(\gssim,\fssim,\X_S,\X_R)+\mu\lssimse(\gssim,\fssim,\X_S,\X_R),
\end{align*}
where $\gssim:\X_S\to \X_T$ is the generator from synthetic to real domains, $\fssim:\X_T\to\X_S$ works in the opposite direction, $\dssim_T$ and $\dssim_S$ are the corresponding discriminators, $\lssimgan$ is a standard GAN loss function, and $\lssimcyc$ is the cycle consistency loss as defined above, while $\lssimse$ is a special loss function designed to improve SSIM:
	\begin{align*}
	\lssimse(\gssim,\fssim,\X_S,\X_R) =& \EEE{\x_S\sim\psyn}{1-\mathrm{SSIM}(\x_S,\fssim(\gssim(\x_S))} \\
	& + \EEE{\x_T\sim\pdata}{1-\mathrm{SSIM}(\x_T,\gssim(\fssim(\x_T))}.
	\end{align*}

Bak et al.~\cite{Bak2018DomainAT} (see Section~\ref{sec:datapeople}) use domain translation with synthetic people as a method for person re-identification. The domains in this model are represented by illumination conditions for the images; the model has access to $M$ real source domains and $N$ synthetic domains, with $N\gg M$, and the objective is to perform re-identification in an unknown target domain. To achieve this, Bak et al. first learned a generic feature representation from all domains, but the resulting model, even trained on a hybrid dataset, did not generalize well. Therefore, Bak et al. proceeded to domain adaptation done as follows: first choose the nearest synthetic with a separately trained domain identification network fine-tuned for illumination classification, then use the CycleGAN architecture (as shown above) to do domain translation from this synthetic domain to the target domain, and then use the to fine-tune the re-identification network. Bak et al. report improved results from the entire pipeline as compared to any individual parts in the ablation study.

In robotics, domain adaptation of synthetic imagery is not yet common, but some applications have appeared there as well. For example, Pecka et al.~\cite{Pecka2018DataDrivenPT} train a CycleGAN-based domain adaptation model to learn a transformation from the data observed in a non-differentiable physics simulator and the data from a real robotic platform, showing improved sim-to-real policy transfer results.

\subsubsection{Making synthetic data from real with GANs}\label{sec:synfromreal}

A related idea is to generate synthetic data from real data by learning to transform real data with conditional GANs. This could either simply serve as ``smart augmentation'' to extend the dataset or, more interestingly, could ``fill in the holes'' in the data distribution, obtaining synthetic data for situations that are lacking in the original dataset.

Zhao et al.~\cite{Zhao2018,ijcai2018-165} concentrated on applying this idea to face recognition in the wild, with different poses rather than by a frontal image. They continued the work of Tran et al.~\cite{8099624} (we do not review it in detail) and Huang et al.~\cite{8237529}, who presented a TP-GAN (two-pathway GAN) architecture for frontal view synthesis: given a picture of a face, generate a frontal view picture. TP-GAN's generator $\gtp_{\theta}$ has two pathways: a global network $\gtp_{\theta^g}$ that rotates the entire face and four local patch networks $\gtp_{\theta^l_i}$, $i=1,\ldots,4$, that process local textures around four facial landmarks (eyes, nose, and mouth). Both $\gtp_{\theta^g}$ and $\gtp_{\theta^l_i}$ have encoder-decoder architectures with skip connections for multi-scale feature fusion. The discriminator $\dtp_{\phi}$ learns to distinguish real frontal face images $\x_{\mathrm{front}}\sim \D_{\mathrm{front}}$ from synthesized images $\gtp_{\theta}(\x)$, $\x\sim\pdata$:
$$\min_{\theta}\max_{\phi}\left[ \EEE{\x_{\mathrm{front}}\sim \D_{\mathrm{front}}}{\log\dtp_{\phi}(\x_{\mathrm{front}})} + \EEE{\x\sim\pdata}{\log\left(1-\dtp_{\phi}(\gtp_{\theta}(\x))\right)}\right].$$
The synthesis loss function in TP-GAN is a sum of four loss functions:
\begin{itemize}
	\item pixel-wise $L_1$-loss between the ground truth frontal image and rotated image:
	$$\ltp_{\mathrm{pixel}}(\x,\x^{\mathrm{gt}}) = \frac{1}{W\times H} \sum_{i=1}^W\sum_{j=1}^H\left|\gtp_{\theta}(\x)_{i,j}-\x^{\mathrm{gt}}_{i,j}\right|,$$
	where $\x^{\mathrm{gt}}$ is the ground truth frontal image corresponding to $\x$, and $W$ and $H$ are an image's width and height; this loss is measured at several different places of the network: output of $\gtp_{\theta^g}$, outputs of $\gtp_{\theta^l_i}$, and the final output of $\gtp_{\theta}$;
	\item symmetry loss 
	$$\ltp_{\mathrm{sym}} =\frac{1}{W/2\times H} \sum_{i=1}^{W/2}\sum_{j=1}^H\left|\gtp_{\theta}(\x)_{i,j}-\gtp_{\theta}(\x)_{W-(i-1),j}\right|,$$ intended to preserve the symmetry of human faces;
	\item adversarial loss 
	$$\ltp_{\mathrm{adv}}=\frac 1N\sum_{n=1}^N-\log\dtp_{\phi}(\gtp_{\theta}(\x_n));$$
	\item identity preserving loss 
	$$\ltp_{\mathrm{ip}}=\sum_{l}\frac{1}{W_l\times H_l} \sum_{i=1}^{W_l}\sum_{j=1}^{H_l}\left|F^l(\gtp_{\theta}(\x))_{i,j}-F^l(\x^{\mathrm{gt}})_{i,j}\right|,$$
	where $F^l$ denotes the output of the $l$th layer of a face recognition network applied to $\x$ and $\x^{\mathrm{gt}}$ (Huang et al. used Light CNN~\cite{Wu2018ALC}, and only used its last two layers in $\ltp_{\mathrm{ip}}$); this idea is based on \emph{perceptual losses}~\cite{Johnson2016PerceptualLF}, a popular idea in GANs designed to preserve high-level features when doing low-level transformations; in this case, it serves to preserve the person's identity when rotating the face.
\end{itemize}
As usual, the final loss function is a linear combination of the four losses above and a regularization term.

In~\cite{Zhao2018}, Zhao et al. propose the DA-GAN (Dual-Agent GAN) model that also works with faces but in the opposite scenario: while TP-GAN rotates every face into the frontal view, DA-GAN aims to fill in the ``holes'' in the real data distribution, rotating real faces so that the distribution of angles becomes more uniform. They begin with a 3D morphable model (see also Section~\ref{sec:datapeople}) from~\cite{7163096}, extracting 68 facial landmarks with the Recurrent Attentive-Refinement (RAR) model~\cite{10.1007/978-3-319-46448-0_4} and estimating the transformation matrix with 3D-MM~\cite{4409029}. However, the authors report that simulation quality dramatically decreases for large yaw angles, necessitating further improvement with the DA-GAN framework.

Again, DA-GAN's generator $\gda$ maps a synthetized image to a refined one, $\hxs=\gda(\xs)$. It is trained on a linear combination of three loss functions
$$\lda_{G}=-\lda_{\mathrm{adv}}+\lambda_1\lda_{\mathrm{ip}}+\lambda_2\lda_{\mathrm{pp}},$$ and the discriminator $\dda$ consists of two parallel branches (agents) that optimize $\lda_{\mathrm{adv}}$ and $\lda_{\mathrm{ip}}$ respectively. The loss functions are defined as follows:
\begin{itemize}
	\item the adversarial loss $\lda_{\mathrm{adv}}$ follows the BEGAN architecture introduced in~\cite{DBLP:journals/corr/BerthelotSM17}: this branch of $\dda$ is an autoencoder that minimizes the Wasserstein distance with a boundary equilibrium regularization term:
	$$\lda_{\mathrm{adv}} = \sum_j\left|\xtsub{j}-\dda(\xt)_j\right| - k_t\sum_i\left|\hxssub{i}-\dda(\hxs)_i\right|,$$
	where, again, $\xt$ is a real image, $\hxs$ is a refined image, and $k_t$ is a boundary equilibrium regularization term continuously trained to maintain the equilibrium 
	$$\EE{\sum_i\left|\hxssub{i}-\dda(\hxs)_i\right|}=\gamma\EE{\sum_j\left|\xtsub{j}-\dda(\xt)_j\right|}$$
	for some diversity ratio $\gamma$ (see~\cite{DBLP:journals/corr/BerthelotSM17,Zhao2018} for more details); in general, $\lda_{\mathrm{adv}}$ is designed to keep the refined face in the manifold of real faces;
	\item the identity preservation loss $\lda_{\mathrm{ip}}$, similar to $\ltp_{\mathrm{ip}}$, aims to make the refinement respect the identities, but does it in a different way; here the idea is to put both $\xt$ and $\xs$ through the same (relatively simple) face recognition network and bring its features together; DA-GAN uses for this purpose a classifier $\fda$ trained on the bottleneck layer of $\dda$:
	\begin{align*}\lda_{\mathrm{ip}} =& \frac1N\sum_j\left[-\y_j\log\fda(\xtsub{j}) + (1-\y_j)\log(1-\fda(\xtsub{j}))\right] \\
	+& \frac1N\sum_j\left[-\y_j\log\fda(\hxssub{j}) + (1-\y_j)\log(1-\fda(\hxssub{j}))\right],
	\end{align*}
	where $\y_j$ is the ground truth label;
	\item the pixel-wise loss $\lda_{\mathrm{pp}}$ is the $L_1$-loss intended to make sure that the pose (angle of inclination for the head) remains the same after refinement:
	$$\lda_{\mathrm{pp}} = \frac{1}{W\times H}\sum_{i=1}^W\sum_{j=1}^H\left|\xssub{{}i,j}-\hxssub{{}i,j}\right|.$$
\end{itemize}
In total, during training DA-GAN alternatively optimizes $\gda$ and $\dda$ with loss functions $\lda_G$ and $\lda_D=\lda_{\mathrm{adv}}+\lambda_1\lda_{\mathrm{ip}}$. Following~\cite{DBLP:journals/corr/BerthelotSM17}, to measure convergence DA-GAN tests the reconstruction quality together with proportion control theory, evaluating
{\footnotesize
$$\lda_{\mathrm{con}}=\sum_j\left|\xtsub{j}-\dda(\xtsub{j})\right| + \left|\gamma\sum_j\left|\xtsub{j}-\dda(\xtsub{j})\right|-\sum_i\left|\hxssub{j}-\dda(\hxssub{j})\right|\right|.$$}

Apart from experiments done by the authors, DA-GAN was verified in a large-scale NIST IJB-A competition~\cite{nist2017} where a model based on DA-GAN won the face verification and face identification tracks. This result heavily supports the general premise of using synthetic data: augmenting the dataset and balancing out the training data distribution with synthetic images proved highly beneficial in this case.

Inoue et al.~\cite{8451064} try to find a middle ground between synthetic and real data. They use two variational autoencoders (VAE) to reduce both synthetic and real data to a common pseudo-synthetic image space, and then train CNNs on images from this common space. The training sequence is as follows:
\begin{itemize}
	\item train $\mathrm{VAE}_1:\X_S\to\X_S$ as an autoencoder;
	\item train $\mathrm{VAE}_2:\X_T\to\X_S$ where the decoder is fixed and shares weights with the decoder from $\mathrm{VAE}_1$; as a result, $\mathrm{VAE}_2$ has to learn to generate pseudo-synthetic images from real images;
	\item train a CNN for the task in question on synthetic data, using $\mathrm{VAE}_1$ to map it to the common image space;
	\item during inference, use a composition of $\mathrm{VAE}_2$ and CNN.
\end{itemize}

In the context of robotics, this kind of \emph{real-to-sim} approach was continued by Zhang et al.~\cite{DBLP:journals/corr/abs-1802-00265} in a framework called ``VR-Goggles for Robots''. It is based on the CycleGAN ideas as they were continued in CyCADA~\cite{DBLP:journals/corr/abs-1711-03213}, a popular domain adaptation model that adds semantic losses to CycleGAN. The \emph{VR-Goggles} model has two generators, $\gvrg_S:\X_T\to\X_S$ with discriminator $\dvrg_S$ that distinguishes fake synthetic images and $\gvrg_T:\X_S\to\X_T$ with discriminator $\dvrg_T$ that is defined in the domain of real images. The overall loss function is
\begin{align*}
\lvrg =& \lvrgg(\gvrg_T,\dvrg_T;\X_S,\X_T) + \lvrgg(\gvrg_S,\dvrg_S;\X_T,\X_S) \\
  +& \lambda_1\left( \lvrgcyc(\gvrg_S,\gvrg_T;\X_T) + \lvrgcyc(\gvrg_T,\gvrg_S;\X_S) \right) \\
  +& \lambda_2\left( \lvrgsem(\gvrg_S;\X_T, \fvrg_S) + \lvrgsem(\gvrg_S;\X_S, \fvrg_S) \right) \\
  +& \lambda_3\left( \lvrgshift(\gvrg_T;\X_S) + \lvrgshift(\gvrg_S;\X_T) \right),\text{ where}
\end{align*}
\begin{itemize}
	\item $\lvrgg$ is the standard GAN loss:
	\begin{align*}
	\lvrgg(\gvrg_T,\dvrg_T;\X_S,\X_T) = & \EEE{\x_T\sim\preal}{\log\dvrg_T(\xt)} + \\
	 & \EEE{\x_S\sim\psyn,\z}{\log\left(1-\dvrg_T(\gvrg_T(\xs))\right)}
	\end{align*}
	and similarly for $\lvrgg(\gvrg_S,\dvrg_S;\X_T,\X_S)$;
	\item $\lvrgsem$ is the semantic loss as introduced in CyCADA~\cite{DBLP:journals/corr/abs-1711-03213}; the idea is that if we have ground truth labels for the synthetic domain $\X_S$ (in this case, we are doing semantic segmentation), we can train a network $\fvrg_S$ on $\X_S$ and then use it to generate pseudolabels for the domain $\X_T$ where ground truth is not available; the semantic loss now makes sure that the results (segmentation maps) remain the same after image translation:
	\begin{align*}
	\lvrgsem(\gvrg_S;\X_T, \fvrg_S) &= \EEE{\xt\sim\preal}{\CE\left(\fvrg_S(\xt),\fvrg_S(\gvrg_S(\xt))\right)}, \\
	\lvrgsem(\gvrg_T;\X_S, \fvrg_S) &= \EEE{\xs\sim\psyn}{\CE\left(\fvrg_S(\xs),\fvrg_S(\gvrg_T(\xs))\right)}, \\
	\end{align*}
	where $\mathrm{CE}$ denotes cross-entropy;
	\item $\lvrgshift$ is the \emph{shift loss} that makes the image translation result invariant to shifts:
	\begin{align*}
	\lvrgshift(\gvrg_T;\X_S) &= \EEE{\xs,i,j}{\left\|\gvrg_T(\xs)_{\left[\begin{smallmatrix} x\to i \\ y\to j\end{smallmatrix}\right]} - \gvrg_T\left(\x_{S,\left[\begin{smallmatrix} x\to i \\ y\to j\end{smallmatrix}\right]}\right)\right\|^2_2},\\
	\lvrgshift(\gvrg_S;\X_T) &= \EEE{\xt,i,j}{\left\|\gvrg_S(\xt)_{\left[\begin{smallmatrix} x\to i \\ y\to j\end{smallmatrix}\right]} - \gvrg_S\left(\x_{T,\left[\begin{smallmatrix} x\to i \\ y\to j\end{smallmatrix}\right]}\right)\right\|^2_2},\\
	\end{align*}
	where $\x_{\left[\begin{smallmatrix} x\to i \\ y\to j\end{smallmatrix}\right]}$ denotes the shifting operation by $i$ pixels along the X-axis and $j$ pixels along the Y-axis, and $i$ and $j$ are chosen uniformly at random up to the total downsampling factor of the network $K$ (since the result will always be invariant to shifts of multiples of $K$).
\end{itemize}
Zhang et al. test their solution on the CARLA navigation benchmark~\cite{Dosovitskiy17} and show significant improvements.

James et al.~\cite{James2018SimtoRealVS} consider the same kind of approach for robotic grasping. Their model, \emph{Randomized-to-Canonical Adaptation Networks} (RCAN), learn to map heavily randomized simulation images (with random textures) to a canonical (much simpler) rendered image and also map real images to canonical rendered images; interestingly, they achieve good results with a much simpler GAN architecture where additional losses simply bring together the segmentation masks and depth maps for simulated images, and there are no cycle consistency losses. An even simpler approach is taken by Yang et al.~\cite{DBLP:journals/corr/abs-1801-03458} who introduce \emph{domain unification} for autonomous driving. Their model, called \emph{DU-Drive}, consists of a generator that translates real images to simplified synthetic images and a discriminator that distinguishes them from actual synthetic images; the driving policy is then trained in the simulator.

Another idea for generating synthetic data from real is to compose parts of real images to produce synthetic ones. We have discussed the cut-and-paste approaches in~\ref{sec:cutpaste}; a natural continuation of these ideas would be to use more complex, semantic conditioning with a GAN-based architecture. For example, Joo et al.~\cite{8578274} provide a GAN-based architecture for generating a fusion image, where, say, one input $\x$ provides the identity of a person, another input $\y$ provides the shape (pose) of a person, and the result is $\hx$ which has the identity of $\x$ and the shape of $\y$. Their FusionGAN architecture extends CycleGAN-like ideas to losses that distinguish between identity and shape of an image, introduceing the concepts of \emph{identity loss} and \emph{shape loss}. FusionGAN relies on a dataset where there are several images with different shapes but the same identity (e.g., the same person in different poses; a dataset of videos would provide a simple example); its overall loss function is
$$\lfuse = \lfuse_I + \lambda\lfuse_S,\text{ where}$$
\begin{itemize}
	\item $\lfuse_I$ is the \emph{identity loss}
	\begin{align*}
	\lfuse_I(\gfuse,\dfuse) =& \EEE{\x,\x'\sim\preal(\x)}{\left\|1-\dfuse(\x,\x')\right\|_2} \\
			+ & \EEE{\x\sim\preal(\x),\y\sim\preal(\y)}{\left\|\dfuse(\x,\gfuse(\x,\y))\right\|_2},
	\end{align*}
	i.e., the discriminator $\dfuse$ learns to distinguish real pairs of images $(\x,\x')$ with the same identity (but different shapes) and fake pairs of images $(\x,\gfuse(\x,\y))$ where $\gfuse$ is supposed to take the identity from $\x$;
	\item $\lfuse_S$ is the \emph{shape loss} defined as
	$$\lfuse_{S_1}(\gfuse) = \EEE{\x,\x'\sim\preal(\x)}{\left\|\x'-\gfuse(\x,\x')\right\|_1}$$
	when $\x$ and $\x'$ have the same identity, and
	\begin{align*}
		\lfuse_{S_2a}(\gfuse) &= \EEE{\x\sim\preal(\x),\y\sim\preal(\y)}{\left\|\y-\gfuse(\y,\gfuse(\x,\y))\right\|_1}, \\
		\lfuse_{S_2b}(\gfuse) &= \EEE{\x\sim\preal(\x),\y\sim\preal(\y)}{\left\|\gfuse(\x,\y)-\gfuse(\gfuse(\x,\y),\y))\right\|_1}, \\
	\end{align*}
	i.e., $\gfuse(\y,\gfuse(\x,\y))$ should be the same as $\y$, with identity from $\y$ and shape also from $\y$, and $\gfuse(\gfuse(\x,\y),\y))$ should be the same as $\gfuse(\x,\y)$, with identity from $\x$ and shape from $\y$.
\end{itemize}

Similar ideas have been extended to animating still images~\cite{Siarohin2018AnimatingAO}, motion transfer~\cite{Chan2018EverybodyDN}, and image-to-image translation~\cite{Ma2019ANB}. In general, these works belong to an interesting field of generative semantic manipulation with GANs. Important works in this direction include Mask-Contrasting GAN~\cite{Liang2017GenerativeSM} that can modify an object to a different suitable category inside its segmentation mask (e.g., replace a cat with a dog), Attention-GAN~\cite{10.1007/978-3-030-01216-8_11} that performs the same task with an attention-based architecture, IterGAN~\cite{Galama2018IterGANsIG} that attempts iterative small-scale 3D manipulations such as rotation from 2D images, and others. However, while this field produces very interesting works, so far we have not seen direct applications of such architectures to generating synthetic data. We believe that ideas similar to TP-GAN can also be fruitful in other domains, especially in situations where one- or few-shot learning is required so ``smart augmentations'' such as rotation can bring significant improvements.

\subsection{Domain adaptation at the feature/model level}\label{sec:damodel}

In the previous sections, we have considered models that perform domain adaptation (DA) at the data level, i.e., one can extract a part of the model that takes as input a data point from the source domain (in our case, a synthetic image) and map it to the target domain (domain of real images). However, the final goal of model design rarely involves the generation of more realistic synthetic images; they are merely a stepping stone to producing models that work better, e.g., in the absence of supervision in the target domain. Therefore, to make better use of synthetic data it makes sense to also consider \emph{feature-level} or \emph{model-level} domain adaptation, i.e., methods that work in the space of features or model weights and never go back to change the actual data.

The simplest approach to domain adaptation would be to share the weights between networks operating on different domains or learn an explicit mapping between them~\cite{1467314,Glorot:2011:DAL:3104482.3104547}. While we mostly discuss other approaches, we note that simpler techniques based on weight sharing remain relevant for domain adaptation. In particular, Rozantsev et al.~\cite{8310033} recently presented a domain adaptation approach where two similar networks are trained on the source and target domain with special regularizers that bring their weights together; the authors evaluate their approach on synthetic-to-real domain adaptation for drone detection with promising results.

Another approach to model-level domain adaptation is related to mining relatively strong priors from real data that can then inform a model trained on synthetic data, helping fix problematic cases or incongruencies between the synthetic and real datasets. For example, Zhang et al.~\cite{DBLP:journals/corr/ZhangDG17,DBLP:journals/corr/abs-1812-09953} present a curriculum learning approach to domain adaptation for semantic segmentation of urban scenes. They train a segmentation network on synthetic data (specifically on the GTA dataset; see also Section~\ref{sec:dataoutdoor}) but with a special component in the loss function related to the general label distribution in real images:
$$\LL^{\textsc{Curr}}=\frac{1}{|\Xs|}\sum_{\xs\in\Xs}\LL\left(\ys,\hy_S\right) + \lambda\frac{1}{|\Xt|}\sum_{\xt\in\Xt}\sum_k{\mathcal C}\left(p^k(\xt),{\hat p}^k(\xt)\right),$$
where $\LL\left(\ys,\hy_S\right)$ is the pixel-wise cross-entropy, a standard segmentation loss, and ${\mathcal C}\left(p^k(\xt),{\hat p}^k(\xt)\right)$ is the cross-entropy between the distribution of labels ${\hat p}(\xt)$ in a real image $\xt$ that the network produces and $p(\xt)$ is the real label distribution (superscript $k$ denotes different kinds of label distributions). Note that $p(\xt)$ is not available in the real data, so this is where curriculum learning comes in: the authors first train on synthetic a simpler model to estimate $p(\xt)$ from image features and then use it to inform the segmentation model. Recent developments of this interesting direction shift from merely enforcing the label distribution to matching features on multiple different levels~\cite{Huang2018DomainTT}. In particular, recent works~\cite{Lian2019ConstructingSP,Yue2019DomainRA} have introduced the so-called \emph{pyramid consistency loss} instead of ${\mathcal C}\left(p(\xt),{\hat p}(\xt)\right)$ that tries to enforce consistency across domains on the activation maps of later layers of the network.

\begin{figure}[t]\centering
\input{tikz_ganin}

\caption{The high-level architecture of model-level domain adaptation from~\cite{Ganin:2015:UDA:3045118.3045244,Ganin:2016:DTN:2946645.2946704}: the gradient flow (green) from the domain classification loss is reversed (becomes red) at the features.}\label{fig:ganin}
\end{figure}
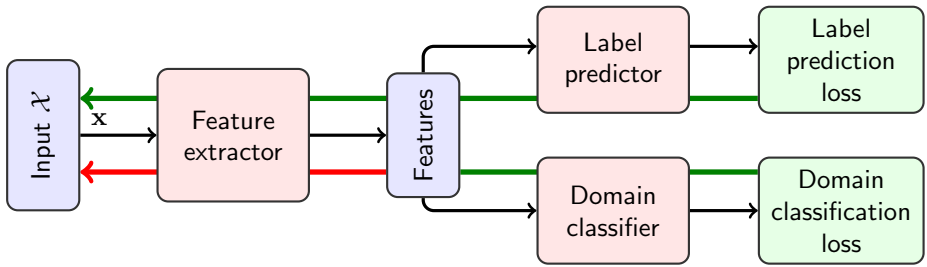

One of the main directions in model-level domain adaptation was initiated by Ganin and Lempitsky~\cite{Ganin:2015:UDA:3045118.3045244} who present a generic framework for unsupervised domain adaptation. Their approach, illustrated on Figure~\ref{fig:ganin}, consists of a \emph{feature extractor}, a \emph{label predictor} that performs the necessary task (e.g., classification) on extracted features, and a \emph{domain classifier} that takes the same features and attempts to classify which domain the original input belonged to. The idea is to train the label predictor to perform as well as possible and at the same time train the domain classifier to perform as badly as possible; this is achieved with \emph{gradient reversal}, i.e., multiplying the gradients by a negative constant as they pass from the domain classifier to the feature extractor. In a subsequent work, Ganin et al.~\cite{Ganin:2016:DTN:2946645.2946704} generalized this domain adaptation approach to arbitrary architectures and experimented with DA in different domains, including image classification, person re-identification, and sentiment analysis. We also note extensions and similar approaches to domain adaptation developed in~\cite{Long:2015:LTF:3045118.3045130,Tzeng:2015:SDT:2919332.2919970,Long:2016:UDA:3157096.3157112} and the domain confusion metric that helps produce domain-invariant representation~\cite{DBLP:journals/corr/TzengHZSD14}, but proceed to highlight the works that perform specifically synthetic-to-real domain adaptation.

Many general model-level domain adaptation approaches have been validated or subsequently extended to synthetic-to-real domain adaptation. Xu et al.~\cite{6595948} consider the pedestrian detection problem (this work is a continuation of~\cite{6587038}, see Section~\ref{sec:visiongeneral}). They adapt detectors trained on virtual datasets with a boosting-based procedure, assigning larger weights to samples that are similar to target domain ones. Sun and Saenko~\cite{BMVC.28.82} propose a domain adaptation approach based on decorrelating the features of a classifier, both in unsupervised and supervised settings. Later, L{\'opez} et al.~\cite{Lopez2017} extended the SA-SVVM domain adaptation used in~\cite{6587038} to train deformable part-based models, using synthetic pedestrians from the SYNTHIA dataset (see Section~\ref{sec:dataoutdoor}) as the main example. In a parallel paper, the authors of SYNTHIA Ros et al.~\cite{Ros2017} used a simple domain adaptation technique called Balanced Gradient Contribution~\cite{DBLP:journals/corr/RosSAW16}, where training on synthetic data is regularized by the gradient obtained on a (small) real dataset, to further improve their results on segmentation aided by synthetic data. Ren et al.~\cite{Ren2018CrossDomainSM} perform cross-domain self-supervised multi-task learning with synthetic images: their model predicts several parameters of an image (surface normal, depth, and instance contour) and at the same time tries to minimize the difference between synthetic and real data in feature space.

\begin{figure}[t]\centering
\input{tikz_dsn}

\caption{Architecture of the domain separation network~\cite{Bousmalis:2016:DSN:3157096.3157135}. Blocks with identical labels have shared weights.}\label{fig:dsn}
\end{figure}
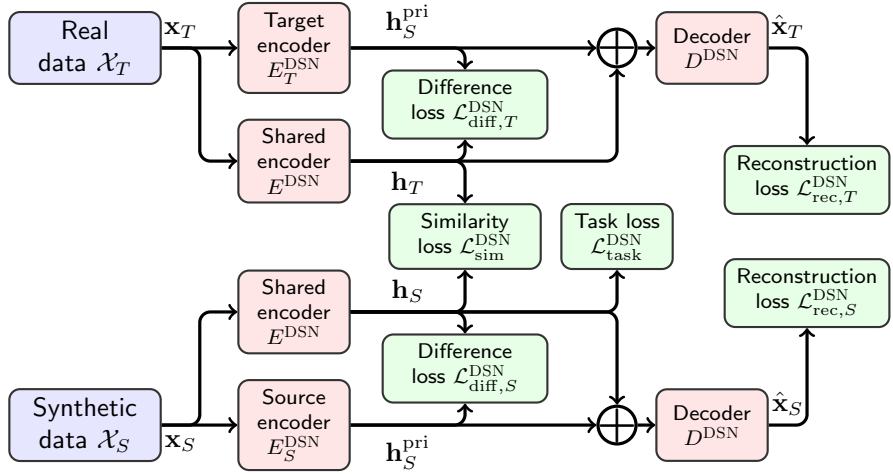

\emph{Domain separation networks} by Bousmalis et al.~\cite{Bousmalis:2016:DSN:3157096.3157135}, illustrated on Fig.~\ref{fig:dsn}, explicitly separate the shared and private components of both source and target domains. Specifically, they introduce a shared encoder $\edsn(\x)$ and two private encoders, $\edsn_S$ for the source domain and $\edsn_T$ for the target domain. The total objective function for a domain separation network is
$$\ldsn = \ldsn_{\mathrm{task}} + \lambda_1\ldsn_{\mathrm{rec}} + \lambda_2\ldsn_{\mathrm{diff}} + \lambda_1\ldsn_{\mathrm{sim}},\text{ where}$$
\begin{itemize}
	\item $\ldsn_{\mathrm{task}}$ is the supervised task loss in the source domain, e.g., for the image classification task it is 
	$$\ldsn_{\mathrm{task}}=-\EEE{\xs\sim\psyn}{\y^\top\log\fdsn(\edsn(\xs))},$$
	where $\fdsn$ is the classifier operating on the output of the shared encoder;
	\item $\ldsn_{\mathrm{rec}}$ is the reconstruction loss defined as the difference between the original samples $\xs$ and $\xt$ and the results of a shared decoder $\ddsn$ that tries to reconstruct the images from a combination of shared and private representations:
	\begin{align*}
	\ldsn_{\mathrm{rec}} = & -\EEE{\xs\sim\psyn}{\LL_{\mathrm{sim}}(\xs,\ddsn(\edsn(\xs)+\edsn_S(\xs)))} \\
	& -\EEE{\xt\sim\preal}{\LL_{\mathrm{sim}}(\xt,\ddsn(\edsn(\xt)+\edsn_T(\xt)))}
	\end{align*}
	for some similarity metric $\LL_{\mathrm{sim}}$;
	\item $\ldsn_{\mathrm{diff}}$ is the difference loss that encourages the hidden shared representations of instances from the source and target domains $\edsn(\xs)$ and $\edsn(\xt)$ to be orthogonal to their corresponding private representations $\edsn_S(\xs)$ and $\edsn_T(\xt)$; in~\cite{Bousmalis:2016:DSN:3157096.3157135}, the difference loss is defined as
	$$\ldsn_{\mathrm{diff}} = \left\|H_S^\top H_S^{\mathrm{pri}}\right\|^2_F + \left\|H_T^\top H_T^{\mathrm{pri}}\right\|^2_F,$$
	where $H_S$ is the matrix of $\edsn(\xs)$, $H_S^{\mathrm{pri}}$ is the matrix of $\edsn_S(\xs)$, and similarly for $H_T$ and $H_T^{\mathrm{pri}}$;
	\item $\ldsn_{\mathrm{sim}}$ is the similarity loss that encourages the hidden shared representations from the source and target domains $\edsn(\xs)$ and $\edsn(\xt)$ to be similar to each other, i.e., indistinguishable by a domain classifier trained through the gradient reversal layer as in~\cite{Ganin:2015:UDA:3045118.3045244}; in~\cite{Bousmalis:2016:DSN:3157096.3157135}, this loss is composed of the cross-entropy for the domain classifier and maximal mean discrepancy (MMD)~\cite{Gretton:2012:KTT:2503308.2188410} for the hidden representations themselves.
\end{itemize}
Bousmalis et al. evaluate their model on several synthetic-to-real scenarios, e.g., on synthetic traffic signs from~\cite{10.1007/978-3-319-02895-8_52} and synthetic objects from the \emph{LineMod} dataset~\cite{Wohlhart2015LearningDF}. 

Domain separation networks became one of the first major examples in domain adaptation with \emph{disentanglement}, where the hidden representations are domain-invariant and some of the features can be changed to transition from one domain to another. Further developments include asymmetric training for unsupervised domain adaptation~\cite{Saito:2017:ATU:3305890.3305990}, DistanceGAN for one-sided domain mapping~\cite{Benaim:2017:OUD:3294771.3294843}, co-regularized alignment~\cite{Kumar:2018:CAU:3327546.3327607},  cross-domain autoencoders~\cite{Gonzalez-Garcia:2018:ITC:3326943.3327062}, multisource domain adversarial networks~\cite{Zhao:2018:AMS:3327757.3327947}, continuous cross-domain translation~\cite{Liu:2018:UFD:3327144.3327184}, face recognition adaptation from images to videos with the help of synthetic augmentations~\cite{DBLP:journals/corr/abs-1708-02191}, and more~\cite{Zhang:2019:RAT:3309872.3291124}; all of these advances may be relevant for synthetic-to-real domain adaptation but we will highlight some works that are already doing adaptation between these two domains.

The popular and important domains for feature-based domain adaptation are more or less the same as in domain adaptation on the data level, but feature-based DA may be able to handle higher-dimensional inputs and more complex scenes because the adaptation itself is done in an intermediate lower-dimensional space. As an illustrative example, let us consider feature-based DA for computer vision problems for outdoor scenes (see also Section~\ref{sec:dataoutdoor}). In their \emph{FCNs in the Wild} model, Hoffman et al.~\cite{DBLP:journals/corr/HoffmanWYD16} consider feature-based DA for semantic segmentation with fully convolutional networks (FCN) where ground truth is available for the source domain (synthetic data) but unavailable for the target domain (real data). Their unsupervised domain adaptation framework contains a feature extractor $\ffcnw$ and the joint objective function
$$\lfcnw = \lfcnw_{\mathrm{seg}} + \lfcnw_{\mathrm{DA}} + \lfcnw_{\mathrm{MI}},\text{ where}$$
\begin{itemize}
	\item $\lfcnw_{\mathrm{seg}}$ is the standard supervised segmentation objective on the source domain, where supervision is available;
	\item $\lfcnw_{\mathrm{DA}}$ is the domain alignment objective that minimizes the observed source and target distance in the representation space by training a discriminator (domain classifier) to distinguish instances from source and target domains; an interesting new idea here is to take as an instance for this objective not the entire image but a cell from a coarse grid that corresponds to the of high-level features that domain adaptation is supposed to bring together;
	\item $\lfcnw_{\mathrm{MI}}$ is the multiple instance loss that encourages pixels to be assigned to class $c$ in such a way that the percentage of an image labeled with $c$ remains withing the expected range derived from the source domain.
\end{itemize}

Another direction of increasing the input dimension is to move from images to videos. Xu et al.~\cite{Xu2019AdversarialAF} use adversarial domain adaptation to transfer object detection models---single-shot multi-box detector (SSD)~\cite{10.1007/978-3-319-46448-0_2} and multi-scale deep CNN (MSCNN)~\cite{DBLP:journals/corr/CaiFFV16}---from synthetic samples to real videos in the smoke detection problem.

Chen et al.~\cite{DBLP:journals/corr/ChenCCTWS17} construct the \emph{Cross City Adaptation} model that brings together the features from different domains, again with semantic segmentation of outdoor scenes in mind. Their framework optimizes the joint objective function
$$\lcca = \lcca_{\mathrm{task}} + \lcca_{\mathrm{global}} + \lcca_{\mathrm{class}},\text{ where}$$
\begin{itemize}
	\item $\lcca_{\mathrm{task}}$ is the task loss, in this case cross-entropy between predicted and ground truth segmentation masks in the source domain;
	\item $\lcca_{\mathrm{global}}$ is the global domain alignment loss, again defined as fooling the domain discriminator similar to \emph{FCNs in the Wild};
	\item $\lcca_{\mathrm{class}}$ is the class-wise domain alignment loss, where grid cells are assigned soft class labels (extracted from the truth in the source domain and predicted in the target domain), and the domain classifiers and discriminators are trained and applied \emph{class-wise}, separately.
\end{itemize}
As the title suggests, \emph{Cross City Adaptation} is intended to adapt outdoor segmentation models trained on one city to other cities, but Chen et al. also apply it to synthetic-to-real domain adaptation from SYNTHIA to Cityscapes (see Section~\ref{sec:dataoutdoor}), achieving noticeable gains in segmentation quality.

Hong et al.~\cite{8578243} provide one of the most direct and most promising applications of feature-level synthetic-to-real domain adaptation. In their \emph{Structural Adaptation Network}, the conditional generator $\gsda_{\btheta}(\xs,\z)$ takes as input the features $\fsda_l(\xs)$ from a low-level layer of the feature extractor (i.e., features with fine-grained details) and random noise $\z$ and produces transformed feature maps that should be simlar to feature maps extracted from real images. To achieve that, $\gsda$ produces a noise map $\ghssda_{\btheta}(\fsda_l(\xs),\z)$ and then adds it to high-level features: $\gsda_{\btheta}(\xs,\z) = \fsda_h(\xs) + \ghssda_{\btheta}(\fsda_l(\xs),\z)$. The optimization problem is
$$\min_{\btheta,\btheta'}\max_{\bphi}\left(\lsdag(\gsda_{\btheta},\dsda_{\bphi})+\lambda\lsda_{\mathrm{task}}(\gsda_{\btheta},\tsda_{\btheta'})\right),\text{ where}$$
\begin{itemize}
	\item $\lsdag$ is the GAN loss in the feature space:
	\begin{align*}
	\lsdag(\gsda_{\btheta},\dsda_{\bphi}) = & \EEE{\x_T\sim\preal}{\log\dsda_{\bphi}(\xt)} + \\
	 & \EEE{\x_S\sim\psyn,\z}{\log\left(1-\dsda_{\bphi}(\gsda_{\btheta}(\xs,\z))\right)};
	\end{align*}
	\item $\lsda_{\mathrm{task}}$ is the task loss for the pixel-wise classifier $\tsda_{\btheta'}$ which is trained end-to-end, together with the rest of the architecture; the task loss is defined as the pixel-wise cross-entropy between the segmentation mask $\tsda_{\btheta'}(\gsda_{\btheta}(\xs,\z))$ produced by $\tsda_{\btheta'}$ on adapted features and the ground truth synthetic segmentation mask $\ys$.
\end{itemize}
Hong et al. compare the Structural Adaptation Network with other state of the art approaches, including FCNs in the Wild~\cite{DBLP:journals/corr/HoffmanWYD16} and cross-city adaptation~\cite{DBLP:journals/corr/ChenCCTWS17}, with source domain datasets SYNTHIA and GTA and target domain dataset Cityscapes; they conclude that this adaptation significantly improves the results for semantic segmentation of urban scenes.

To summarize, feature-level domain adaptation provides interesting opportunities for synthetic-to-real adaptation, but these methods still mostly represent work in progress. In our experience, feature- and model-level DA is usually a simpler and more robust approach, easier to get to work, so we expect new exciting developments in this direction and recommend to try this family of methods for synthetic-to-real DA (unless actual refined images are required).

\subsection{Domain adaptation for control and robotics}\label{sec:darobot}

In the field of control, joint domain adaptation is usually intended to transfer control policies learned in a simulated environment to a real setting. As we have already discussed in Sections~\ref{sec:dataindoor} and~\ref{sec:dataoutdoor}, simulated environments are almost inevitable in reinforcement learning for robotics, as they allow to scale the datasets up compared to real data and cover a much wider range of situations than real data that could be used for imitational learning (see also the survey~\cite{DBLP:journals/corr/TaiL16a}). In this setting, domain adaptation is performed either for the control itself or jointly for the control and synthetic data. The field began even before deep learning; for instance, Saxena et al.~\cite{doi:10.1177/0278364907087172} learned a model for estimating grasp locations for previously unseen objects on synthetic data. 

In Cutler et al.~\cite{7139550}, the results of training on a simulator serve as a prior for subsequent learning in the real world. Moreover, in~\cite{Cutler14_ICRA,Cutler15_TRO} Cutler et al. proceed to multifidelity simulators, training the reinforcement learning agent in a series of simulators with increasing realism; we note this idea as potentially fruitful for other domains as well.

\emph{DeepMind} researchers Rusu et al.~\cite{Rusu2017SimtoRealRL} studied the possibility for transfer learning from simulated environments to the real world in the context of end-to-end reinforcement learning for robotics. They use the general idea of \emph{progressive networks}~\cite{DBLP:journals/corr/RusuRDSKKPH16}, an architecture designed for multitask learning and transfer where each subsequent column in the network solves a new task and receives as input the hidden activations from previous columns. Rusu et al. present a modification of this idea for robot transfer learning, then train the first column in the MuJoCo physics simulator~\cite{Todorov2012MuJoCoAP}, and then transfer to a real \emph{Jaco} robotic arm, using the \emph{Asynchronous Advantage Actor-Critic} (A3C) framework for reinforcement learning~\cite{pmlr-v48-mniha16}. The authors report improved results for progressive networks compared to simple transfer via fine-tuning.

There are plenty of works that consider similar kinds of transfer learning, known in robotics as closing the \emph{reality gap}. In particular, Bousmalis et al.~\cite{Bousmalis2018UsingSA} use a simulated environment to learn robotic grasping with a domain adaptation model called GraspGAN that makes synthetic images more realistic with a refiner (see Section~\ref{sec:synrefine}); they argue that the added realism improves the results for control transfer. Tzeng et al.~\cite{DBLP:journals/corr/TzengDHFPLSD15} propose a framework that combines supervised domain adaptation (that requires paired images) and unsupervised DA (that aligns the domains on the level of distributions); to do that, they introduce the notion of a ``weak pairing'' between images in the source and target domains and learn to find matching synthetic images to produce aligned data. The resulting model is successfully applied to training a visuomotor policy for real robots. Pan et al.~\cite{DBLP:journals/corr/YouPWL17} consider sim-to-real translation for autonomous driving; they convert synthetic images to a scene parsing representation and then generate a realistic image by a generator corresponding to this parsing representation; the reinforcement learning agent receives this image as part of its driving environment. An even simpler approach, taken, e.g., by Xu et al~\cite{Tan2018AutonomousDI}, would be to directly use the segmentation masks as input for the RL agent.

Researchers from \emph{Wayve} Bewley et el.~\cite{Bewley2018LearningTD} perform domain adaptation for learning to drive from simulation; they claim to present the first end-to-end driving policy transferred from an (obviously supervised) synthetic setting to the fully unsupervised real domain. Their model does image translation and control transfer at the same time, learning the control on a jointly learned latent embedding space. The architecture consists of two encoders $\eswayve$ and $\etwayve$, two generators $\gswayve$ and $\gtwayve$, two discriminators $\dswayve$ and $\dtwayve$, and a controller $\cwayve$. The image translator follows the MUNIT architecture~\cite{NIPS2017_6672}, with two convolutional variational autoencoder networks that swap the latent embeddings to translate between domains, i.e., for $\xs\sim\Xs$, $\xt\sim\Xt$
	$$\begin{array}{cc}
		\zs = \eswayve(\xs) + \epsilon, & \hxs = \gswayve(\zs), \\
		\zt = \etwayve(\xt) + \epsilon, & \hxt = \gtwayve(\zt), \\
	\end{array}$$
and the translation is to compute $\zs$ with $\eswayve$ and then predict $\hx$ with $\gtwayve$ and vice versa. The overall generator loss function is
\begin{align*}
\lwayve &= \lambda_0\lrecwayve + \lambda_1\lcycwayve + \lambda_2\lcontrolwayve + \lambda_3\lccycwayve + \\
        &+ \lambda_4\llswayve + \lambda_5\lpercwayve + \lambda_6\lzrecwayve,\quad\text{where}
\end{align*}
\begin{itemize}
	\item $\lrecwayve$ is the $L_1$ image reconstruction loss in both domains: 
	\begin{align*}
	\lrecwayve(\xs)&=\|\gswayve(\eswayve(\xs)) - \xs\|_1,\\
	\lrecwayve(\xt)&=\|\gtwayve(\etwayve(\xt)) - \xt\|_1;
	\end{align*}
	\item $\lcycwayve$ is the cycle consistency loss for both domains: 
	$$\lcycwayve(\xs)=\|\gswayve(\etwayve(\gtwayve(\eswayve(\xs)))) - \xs\|_1$$ and similar for $\Xt$;
	\item $\lcontrolwayve$ is the control loss that compares the controls produced on the training set with the autopilot: $\lcontrolwayve(\xs) = \| \cwayve(\eswayve(\xs)) - \mathbf{c} \|_1$ for ground truth control $\mathbf{c}$, and similar for $\Xt$;
	\item $\lccycwayve$ is the control cycle consistency loss that makes the controls similar for images translated to another domain: 
	$$\lccycwayve(\xs) = \| \cwayve(\etwayve(\gswayve(\eswayve(\xs)))) - \cwayve(\eswayve(\xs)) \|_1,$$ and similar for $\Xt$;
	\item $\llswayve$ is the LSGAN adversarial loss applied to both generator-disc\-ri\-mi\-nator pairs (see above);
	\item $\lpercwayve$ is the perceptual loss (see above) for both image translation directions with instance normalization applied before, as shown in~\cite{DBLP:journals/corr/abs-1804-04732};
	\item $\lzrecwayve$ is the latent reconstruction loss: $\lzrecwayve(\zs) = \|\etwayve(\gtwayve(\zs))-\zs\|_1$ and similar for $\Xt$.
\end{itemize}
Bewley et al. compare their approach with a number of transfer learning baselines, show excellent results for end-to-end learning to drive, and even perform real world experiments with the trained policy.
Similar techniques have been used without synthetic data in the loop as well; e.g., Wulfmeier et al.~\cite{Wulfmeier2017AddressingAC} use a similar model for domain adaptation to handle appearance changes in outdoor robotics, i.e., changes in weather conditions, lighting, and the like.

We have already discussed the works of Inoue et al.~\cite{8451064}, Zhang et al.~\cite{DBLP:journals/corr/abs-1802-00265}, James et al.~\cite{James2018SimtoRealVS}, and Yang et al.~\cite{DBLP:journals/corr/abs-1801-03458} who make real data more similar to synthetic for computer vision problems related to robotic grasping and visual navigation (see Section~\ref{sec:synfromreal}). Importantly, these models are not merely translating images but are also tested on real world robots. Zhang et al. not only show improvements in semantic segmentation results but also conduct real-world robotic experiments for indoor and outdoor visual navigation tasks, first training a navigation policy in a simulated environment and then directly deploying it on a robot in a real environment, while James et al. test their solution on a real robotic hand, training the \emph{QT-Opt} policy~\cite{pmlr-v87-kalashnikov18a} to grasp from a simulation with 5000 additional real life grasping episodes better than the same policy trained on 580{,}000 real episodes, a more than 99\% reduction in required real world input.

Another direction where synthetic data might be useful for learning control is to generate synthetic behaviours to improve imitation learning~\cite{DBLP:journals/corr/abs-1811-06711}. Bansal et al.~\cite{Bansal2018ChauffeurNetLT} discuss the insufficient data problem in imitation learning: for learning to drive, even 30 million real-world expert driving examples that combine into more than 60 days of driving is not sufficient to train an end-to-end driving model. To alleviate this lack of data, they present their imitation learning framework \emph{ChauffeurNet} with data where synthetic perturbations have been introduced to expert driving examples. This allows to cover corner cases such as collisions and off-road driving, i.e., bad examples that should be avoided but that are lacking in expert examples altogether. Interestingly, perturbations are introduced into intermediate representations rather than in raw sensor input or controller outputs.

To sum up, closing the reality gap is one of the most important problems in the field of control and robotics. Important breakthroughs in this direction appear constantly, but there is still some way to go before self-driving cars and robotic arms are able to train in a simulated environment and then perfectly transfer these skills to the real world.

\subsection{Case study: GAN-based domain adaptation for medical imaging}\label{sec:medical}

Medical imaging is a field where labeled data is especially hard to come by. First, while manual labeling is hard and expensive enough for regular computer vision problems, in medical imaging it is far more expensive because it cannot be crowdsourced to anonymous annotators: for most problems, medical imaging data can only be reliably labeled by a trained professional, often with a medical degree. Second, for obvious privacy reasons it is very hard to arrange for publishing real datasets, and collecting a large enough labeled dataset to train a standard object detection or segmentation model would in many cases require a concerted effort from several different hospitals; thus, with the exception of public competitions, most papers in the field use private datasets and are not allowed to share their data. Third, some pathologies simply do not have sufficiently large and diverse datasets collected yet. At the same time, often there are relatively large generic datasets available, e.g., of healthy tissue but not of a specific pathology of interest.

While we emphasize GAN-based generation methods, we note that there have been successful attempts to use rendered synthetic data for medical imaging tasks that are based on recent developments in medical visualization and rendering tools. For example, Mahmood et al.~\cite{DBLP:journals/corr/abs-1805-08400} use the recently developed cinematic rendering technique for CT~\cite{Eid2017CinematicRI} (a photorealistic simulation of the propagation of light through tissue) to train a CNN for depth estimation in endoscopy data.

GANs have been widely applied to generating realistic medical images~\cite{8310638,DBLP:journals/corr/abs-1807-03401,DBLP:journals/corr/abs-1805-03144,DBLP:journals/corr/abs-1804-04338,DBLP:journals/corr/abs-1804-11024,DBLP:journals/corr/abs-1809-07294}. Moreover, since the images are domain-specific, the quality of GAN-produced images has relatively quickly reached the level where it can in many applications pass the ``visual Turing test'', fooling even trained specialists; see, e.g., lung nodule samples generated in~\cite{8363564} or magnetic resonance (MR) images of the brain in~\cite{10.1007/978-3-319-68612-7_71,8363678}. Therefore, it is no wonder that GAN-based domain adaptation (DA) techniques, especially based on fusing and augmenting real images, are increasingly finding their way into medical imaging. In this section, we give a brief overview of recent work in this domain.

In some works, synthetic data is generated from scratch, i.e., GANs are trained to convert random noise into synthetic images (see also Section~\ref{sec:datagan}). Frid-Adar et al.~\cite{FRIDADAR2018321,FridAdar2018GANbasedDA} used two standard GAN architectures, Deep Convolutional GAN (DCGAN)~\cite{DBLP:journals/corr/RadfordMC15} and Auxiliary Classifier GAN (ACGAN)~\cite{Odena:2017:CIS:3305890.3305954} with class label auxiliary information, to generate synthetic computed tomography (CT) images of liver lesions. They report significantly improved results in image classification with CNNs when training on synthetic data compared to standard augmentations of their highly limited dataset (182 2D scans divided into three types of lesions). Baur et al.~\cite{DBLP:journals/corr/abs-1804-11024} attempt high resolution skin lesion synthesis, comparing several GAN architectures and obtaining highly realistic results even with a small training dataset. Han et al.~\cite{DBLP:journals/corr/abs-1903-12564,2019arXiv190513456H} concentrate on brain magnetic resonance (MR) images. They use progressively growing GANs (PGGAN)~\cite{Karras2018ProgressiveGO} to generate $256\times 256$ MR images and then compare two different refinement approaches: SimGAN~\cite{Shrivastava2017LearningFS} as discussed in Section~\ref{sec:gaze} and UNIT~\cite{NIPS2017_6672}, an unsupervised image-to-image translation architecture that maps each domain into a shared latent space with a VAE-GAN architecture~\cite{pmlr-v48-larsen16} (we remark that the original paper~\cite{NIPS2017_6672} also applies UNIT, among other things, to synthetic-to-real translation). Han et al. report improved results when combining GAN-based synthetic data with classic domain adaptation techniques. Neff~\cite{Neff18} uses a slightly different approach: to generate synthetic data for segmentation, he uses a standard WGAN-GP architecture~\cite{NIPS2017_7159} but generates image-segmentation pairs, i.e., images with an additional channel that shows the segmentation mask. Neff reports improved segmentation results with U-Net~\cite{RFB15a} after augmenting a real dataset with synthetic image-segmentation pairs. Mahmood et al.~\cite{DBLP:journals/corr/abs-1711-06606} show an interesting take on the problem by doing the reverse: they make real medical images look more like synthetic images in order to then apply a network trained on synthetic data (see also Section~\ref{sec:synfromreal}). With this approach, they improve state of the art results in depth estimation for endoscopy images.

In general, segmentation problems in medical imaging are especially hard to label, and segmentation data is especially lacking in many cases. In this context, recent works have often employed conditional GANs and pix2pix models to generate realistic images from randomized segmentation masks. For example, Bailo et al.~\cite{DBLP:journals/corr/abs-1901-06219} consider red blood cell image generation with the \emph{pix2pixHD} model~\cite{8579015}. Namely, their conditional GAN optimizes
\begin{multline*}
\min_{\gptp}\left[\max_{\dptp_1,\dptp_2}\left[\lptpg(\gptp,\dptp_1)+\lptpg(\gptp,\dptp_2)\right] + \right. \\ \left. \vphantom{\max_{\dptp_1}\left[\lptpg(\gptp,\dptp_1)\right]} + \lambda_1\left(\lptpfm(\gptp,\dptp_1)+\lptpfm(\gptp,\dptp_2)\right) + \lambda_2\lptppr(\gptp(\s,\eptp(\x)), \x)\right],\end{multline*}
where:
\begin{itemize}
	\item $\x$ is an input image, $\s$ is a segmentation mask (it serves as input to the generator), $\dptp_1$ and $\dptp_2$ are two discriminators that have the same architecture but operate on different image scales (original and 2x downsampled), $\lptpg(G, D)$ is the regular GAN loss;
	\item $\eptp(\x)$ is the feature encoder network that encodes low-level features of the objects with instance-wise pooling; its output is fed to the generator $\gptp(\s,\eptp(\x))$ and can be used to manipulate object style in generated images (see~\cite{8579015} for more details);
	\item $\lptpfm$ is the \emph{feature matching loss} that makes features at different layers of the discriminators (we denote the input to the $i$th layer of $D$ as $D^{(i)}$) match for $\x$ and $G(\s)$:
	$$\lptpfm(G,D)=\EEE{(\s,\x)}{\sum_{i=1}^L\frac{1}{N_i}\left\|D^{(i)}(\s,\x)-D^{(i)}(\s,G(\s, \eptp(\x))\right\|_1};$$
	\item $\lptppr$ is the \emph{perceptual reconstruction loss} for some feature encoder $F$:
	$$\lptppr(G,D)=\EEE{(\s,\x)}{\sum_{i=1}^{L'}\frac{1}{M_i}\left\|F^{(i)}(\x)-F^{(i)}(G(\s, \eptp(\x)))\right\|_1}.$$
\end{itemize}
The real dataset in~\cite{DBLP:journals/corr/abs-1901-06219} consisted of only $60$ manually annotated $1920\times 1200$ RGB images (with another $40$ images used for testing), albeit with a lot of annotated objects ($669$ blood cells per image on average). Bailo et al. also developed a scheme for sampling randomized but realistic segmentation masks to use for synthetic data generation. They report improved segmentation results with FCN and improved detection with Faster R-CNN when trained on a combination of real and synthetic data. 

Zhao et al.~\cite{DBLP:journals/corr/ZhaoLC17} consider the problem of generating filamentary structured images, such as retinal fundus and neuronal images, from a ground truth segmentation map, with an emphasis on generating images in multiple different styles. Their FILA-sGAN approach is based on GAN-based image style transfer ideas~\cite{Ulyanov:2016:TNF:3045390.3045533,Johnson2016PerceptualLF}. Its generator loss function is
$$\lfil = \lfilgan(\gfil,\dfil) + \lambda_1\lfilcont(\gfil) +\lambda_2\lfilsty(\gfil) + \lambda_3\lfiltv(\gfil),\text{ where}$$
\begin{itemize}
	\item $\gfil:\y\to\hx$ is a generator that takes a binary image and produces a ``phantom'' $\hx$, $\dfil$ is a synthetic vs. real discriminator, and $\lfilgan$ is the standard GAN loss;
	\item $\lfilcont$ is the \emph{content loss} that makes the filamentary structure of a generated phantom $\hx$ match the real raw image $\x$, evidenced through the features $\phi^{(i)}$ for some standard CNN feature extractor such as VGG-19:
	$$\lfilcont(\gfil)=\sum_{l}\frac{1}{W_lH_l}\left\|\phi^{(l)}(\x) - \phi^{(l)}(\hx)\right\|^2_F,$$
	where $\x$ is the real raw image, $l$ spans the CNN blocks and layers, $W_l$ and $H_l$ are the width and height of the corresponding feature maps, and $\|\cdot\|_F$ is the Frobenius matrix norm;
	\item $\lfilsty$ is the \emph{style loss} that minimizes the textural difference between $\hx$ and a style image $\x_s$:
	$$\lfilsty(\gfil) = \sum_l\frac{\omega_l}{W_lH_l}\left\|\mathbb{G}^{(l)}(\x_s) - \mathbb{G}^{(l)}(\hx)\right\|^2_F,$$
	where $\x_s$ is the style image, $\mathbb{G}^{(l)}$ is the Gram matrix of the features in CNN block $l$, and $\omega_l$ is its weight (a hyperparameter);
	\item $\lfiltv$ is the \emph{total variation loss} that serves as a regularizer and encourages $\hx$ to be smooth:
	$$\lfiltv(\gfil) = \sum_{i,j}\left(\left\|\hx_{i,j+1}-\hx_{i,j}\right\|^2_2 + \left\|\hx_{i+1,j}-\hx_{i,j}\right\|^2_2\right).$$
\end{itemize}
As a result, Zhao et al. report highly realistic filamentary structured images generated from a segmentation map and a single style image in a variety of different styles. Importantly for us, they also report improved segmentation results with state of the art approaches to the corresponding segmentation task.

In other works, Hou et al.~\cite{DBLP:journals/corr/abs-1712-05021} use GANs to refine synthesized histopathology images (similar to SimGAN discussed in Section~\ref{sec:gaze}), with improved nucleus segmentation and glioma classification results. Tang et al.~\cite{Tang19} use the \emph{pix2pix} model~\cite{Isola2017ImagetoImageTW} to generate realistic computed tomography (CT) images from customized lymph node masks, reporting improved lymph node segmentation with U-Net~\cite{RFB15a}. In~\cite{DBLP:journals/corr/abs-1904-09229}, the same researchers use the MUNIT (multimodal image-to-image translation) model~\cite{DBLP:journals/corr/abs-1804-04732} to generate realistic chest X-rays with custom abnormalities, reporting improved segmentation with both U-Net and their developed model XLSor. Han et al.~\cite{2019arXiv190604962H} were the first to apply 3D GAN-based DA to produce data for 3D object detection, i.e., bounding boxes; they use it in the context of synthetizing CT images of lung nodules.

In a related approach, synthetic data can be generated from real data, but in a different domain. For example, Zhang et al.~\cite{Zhang2018TranslatingAS} learn a CycleGAN-based architecture~\cite{8237506} to learn volume-to-volume (i.e., 3D) translation for unpaired datasets of CT and MR images (domain $A$ and domain $B$ respectively). Moreover, they augment the basic CycleGAN with segmentors $\svol_A$ and $\svol_B$ that help preserve segmentation mask consistency. In total, their model optimizes the loss function
\begin{multline*}
\lvol = \lvolg(\gvol_A,\dvol_A) + \lvolg(\gvol_B,\dvol_B) + \lambda_1\lvolcyc(\gvol_A,\gvol_B)+ \\ + \lambda_2\lvolshape(\svol_A,\svol_B,\gvol_A,\gvol_B),\quad\text{where}
\end{multline*}
\begin{itemize}
	\item $\gvol_A: B\to A$ and $\gvol_B: A\to B$ are CycleGAN generators, and $\dvol_A$ and $\dvol_B$ are discriminators in domains $A$ and $B$ respectively, trained to distinguish between real and synthetic (generated by $\gvol$) images by the standard GAN loss function $\lvolg$;
	\item $\lvolcyc$ is the cycle consistency loss
	\begin{align*}\lvolcyc(\gvol_A,\gvol_B)= & \EEE{\x_A}{\left\|\gvol_A(\gvol_B(\x_A))-\x_A\right\|_1} \\ +& \EEE{\x_B}{\left\|\gvol_B(\gvol_A(\x_B))-\x_B\right\|_1},\end{align*}
	\item $\svol_A: A\to Y$ and $\svol_B: B\to Y$ are segmentors that produce 3D segmentation masks, and $\lvolshape$ is the shape consistency loss
	\begin{multline*}
	\lvolshape(\svol_A,\svol_B,\gvol_A,\gvol_B) =\\ \EEE{\x_B}{-\frac 1N\sum_i \y_B^i\log\svol_A(\gvol_A(\x_B))_i} \\
	 + \EEE{\x_A}{-\frac 1N\sum_i \y_B^i\log\svol_A(\gvol_A(\x_B))_i},
	\end{multline*}
where $\y_A$ and $\y_B$ are ground truth segmentation results for $\x_A$ and $\x_B$ respectively.
\end{itemize}
Zhang et al. report that 3D segmentation in their architecture improves not only over the baseline model trained only on real data but also over the standard approach of fine-tuning $\svol_A$ and $\svol_B$ separately on generated synthetic data. Ben-Cohen et al.~\cite{DBLP:journals/corr/abs-1802-07846} present a similar architecture for cross-modal synthetic data generation of PET scans from CT images, also with improved segmentation results for lesion detection. Similar image-to-image translation techniques have been applied to generating images from 2D MR brain images to CT and back~\cite{DBLP:journals/corr/NieTPRS16,DBLP:journals/corr/abs-1708-01155,DBLP:journals/corr/abs-1805-10790}, PET to CT~\cite{Armanious2018MedGANMI}, cardiac CT to MR~\cite{Chartsias2017AdversarialIS}, virtual H\&E staining, including transformation from unstained to stained lung histology images~\cite{8265226} and stain style transfer~\cite{DBLP:journals/corr/abs-1804-01601}, multi-contract MRI (from contrast to contrast)~\cite{DBLP:journals/corr/abs-1802-01221}, 3D cross-modality MRI~\cite{8363653}, different styles of prostate histopathology~\cite{DBLP:conf/miccai/RenHSFQ18}, different datasets of chest X-rays~\cite{DBLP:journals/corr/abs-1806-00600}, and others.

Model-based domain adaptation (Section~\ref{sec:damodel}) has also been applied in the context of medical imaging. Often it has been used to do domain transfer between different types of real images, e.g. between different parts of the brain~\cite{10.1007/978-3-319-46723-8_38} or from \emph{in vitro} to \emph{in vivo} images~\cite{CONJETI20161}, but synthetic-to-real DA has also been a major topic. As early as 2013, Heimann et al.~\cite{10.1007/978-3-642-40760-4_7} generated synthetic training data in the form of digitally reconstructed radiographs for ultrasound transducer localization. To close the domain gap between synthetic and real images, they used standard instance weighting and found significant improvements in the resulting detections. Kamnitsas et al.~\cite{10.1007/978-3-319-59050-9_47} use unsupervised DA for brain lesion segmentation in 3D, switching from one type of MR images to another in domain adaptation. They use a state of the art 3D multi-scale fully convolutional segmentation network~\cite{KAMNITSAS201761} and a domain discriminator that makes intermediate feature representations of the segmentation networks indistinguishable between the domains.

In general, GAN-based architectures for medical imaging, either generating synthetic data, adapting real data from other domains, or , represent promising directions of further research and will, in our opinion, define state of the art in the field for years to come. However, at present the architectures used in different works differ a lot, and comparisons across different GAN-based architectures are usually lacking: each work compares their architecture only with the baselines. Further research and large-scale experimental studies are needed to determine which architectures work best for various domain adaptation problems related to medical imaging.

%% file: tikz_simgan.tex
\begin{tikzpicture}[node distance=.3cm]
	\node[diablo2,text width=5em,fill=blue!10] (n1) at (0,0) {Source domain data $\X_S$};
	\node[diablo2,text width=5em,fill=blue!10] (n2) at (0,2) {Target domain data $\X_T$};
	\node[diablo2,text width=5em,rotate=90] (n3) at (3.8,0.5) {Refiner $\grt$};
	\node[dialoss,text width=3.5em,minimum height=7em] (l1) at (8.75,0.75) {Loss\\[1mm] $\lgreal(\btheta)$,\\[1mm] $\l^{\textsc{Ref}}_{D}$};
	\node[diablo2,text width=12em,rotate=90] (n4) at (7,1) {Discriminator $\ddp$};

	\node[diablo2,text width=3em,fill=green!10] (l2) at (3.8,-2) {Loss $\lgreg$};

	\node[diablo2,text width=2em] (n5) at (2.1,-2) {$\psi$};
	\node[diablo2,text width=2em] (n6) at (5.5,-2) {$\psi$};

	\path[draw] (n2) -- node[ne,above,near start] {$\x_T$} (n4.north |- n2);
	\draw (n1) -- node[ne,below,near start] {$\x_S$} (n3.north |- n1);
	\draw (n5) -- (l2) (n6) -- (l2) (n4.south |- l1.west) -- (l1.west);
	\path[draw] (n3.south |- n1.east) -- node[ne,below,near start] {$\hx_S$} (n1.east -| n4.north);
	\draw (n1) -| (n5);
	\draw (n3.south |- n1.east) -| (n6);

	\begin{pgfonlayer}{background}
		\draw[lineblue,draw=green!50!black] (l1.west |- n4.south) -> (n3.south |- n4.south);
		\draw[lineblue,draw=green!50!black] (l1.west |- n4.310) -> (n4.south |- n4.310);
		\draw[lineblue,draw=green!50!black] (l2) -> (n3);
	\end{pgfonlayer}
\end{tikzpicture}\vspace{.2cm}

%% file: tikz_gazegan.tex
\begin{tikzpicture}[node distance=.3cm]
	\node[diablo2,text width=5em,fill=blue!10] (sd) at (0,4) {Synthetic data $\X_S$};
	\node[diablo2,text width=5em,fill=blue!10] (td) at (0,0) {Real data $\X_T$};

	\node[diablo2,text width=5em] (rg) at (3,4) {Refiner $\ggaze$};
	\node[diablo2,text width=5em] (rf) at (3,0) {Refiner $\fgaze$};

	\node[diablo2,text width=4.75em] (rf2) at (4.25,2.7) {Refiner $\fgaze$};
	\node[diablo2,text width=4.75em] (rg2) at (4.25,1.3) {Refiner $\ggaze$};

	\node[diablo2,text width=6em] (ds) at (5.8,4) {Discriminator $\drgaze$};
	\node[diablo2,text width=6em] (dr) at (5.8,0) {Discriminator $\dsgaze$};

	\node[dialoss,text width=10.5em,font=\footnotesize\sffamily] (lg) at (9.25,4) {LSGAN loss $\lgaze_{\mathrm{LSGAN}}(\ggaze,\drgaze,\X_S,\X_T)$};
	\node[dialoss,text width=10.5em,font=\footnotesize\sffamily] (lf) at (9.25,0) {LSGAN loss $\lgaze_{\mathrm{LSGAN}}(\fgaze,\dsgaze,\X_T,\X_S)$};
	\node[dialoss,text width=3.5em,minimum height=6em] (lc) at (7,2) {Loss $\lgaze_{\mathrm{Cyc}}$};
	\node[dialoss,text width=3.75em,minimum height=6em] (lgc) at (-0.2,2) {Loss $\lgaze_{\mathrm{GazeCyc}}$};
	
	\node[diablo2,text width=2em] (eg1) at (1.5,1.3) {$\egaze$};
	\node[diablo2,text width=2em] (eg2) at (1.5,2.7) {$\egaze$};

	\path[draw] (td) -- node[ne,above] {$\x_T$} (rf);
	\path[draw] (sd) -- node[ne,above] {$\x_S$} (rg);
	\draw (rg) -- (ds);
	\draw (rf) -- (dr);
	\draw (ds) -- (lg);
	\draw (dr) -- (lf);
	\draw (rg) -| (rf2);
	\draw (rf) -| (rg2);
	\draw (rf2) -- (lc.west |- rf2);
	\draw (rg2) -- (lc.west |- rg2);

	\draw (sd) -| (eg2);
	\draw (rg2) -- (eg1);

	\draw (eg1) -- (lgc.east |- eg1);
	\draw (eg2) -- (lgc.east |- eg2);




\end{tikzpicture}\vspace{.2cm}

%% file: tikz_t2net.tex
\begin{tikzpicture}[node distance=.3cm]
	\node[diablo2,text width=5em,fill=blue!10] (sd) at (-1,3.5) {Synthetic data $\X_S$};
	\node[diablo2,text width=5em,fill=blue!10] (td) at (-1,0) {Real data $\X_T$};

	\node[diablo2,text width=5em] (rg) at (2,3.5) {Refiner $\gsttwo$};	

	\node[diablo2,text width=4.75em] (rg2) at (3,0) {Refiner $\gsttwo$};

	\node[diablo2,text width=6em] (ds) at (4,2) {Discriminator $\drttwo$};
	\node[diablo2,text width=6em] (df) at (8,2.25) {Discriminator $\dfttwo$};

	\node[dialoss,text width=6.5em,font=\footnotesize\sffamily] (lg) at (1,2) {GAN loss $\lttwog(\gsttwo,\drttwo)$};
	\node[dialoss,text width=6.5em,font=\footnotesize\sffamily] (lr) at (6,0) {Reconstruction loss $\lttwo_r(\gsttwo)$};
	\node[dialoss,text width=6.5em,font=\footnotesize\sffamily] (lt) at (8,3.5) {Task loss $\lttwo_r(\gsttwo)$};
	\node[dialoss,text width=6.5em,font=\footnotesize\sffamily] (lf) at (8,1) {Feature GAN loss $\lttwogf$};

	\node[diablo2,text width=2em] (ft1) at (5,3.5) {$\fttwo$};
	\node[diablo2,text width=2em] (ft2) at (6,2) {$\fttwo$};

	\path[draw] (td) -- node[ne,above,near start] {$\x_T$} (rg2);
	\path[draw] (sd) -- node[ne,above] {$\x_S$} (rg);
	\draw (rg) -| node[ne,above,near start] {$\hxs$} (ds);
	\draw (rg2) -- node[ne,above] {$\hxt$} (lr);
	\draw (td.east) -| ++(1,1) -| (ds.south);
	\draw (td.east) -| ++(1,1) -| (lr.north);
	\draw (td.east) -| ++(1,1) -| (ft2.south);
	\draw (ds) -- (lg);
	\draw (rg) -- (ft1);
	\draw (ft1) -- (lt);
	\draw (sd.north) |- node[left,near start] {$\y_S$} ++(1,.5) -| (lt.north);
	\draw (ft2) -- (df.west |- ft2.east);
	\draw (ft1.east) -| ++(.75,-.88) -- (df.162);
	\draw (df) -- (lf);


\end{tikzpicture}\vspace{.2cm}

%% file: tikz_wanggan.tex
\begin{tikzpicture}[node distance=.3cm]
	\node[diablo2,text width=5em,fill=blue!10] (sd) at (0,4.5) {Synthetic data $\X_S$};
	\node[diablo2,text width=5em,fill=blue!10] (td) at (0,0) {Real data $\X_T$};

	\node[diablo2,text width=5em] (rg) at (3,4.5) {Refiner $\god$};
	\node[diablo2,text width=5em] (rf) at (3,0) {Refiner $\fod$};

	\node[diablo2,text width=4.75em] (rf2) at (5.2,3.3) {Refiner $\fod$};
	\node[diablo2,text width=4.75em] (rg2) at (5.2,1.2) {Refiner $\god$};

	\node[diablo2,text width=6em] (ds) at (5.8,4.5) {Discriminator $\dod_T$};
	\node[diablo2,text width=6em] (dr) at (5.8,0) {Discriminator $\dod_S$};

	\node[dialoss,text width=10.5em,font=\footnotesize\sffamily] (lg) at (9.25,4.5) {GAN loss $\lod_{\mathrm{GAN}}(\god,\dod_T,\X_S,\X_T)$};
	\node[dialoss,text width=10.5em,font=\footnotesize\sffamily] (lf) at (9.25,0) {GAN loss $\lod_{\mathrm{GAN}}(\fod,\dod_S,\X_T,\X_S)$};
	\node[dialoss,text width=3.2em] (lc) at (10.5,2.25) {Loss $\lodcyc$};
	\node[dialoss,text width=5em,minimum height=2.7em] (lfg) at (8.25,2.9) {Loss $\lodfg$};
	\node[dialoss,text width=5em,minimum height=2.7em] (lbg) at (8.25,1.6) {Loss $\lodbg$};

	\node[diablo2,text width=5.5em,minimum height=0,fill=blue!10] (sfg) at (2,2.9) {Segmentation $\m_{\mathrm{fg}}$};	
	\node[diablo2,text width=5.5em,minimum height=0,fill=blue!10] (sbg) at (2,1.6) {Segmentation $\m_{\mathrm{bg}}$};	

	\path[draw] (td) -- node[ne,above] {$\x_T$} (rf);
	\path[draw] (sd) -- node[ne,above] {$\x_S$} (rg);
	\draw (rg) -- (ds);
	\draw (rf) -- (dr);
	\draw (ds) -- (lg);
	\draw (dr) -- (lf);
	\draw (rg) -- (rf2);
	\draw (rf) -- (rg2);
	\draw (rf2.east) -- (lbg);
	\draw (rf2.east) -- (lfg);
	\draw (rg2.east) -- (lbg);
	\draw (rg2.east) -- (lfg);
	\draw (rf2.15) -| (lc.north);
	\draw (rg2.-15) -| (lc.south);
	\draw (sfg.-10) -- node[ne,at start,below,anchor=north west,font=\scriptsize\sffamily] {$\m_{\mathrm{fg}}(\x_S)$, $\m_{\mathrm{fg}}(\x_T)$} (lfg.west |- sfg.-10);
	\draw (sbg.10) -- node[ne,at start,above,anchor=south west,font=\scriptsize\sffamily] {$\m_{\mathrm{bg}}(\x_S)$, $\m_{\mathrm{bg}}(\x_T)$} (lbg.west |- sbg.10);
	\draw (sd.245) |- node[ne,near start,right] {$\x_S$} (sfg.165);
	\draw (sd.245) |- (sbg.165);
	\draw (td.75) |- node[ne,near start,right] {$\x_T$} (sbg.195);
	\draw (td.75) |- (sfg.195);






\end{tikzpicture}\vspace{.2cm}

%% file: tikz_ganin.tex
\begin{tikzpicture}[node distance=.3cm]
	\node[diablo2,text width=5em,fill=blue!10,rotate=90] (n1) at (-.5,0) {Input $\X$};
	\node[diablo2,text width=5em,minimum height=5em] (ref) at (2,0) {Feature extractor};

	\node[diablo2,text width=4em,fill=blue!10,rotate=90] (feat) at (4.5,0) {Features};
	
	\node[diablo2,text width=5em,minimum height=4em] (dc) at (7,-1.0) {Domain classifier};
	\node[diablo2,text width=5em,minimum height=4em] (lp) at (7,1.0) {Label predictor};
	
	\node[dialoss,text width=5.5em,minimum height=4em] (l1) at (10,1.0) {Label\\ prediction\\ loss};
	\node[dialoss,text width=5.5em,minimum height=4em] (l2) at (10,-1.0) {Domain\\ classification\\ loss};



	\draw (n1) -- node[ne,above,near start] {$\x$} (ref);
	\draw (ref) -- (feat);
	\draw (feat.west) |- (dc.west);
	\draw (feat.east) |- (lp.170);
	\draw (dc) -- (l2);
	\draw (lp.170 -| lp.east) -- (l1.west |- lp.170);

	\begin{pgfonlayer}{background}
		\draw[lineblue,draw=green!50!black,->] (l1.205) -> (n1.south |-l1.205);
		\draw[lineblue,draw=green!50!black,-] (l2.155) -- (feat.south |- l2.155);
		\draw[lineblue,draw=red,->] (feat.north |- l2.155) -> (n1.south |-l2.155);
	\end{pgfonlayer}
\end{tikzpicture}

%% file: tikz_dsn.tex
\begin{tikzpicture}[node distance=.3cm]
	\node[diablo2,text width=5em,fill=blue!10] (sd) at (-1,0) {Synthetic data $\X_S$};
	\node[diablo2,text width=5em,fill=blue!10] (td) at (-1,5) {Real data $\X_T$};

	\node[diablo2,text width=3.5em,font=\footnotesize\sffamily] (te) at (1.75,5) {Target encoder $\edsn_T$};
	\node[diablo2,text width=3.5em,font=\footnotesize\sffamily] (se) at (1.75,0) {Source encoder $\edsn_S$};
	\node[diablo2,text width=3.5em,font=\footnotesize\sffamily] (e1) at (1.75,1.5) {Shared encoder $\edsn$};
	\node[diablo2,text width=3.5em,font=\footnotesize\sffamily] (e2) at (1.75,3.5) {Shared encoder $\edsn$};

	\node[dialoss,text width=5em,minimum height=1em,font=\footnotesize\sffamily] (lsim) at (4,2.5) {Similarity loss $\ldsn_{\mathrm{sim}}$};
	\node[dialoss,text width=5.5em,minimum height=1em,font=\footnotesize\sffamily] (lds) at (4,0.75) {Difference loss $\ldsn_{\mathrm{diff},S}$};
	\node[dialoss,text width=5.5em,minimum height=1em,font=\footnotesize\sffamily] (ldt) at (4,4.25) {Difference loss $\ldsn_{\mathrm{diff},T}$};
	\node[dialoss,text width=3.5em,minimum height=1em,font=\footnotesize\sffamily] (ltask) at (6,2.5) {Task loss $\ldsn_{\mathrm{task}}$};
	\node[dialoss,text width=5.5em,minimum height=1em,font=\footnotesize\sffamily] (lrs) at (8.5,1.75) {Reconstruction loss $\ldsn_{\mathrm{rec},S}$};
	\node[dialoss,text width=5.5em,minimum height=1em,font=\footnotesize\sffamily] (lrt) at (8.5,3.25) {Reconstruction loss $\ldsn_{\mathrm{rec},T}$};
	
	\node[ne,circle,minimum size=5pt,inner sep=-2pt,draw=black,font=\LARGE] (op2) at (6,5) {$\boldsymbol{+}$};
	\node[ne,circle,minimum size=5pt,inner sep=-2pt,draw=black,font=\LARGE] (op1) at (6,0) {$\boldsymbol{+}$};
	\node[diablo2,text width=3.5em,font=\footnotesize\sffamily] (d1) at (7.25,0) {Decoder $\ddsn$};
	\node[diablo2,text width=3.5em,font=\footnotesize\sffamily] (d2) at (7.25,5) {Decoder $\ddsn$};
	
	



	\draw (sd) -- node[ne,below,near start] {$\xs$} (se);
	\draw (sd.east) -- ++(0.5,0) |- (e1.west);
	\draw (td) -- node[ne,above,near start] {$\xt$} (te);
	\draw (td.east) -- ++(0.5,0) |- (e2.west);
	\draw (e1.east) -| node[ne,above,near start] {$\h_S$} (lsim.south);
	\draw (e2.east) -| node[ne,below,near start] {$\h_T$} (lsim.north);
	\draw (e2.east) -| (ldt.south);
	\draw (e1.east) -| (lds.north);
	\draw (se.east) -| node[ne,below,near start] {$\h_S^{\mathrm{pri}}$} (lds.south);
	\draw (te.east) -| node[ne,above,near start] {$\h_S^{\mathrm{pri}}$} (ldt.north);
	\draw (se) -- (op1);
	\draw (e1) -| (op1);
	\draw (e1) -| (ltask);
	\draw (te) -- (op2);
	\draw (e2) -| (op2);
	\draw (op1) -- (d1);
	\draw (op2) -- (d2);
	\draw (d1) -| node[ne,above,near start] {$\hxs$} (lrs);
	\draw (d2) -| node[ne,above,near start] {$\hxt$} (lrt);

\end{tikzpicture}

%% file: privacy.tex
\section{Privacy guarantees in synthetic data}\label{sec:privacy}

\subsection{Differential privacy in deep learning}\label{sec:privdl}

In many domains, real data is not only valuable but also sensitive; it should be protected by law, commercial interest, and common decency. The unavailability of real data is exactly what makes synthetic data solutions attractive in these domains---but models for generating synthetic data have to train on real datasets anyway, how do we know we are not revealing it? A famous paper by Dinur and Nissim~\cite{Dinur:2003:RIW:773153.773173} showed that a few database queries (e.g., taking sums or averages of subsets) suffice to bring about strong violations of privacy. If a machine learning model has trained on a dataset with a few outliers, how do we know it doesn't ``memorize'' these outliers directly? For a sufficiently expressive model, such memorization is quite possible, and note that the outliers are usually the most sensitive data points. For instance, Carlini et al.~\cite{DBLP:journals/corr/abs-1802-08232} show that state of the art language models do memorize specific sequences of symbols, and one can extract, e.g., a secret string of numbers from the original dataset with a reasonably high success rate.

The field of differential privacy, pioneered by Dwork et al.~\cite{Dwork:2006:CNS:2180286.2180305,Dwork:2014:AFD:2693052.2693053,10.1007/978-3-540-79228-4_1} (the work~\cite{Dwork:2006:CNS:2180286.2180305} received the G{\"o}del Prize in 2017), was largely motivated by considerations such as above. In the main definition of the field, a randomized algorithm $A$ is called \emph{$(\epsilon,\delta)$-differentially private} if for any two databases $D$ and $D'$ that differ in only a single point and any subset of outputs $S$
$$p(A(D)\in S)\le e^{\epsilon}p(A(D')\in S)+\delta,$$
or, equivalently, for every point $s$ in the output range of $A$
$$\left|\frac{p(A(D)=s)}{p(A(D')=s)}\right|\le \epsilon\quad\text{with probability }1-\delta.$$
The intuition here is that an adversary who receives only the outputs of $A$ should have a hard time learning anything about any single point in $D$. Differential privacy has many desirable qualities such as composability (allowing for modular design of architectures), group privacy (graceful degradation when independency assumptions break), and robustness to auxiliary information that an adversary might have. 

In this section, we review the applications of differential privacy and related concepts to synthetic data generation. The purpose is similar: the release of a synthetic dataset generated by some model trained on real data should not disclose information regarding the individual points in this real dataset. Our review is slanted towards deep learning; for a more complete picture of the field we refer to the surveys in~\cite{Bindschadler18,10.1007/978-3-540-79228-4_1}. However, we do note the efforts devoted to generating differentially private synthetic datasets in classical machine learning. Lu et al.~\cite{6816689} develop a model for making sensitive databases private by fixing a set of queries to the database and perturbing the outputs to ensure differential privacy. Zhang et al. present the \emph{PrivBayes} approach~\cite{Zhang:2017:PPD:3155316.3134428}: construct a Bayesian network that captures the correlations and dependencies between data attributes, inject noise into the marginals that constitute this network, and then sample from the perturbed network to produce the private synthetic dataset. In a similar effort, the \emph{DataSynthetizer} model by Ping et al.~\cite{Ping:2017:DPS:3085504.3091117} is able to take a sensitive dataset as input and generate a synthetic dataset that has the same statistics and structure but at the same time provides differential privacy guarantees.

We also note some privacy-related applications of synthetic data that are not about differential privacy. For example, Ren et al.~\cite{10.1007/978-3-030-01246-5_38} present an adversarial architecture for video face anonymization; their model learns to modify the original real video to remove private information while at the same time still maximizing the performance of action recognition models (see also Section~\ref{sec:datapeople}).

As for deep learning, we begin with a brief overview of how to make complex high-dimensional optimization, such as training deep neural networks, respect privacy constraints. The basic approach to achieving differential privacy is to add noise to the output of $A$; many classical works on the subject focus on estimating and reducing the amount of noise necessary to ensure privacy under various assumptions~\cite{Dwork:2006:CNS:2180286.2180305,10.1007/11761679_29,10.1007/978-3-540-28628-8_32,10.1007/978-3-540-79228-4_1,NIPS2017_6850}. However, it is not immediately obvious how to apply this idea to a deep neural network. There are two major approaches to achieving differential privacy in deep learning.

Abadi et al.~\cite{Abadi:2016:DLD:2976749.2978318} suggest a method for controlling the influence of the training data during stochastic gradient descent called Differentially Private SGD (DP-SGD). Specifically, they clip the gradients on each SGD iteration to a predefined value of the $L_2$-norm and add Gaussian noise to the resulting gradient value. By careful analysis of the privacy loss variable, i.e., $\log\frac{p(A(D)=s)}{p(A(D')=s)}$ above, Abadi et al. show that the resulting algorithm preserves differential privacy under reasonable choices of the clipping and random noise parameters. Moreover, this is a general approach that is agnostic to the network architecture and can be extended to various first-order optimization algorithms based on SGD.

A year later, Papernot et al.~\cite{DBLP:conf/iclr/PapernotAEGT17} (actually, mostly the same group of researchers from Google) presented the Private Aggregation of Teacher Ensembles (PATE) approach. In PATE, the final ``student'' model is trained from an ensemble of ``teacher'' models that have access to sensitive data, while the ``student'' model only has access to (noisy) aggregated results of ``teacher'' models, which allows to control the disclosure and preserve privacy. A big advantage of this approach is that the ``teacher'' models can be treated as black-box while still providing rigorous differential privacy guarantees based on the same moments accounting technique from~\cite{Abadi:2016:DLD:2976749.2978318}. Incidentally, the best results were obtained with adversarial training for the ``student'' in a semi-supervised fashion, where the entire dataset is available for the ``student'' but labels are only provided for a subset of it, preserving privacy.

\subsection{Differential privacy guarantees for synthetic data generation}\label{sec:privgan}

The general approaches we have discussed in the previous section have been modified and applied for producing synthetic data with generative models, mostly, of course, with generative adversarial networks. Although the methods are similar, we note an important conceptual difference that synthetic data brings in this case. \emph{Model release} approaches in the previous section assumed access to and full control of model training. \emph{Data release} approaches (here we use the terminology from~\cite{DBLP:journals/corr/abs-1803-03148}) that perform synthetic data generation have the following advantages:
\begin{itemize}
	\item they can provide private data to third parties to construct better models and develop new techniques or use computational resources that might be unavailable to the holders of sensitive data;
	\item moreover, these third parties are able to pool synthetic data from different sources, while in the model release framework this would require a transfer of sensitive data;
	\item synthetic data can be either traded of freely made public, which is an important step towards reproducibility of research, especially in such fields as bioinformatics and healthcare, where reproducibility is an especially important problem and where, at the same time, sensitive data abounds.
\end{itemize}

In this section, we discuss existing constructions of GANs that provide rigorous privacy guarantees for the resulting generated data. Basically, in the ideal case a differentially private GAN has to generate an artificial dataset that would be sampled from the same distribution $\pdata$ but with differential privacy guarantees as discussed above. One general remark that is used in most of these works is that in a GAN architecture, it suffices to have privacy guarantees or additional privacy-preserving modifications (such as adding noise) only in the discriminator since gradient updates for the generator are functions of discriminator updates. Another important remark is that in cases when we generate differentially private synthetic data, a drop in quality for subsequent ``student'' models trained on synthetic data is expected in nearly all cases, not because of any deficiencies of synthetic data vs. real in general but because the nature of differential privacy requires adding random noise to the generative model training.

Xie et al.~\cite{DBLP:journals/corr/abs-1802-06739} present the differentially private GAN (DPGAN) model, which is basically the already classical Wasserstein GAN~\cite{pmlr-v70-arjovsky17a,NIPS2017_7159} but with additional noise on the gradient of the Wasserstein distance, in a fashion following the DP-SGD approach (Section~\ref{sec:privdl}). They apply DPGAN to generate electronic health records, showing that classifiers trained on synthetic records have accuracy approaching that of classifiers trained on real data, while guaranteeing differential privacy. This was further developed by Zhang et al.~\cite{DBLP:journals/corr/abs-1801-01594} who used the Improved WGAN framework~\cite{NIPS2017_7159} and obtained excellent results on the synthetic data generated from various subsets of the LSUN dataset~\cite{DBLP:journals/corr/abs-1801-01594}, which is already a full-scale image dataset, albeit low-resolution ($64\times 64$). Beaulieu-Jones et al.~\cite{Beaulieu-Jones159756} apply the same idea to generating electronic health records, specifically training on the data of the Systolic Blood Pressure Trial (SPRINT) data analysis challenge~\cite{doi:10.1056/NEJMoa1511939,doi:10.1056/NEJMe1513991}, which are in nature low-dimensional time series. They used the DP-SGD approach for the Auxiliary Classifier GAN (AC-GAN) architecture~\cite{Odena:2017:CIS:3305890.3305954} and studied how the accuracy of various classifiers drops when passing to synthetic data. Triastcyn and Faltings~\cite{DBLP:journals/corr/abs-1803-03148} continue this line of work and show that differential privacy guarantees can be obtained by adding a special Gaussian noise layer to the discriminator network. They show good results for ``student'' models trained on synthetically generated data for MNIST, but already at the SVHN dataset the performance degrades more severely. 

Bayesian methods are a natural fit for differential privacy since they deal with entire distributions of parameters and lend themselves easily to adding extra noise needed for DP guarantees. A Bayesian variant of the GAN framework, which provides representations of full posterior distributions over the parameters, was provided by Saatchi and Wilson~\cite{NIPS2017_6953}. Arnold et al.~\cite{arnold19} adapted the BayesGAN framework for differential privacy by injecting noise into the gradients during training, shown by Wang et al.~\cite{Wang:2015:PFP:3045118.3045383} to lead to DP guarantees. They apply the resulting DP-BayesGAN framework to microdata, i.e., medium-dimensional samples of 40 explanatory variables of different nature and one dependent variable.

As for the PATE framework, it cannot be directly applied to GANs since noisy aggregation of a PATE ensemble is not a differentiable function that could serve as part of a GAN discriminator. {\'{A}}cs et al.~\cite{DBLP:journals/corr/abs-1709-04514} proposed to use a differentially private clustering method to split the data into $k$ clusters, then train a separate generative models (the authors tried VAE) on their own clusters, and then create a mixture of the resulting models that will inherit differential privacy properties as well. A recent work by Jordon et al.~\cite{yoon2018pategan} circumvents the non-differentiability problem by training a ``student-discriminator'' on already differentially private synthetic data produced by the generator. The learning procedure alternates between updating ``teacher'' classifiers for a fixed generator on real samples and updating the ``student-discriminator'' classifier and the generator for fixed ``teachers''. PATE-GAN works well on low-dimensional data but begins to lose ground on high-dimensional datasets such as, e.g., the UCI Epileptic Seizure Recognition dataset (with $184$ features).

However, these results are still underwhelming; it has proven very difficult to stabilize GAN training with the additional noise necessary for differential privacy guarantees, which has not allowed researchers to progress to, say, higher resolution images so far. In a later work, Triastcyn and Faltings~\cite{2018arXiv180303148T} consider a different approach: they use the \emph{empirical DP} framework~\cite{Charest2017,AbowdSV13,doi:10.1111/rssa.12100,6547431}, an approach that empirically estimates the privacy of a posterior distribution, and the modification that ensures privacy is usually a sufficiently diffuse prior. In this framework, evaluating the privacy would reduce to training a GAN on the original dataset $D$, removing one sample from $D$ to obtain $D'$, retraining the GAN and comparing the probabilities of all outcomes, and so on, repeating these experiments enough times to obtain empirical estimates for $\epsilon$ and $\delta$. For realistic GANs, a large number of retrainings is impractical, so Triastcyn and Faltings modify this procedure to make it operate directly on the generated set $\tilde D$ rather than the original dataset $D$. They study the tradeoff of privacy vs. accuracy of the ``student'' models trained on synthetic data and show that GANs can fall into the region of practical values for both privacy and accuracy. Their proposed modification of the architecture (a single randomizing layer close to the end of the discriminator) strenghens DP guarantees while preserving good generation quality for datasets up to \emph{CelebA}~\cite{liu2015faceattributes}; in fact, it appears to serve as a regularizer and improve generation.

Frigerio et al.~\cite{10.1007/978-3-030-22312-0_11} extend the DPGAN framework to continuous, categorical, and time series data. They use the Wasserstein GAN loss function~\cite{NIPS2017_7159}, extending the moment accountant to this case. To handle discrete variables, the generator produces an output for every possible value with a softmax layer on top, and its results are sent to the discriminator. Bindschadler~\cite{Bindschadler18} presents a \emph{seedbased} modification of synthetic data generation: an algorithm that produces data records through a generative model conditioned on some seed real data record; this significantly improves quality but introduces correlations between real and synthetic data. To avoid correlations, Bindschadler introduces privacy tests that reject unsuitable synthetic data points. The approach can be used in complex models based on encoder-decoder architectures by adding noise to a seed in the latent space; it has been evaluated across different domains from census data to celebrity face images, the latter through a VAE/GAN architecture~\cite{pmlr-v48-larsen16}.

Finally, we note that synthetic data produced with differential privacy guarantees is also starting to gain legal status; in a technical report~\cite{Bellovin18}, Bellovin et al. from the Stanford Law School discuss various definitions of privacy from the point of view of what kind of data can be released. They conclude: ``...as we recommend, synthetic data may be combined with differential privacy to achieve a best-of-both-worlds scenario'', i.e., combining added utility of synthetic data produced by generative models with formal privacy guarantees.

\subsection{Case study: synthetic data in economics, healthcare, and social sciences}\label{sec:finance}

Synthetic data is increasingly finding its way into economics, healthcare, and social sciences in a variety of applications. We discuss this set of models and applications here since often the main concern that drives researchers in these fields to synthetic data is not lack of data \emph{per se} but rather privacy issues. A number of models that guarantee differential privacy have already been discussed above, so in this section we concentrate on other approaches and applications.

As far back as 1993, Rubin~\cite{Rubin93} discussed the dangers of releasing microdata (i.e., information about individual transactions) and the extremely complicated legal status of data releases, as the released data might be used to derive protected information even if it had been masked by standard techniques. To avoid these complications, Rubin proposed to use imputed synthetic data instead: given a dataset with confidential information, ``forget'' and impute confidential values for a sample from this dataset, using the same background variables but drawing confidential data from the predictions of some kind of imputation model. Repeating the process for several samples, we get a multiply-imputed population that can then be released. In the same year (actually, the same special issue of the \emph{Journal of Official Statistics}), Little~\cite{Little93} suggested to also keep the non-confidential part of the information to improve imputation. By now, synthetic datasets produced by multiple imputation are a well-established field of statistics, with applications to finance and economics~\cite{RD2007,DiCesare06}, healthcare~\cite{doi:10.1002/sim.3974}, social sciences~\cite{Burgard2017}, survey statistics~\cite{Abowd2006,Alfons2011}, and other domains. Since the main emphasis of the present survey is on synthetic data for deep learning, we do not go into details about multiple imputation and refer to the book~\cite{Drechsler11} and the main recent sources in the field~\cite{Raghunathan03,RePEc:bes:jnlasa:v:102:y:2007:m:december:p:1462-1471,Reiter05,Drechsler10}.

In a very recent work, Heaton and Witte~\cite{HW19} propose another interesting take on synthetic data in finance. They begin with the well-known problem of overfitting during backtesting: since there is a very large number of financial products and relatively short time series available for them, one can always find a portfolio (subset of products) that works great during backtesting, but it does not necessarily reflect future performance. The authors suggest to use synthetic data not to train financial strategies (they regard this as infeasible) but rather to \emph{evaluate} developed strategies, generating synthetic data with a different distribution of abnormalities and testing strategies for robustness in these altered circumstances. Interestingly, the motivation here is not to improve or choose the best strategies but to obtain evidence of their robustness that could be used for regulatory purposes. As a specific application, the authors use existing fraud detection algorithms to find anomalies in the Kaggle Credit Card Fraud Detection Dataset~\cite{PCBWB14} and generate synthetic data that balances the found abnormalities.

However, at present, we know of no direct applications where synthetically generated financial time series that would lead to improved results in financial forecasting, developing financial strategies, and the like. In general, financial time series are notorious for not being amenable to either prediction or accurate modeling, and even with current state of the art economic models we can hardly hope to generate useful synthetic financial time series any more than we can hope to generate meaningful text (see Section~\ref{sec:nlp}).

As for healthcare, this is again a field where the need for synthetic data was understood very early, and this need was mostly caused by privacy concerns: hospitals are required to protect the confidentiality of their patients. Ever since the first works in this direction, dating back to early 1990s, researchers mostly concentrated on generating synthetic electronic medical records (EMR) in order to preserve privacy~\cite{Barrows1996PrivacyCA}. In more recent work, MDClone~\cite{mdclone} is a system that samples synthetic EMRs from the distributions learned on existing cohorts, without actually reusing original data points. Walonoski et al.~\cite{10.1093/jamia/ocx079} present the \emph{Synthea} software suite designed to simulate the lifespans of synthetic patients and produce realistic synthetic EMRs. McLaghlan~\cite{McLachlan17} discusses realism in synthetic EMR generation and methods for its validation, and in another work presents a state transition machine that incorporates domain knowledge and clinical practice guidelines to generate realistic synthetic EMRs~\cite{7776401}.

Another related direction of research concentrates not on individual EMRs but on modeling entire populations of potential patients. Synthetic micro-populations produced by Smith et al.~\cite{doi:10.1068/a4147} are intended to match various sociodemographic conditions found in real cities and use them in imitational modeling to estimate the effect of interventions. Moniz et al.~\cite{Moniz2009ConstructionAV,LM08} create synthetic EMRs made available on the CDC Public Health grid for imitational modeling. Buczak et al.~\cite{Buczak2010} generate synthetic EMRs for an outbreak of a certain disease (together with background records). Kartoun~\cite{DBLP:journals/corr/Kartoun16} progressed from individual EMRs to entire virtual patient repositories, concentrating on preserving the correct general statistics while using simulated individual records. However, most of this work does not make use of modern formalizations of differential privacy or recent developments in generative models, and only very recently researchers have attempted to bring those into the healthcare microdata domain as well.

A direct appication of GANs for synthetic EMR generation was presented by Choi et al.~\cite{pmlr-v68-choi17a}. Their \emph{medGAN} model consists of a generator $\gmed$, discriminator $\dmed$, and an autoencoder with encoder $\encmed$ and decoder $\decmed$. The autoencoder is trained to reconstruct real data $\xt\sim\Xt$, while $\gmed$ learns to generate latent representations $\gmed(\z)$ from a random seed $\z$ such that $\dmed$ will not be able to differentiate between $\decmed(\gmed(\z))$ and a real sample $\xt\sim\Xt$. The privacy in the \emph{medGAN} model is established empirically, and the main justification for privacy is the fact that \emph{medGAN} uses real data only for the discriminator and never trains the generator on any real samples. We note, however, that in terms of generation \emph{medGAN} is not perfect: for example, Patel et al.~\cite{Patel2018CorrelatedDD} present the \emph{CorrGAN} model for correlated discrete data generation (with no regard for privacy) and show improvements over \emph{medGAN} with a relatively straightforward architecture.

The DP-SGD framework has also been applied to GANs in the context of medical data. We have discussed Beaulieu-Jones et al.~\cite{Beaulieu-Jones159756} above. Another important application for synthetic data across many domains, including but not limited to finance, would be to generate synthetic time series. This, however, has proven to be a more difficult problem, and solutions are only starting to appear. In particular, Hyland et al.~\cite{2017arXiv170602633E} present the Recurrent GAN (RGAN) and Recurrent Conditional GAN architectures designed to generate realistic real-valued multi-dimensional time series. They applied the architecture to generating medical time series (vitals measured for ICU patients) and reported successful generation and ability to train classifiers on synthetic data, although there was a significant drop in quality when testing on real data. Hyland et al. also discuss the possibility to use a differentially private training procedure, applying the DP-SGD framework to the discriminator and thus achieving differential privacy for the RGAN training. The authors report that after this procedure, synthetic-to-real test results deteriorate significantly but remain reasonable in classification tasks on ICU patient vitals.

Finally, we note another emerging field of research related to generating synthetic EMRs for the sake of privacy: generating clinical notes and free-text fields in EMRs with neural language models (see also Section~\ref{sec:nlp}). Latest advances in deep learning for natural language processing have led to breakthroughs in large-scale language modeling~\cite{radford2018improving,devlin2018bert,DBLP:journals/corr/abs-1801-06146}, and this has been applied to smaller datasets of clinical notes as well. Lee~\cite{Lee2018NaturalLG} uses an encoder-decoder architecture to generate chief complaint texts for EMRs. Guan et al.~\cite{8621223} propose a GAN architecture called \emph{mtGAN} (medical text GAN) for the generation of synthetic EMR text. It is based on the SeqGAN architecture~\cite{Yu:2017:SSG:3298483.3298649} and is trained with the REINFORCE algorithm; the primary difference is a condition added by Guan et al. to be able to generate EMRs for a specific disease or other features. Melamud and Shivade~\cite{Melamud2019TowardsAG} compare LSTM-based language models for generating synthetic clinical notes, suggesting a new privacy measure and showing promising results. Further advances in this direction may be related to the recently developed differentially private language models~\cite{brendan2018learning}.

%% file: future.tex
\section{Promising directions for future work}\label{sec:future}

\subsection{Procedural generation of synthetic data}\label{sec:procedural}

The first direction that we highlight as important for further study in the field of synthetic data is \emph{procedural generation}. Take, for instance, synthetic indoor scenes that we discussed in Section~\ref{sec:dataindoor}. Note that in the main synthetic dataset for indoor scenes, SUNCG, the 3D scenes and their layouts were created manually. While we have seen that this approach has an advantage of several orders of magnitude in terms of labeled images over real datasets, it still cannot scale up to millions of 3D scenes. The only way to achieve that would be to learn a model that can generate the \emph{contents} of a 3D scene (in this case, an indoor environment with furniture and supported objects), varying it stochastically according to some random input seed. This is a much harder problem than it might seem: e.g., a bedroom is much more than just a predefined set of objects placed at random positions.

There is a large related field of procedural content generation for video games~\cite{Shaker:2016:PCG:3029335,Togelius:2011:PCG:2000919.2000922,Hendrikx:2013:PCG:2422956.2422957}, but we highlight a recent work by Qi et al.~\cite{8578716} as representative for state of the art and more directly related to synthetic data. They propose a human-centric approach to modeling indoor scene layout, learning a stochastic scene grammar based on an attributed spatial AND-OR graph~\cite{Zhu:2006:SGI:1315336.1315337} that relates scene components, objects, and corresponding human activities. A scene configuration is represented by a parse graph whose probability is modeled with potential functions corresponding to functional grouping relations between furniture, relations between a supported object and supporting furniture, and human-centric relations between the furniture based on the map of sampled human trajectories in the scene. After learning the weights of potential functions corresponding to the relations between objects, MCMC sampling can be used to generate new indoor environments. Qi et al. train their model on the very same SUNCG dataset and show that the resulting layouts are hard to distinguish (by an automated state of the art classifier trained on layout segmentation maps) from the original SUNCG data.

In Section~\ref{sec:dataoutdoor}, we have already discussed \emph{ProcSy} by Khan et al.~\cite{Khan_2019_CVPR_Workshops}. In addition to randomizing weather and lighting conditions, another interesting part of their work is the procedural generation of cities and outdoor scenes. They base this procedural generation on the method of Parish and M\"{u}ller~\cite{Parish:2001:PMC:383259.383292}, which is in turn based on the notion of Lindenmayer systems (L-systems)~\cite{Prusinkiewicz13} and embodied in the \emph{CityEngine} tool~\cite{6737786} (see Section~\ref{sec:dataoutdoor}). They use a part of real \emph{OpenStreetMaps} data for the road network and buildings, but we hope that future work based on the same ideas can offer fully procedural modeling of cities and road networks.

We believe that procedural generation can lead to an even larger scale of synthetic data adoption in the future, covering cases when simply placing the objects at random is not enough. The nature of the model used for procedural generation may differ depending on the specific field of application, but in general for procedural generation one needs to train probabilistic generative models that allow for both learning from real or manually prepared synthetic data and then generating new samples.

\subsection{From domain randomization to the generation feedback loop}\label{sec:feedback}

The work~\cite{DBLP:journals/corr/abs-1810-05687} makes one of the first steps in a very interesting direction. They are also working on domain transfer, transferring continuous control policies for robotic arms from synthetic to real domain. But importantly, they attempt to close the feedback loop between synthetic data generation and domain transfer via domain randomization. Previous works on domain randomization (see above) manually tuned the distribution of simulation parameters $p_{\phi}(\xi)$ such that a policy trained on $D_{\xi\sim p_{\phi}}$ would perform well. In~\cite{DBLP:journals/corr/abs-1810-05687}, the parameters of $p_{\phi}(\xi)$ are learned automatically via a feedback loop from the results of real observations.

A similar approach on the level of data augmentation was presented in~\cite{NIPS2017_6916}. As we discussed in Section~\ref{sec:intro}, data augmentation differs from synthetic data in that it modifies real data rather than creates new; the modifications are usually done with predefined transformation functions (TFs) that do not change the target labels. This assumption is somewhat unrealistic: e.g., if we augment by shifting or cropping the image for image classification, we might crop out exactly the object that determines the class label. The work~\cite{NIPS2017_6916} relaxes this assumption, treating TFs as black boxes that might move the data point out of all necessary classes, into the ``null'' class, but cannot mix up different classes of objects. The authors train a generative sequence model with an adversarial objective that learns a sequence of TFs that would not move data points into the ``null'' class by training a null class discriminator $D^{\emptyset}_{\phi}$ and a generator $G_{\theta}$ for sequences of TFs $h_{L}\circ\ldots\circ h_1$:
$$\min_{\theta}\max_{\phi}\E_{\tau\sim G_{\theta}}\EEE{x\sim \U}{\log(1-D^{\emptyset}_{\phi}(h_{\tau_{L}}\circ\ldots\circ h_{\tau_1}(x)))}+\EEE{x'\sim\U}{\log(D^{\emptyset}_{\phi}(x'))},$$
where $\U$ is some distribution of (possibly unlabeled) data. Since TFs are not necessarily differentiable or deterministic, learning $G_{\theta}$ is defined in the syntax of reinforcement learning. Pashevich et al.~\cite{DBLP:journals/corr/abs-1903-07740} note that the space of augmentation functions is very large (for $8$ different transformations they estimate to have $\approx 3.6\cdot 10^{14}$ augmentation functions), and propose to use Monte Carlo Tree Search (MCTS)~\cite{10.1007/978-3-540-75538-8_7,10.1007/11871842_29} to find the best augmentations by automatic exploration. They apply this idea to augmenting synthetic images for sim-to-real policy transfer for robotic manipulation and report improved results in real world tasks such as cube stacking or cup placing. \emph{Google Brain} researchers Cubuk et al.~\cite{DBLP:journals/corr/abs-1805-09501} continue this work, presenting a framework for learning augmentation strategies from data. Their approach is modeled after recent advances in automated architecture search~\cite{DBLP:journals/corr/ZophL16,DBLP:journals/corr/ZophVSL17,DBLP:journals/corr/BakerGNR16,pmlr-v70-bello17a}: they use reinforcement learning to find the best augmentation policy composed of a number of parameterized operations. As a result, they significantly improve state of the art results on such classical datasets as CIFAR-10, CIFAR-100, and ImageNet.

Zakharov et al.~\cite{DBLP:journals/corr/abs-1904-02750} look at a similar idea from the point of view of domain randomization (see also Section~\ref{sec:domrand}). Their framework consists of a recognition network that does the basic task (say, object detection and pose estimation) and a deception network that transforms the synthetic input with an encoder-decoder architecture. Training is done in an adversarial way, alternating between two phases:
\begin{itemize}
	\item fixing the weights of the deception network, perform updates of the recognition network as usual, serving synthetic images transformed by the deception network as inputs;
	\item fixing the weights of the recognition network, perform updates of the deception network with the same objective but with reversed gradients, updating the deception network so as to make the inputs hard for the recognition network.
\end{itemize}
The deception network is organized and constrained in such a way that its transformations do not change the ground truth labels or change them in predictable ways. This adversarial framework exhibits performance comparable to state of the art domain adaptation frameworks and shows superior generalization capabilities (better avoiding overfitting).

There are two recent works that represent important steps towards closing this feedback loop.
First, the \emph{Meta-Sim} framework~\cite{DBLP:journals/corr/abs-1904-11621}, which we discussed in Section~\ref{sec:cgi} in the context of high-level procedural scene generation, also makes inroads in this direction: the distribution parameters for synthetic data generation are tuned not only to bring the synthetic data distribution closer to the real one but also to improve the performance on downstream tasks such as object detection. The difference here is that instead of low-level parameters of image augmentation functions \emph{Meta-Sim} adapts high-level parameters such as the synthetic scene structure captured as a scene graph.

Second, the \emph{Visual Adversarial Domain Randomization and Augmentation} (VADRA) model by Khirodkar et al.~\cite{Khirodkar2018VADRAVA} makes the next step in developing domain randomization ideas: instead of simply randomizing synthetic data or making it similar to real, let's learn a policy $\pi_{\omega}$ that generates rendering parameters in such a way that the downstream model learns best. They use the REINFORCE algorithm to obtain stochastic gradients for the objective $J(\omega)$ which consists of the downstream model performance (for supervised data) and the errors of a domain classifier (for unsupervised data; this is the ``adversarial'' part of VADRA). As a result, VADRA works much better for the syn-to-real transfer on problems such as object detection and segmentation than regular domain randomization. 

Similar ideas have been recently explored by Mehta et al.~\cite{Mehta2019ActiveDR}, who present \emph{Active Domain Randomization}, again learning a policy for generating better simulated instance, but this time in the context of generating Markov decision processes for reinforcement learning, Ruiz et al.~\cite{ruiz2018learning}, who also learn a policy $\pi_{\omega}$ that outputs the parameters for a simulator, learning to generate data to maximize validation accuracy, with reinforcement learning techniques, and Louppe et al.~\cite{2017arXiv170707113L}, who provide an inference algorithm based on variational approximations for fitting the parameters of a domain-specific non-differentiable simulator.

We believe that this meta-approach to automatically learning the best way to generate synthetic data, both high-level and low-level, is an important new direction that might work well for other applications too. In our opinion, this idea might be further improved by methods such as the SPIRAL framework by Ganin et al.~\cite{DBLP:journals/corr/abs-1804-01118} or neural painter models~\cite{Zheng2019StrokeNetAN,Huang2019LearningTP,Nakano2019NeuralPA} that train adversarial architectures to generate images in the form of sequences of brushstrokes or higher-level image synthesis programs with instructions such as ``place object $X$ at location $Y$''; these or other kinds of high-level descriptions for images might be more convenient for the generation feedback loop. We expect further developments in this direction in the nearest future.

\subsection{Improving domain adaptation with domain knowledge}\label{sec:knowledge}

To showcase this direction, we consider one more work on gaze estimation (see Section~\ref{sec:gaze}) that presents a successful application of a hybrid approach to image refinement. Namely, Wang et al.~\cite{8578151} propose a very different approach to generating synthetic data for gaze estimation: a hierarchical generative model (HGM) that is able to operate both top-down, generating new synthetic images of eyes, and bottom-up, performing Bayesian inference to estimate the gaze in a given new image.

\begin{figure}[!t]
\input{tikz_hgm}

\caption{General structure of the hierarchical generative model for eye image synthesis and gaze estimation~\cite{8578151}. Left to right: top-down image synthesis pipeline; right to left: bottom-up eye gaze estimation pipeline.}\label{fig:hgm}
\end{figure}
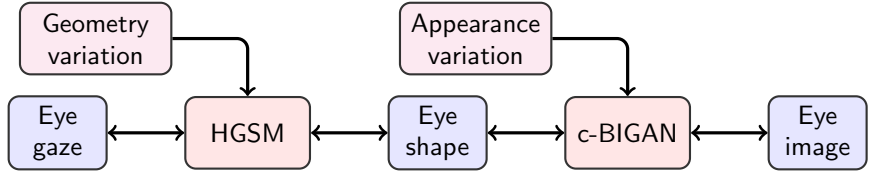

The general structure of their approach is shown on Fig.~\ref{fig:hgm}. Specifically, Wang et al. design a probabilistic hierarchical generative shape model (HGSM) based on 27 eye-related landmarks that together represent the shape of a human eye. The model connects personal parameters that define variation between humans, visual axis parameters that define eye gaze, and eye shape parameters. The structure of HGSM is based on anatomical studies, and its parameters are learned from the UnityEyes dataset~\cite{Wood:2016:LAG:2857491.2857492}. During generation, HGSM generates eye shape parameters (positions of the 27 landmarks in the eyeball's spherical coordinate system) based on the given gaze direction.

The second part of the pipeline generates the actual images with a conditional BiGAN (bidirectional GAN) architecture~\cite{DBLP:journals/corr/DonahueKD16}. Bidirectional GAN is an architecture that learns to transform data in both directions, from latent representations to the objects and back, while regular GAN's learn to generate only the objects from latent representations. The conditional BiGAN (c-BiGAN) modification developed in~\cite{8578151} does the same with a condition, which in this case are the eye shape parameters produced by HGSM. As a result, the model by Wang et al. can work in both directions:
\begin{itemize}
	\item generate eye images by sampling the gaze from a prior distribution, sampling 2D eye shape parameters from HGSM, and then using the generator $G$ of c-BiGAN to generate a refined image;
	\item infer gaze parameters from an eye image by first estimating the eye shape through the encoder $E$ of c-BIGAN and then performing Bayesian inference in the HGSM to find the posterior distribution of gaze parameters.
\end{itemize}
Wang et al. report performance improvements of the model itself applied to gaze estimation over~\cite{Shrivastava2017LearningFS} for sufficiently large training sets, and also show that the synthetic data generated by the model improves the results of standard gaze estimators (\emph{LeNet}, as used in~\cite{7299081}).

In general, we mark the approach of~\cite{8578151} to combining probabilistic generative models that incorporate domain knowledge and GAN-based architectures as a very interesting direction for further studies. We believe this approach can be suitable for applications other than gaze estimation.

\subsection{Additional modalities for domain adaptation architectures}\label{sec:futureadd}

Another natural idea that has not been used too widely yet is to use the additional data modalities such as depth maps or 3D volumetric data, which are available for free in synthetic datasets but usually unavailable in real ones, to improve 

Pioneering work in this direction has been recently done by Chen et al.~\cite{Chen_2019_CVPR}. They note that depth estimation and segmentation are related tasks that are increasingly learned together with multitask learning architectures~\cite{DBLP:journals/corr/KendallGC17,7298897,Xu2018PADNetMG} and then propose to use this idea to improve synthetic-to-real domain adaptation. Their model, called \emph{Geometrically Guided Input-Output Adaptation} (GIO-Ada), is based on the PatchGAN architecture~\cite{Isola2017ImagetoImageTW} intended for image translation.

\begin{figure}[!t]
\scalebox{.9}{
\input{tikz_gio}
}

\caption{General structure of the GIO-Ada model with input- and output-level domain adaptation~\cite{Chen_2019_CVPR}.}\label{fig:gioada}
\end{figure}
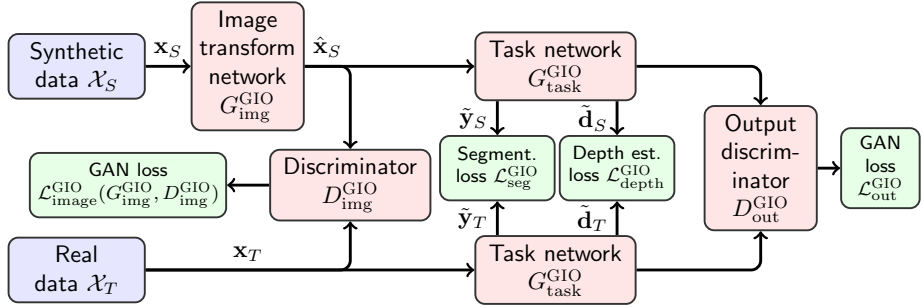

We have illustrated the GIO-Ada model on Figure~\ref{fig:gioada}. Similar to the refiners considered in Section~\ref{sec:synrefine}, they train a GAN generator to refine the synthetic image, but the refiner takes as input not just a synthetic image $\xs$ but an input triple $(\xs,\y,\bd)$, where $\xs$ is a synthetic image, $\y$ is its segmentation map, and $\bd$ is its depth map. Moreover, they also incorporate \emph{output-level adaptation}, where a separate generator predicts segmentation and depth maps, and a discriminator tries to distinguish whether these maps came from a synthetic transformed image or from a real one. In this way, the model can use the depth information to obtain additional cues to further improve segmentation on real images, and the output-level adaptation brings segmentation results on synthetic and real domains closer together.

Specifically, the GIO-Ada model optimizes, in an adversarial way, the following objective:
$$\min_{\ggioimg,\ggiotask}\max_{\dgioimg,\dgioout}\left[\lgioseg + \lambda_1\lgiodepth + \lambda_2\lgioimage + \lambda_3\lgioout\right],\text{ where:}$$
\begin{itemize}
	\item $\ggioimg$ is an image transformation network that performs input-level adaptation, i.e., produces a transformed image $\hx=\ggioimg(\x,\y,\bd)$, and $\ggiotask$ is the output-level adaptation network that predicts the segmentation and depth maps $(\ty, \td) = \ggiotask(\x)$;
	\item $\dgioimg$ is the image discriminator that tries to distinguish between $\hxs$ and $\xt$, and $\dgioout$ is the output discriminator that distinguishes between $(\ty_S,\td_S)=\ggiotask(\hxs)$ and $(\ty_T,\td_T)=\ggiotask(\xt)$;
	\item $\lgioseg$ is the segmentation cross-entropy loss $\lgioseg = \EEE{\xs\sim\psyn}{\mathrm{CE}\left(\ys,\ty_S\right)}$;
	\item $\lgiodepth$ is the depth estimation $L_1$-loss $\lgiodepth = \EEE{\xs\sim\psyn}{\left\|\bd_S-\td_S\right\|_1}$;
	\item $\lgioimage$ is the GAN loss for $\dgioimg$: 
	$$\lgioimage = \EEE{\xt\sim\preal}{\log\dgioimg(\xt)} + \EEE{\xs\sim\psyn}{\log(1 - \dgioimg(\hxs))};$$
	\item $\lgioout$ is the GAN loss for $\dgioout$:
	$$\lgioout = \EEE{\xt\sim\preal}{\log\dgioout(\ty_T,\td_T)} + \EEE{\xs\sim\psyn}{\log(1 - \dgioout(\ty_S,\td_S))}.$$
\end{itemize}

Chen et al. report promising results on standard synthetic-to-real adaptations: Virtual KITTI to KITTI and SYNTHIA to Cityscapes (see Section~\ref{sec:dataoutdoor}). This, however, looks to us as merely a first step in the very interesting direction of using additional data modalities easily provided by synthetic data generators to further improve domain adaptation.



%% file: tikz_hgm.tex
\begin{tikzpicture}[node distance=.3cm]
	\node[diablo2,text width=3em,fill=blue!10] (eg) at (0,0) {Eye gaze};
	\node[diablo2,text width=3em,fill=blue!10] (es) at (5,0) {Eye shape};
	\node[diablo2,text width=3em,fill=blue!10] (ei) at (10,0) {Eye image};
	\node[diablo2,text width=5em,fill=magenta!10] (gv) at (0.5,1.25) {Geometry variation};
	\node[diablo2,text width=5em,fill=magenta!10] (av) at (5.5,1.25) {Appearance variation};

	\node[diablo2,text width=4em] (h) at (2.5,0) {HGSM};
	\node[diablo2,text width=4em] (g) at (7.5,0) {c-BIGAN};

	\draw[<->] (eg) -- (h);
	\draw[<->] (h) -- (es);
	\draw[<->] (es) -- (g);
	\draw[<->] (g) -- (ei);
	\draw (gv) -| (h);
	\draw (av) -| (g);
\end{tikzpicture}\vspace{.2cm}

%% file: tikz_gio.tex
\begin{tikzpicture}[node distance=.3cm]
	\node[diablo2,text width=5em,fill=blue!10] (sd) at (-1,3.5) {Synthetic data $\X_S$};
	\node[diablo2,text width=5em,fill=blue!10] (td) at (-1,0.5) {Real data $\X_T$};

	\node[diablo2,text width=4em] (rg) at (1.5,3.5) {Image transform network $\ggioimg$};	
	\node[diablo2,text width=6em] (tn) at (6,3.5) {Task network $\ggiotask$};
	\node[diablo2,text width=6em] (tn2) at (6,0.5) {Task network $\ggiotask$};

	\node[diablo2,text width=4em] (do) at (9,2) {Output discriminator $\dgioout$};

	\node[diablo2,text width=6em] (ds) at (3,1.75) {Discriminator $\dgioimg$};

	\node[dialoss,text width=7.5em,font=\footnotesize\sffamily] (lg) at (-.25,1.75) {GAN loss $\lgioimage(\ggioimg,\dgioimg)$};
	\node[dialoss,text width=4.1em,font=\footnotesize\sffamily] (ls) at (5.15,2) {Segment. loss $\lgioseg$};
	\node[dialoss,text width=4.1em,font=\footnotesize\sffamily] (ld) at (6.9,2) {Depth est. loss $\lgiodepth$};
	\node[dialoss,text width=2.6em,font=\footnotesize\sffamily] (lt) at (10.75,2) {GAN loss $\lgioout$};


	\path[draw] (td) -| node[above,near start] {$\x_T$} (ds);
	\path[draw,very thick,->] (sd) -- node[ne,above] {$\x_S$} (rg);
	\draw (rg) -| node[ne,above,near start] {$\hxs$} (ds);
	\draw (rg) -- (tn);
	\draw (td) -- (tn2);
	\draw (do) -- (lt);
	\draw (tn) -| (do);
	\draw (tn2) -| (do);
	\draw (tn.south-|ls.north) -- node[left] {$\ty_S$} (ls);
	\draw (tn2.north-|ls.south) -- node[left] {$\ty_T$} (ls);
	\draw (tn.south-|ld.north) -- node[left] {$\td_S$} (ld);
	\draw (tn2.north-|ld.south) -- node[left] {$\td_T$} (ld);
	\draw (ds) -- (lg);


\end{tikzpicture}\vspace{.2cm}

%% file: conclusion.tex
\section{Conclusion}\label{sec:concl}

In this work, we have attempted a survey of one of the most promising general techniques on the rise in modern deep learning, especially computer vision: synthetic data. This source of virtually limitless perfectly labeled data has been explored in many problems, but we believe that many more potential use cases still remain.

In direct applications of synthetic data, we have discussed many different domains and use cases, from basic computer vision tasks such as stereo disparity estimation or semantic segmentation to full-scale simulated environments for autonomous driving, unmanned aerial vehicles, and robotics. In the domain adaptation part, we have surveyed a wide variety of generative models for synthetic-to-real refinement and for feature-level domain adaptation. As another important field of synthetic data applications, we have considered data generation with differential privacy guarantees. We have also reviewed the works dedicated to improving synthetic data generation and outlined potential promising directions for further research.

In general, throughout this survey we have seen synthetic data work well across a wide variety of tasks and domains. We believe that synthetic data is essential for further development of deep learning: many applications require labeling which is expensive or impossible to do by hand, other applications have a wide underlying data distribution that real datasets do not or cannot fully cover, yet other applications may benefit from additional modalities unavailable in real datasets, and so on. Moreover, we believe that synthetic data applications with be extended in the future. For example, while this survey does not yet have a section devoted to sound and speech processing, works that use synthetic data in this domain are already beginning to appear~\cite{DBLP:journals/corr/abs-1811-00707,Rygaard2015}. As synthetic data becomes more realistic (where necessary) and encompasses more use cases and modalities, we expect it to play an increasingly important role in deep learning.